%% file: main.tex
\pdfoutput=1

\documentclass[11pt]{article}

\usepackage[]{EMNLP2022}

\usepackage{times}
\usepackage{latexsym}

\usepackage[T1]{fontenc}

\usepackage[utf8]{inputenc}

\usepackage{microtype}

\usepackage{inconsolata}

\usepackage{amsfonts,amssymb}

\usepackage{amsmath}
\usepackage{cleveref}
\usepackage{titlesec}
\crefname{section}{§}{§§}
\Crefname{section}{§}{§§}
\usepackage[utf8]{inputenc} 
\usepackage[T1]{fontenc}    
\usepackage{hyperref}       
\usepackage{url}            
\usepackage{booktabs}       
\usepackage{multirow}
\usepackage{tabularx}
\usepackage{bm} 
\usepackage{amsfonts}       
\usepackage{nicefrac}       
\usepackage{microtype}      
\usepackage{xcolor}         
\usepackage{subfigure}

\usepackage[tikz]{bclogo}
\definecolor{msftBlue}{RGB}{0,164,239}
\definecolor{msftGreen}{RGB}{127,186,0}
\definecolor{msftYello}{RGB}{255,185,0}
\definecolor{msftBlack}{RGB}{0,0,0}
\newcommand{\finding}[1]{
	\begin{bclogo}[couleur= msftBlack!05, epBord= 1, arrondi=0.1, logo=\bclampe,marge= 2, ombre=true, blur, couleurBord=msftBlack!10, tailleOndu=3, sousTitre ={\em #1}]{} 
	\end{bclogo}
}

\usepackage{enumitem}

\newenvironment{itemize*}%
 {\leftmargini=20pt\begin{itemize}%
  \setlength{\itemsep}{3pt}%
  \setlength{\parskip}{0pt}%
  }%
 {\end{itemize}}
\newenvironment{enumerate*}%
 {\begin{enumerate}%
  \setlength{\itemsep}{0pt}%
  \setlength{\parskip}{0pt}}%
 {\end{enumerate}}

\title{Exploring Mode Connectivity for Pre-trained Language Models}

\author{
 Yujia~Qin$^{1}\thanks{\ \ Indicates equal contribution.}$\hspace{0.5em}, Cheng~Qian$^{1*}$, Jing~Yi$^{1*}$, Weize~Chen$^{1}$, Yankai~Lin$^{2,3}\thanks{\ \  Corresponding author.}$\hspace{0.5em}, Xu~Han$^{1}$, \\  \textbf{Zhiyuan~Liu}$^{1,4,5\dag}$, \textbf{Maosong~Sun$^{1,4,5\dag}$, Jie~Zhou$^6$} \\
 $^1$NLP Group, DCST, IAI, BNRIST, Tsinghua University, Beijing \\
 $^2$Gaoling School of Artificial Intelligence, Renmin University of China, Beijing \\
 $^3$Beijing Key Laboratory of Big Data Management and Analysis Methods, Beijing \\
 $^4$International Innovation Center of Tsinghua University, Shanghai \\
 $^5$Quan Cheng Laboratory $^6$Pattern Recognition Center, WeChat AI, Tencent Inc. \\
\texttt{\{qyj20, qianc20, yi-j20\}@mails.tsinghua.edu.cn}\\
}

\begin{document}
\maketitle

\input{sections/0_abstract}
\input{sections/1_introduction}
\input{sections/2_related_work}
\input{sections/3_methods}
\input{sections/4_experiments}
\input{sections/5_conclusions}

\section*{Limitations}
There are some limitations not well addressed in this paper:

\begin{itemize} [topsep=1pt, partopsep=1pt, leftmargin=12pt, itemsep=-3pt]
    \item The goal of this paper is to investigate the connection among different minima. However, since we use mode connectivity as the analysis tool, we only investigate the connection between two minima at a time. In this regard, it would be interesting to develop more advanced tools to explore the connection among multiple minima simultaneously, which is left as future work.
    \item We do not give a strict mathematical definition for ``good mode connectivity''. For instance, we do not set a specific threshold of performance drop (e.g., $> 5\%$ for the case of ``not well-connected minima''). We argue that this is because, all of our experimental results are significant enough, thus there is no need to follow a strict definition.
\end{itemize}



\bibliography{anthology,custom}
\bibliographystyle{acl_natbib}

\input{sections/appendix}

\end{document}

%% file: sections/0_abstract.tex
\begin{abstract}
Recent years have witnessed the prevalent application of pre-trained language models (PLMs) in NLP. From the perspective of parameter space, PLMs provide generic initialization, starting from which high-performance minima could be found. Although plenty of works have studied how to effectively and efficiently adapt PLMs to high-performance minima, little is known about the connection of various minima reached under different adaptation configurations. In this paper, we investigate the geometric connections of different minima through the lens of \textit{mode connectivity}, which measures whether two minima can be connected with a low-loss path. We conduct empirical analyses to investigate three questions: (1) how could hyperparameters, specific tuning methods, and training data affect PLM's mode connectivity? (2) How does mode connectivity change during pre-training? (3) How does the PLM's task knowledge change along the path connecting two minima? In general, exploring the mode connectivity of PLMs conduces to understanding the geometric connection of different minima, which may help us fathom the inner workings of PLM downstream adaptation. The codes are publicly available at \url{https://github.com/thunlp/Mode-Connectivity-PLM}.
\end{abstract}

%% file: sections/1_introduction.tex
\section{Introduction}
Recent years have witnessed the prevalent application of pre-trained language models (PLMs) in NLP~\citep{han2021pre}, with the state-of-the-art across various NLP tasks consistently being pushed~\citep{devlin2018bert,liu2019roberta,2020t5}. Through large-scale self-supervised training, PLMs acquire versatile semantic~\citep{liu-etal-2019-linguistic} and syntactic~\citep{tenney2019you} knowledge, which could be utilized when conducting transfer learning on downstream tasks.

From the perspective of parameter space, PLMs provide generic initialization for downstream adaptation. Starting from the initialization, many high-performance minima can be found through gradient-based optimization. Up to now, plenty of works have studied how to effectively and efficiently adapt PLMs to high-performance minima, including adjusting hyperparameters~\citep{liu-wang-2021-empirical}, conducting transfer learning using auxiliary training data~\citep{pruksachatkun-etal-2020-intermediate}, tuning PLMs in a parameter-efficient way~\citep{ding2022delta}, etc. Under different adaptation configurations, PLMs may finally reach local minima distributed in highly distinct regions. Although these minima all correspond to excellent performance (low loss), little has been known about their geometric connection in the parameter space.

A straightforward way to explore such geometric connection is to look into the loss landscape around different minima, which is inherently intractable due to the high dimensionality brought by the tremendous parameter size of PLMs. Instead of probing the full landscape, we propose to investigate the relation of different minima through the lens of \textit{mode connectivity}~\citep{garipov2018loss}, which measures whether two different minima can be connected via a parametric path, along which the loss of the downstream task remains low. Exploring the mode connectivity of PLMs contributes to understanding the geometric connection among different minima. Such connection reflects the inherent relation of various adaptation configurations and may help us fathom the inner workings of PLM downstream adaptation under different settings.

To the best of our knowledge, systematic studies for the mode connectivity of PLMs are still lacking. In this paper, we first investigate what factors may affect PLM's mode connectivity by answering the following research questions:

\begin{itemize}
    \item (\textsl{Q1}) How could different adaptation configurations (hyperparameters, tuning methods, and training data) affect PLM's mode connectivity?
    \item (\textsl{Q2}) How does mode connectivity change during pre-training?
\end{itemize}

We first consider the mode connectivity when different minima are trained on the same dataset. We investigate the effects of several hyperparameters (e.g., training data order, initialization of the tunable parameters, training steps, etc.) on PLM's mode connectivity, and find that among these factors, initialization has the greatest impact. In addition, we show that fine-tuning leads to better mode connectivity than parameter-efficient delta tuning (e.g., adapter~\citep{houlsby2019parameter}).

Then we extend the connectivity analysis to minima trained on different datasets. We demonstrate that: (1) the mode connectivity is good for two minima trained on data belonging to the same distribution, but without overlap of specific instances. This means instead of memorizing training data, PLMs learn advanced task-level knowledge during training, and mode connectivity originates from the high overlap of task knowledge of two minima. (2) Although two minima trained on different tasks are inherently disconnected, pre-training gradually pulls the optimal regions of different tasks closer in an implicit way. This phenomenon may help explain PLM's excellent cross-task transferability.

Beyond exploring the effects that could affect PLM's mode connectivity, we also study the intrinsic properties of model solutions between two minima, which leads to the third question:
\begin{itemize}
    \item (\textsl{Q3}) How does the PLM's task knowledge change along the path connecting two minima?
\end{itemize}

In the experiments, we observe that for two minima obtained independently on two tasks, when traversing from the minima trained on a source task to that of a target task, a PLM suffers from catastrophic forgetting~\citep{mccloskey1989catastrophic} on the source task, and gradually absorbs the knowledge from the target task. Besides, PLMs prioritize forgetting those elusive source knowledge and acquiring easy-to-grasp target knowledge.

In general, to fathom the connection of minima reached under different settings, we conduct empirical studies on the mode connectivity of PLMs. We also show that our findings may have potential significance in broad research areas, such as designing better ensemble methods for PLMs, understanding the task-level transferability of PLMs and revealing the mechanism of PLM downstream adaptation. We expect our evaluation setup and findings could inspire more future works in this field.

%% file: sections/2_related_work.tex
\section{Related Work}
\paragraph{Adaptation Strategies of PLMs.}
To effectively and efficiently utilize the knowledge learned during pre-training, many strategies have been developed to better tune a PLM, including: (1) \textit{hyperparameter search}, which aims to find an optimal hyperparameter configuration through traditional grid search or modern automated search~\citep{liu-wang-2021-empirical}; (2) \textit{pre-finetuning}, which trains PLMs on intermediate auxiliary tasks before fine-tuning on a target task~\citep{pruksachatkun-etal-2020-intermediate,aghajanyan-etal-2021-muppet}. In this way, PLMs achieve better downstream performance by taking advantage of cross-dataset knowledge transfer; (3) \textit{prompt learning}, which casts downstream tasks into the form of the pre-training objective by leveraging natural language prompts~\citep{NEURIPS2020_1457c0d6,schick-schutze-2021-exploiting,schick-schutze-2021-just}. Prompt learning exhibits superior performance especially under the few-shot and zero-shot scenarios; (4) \textit{delta tuning} (also known as parameter-efficient tuning). Optimizing all the parameters of a PLM is computationally cumbersome. As a lightweight alternative, delta tuning optimizes only a few tunable parameters and keeps other parameters frozen, achieving comparable performance to fine-tuning~\citep{ding2022delta}.

Although plenty of adaptation strategies have been developed to better tune a PLM, little is understood about the connection of local minima reached under different training configurations. In this work, we take the first step by utilizing mode connectivity as the analysis tool.

\paragraph{Mode Connectivity for Neural Networks.}

Despite being extremely high-dimensional, the loss landscape of neural networks exhibits a simple geometric pattern of mode connectivity~\citep{garipov2018loss,freeman2016topology,draxler2018essentially}. It is shown that starting from different initialization, the local minima obtained by gradient-based optimizations are often connected by low-loss paths, along which high-performance solutions could be easily found and ensembled to achieve better performance~\citep{garipov2018loss}. These paths are typically \textit{non-linear} curves, which require a process of curve finding with task supervision. Excellent mode connectivity indicates that different minima are not isolated points in the parameter space, but essentially form a connected manifold~\citep{draxler2018essentially}.

\citet{frankle2020linear} further contend that from the same initialization, local minima obtained with different training data order can be connected by a \textit{linear} low-loss path, reducing the burden of curve finding. Such a phenomenon is dubbed as linear mode connectivity, which is closely related to lottery ticket hypothesis~\citep{frankle2018lottery}, and has direct implications for continual learning \citep{mirzadeh2020linear}. Compared with the non-linear counterpart, linear mode connectivity is a stronger constraint, requiring that the convex combination of two minima stay in the same loss basin.

Previous works typically study mode connectivity using non-pretrained models in the field of computer vision. Until recently, \citet{neyshabur2020being} observe linear mode connectivity on pre-trained vision models. Despite the great efforts spent, a systematic understanding of the mode connectivity of PLMs is still lacking. In this paper, we focus on investigating the effects that would influence PLM's mode connectivity and analyze the knowledge variation along the connecting path. Different from existing works that study mode connectivity for minima trained on the same dataset, we additionally extend the analysis to different datasets.

%% file: sections/3_methods.tex
\section{Mode Connectivity Evaluation}
\label{sec:method}

\paragraph{Preliminaries.} Consider adapting a PLM on a downstream task, let $C_1$ and $C_2$ be two distinct sets of training configurations that may differ in hyperparameters or data. We use $C_1$ and $C_2$ to train two copies of the PLM independently. The specific tuning method determines the tunable parameters $\theta_0$. After training, $\theta_0$ is adapted to $\theta_{C_1} \in \mathbb{R}^{|\theta_0|}$ and $\theta_{C_2} \in \mathbb{R}^{|\theta_0|}$, respectively, where $|\theta_0|$ denotes the total number of tunable parameters.

\paragraph{Connecting Path.} Based on \citet{frankle2020linear}, the same initialization generally leads to a linear low-loss path between two minima. Besides, compared with the non-linear counterpart, linearity is a more favorable property, which indicates a closer connection between different minima. Therefore, our first step is to investigate whether PLMs have good \textit{linear} mode connectivity. Specifically, assume a continuous curve $\phi (\alpha)\!:\! [0,1] \!\rightarrow\! \mathbb{R}^{|\theta_0|}$ connecting $\theta_{C_1}$ and $\theta_{C_2}$, satisfying $\phi(0) = \theta_{C_1}$ and $\phi(1) = \theta_{C_2}$, we consider a linear path as follows:
\begin{equation}
\label{eq:linear}
\small
\begin{aligned}
    \phi (\alpha) = (1-\alpha) \!\cdot\! \theta_{C_1} + \alpha \!\cdot\! \theta_{C_2}.
\end{aligned}
\end{equation}

\paragraph{Connectivity Criterion.} After defining the curve connecting both minima, we traverse along the curve and evaluate the loss and performance of the interpolations. We deem two minima $\theta_{C_1}$ and $\theta_{C_2}$ mode connected if there does not exist a significant loss barrier or performance drop along the defined curve between $\theta_{C_1}$ and $\theta_{C_2}$. In the experiments, we evaluate evenly distributed interpolations on $\phi(\alpha)$.

%% file: sections/4_experiments.tex
\section{Empirical Analysis}
\label{sec:empirical_analysis}
In this section, we conduct experiments to investigate the aforementioned research questions.

\finding{Q1. (a) How could different hyperparameters and the specific tuning method affect the mode connectivity of PLMs?}

We first investigate the effects of several hyperparameters that could affect PLM's mode connectivity, including (1) training data order, initialization of tunable parameters, training step (main paper), (2) learning rate and batch size (\cref{sec:lr_bs}). To explore the effects of the specific tuning method, we experiment with both fine-tuning and a representative delta tuning method, i.e., adapter~\citep{houlsby2019parameter}. Adapter inserts tunable modules into a PLM and keeps other parameters fixed during adaptation. Unless otherwise specified, we mainly conduct the experiments using $\text{T5}_{\texttt{BASE}}$~\citep{2020t5}, and choose two representative NLP tasks (MNLI~\citep{williams-etal-2018-broad} and ReCoRD~\citep{zhang2018record}).

\begin{figure*}[!t]
    \centering
    \subfigure{\includegraphics[width=0.24\textwidth]{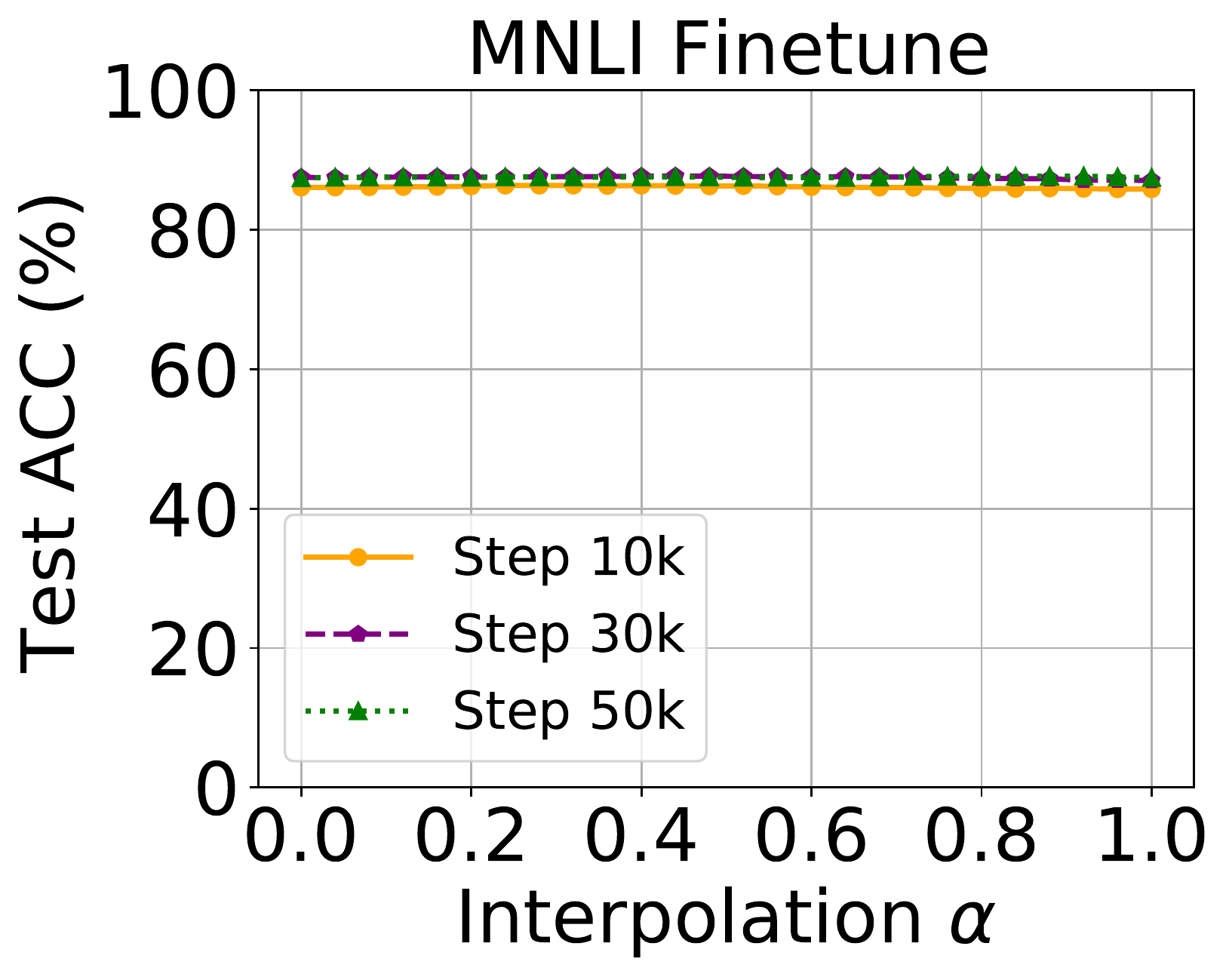}}
    \subfigure{\includegraphics[width=0.24\textwidth]{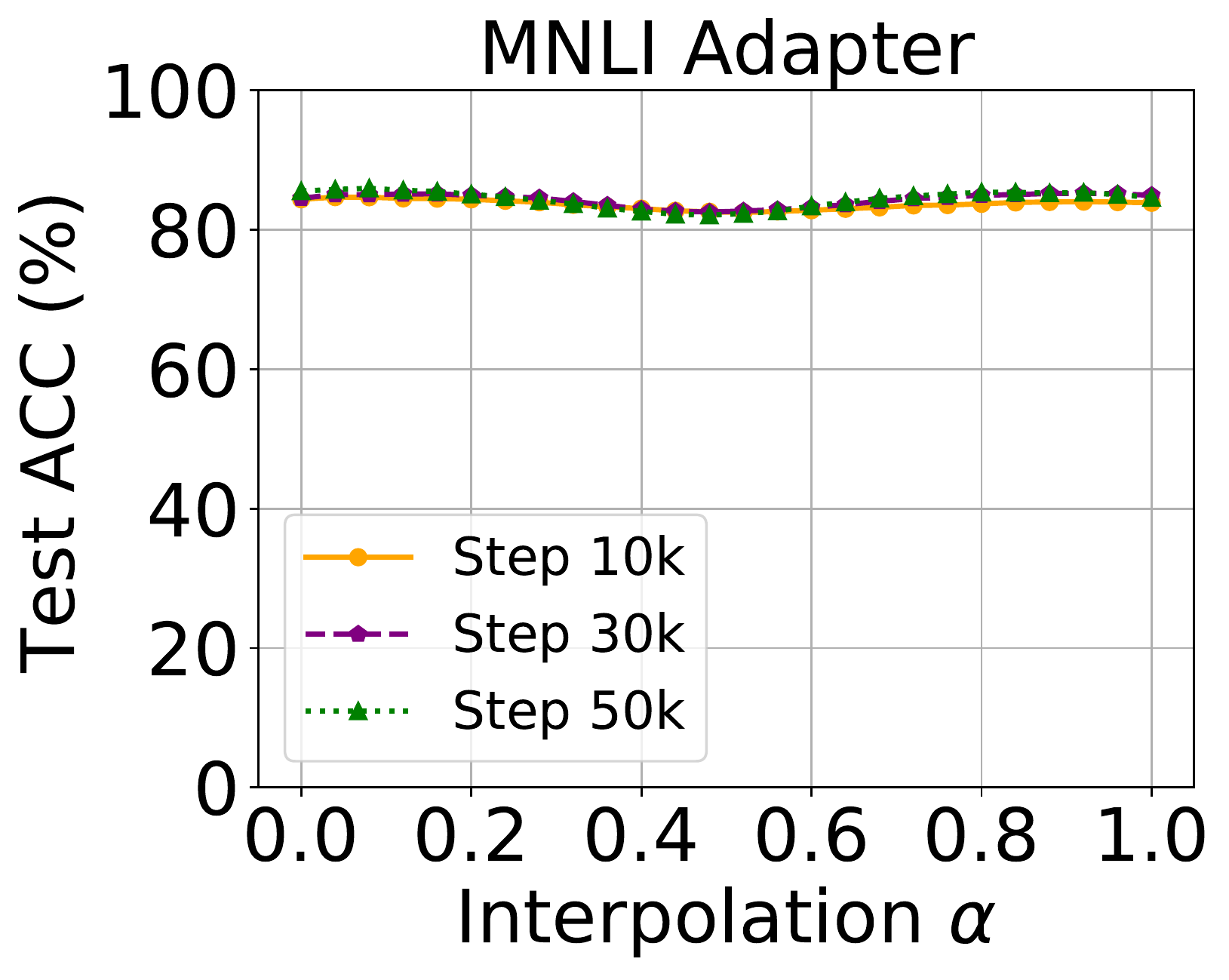}} 
    \subfigure{\includegraphics[width=0.24\textwidth]{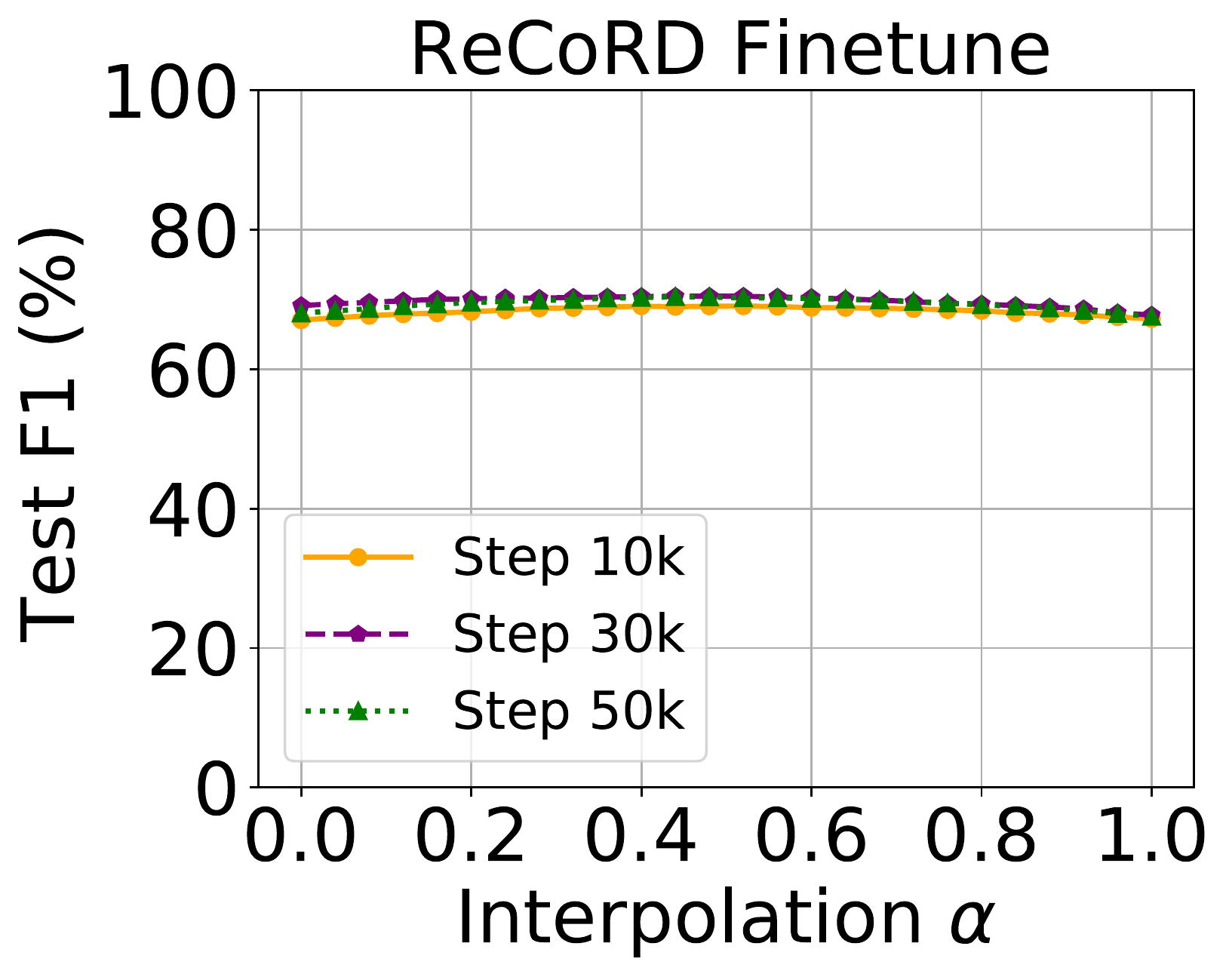}} 
    \subfigure{\includegraphics[width=0.24\textwidth]{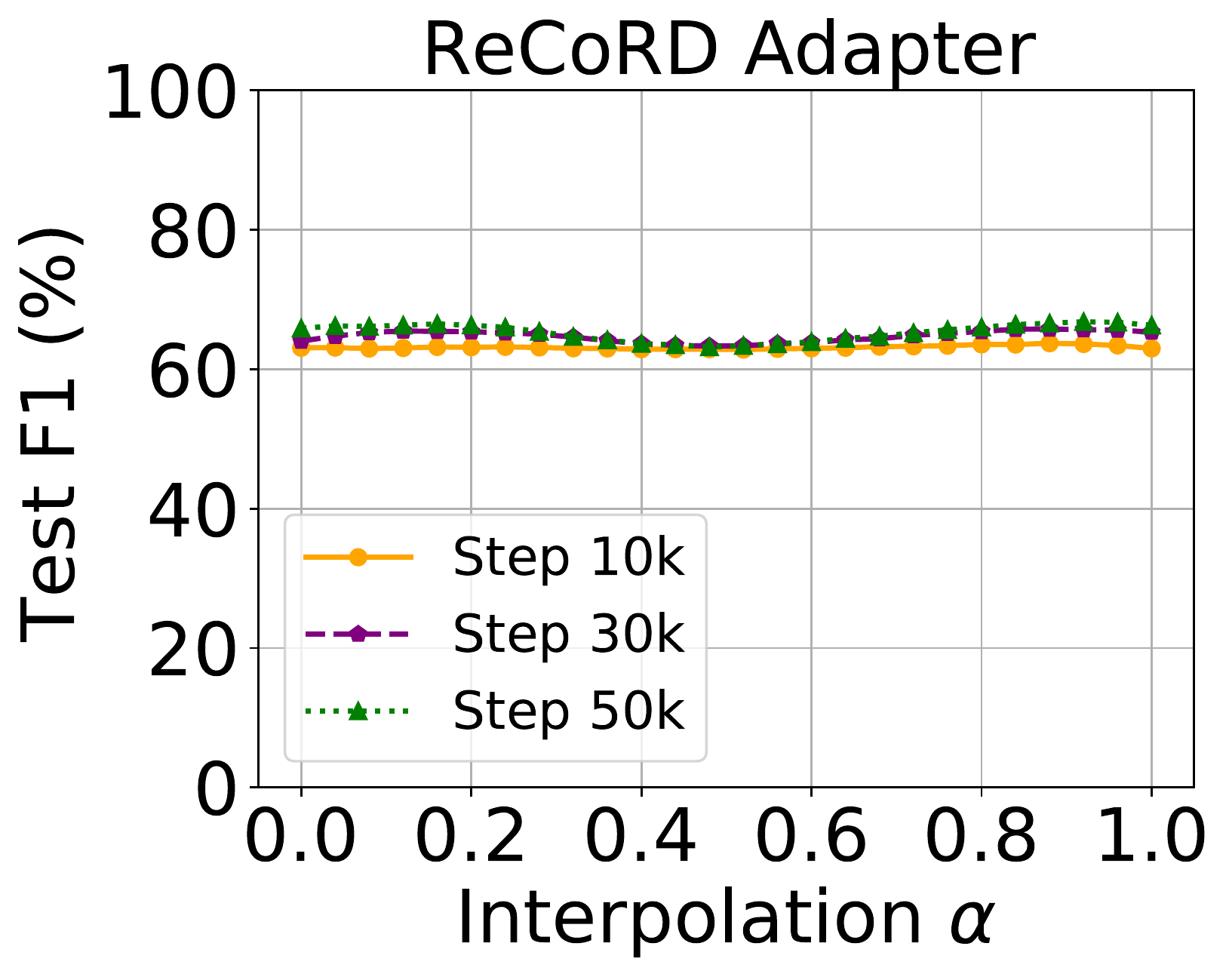}}
    \caption{The performance of linear interpolations between two minima trained with different training data order.}
    \label{fig:order}
\end{figure*}

\begin{figure*}[!t]
    \centering
    \subfigure{\includegraphics[width=0.24\textwidth]{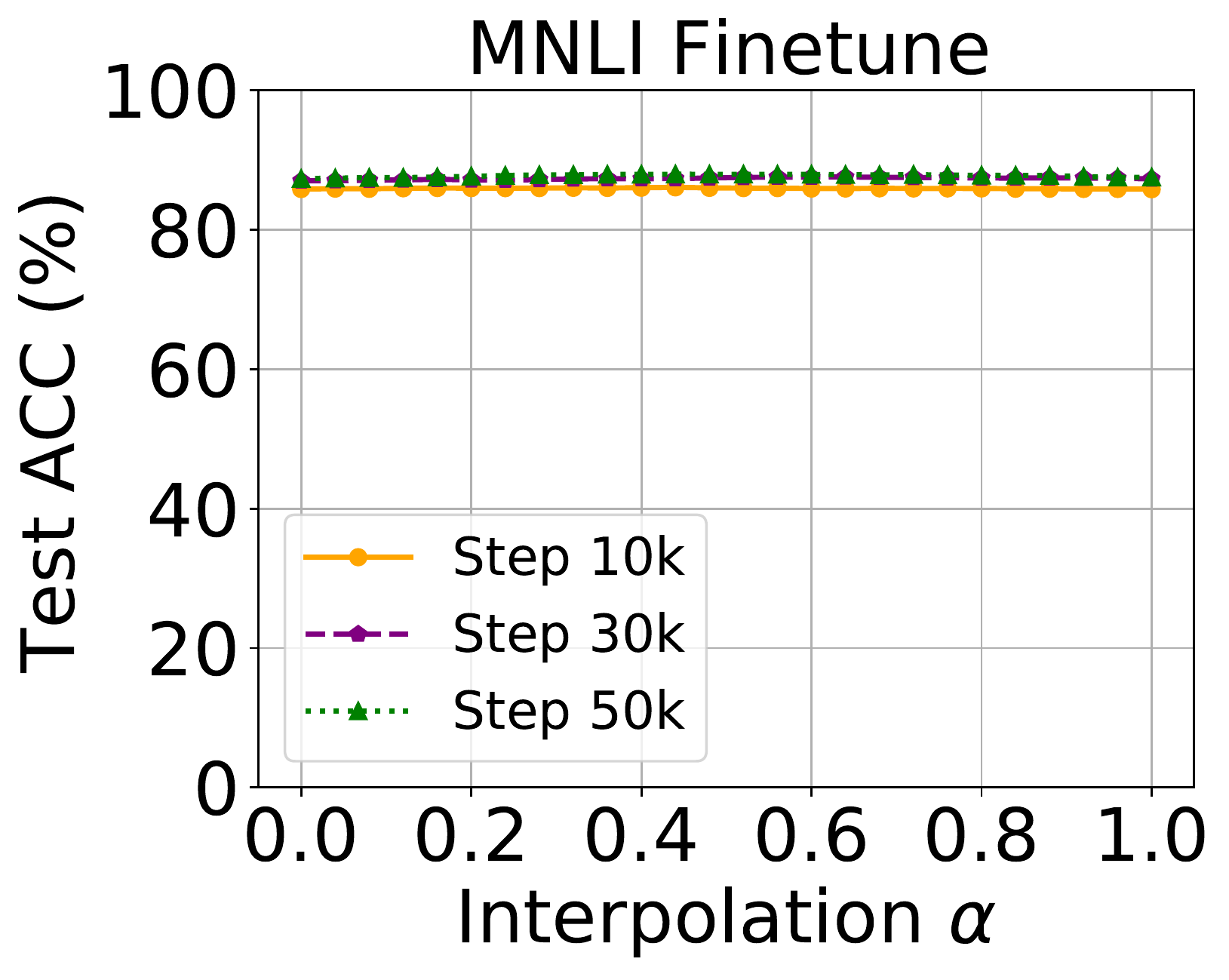}} 
    \subfigure{\includegraphics[width=0.24\textwidth]{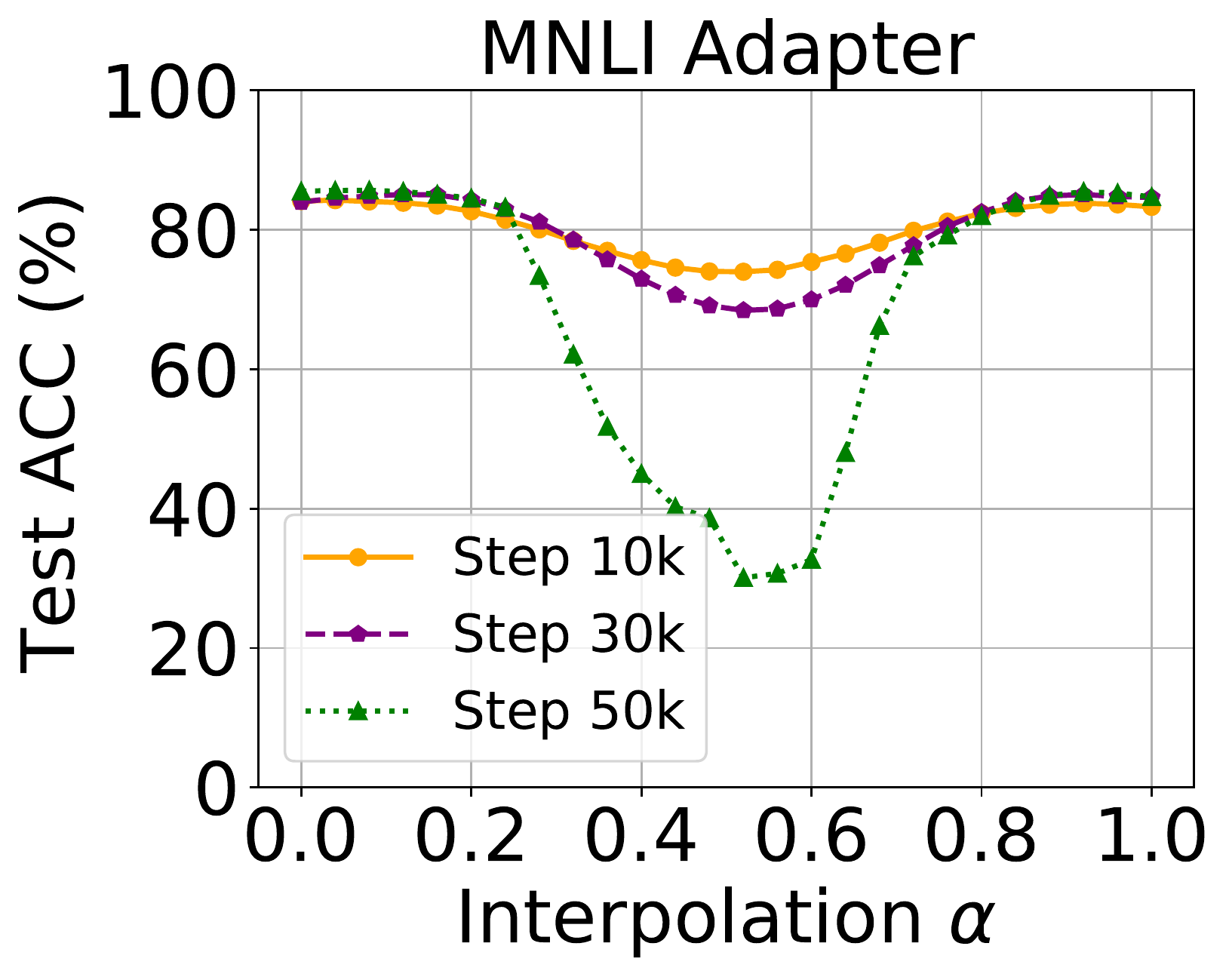}} 
    \subfigure{\includegraphics[width=0.24\textwidth]{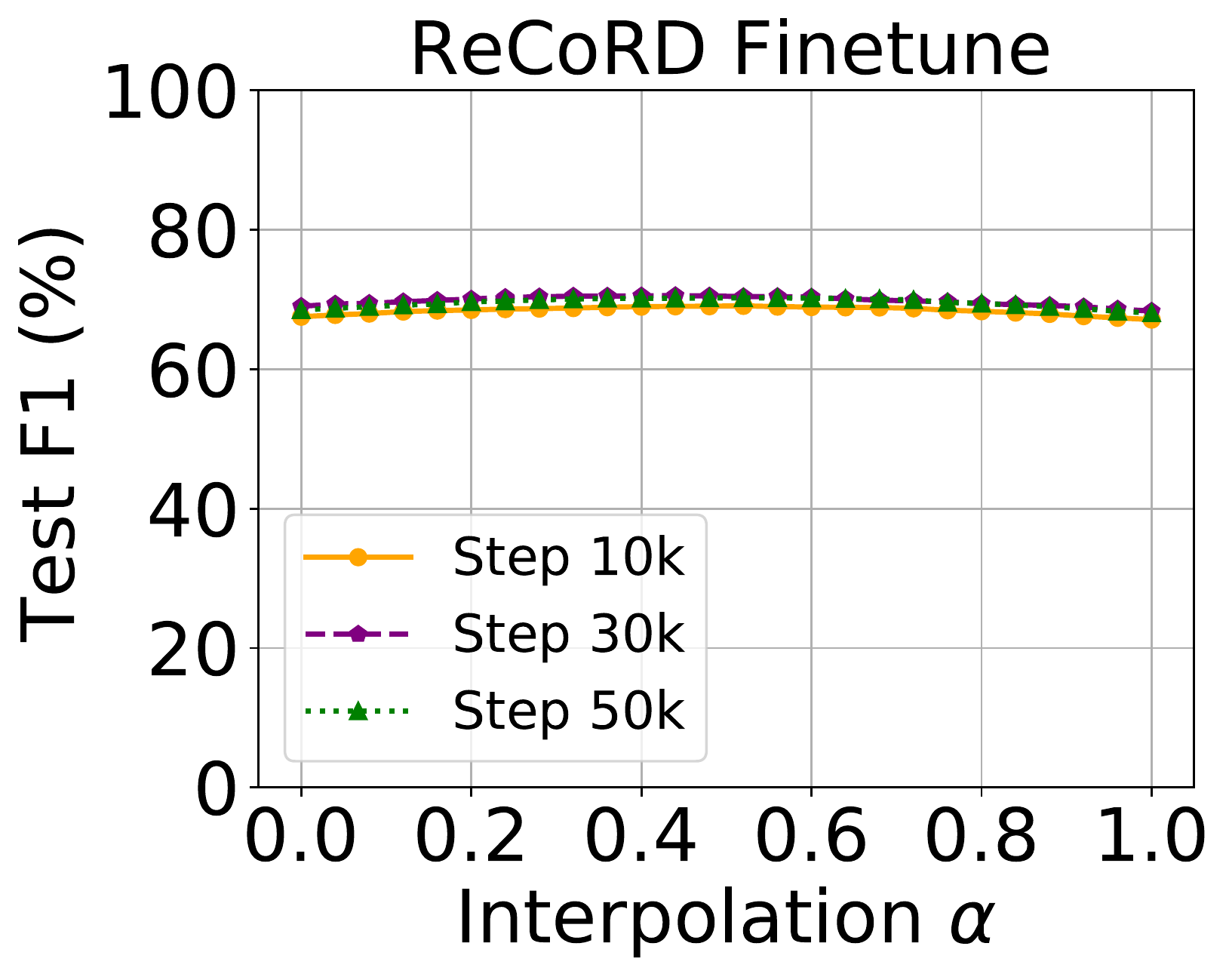}} 
    \subfigure{\includegraphics[width=0.24\textwidth]{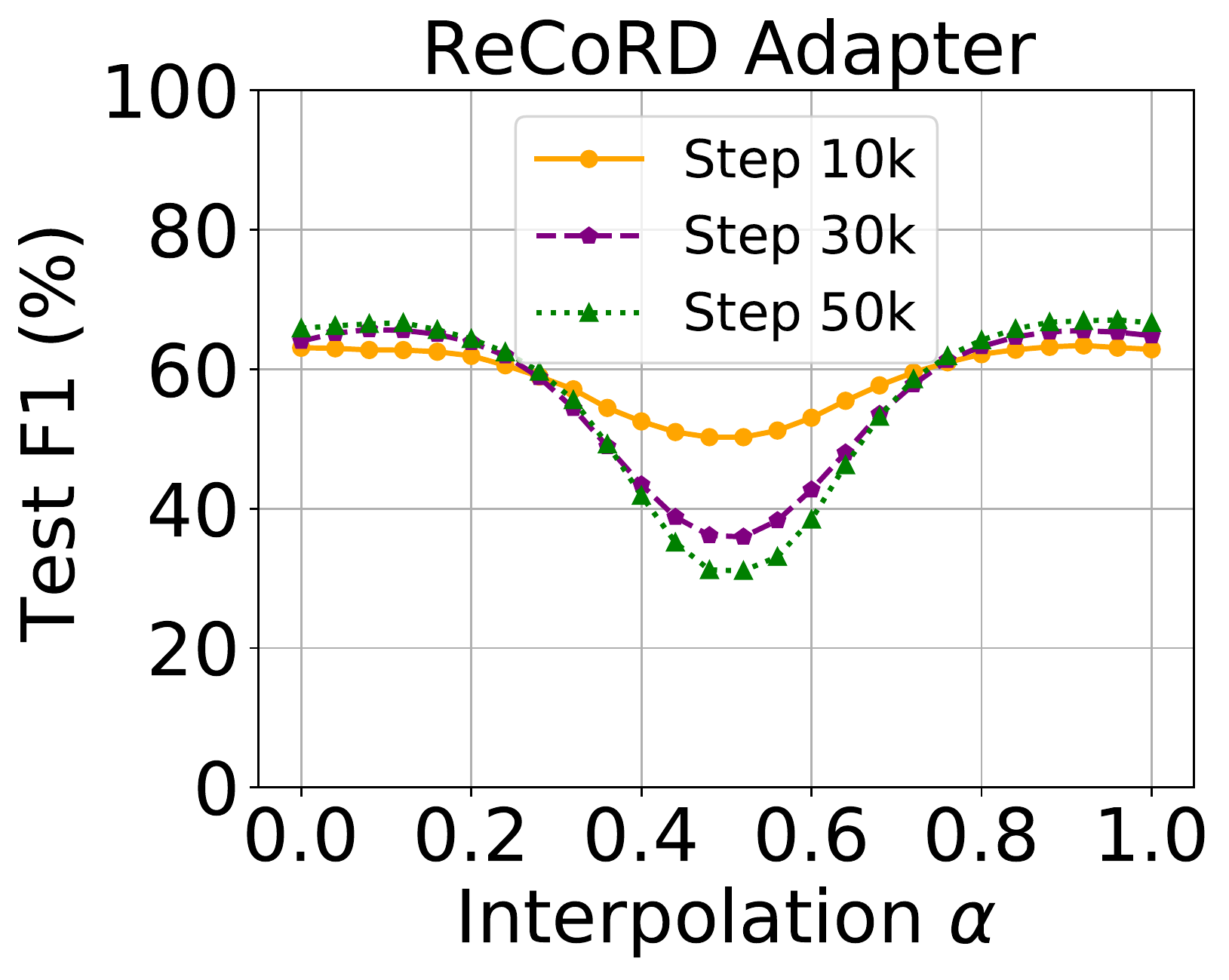}} 
    \caption{The performance of linear interpolations between two minima trained with different initialization.}
    \label{fig:initialization}
\end{figure*}

In each experiment, the training configurations of the two endpoints only differ in one hyperparameter, while other settings are kept the same for a fair comparison. To explore the effects of training steps, we evaluate the performance when both endpoints are trained for \{$10$k, $30$k, $50$k\} steps, respectively. We evaluate $24$ evenly distributed interpolations and $2$ endpoints along a linear path, i.e., we evaluate a series of $\phi(\alpha)$, where $\alpha \in \{\frac{0}{25}, \frac{1}{25}, ..., \frac{25}{25}\}$. Since we find that the trends of loss and performance are generally highly correlated (i.e., a performance drop corresponds to a loss barrier), we report the performance in the main paper and leave the results of loss in \cref{sec:exp_loss}. All experiments are conducted $3$ times with random seeds and we report the average results on test sets. For more training details, please refer to \cref{sec:training_detail}.

\paragraph{Effects of Training Data Order.}
PLM's downstream adaptation generally involves mini-batch gradient-based optimization, where training samples are learned in a random order. To explore its effect, we adapt two copies of a PLM with two different random data order. Then we visualize the performance of linear interpolations in Figure~\ref{fig:order}, from which we observe that for fine-tuning, both endpoints are well connected by a linear path; while for adapter tuning, there exists a slight but negligible performance drop near the midpoint. In general, we conclude that \textbf{local minima are well connected under different random training data order}.

\paragraph{Effects of Initialization.} Before downstream adaptation, additional parameters (e.g., extra modules defined by delta tuning, the classification head, etc.) may be introduced; in addition, \citet{wu-etal-2022-noisytune} recently show that adding noise to the pre-trained weights improves the fine-tuning performance on downstream tasks. Thus, both fine-tuning and delta tuning require proper initialization for the tunable parameters. Since different initialization could lead to distinct optimization trajectories, we explore the effects of initialization on PLM's mode connectivity.

Specifically, for those newly introduced modules, we randomly initialize them with a Gaussian distribution; for those pre-trained weights that require tuning, we add a random Gaussian noise. Two endpoints are initialized with the same configuration (e.g., mean and standard deviation of the Gaussian distribution), but different random seeds. The linear interpolation results are depicted in Figure~\ref{fig:initialization}, from which we observe that the mode connectivity of fine-tuning is generally good; while for adapter tuning, there exists a significant performance drop between two differently initialized minima. This means starting from different initialization, PLMs tend to reach non-connected local minima in the parameter space, especially for delta tuning. In short, \textbf{initialization of tunable parameters has a great impact on mode connectivity}.

\paragraph{Effects of Training Step.}
As mentioned before, the experiments in Figure~\ref{fig:order} and \ref{fig:initialization} are conducted when both minima are trained for \{$10$k, $30$k, $50$k\} steps. Comparing the results at different training steps, we observe that (1) longer training leads to poorer connectivity for adapter tuning under certain cases; while (2) the mode connectivity of fine-tuning is good at different steps. In \cref{sec:exp_training_step}, we further show that (1) the mode connectivity becomes poorer when one endpoint is trained with more steps while the other is trained with fixed steps, and (2) with the training step increasing, the Euclidean distance between two minima is also prolonged, which may partially explain the poorer mode connectivity.

\paragraph{Effects of Tuning Method.}
Comparing the results of fine-tuning and adapter tuning in Figure~\ref{fig:order} and \ref{fig:initialization}, we observe that in general, the linear mode connectivity of fine-tuning is better than adapter tuning. In other words, when using fine-tuning, PLMs are more likely to be optimized to linearly-connected minima. A similar phenomenon also occurs for minima trained with different learning rates or batch sizes (see \cref{sec:lr_bs}). Considering that adapter optimizes only $2.38$\% parameters than fine-tuning, we hypothesize that more tunable parameters may yield better mode connectivity and leave more explorations as future work.

\begin{figure}[!t]
    \centering
    \subfigure{\includegraphics[width=0.235\textwidth]{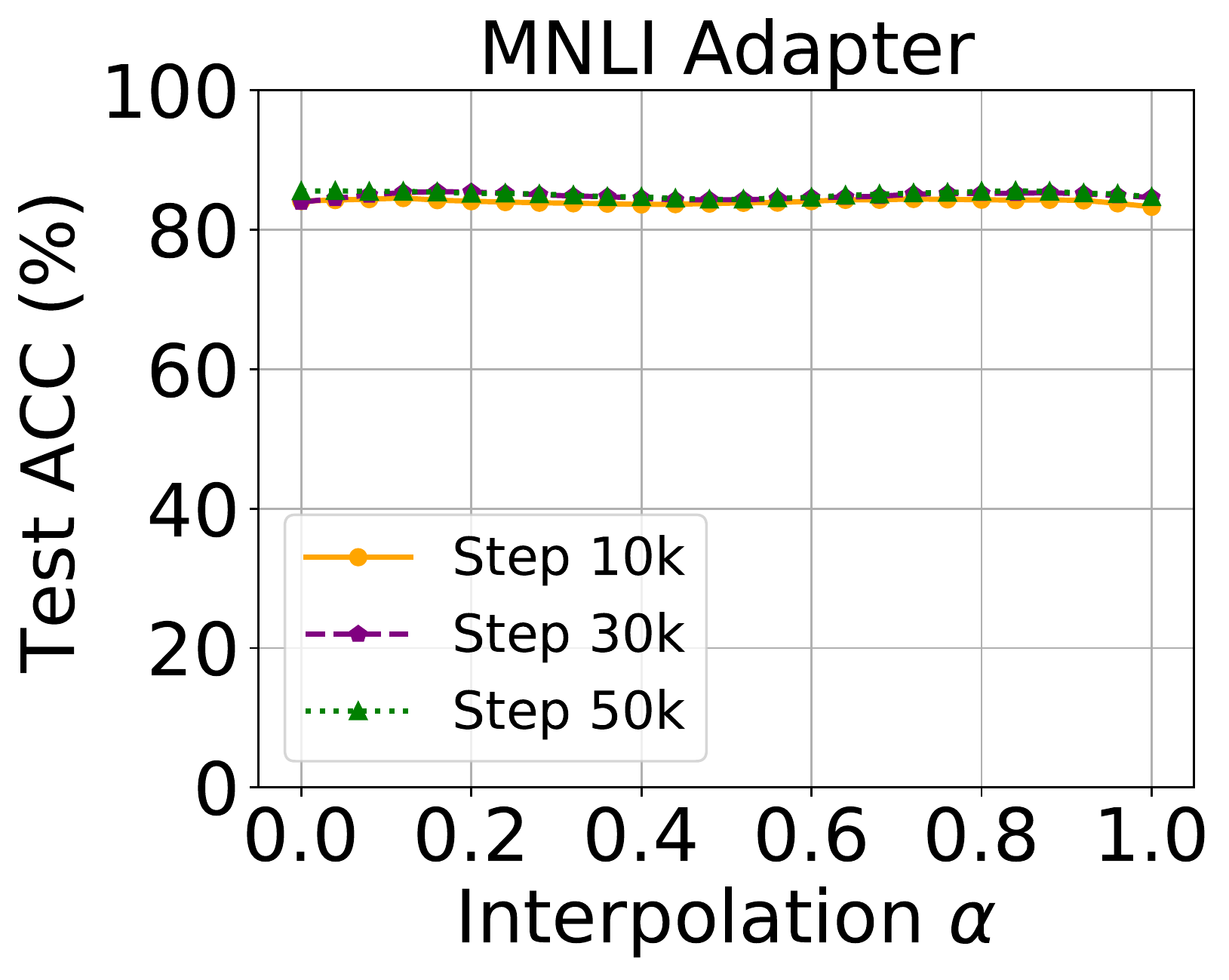}} 
    \subfigure{\includegraphics[width=0.235\textwidth]{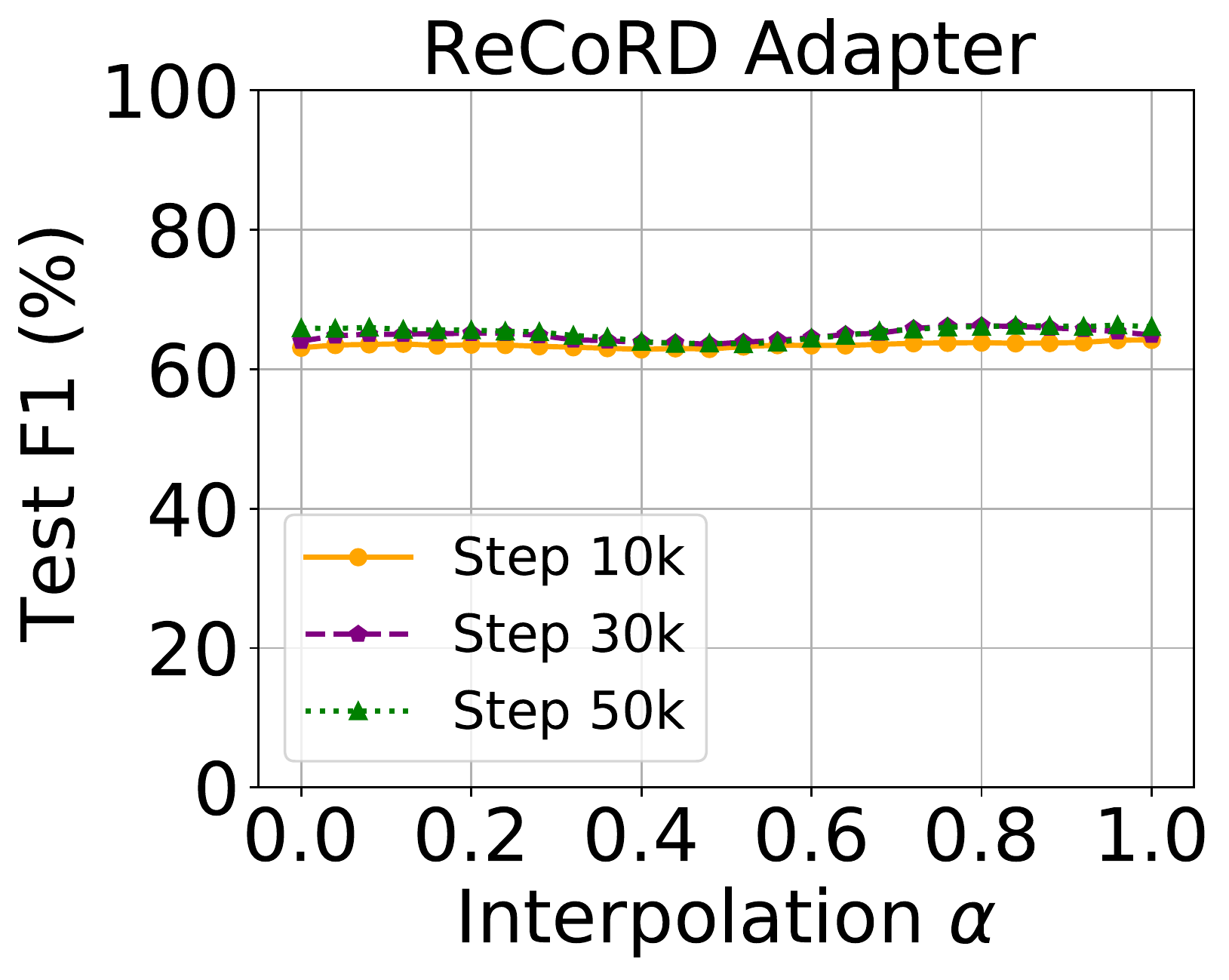}} 
    \caption{The performance of interpolations along a non-linear path connecting two minima, which are trained with adapter tuning from different initialization.}
    \label{fig:curved_main_paper}
\end{figure}

\paragraph{Different Minima are Generally Connected by a Non-linear Path.}
Considering that linearity is a strong constraint for mode connectivity, even if a direct linear path connecting two minima incurs a high loss, both minima may still be connected by a low-loss non-linear path. To explore whether this holds for PLMs, we follow the setting of tuning adapters with different initialization, which has been shown in Figure~\ref{fig:initialization} to have poor linear mode connectivity. We try to use the supervision from the downstream task to find a low-loss parametric path connecting two endpoints $\theta_{C_1}$ and $\theta_{C_2}$. Following \citet{garipov2018loss}, we consider a quadratic Bezier curve defined as follows:
\begin{equation}
\label{eq:curve}
\small
\begin{aligned}
    \phi_\theta (\alpha) = (1-\alpha)^2 \!\cdot\! \theta_{C_1} + 2\alpha(1-\alpha)\theta + \alpha^2 \!\cdot\! \theta_{C_2},
\end{aligned}
\end{equation}
where $\theta \!\in\! \mathbb{R}^{|\theta_0|}$ denotes tunable parameters of the curve. During curve finding, $\theta_{C_1}$ and $\theta_{C_2}$ are kept frozen, and only $\theta$ is optimized. Denote $\mathcal{L}$ as the loss function of the task, the training objective is to minimize the expectation of loss on the curve over a uniform distribution $\text{U}(0,1)$, i.e., $\mathbb{E}_{\alpha \in \text{U}(0,1)} \mathcal{L}(\phi_\theta(\alpha))$. For more details, please refer to \cref{sec:bezier_detail}.

We visualize the performance of the interpolation on the found Bezier curve in Figure~\ref{fig:curved_main_paper}. We observe that the two minima are well-connected by the found Bezier curve, without a significant performance drop. In fact, such a low-loss Bezier curve exists for minima reached under various different settings (see more experiments in \cref{sec:exp_curved}). Given the above results, we conjecture that \textbf{there may exist multiple loss basins which are connected via a low-loss non-linear path, instead of a linear path. For most of the minima within the same loss basin, their convex combination also lies in this basin}. In this sense, if two minima are connected linearly, then both of them probably lie in the same basin; otherwise in different basins (e.g., the case of adapter tuning with different initialization).

\finding{Q1. (b) What are the effects of training data?}

In previous experiments, we focus on the connectivity of two minima trained with the same dataset. From now on, we extend the mode connectivity to two minima trained on different datasets, focusing on two facets: data overlap and data domain.

\begin{figure}[!t]
    \centering
    \subfigure{\includegraphics[width=0.235\textwidth]{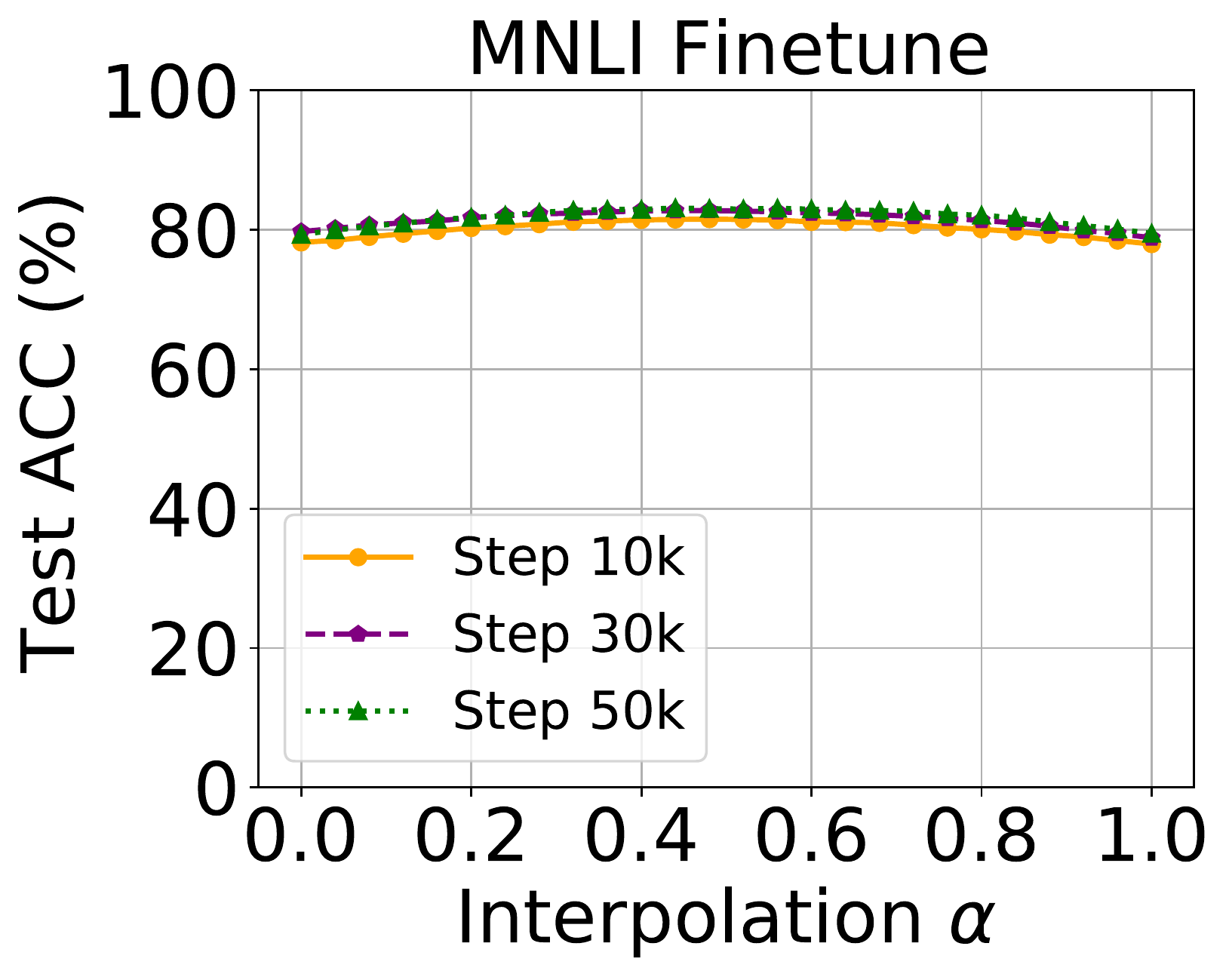}} 
    \subfigure{\includegraphics[width=0.235\textwidth]{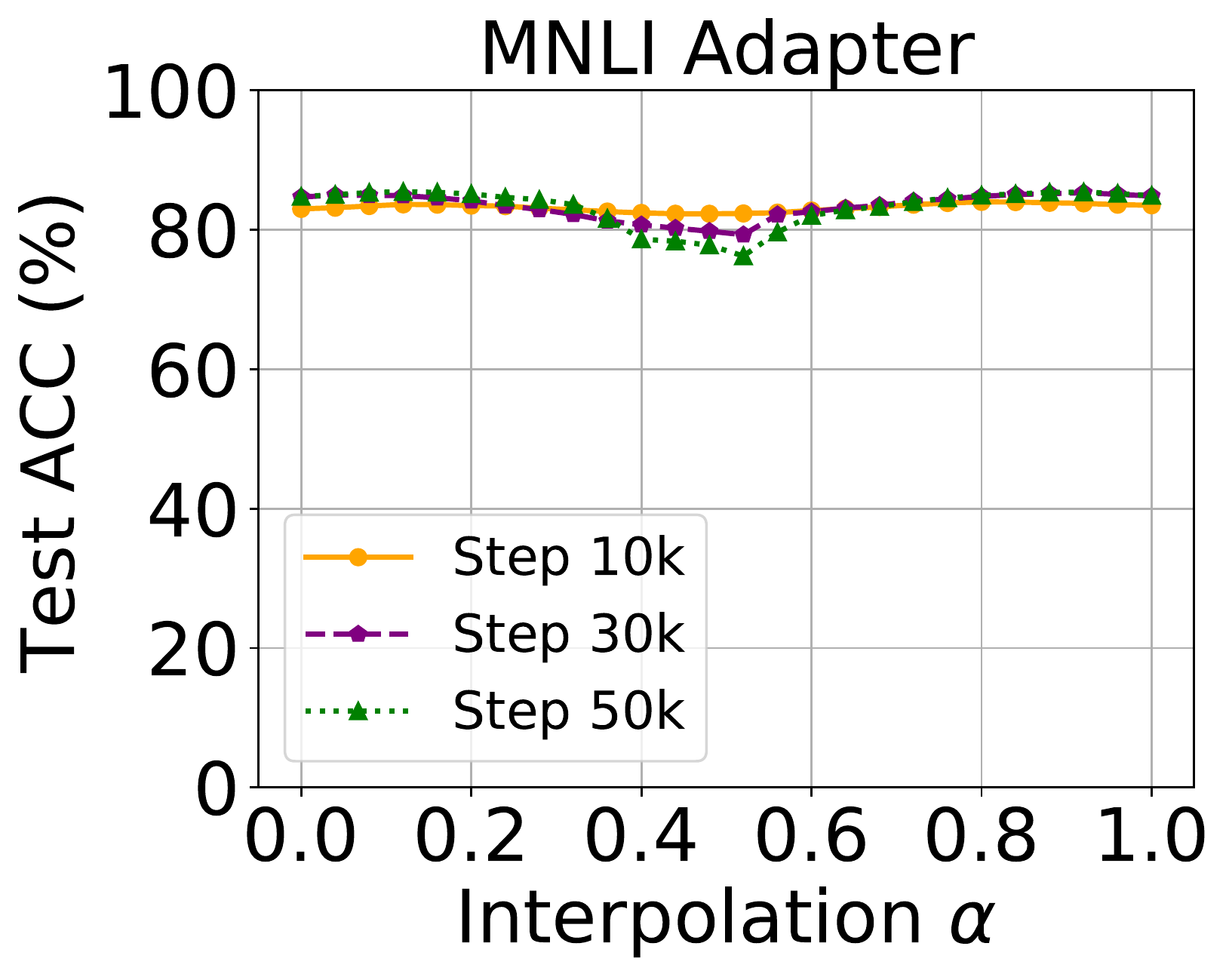}}
    \caption{Linear mode connectivity analysis for two minima trained with in-distribution MNLI data. The results on ReCoRD are left in \cref{sec:additional_overlap}.}
    \label{fig:overlap}
\end{figure}

\paragraph{Effects of Data Overlap.}
PLMs have been demonstrated to be adept at memorizing the training data~\citep{carlini2021extracting,carlini2022quantifying}. To show that the connectivity of both minima does not originate from PLM's memorization, we explore whether such mode connectivity still exists when two minima are obtained on data belonging to the same distribution, but without overlap of specific training samples. Specifically, we partition the original training data of MNLI into two equal splits. Then we adapt two copies of $\text{T5}_\texttt{BASE}$ on either split using the same training configurations. The experiments are conducted using both fine-tuning and adapter tuning.

The performance of linear interpolations is recorded in Figure~\ref{fig:overlap}. The results show that two minima are well connected for both tuning methods, demonstrating that \textbf{mode connectivity does not originate from PLM's memorization of specific training data}; instead, during training, PLMs learn advanced task-level knowledge, and the \textbf{connectivity reflects the high overlap of task knowledge of two local minima}.

\begin{figure}[!t]
    \centering
    \subfigure{\includegraphics[width=0.235\textwidth]{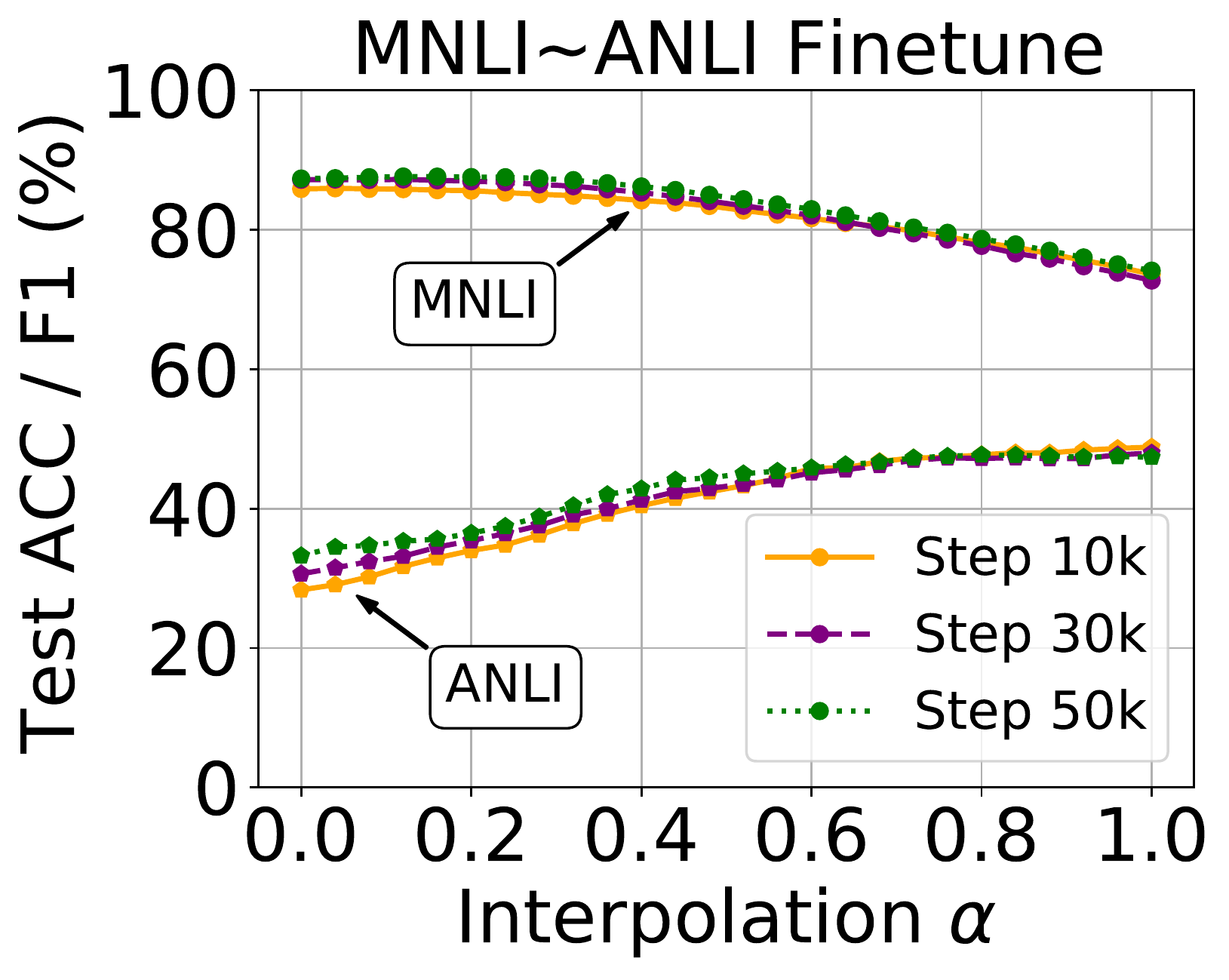}} 
    \subfigure{\includegraphics[width=0.235\textwidth]{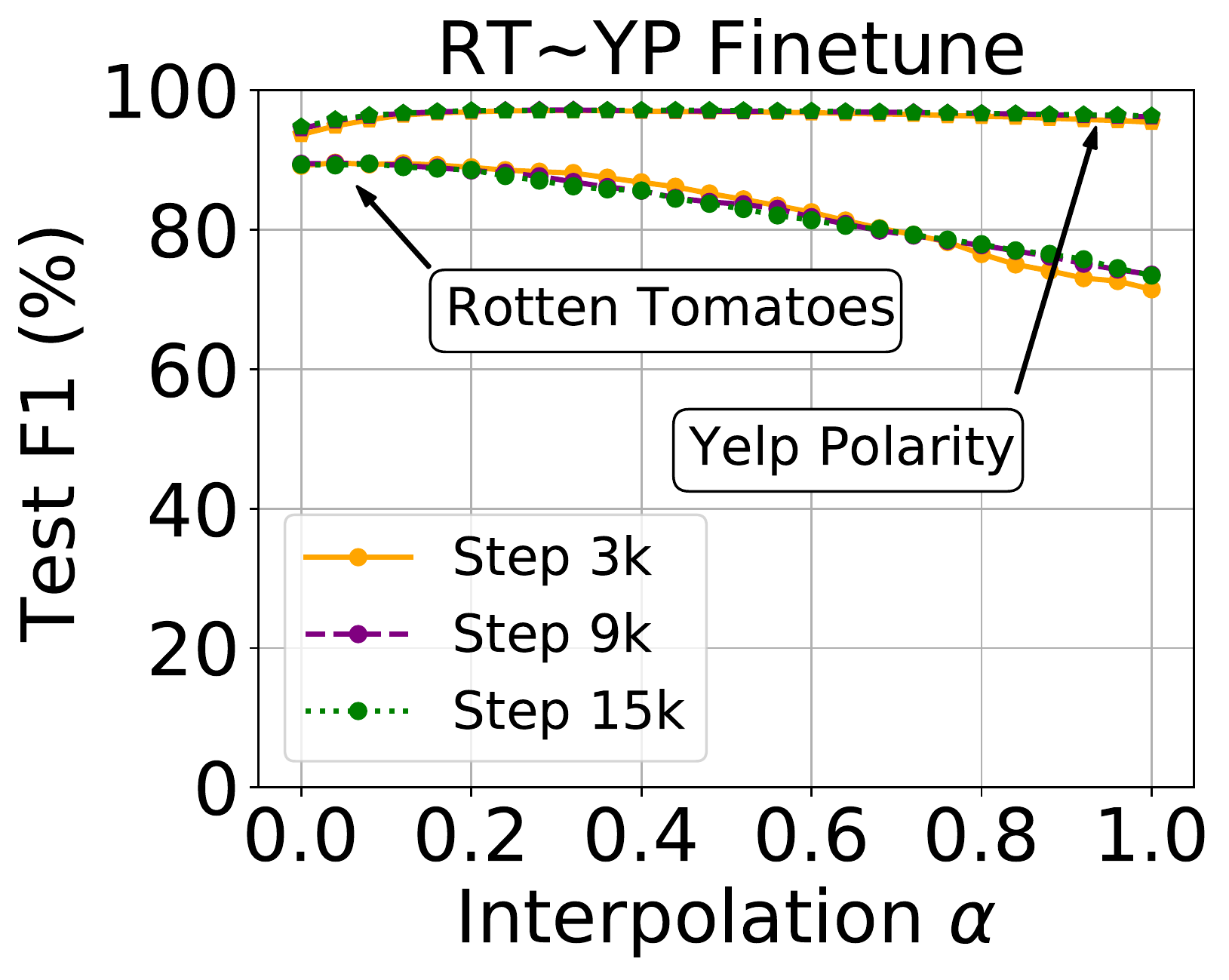}}
    \caption{Linear mode connectivity for two minima fine-tuned on different data distributions of the same task. Left: $\alpha = 0$ / $\alpha = 1$ denotes the minimum of MNLI / ANLI. Right: $\alpha = 0$ / $\alpha = 1$ denotes the minimum of Rotten Tomatoes / Yelp Polarity.}
    \label{fig:domain}
\end{figure}

\paragraph{Effects of Data Domain.}
PLMs are shown to generalize well on out-of-distribution data \citep{hendrycks-etal-2020-pretrained}, implying the connection of minima trained with different data distributions. To gain a deeper understanding, we choose two natural language inference datasets (MNLI and ANLI~\citep{nie-etal-2020-adversarial}), and two sentiment analysis datasets (Rotten Tomatoes~\citep{pang-lee-2005-seeing} and Yelp Polarity~\citep{NIPS2015_250cf8b5}) sourced from different domains. Then we fine-tune two copies of $\text{T5}_\texttt{BASE}$ on two datasets of the same task, and evaluate the linear mode connectivity between two minima. Note previous works typically study mode connectivity on the same dataset; while in our setting, we extend the analysis by evaluating the interpolations on two datasets.

The results are shown in Figure~\ref{fig:domain}. Take the NLI task as an example, starting from one endpoint ($\alpha \!=\! 0$) of a source task (MNLI), with the interpolation approaching the other endpoint ($\alpha \!=\! 1$) of the target task (ANLI), the performance on MNLI / ANLI exhibits almost a monotonous drop / rise. Besides, there does not exist a performance valley where the performance is significantly lower than both endpoints. Intuitively, \textbf{the performance change reflects the variation of the interpolation's task knowledge along the connecting path}. Due to the difference in data domain, the task knowledge of two endpoints only partially overlap with each other. When traversing from the source minimum to the target minimum, PLM suffers from catastrophic forgetting on the source knowledge, but gradually absorbs target knowledge, leading to the performance drop / rise on the source / target task. We defer more in-depth analyses to Q3.

\finding{Q2. How does mode connectivity change during pre-training?}
Previous works demonstrate that compared with random initialization, the initialization obtained by pre-training leads to a wider loss basin after downstream adaptation~\citep{hao-etal-2019-visualizing,neyshabur2020being}. Intuitively, if a local minimum lies in a more flat basin, it should be easier to connect with other minima. In this sense, pre-training may be closely related to mode connectivity. To investigate this, we re-train a $\text{RoBERTa}_{\texttt{BASE}}$ model from scratch and explore how mode connectivity changes at different pre-training steps. We follow the pre-training setting of \citet{liu2019roberta}, with more details described in \cref{sec:pre-training_roberta}.

\paragraph{Pre-training Facilitates Mode Connectivity.}

\begin{figure}[!t]
    \centering
    \subfigure{\includegraphics[width=0.235\textwidth]{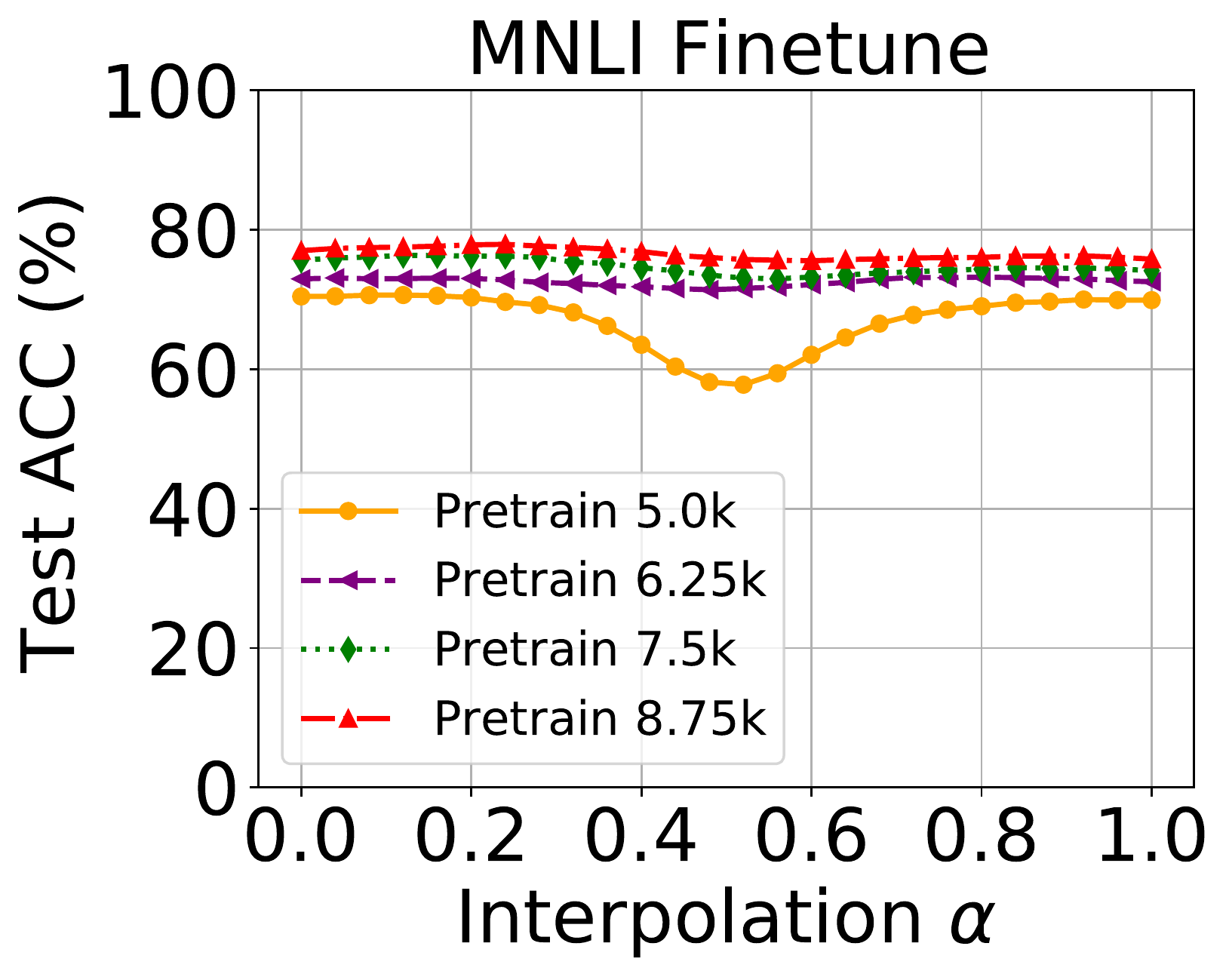}}
    \subfigure{\includegraphics[width=0.235\textwidth]{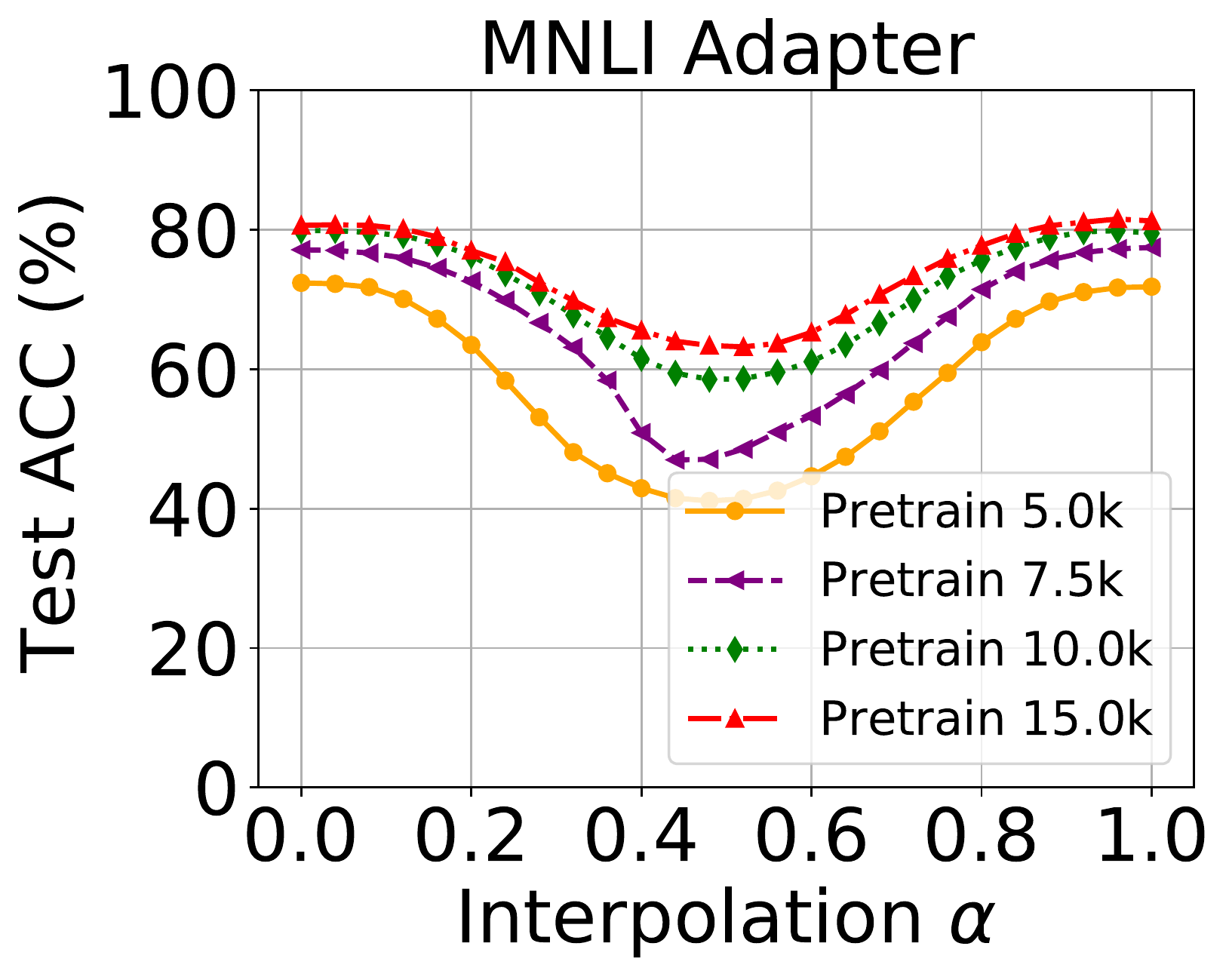}}
    \caption{The change of linear mode connectivity at different pre-training steps. We illustrate the performance of linear interpolations of two minima trained on MNLI using different initialization.}
    \label{fig:pretrain_same_task}
\end{figure}

We select a series of checkpoints at different pre-training steps. For each checkpoint, we adapt two copies on MNLI using different initialization, and evaluate the performance of their linear interpolations. From Figure~\ref{fig:pretrain_same_task}, we observe that for both fine-tuning and adapter tuning, with the pre-training step becoming larger, the mode connectivity of the PLM becomes better. This implies that pre-training implicitly facilitates the mode connectivity. Specifically, when using fine-tuning, there does not exist a performance drop for checkpoints pre-trained with more than $6.25$k steps. Considering that pre-training with a batch size of $2048$ for $6.25$k steps corresponds to almost $5$\% the computational cost of BERT (a batch size of $256$ for $1000$k steps), we conclude that PLMs acquire good mode connectivity at an early stage of pre-training.

\begin{figure}[!t]
    \centering
    \subfigure{\includegraphics[width=0.235\textwidth]{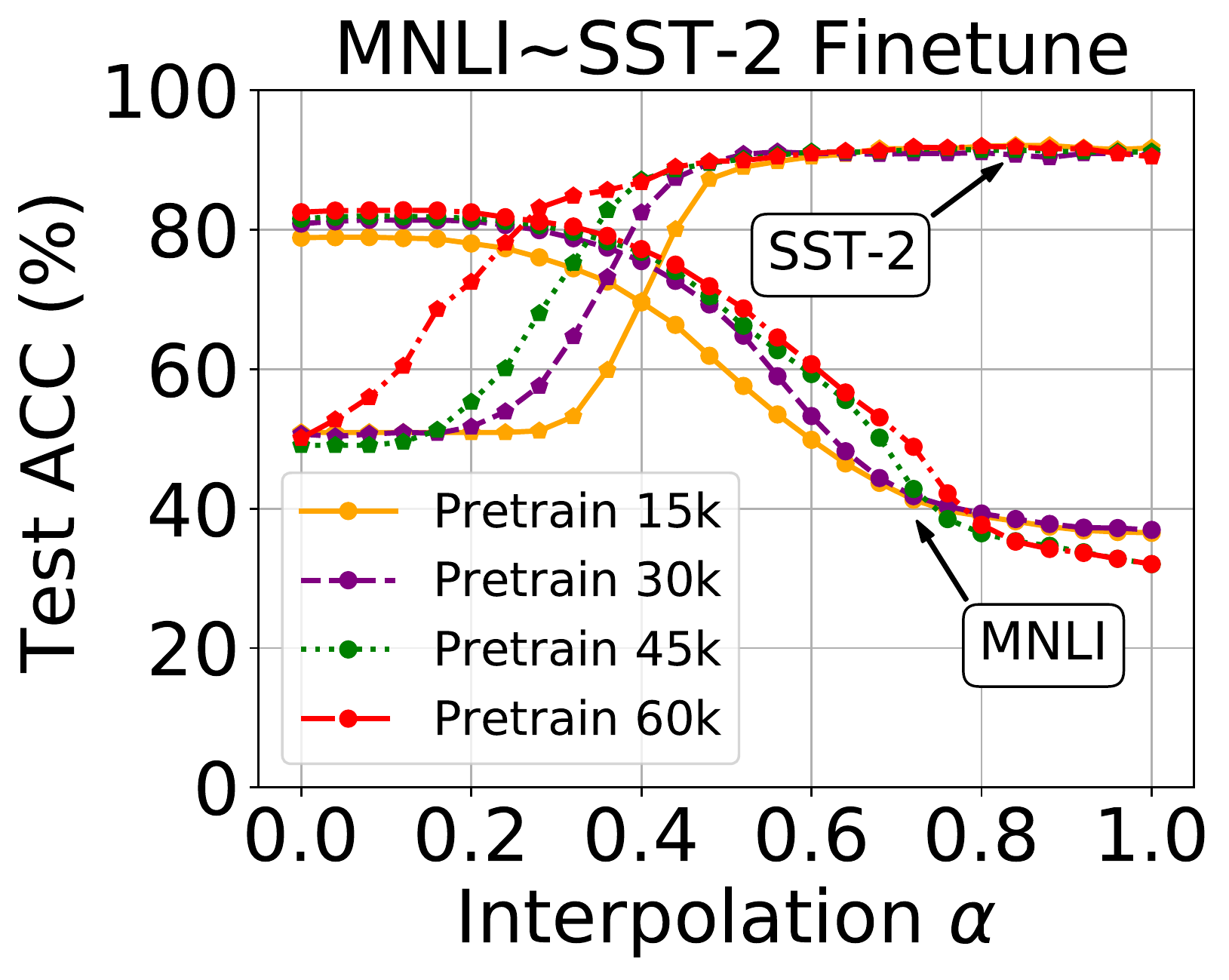}}
    \subfigure{\includegraphics[width=0.235\textwidth]{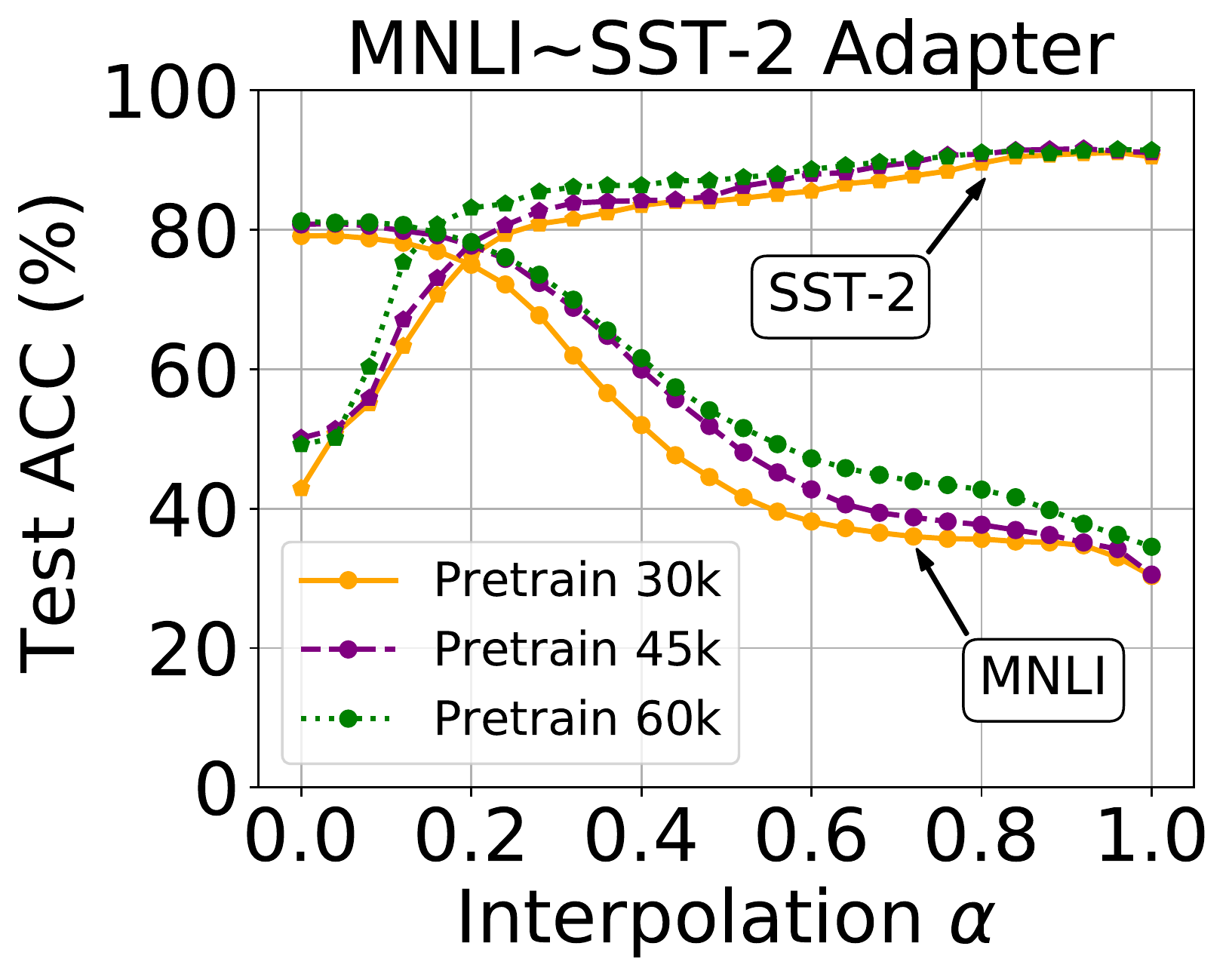}}
    \caption{The change of linear mode connectivity at different pre-training steps. We illustrate the performance of linear interpolations of two minima trained on MNLI and SST-2. $\alpha = 0$ / $\alpha = 1$ denotes the minimum of MNLI / SST-2.}
    \label{fig:pretrain_diff_task}
\end{figure}

\paragraph{Pre-training Pulls Task Boundary Closer.}
Further, we look into the performance variation along a linear path between two minima trained on two different tasks (MNLI and SST-2~\citep{socher-etal-2013-recursive}). Similarly, we choose a series of checkpoints at different pre-training steps. Then for each checkpoint, we adapt two copies on MNLI and SST-2 under the same setting, and conduct linear interpolation between two adapted weights. We also conduct experiments on MNLI and QQP~\href{https://quoradata.quora.com/First-Quora-Dataset-Release-Question-Pairs}{(link)} in \cref{sec:additional_diff_task}. We evaluate the performance of each interpolation on both tasks and illustrate the results in Figure~\ref{fig:pretrain_diff_task}. It can be derived that (1) due to the inherent difference of both tasks, the minimum obtained on one task achieves the performance near random guess ($\approx 50\%$ for SST-2 and $\approx 33.3\%$ for MNLI) on another task. This indicates that minima of different tasks are disconnected. (2) In addition, there is a strong general trend that for a checkpoint pre-trained longer, the intersection of both tasks' high-performance regions becomes wider. In other words, the boundaries of both tasks' optimal regions are gradually pulled closer by pre-training. (3) For the checkpoint pre-trained with $60$k steps, we do not observe a region where the performance on both tasks is poor. This means starting from the initialization of a PLM pre-trained with enough steps, the optimal regions of various downstream tasks are closely packed. This finding may help explain PLM's cross-task transferability, and we leave more discussions in \cref{sec:discussion}.

\finding{Q3. How does the task knowledge change along the path connecting two minima?}

\begin{figure*}[!t]
    \centering
    \subfigure{\includegraphics[width=0.24\textwidth]{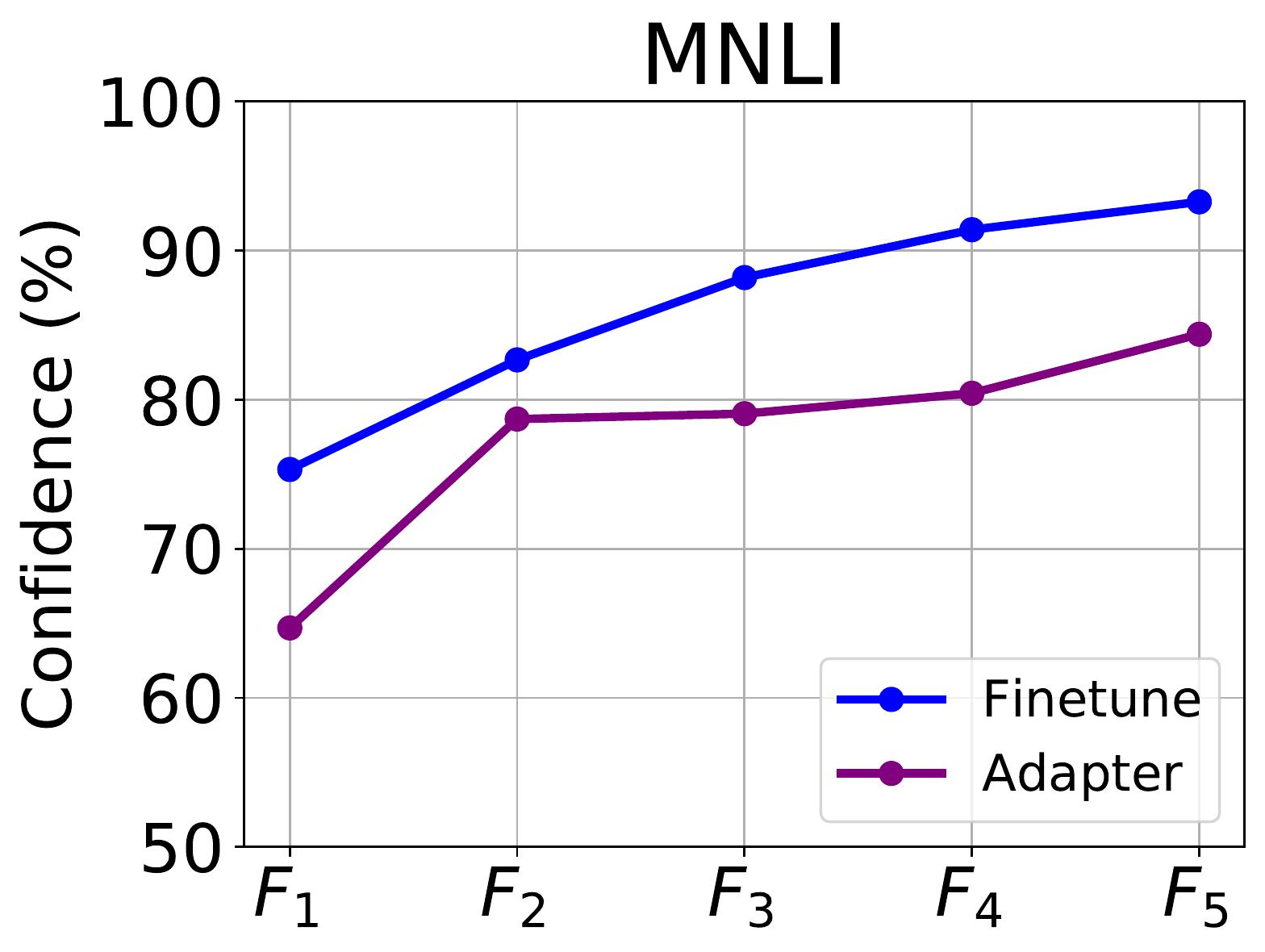}} 
    \subfigure{\includegraphics[width=0.24\textwidth]{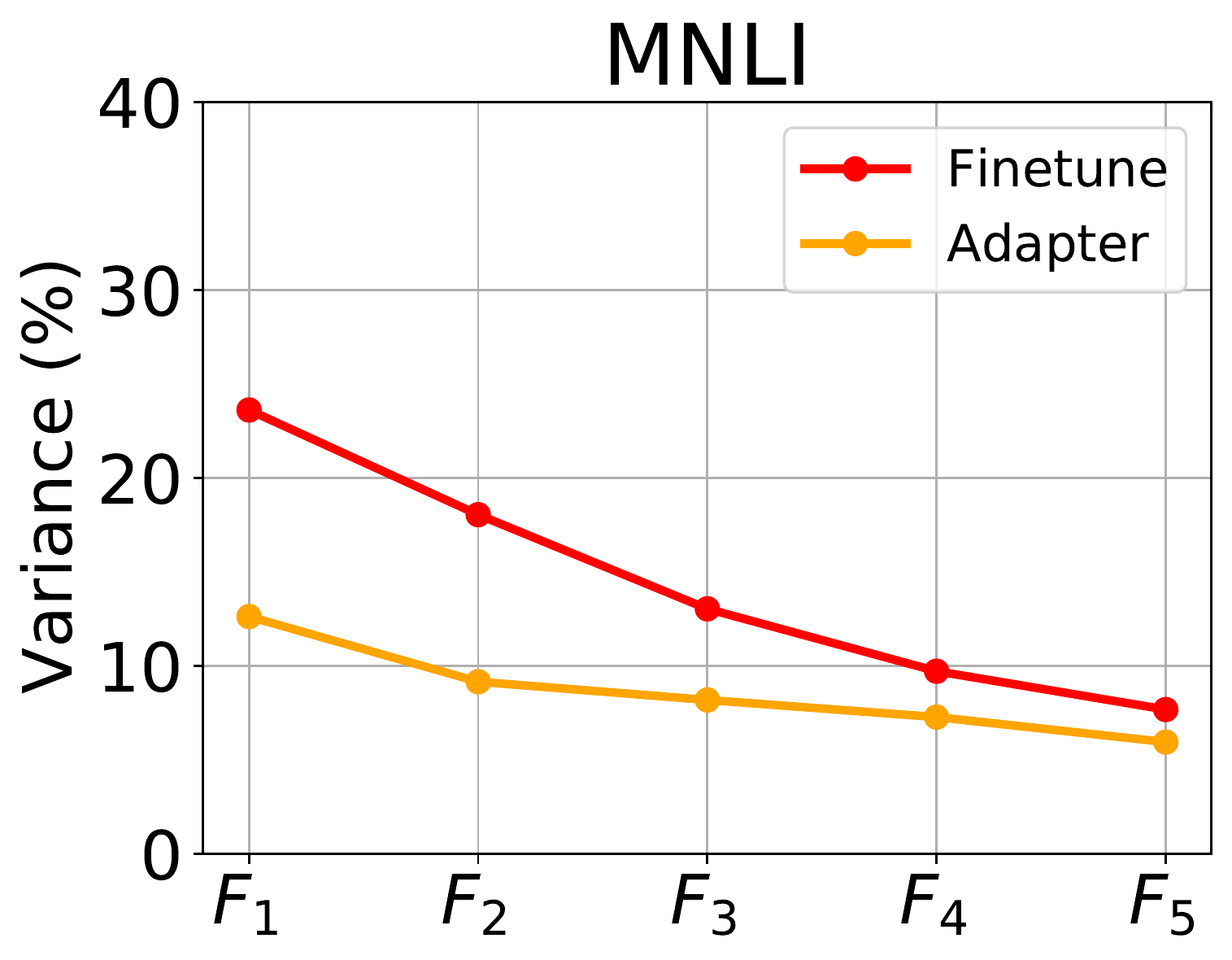}}
    \subfigure{\includegraphics[width=0.24\textwidth]{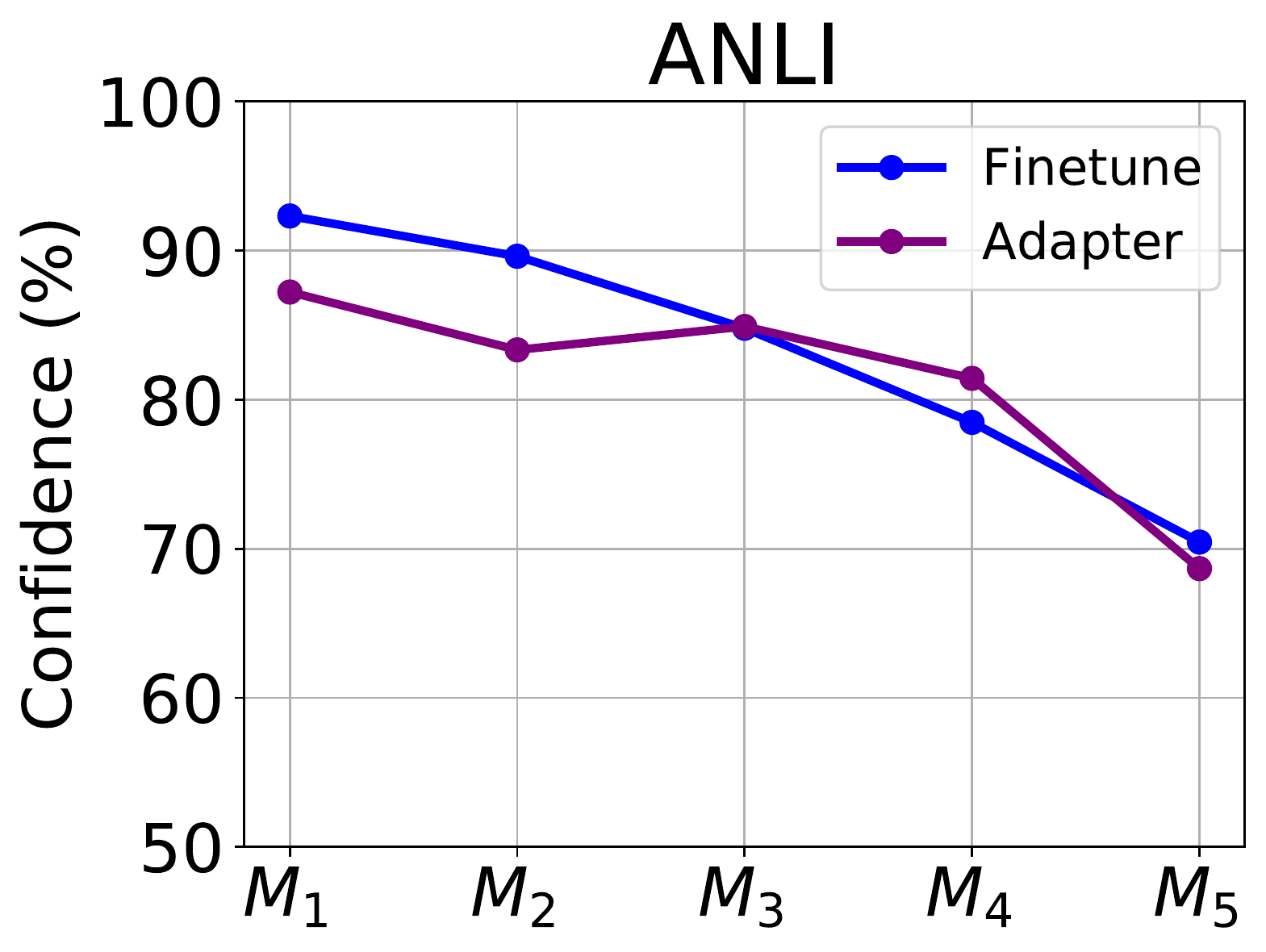}}
    \subfigure{\includegraphics[width=0.24\textwidth]{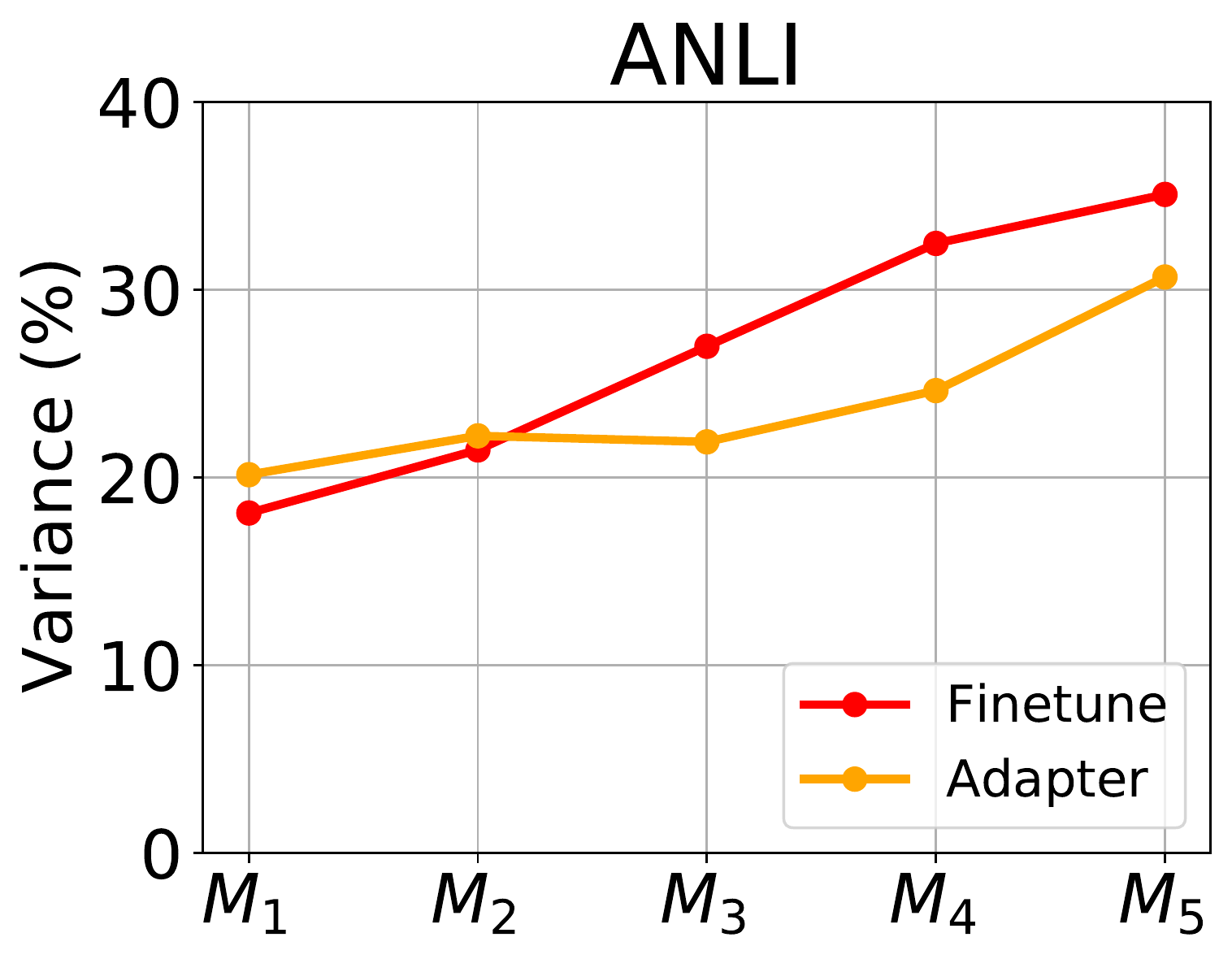}} 
    \caption{The results of knowledge variation for linear interpolations ($\phi_1$, $\phi_2$, $\phi_3$, $\phi_4$) between two minima ($\phi_0$, $\phi_5$) adapted on MNLI and ANLI. We leave experiments on other tasks as future work.}
    \label{fig:forgetting}
\end{figure*}

Having shown that mode connectivity reflects the high overlap of task knowledge of different minima, we further investigate the knowledge variation along the path connecting two minima. To quantify a model's task knowledge, we resort to the \textit{memorization} of the training data as a rough estimate. In experiments, we evaluate two minima obtained on data of different distributions as mentioned in Q1. (b)\footnote{We choose this setting because (1) there does not exist a performance valley between two minima, which means the knowledge is properly combined, and (2) the knowledge of both tasks is diverse enough.}.

Specifically, we adapt two copies of $\text{T5}_\texttt{BASE}$ on MNLI (source task) and ANLI (target task), respectively. Denote $\theta_{s}$ and $\theta_{t}$ as two minima trained on the source dataset $\mathcal{D}_s = \{x_i, y_i\}_{i=1}^{|\mathcal{D}_s|}$ and the target dataset $\mathcal{D}_t = \{x_i, y_i\}_{i=1}^{|\mathcal{D}_t|}$. We investigate the knowledge variation from $\theta_s$ to $\theta_t$ by choosing $4$ evenly distributed linear interpolations ($\phi_1$, $\phi_2$, $\phi_3$, $\phi_4$) and $2$ endpoints ($\phi_0$, $\phi_5$), i.e., $\phi_j = \theta_s + \frac{j}{5} \cdot (\theta_t - \theta_s)$, $j \!\in\! \{0,1,...,5\}$, where $\phi_0 \!=\! \theta_s$, $\phi_5 \!=\! \theta_t$. Then we measure whether each source training sample $x_i \!\in\! \mathcal{D}_s$ is memorized (correctly classified) by $\phi_j$. We find empirically that with $\phi_j$ approaching $\theta_t$, training samples of $\mathcal{D}_s$ are gradually forgotten, but seldom re-memorized under this setting. Therefore, we only record those newly forgotten samples for $\phi_j$ (i.e., those classified correctly by $\phi_{j-1}$ but wrongly by $\phi_j$) and denote them as $\mathcal{F}_{j}$. Similarly, we denote those newly memorized samples of $\mathcal{D}_t$ as $\mathcal{M}_{j}$ (i.e., those classified wrongly by $\phi_{j-1}$ but correctly by $\phi_j$).

After that, we characterize the role of each sample using \textit{dataset cartography}~\citep{swayamdipta-etal-2020-dataset}. For a brief introduction, each sample of $\mathcal{D}_s$ ($\mathcal{D}_t$) is characterized by the training dynamics of $\theta_s$ ($\theta_t$). Take $\mathcal{D}_s$ as an example, assume we train the PLM for $E$ epochs on $\mathcal{D}_s$, and the weights of the PLM are adapted to $\theta_s (e)$ after the $e$-th epoch, where $1\!\leq \! e \! \leq \! E$. For each training instance $(x_i, y_i) \in \mathcal{D}_s$, denote $\mathcal{P}_{\theta_s (e)}(y_i|x_i)$ as the probability $\theta_s (e)$ assigns to the true label, we record the PLM's prediction after each epoch and calculate two statistics:

\noindent $\bullet$ \texttt{confidence} measures how confidently the PLM assigns the true label to a given input, it is defined as the mean probability of the true label:
\begin{equation*}
\small
\begin{aligned}
    \mu_i = \frac{1}{E}\sum_{e=1}^E\mathcal{P}_{\theta_s (e)}(y_i|x_i).
\end{aligned}
\end{equation*}

\noindent $\bullet$ \texttt{variability} captures how consistently the PLM judges each training instance, it is defined using the standard deviation of the true label's probability:
\begin{equation*}
\small
\begin{aligned}
    \sigma_i = \sqrt{\frac{\sum_{e=1}^E(\mathcal{P}_{\theta_s (e)}(y_i|x_i) - \mu_i)^2}{E}}.
\end{aligned}
\end{equation*}

After obtaining both statistics for each training sample of $\mathcal{D}_s$ and $\mathcal{D}_t$, we illustrate the average statistics for newly forgotten / memorized samples ($\mathcal{F}_j$ / $\mathcal{M}_j$) in Figure~\ref{fig:forgetting}. We observe that with $\phi_j$ approaching $\theta_t$, the average confidence of the newly forgotten data gradually increases, while the variability gradually drops; symmetrically, the average statistics of the newly learned data exhibit an opposite trend. According to \citet{swayamdipta-etal-2020-dataset}, instances with high confidence but low variability are generally easy-to-learn ones; while those with low confidence are generally ambiguous or hard-to-learn data. In this regard, when gradually leaving the source minimum, \textbf{the PLM prioritizes forgetting the source knowledge of those difficult instances, and then forgets the source knowledge of the easy-to-learn data}. On the contrary, \textbf{the easy-to-learn target knowledge is learned before the elusive and obscure target knowledge}.

\section{Discussion}
\label{sec:discussion}

\paragraph{Weight Averaging.}
The property of linear mode connectivity is related to recent explorations of weight averaging~\citep{wortsman2021robust,wortsman2022model,matena2021merging}, which combines independently fine-tuned models in the parameter space. In this way, the knowledge from multiple models can be merged. Our findings have direct implications for designing better weight averaging methods: (1) for two minima, weight averaging can be seen as choosing the midpoint on the linear path. We have shown that a non-linear curve may have better mode connectivity under certain cases. This implicates that linear interpolation may not find the optimal combination despite its simplicity; instead, there may exist better methods to ensemble weights (see experiments in \cref{sec:weight_ensemble}); (2) our findings on the effects of different training configurations can also inspire choosing more appropriate models (with better mode connectivity) to ensemble.

\paragraph{Task-level Transferability.}
Although PLMs are demonstrated to have excellent cross-task transferability~\citep{vu-etal-2020-exploring,pruksachatkun-etal-2020-intermediate,poth-etal-2021-pre,su-etal-2022-transferability}, it is still under-explored why PLMs have such an ability. Our findings that pre-training implicitly pulls the task boundary closer may help explain this phenomenon. Since the optimal regions of various tasks are packed closely, PLMs are easier to traverse across the task boundary, without getting blocked by a loss barrier.
 
\paragraph{Knowledge Quantification.}
Investigating the knowledge variation along the connecting path helps better understand how different model knowledge is merged. Quantifying the task knowledge of various models may also provide insights for research topics like knowledge distillation~\citep{hinton2015distilling} and knowledge transfer~\citep{weiss2016survey}. While we use training data memorization as a rough estimate for task knowledge, it would be interesting to explore whether there exist more granular methods, such as knowledge probing~\citep{petroni-etal-2019-language,liu-etal-2019-linguistic}.

%% file: sections/5_conclusions.tex
\section{Conclusions}
In this paper, we conduct empirical analyses on the mode connectivity of PLMs, aiming to fathom the connection of minima reached under different settings. We investigate how different downstream adaptation configurations and pre-training affect PLM's mode connectivity. In addition, we explore the knowledge variation along the path connecting different minima. In general, exploring the mode connectivity of PLMs contributes to understanding the inner workings of PLM downstream adaptation. We expect our evaluation setup and analyses could inspire more future explorations in this field.

\section*{Acknowledgments}
This work is supported by the National Key R\&D Program of China (No. 2020AAA0106502) and Institute Guo Qiang at Tsinghua University.

Yujia Qin designed the experiments and wrote the paper. Cheng Qian and Jing Yi conducted the experiments. Yankai Lin, Zhiyuan Liu, Maosong Sun, and Jie Zhou advised the project. All authors participated in the discussion.

The authors would like to thank anonymous reviewers for their valuable feedback.

%% file: sections/appendix.tex
\clearpage
\appendix

\section*{Appendices}
\label{sec:appendix}

\section{Details for Finding a Bezier Curve}
\label{sec:bezier_detail}
We follow \citet{garipov2018loss} to find a quadratic Bezier curve connecting two endpoints $\theta_{C_1}$ and $\theta_{C_2}$. The Bezier curve is defined as follows:
\begin{equation*}
\small
\begin{aligned}
    \phi_\theta (\alpha) = (1-\alpha)^2 \!\cdot\! \theta_{C_1} + 2\alpha(1-\alpha)\theta + \alpha^2 \!\cdot\! \theta_{C_2}.
\end{aligned}
\end{equation*}
During curve finding, only $\theta \in \mathbb{R}^{|\theta_0|}$ is optimized to minimize the expectation of loss over a uniform distribution on the curve as follows:
\begin{equation*}
\small
\begin{aligned}
    \mathcal{L}_{\text{curve}} (\theta) \!=\! \frac{\int \mathcal{L}(\phi_\theta)d\phi_\theta}{\int d\phi_\theta} \!=\! \frac{\int_0^1 \mathcal{L}(\phi_\theta(\alpha))||\phi_\theta'(\alpha)||d\alpha}{\int_0^1 ||\phi_\theta'(\alpha)||d\alpha} \\
    =\! \int_0^1 \mathcal{L}(\phi_\theta(\alpha))q_\theta(\alpha)d\alpha \!=\! \mathbb{E}_{\alpha \sim q_\theta(\alpha)}\mathcal{L}(\phi_\theta(\alpha)),
\end{aligned}
\end{equation*}
where $q_\theta(\alpha) = \frac{||\phi_\theta'(\alpha)||d\alpha}{\int_0^1 ||\phi_\theta'(\alpha)||d\alpha}$. Since $q_\theta(\alpha)$ is dependent on $\theta$, it is generally intractable to compute the original loss $\mathcal{L}_{\text{curve}} (\theta)$. To this end, \citet{garipov2018loss} suggest optimizing a more computationally tractable loss as follows:
\begin{equation*}
\small
\begin{aligned}
    \mathcal{L}_{\text{curve}}'(\theta) = \mathbb{E}_{\alpha \in \text{U}(0,1)} \mathcal{L}(\phi_\theta(\alpha)),
\end{aligned}
\end{equation*}
where $\alpha$ is sampled from a uniform distribution $\text{U}(0,1)$ instead of $q_\theta(\alpha)$. In experiments, we initialize $\theta$ with $\frac{1}{2}\theta_{C_1} + \frac{1}{2}\theta_{C_2}$ (i.e., starting from a linear curve), which makes training more stable than using randomly initialized weights or the pre-trained weights.

\section{Additional Experiments}
\label{sec:additional_exp}

\subsection{Effects of the Learning Rate and the Batch Size}
\label{sec:lr_bs}

\begin{figure*}[!t]
    \centering
    \subfigure[]{\includegraphics[width=0.24\textwidth]{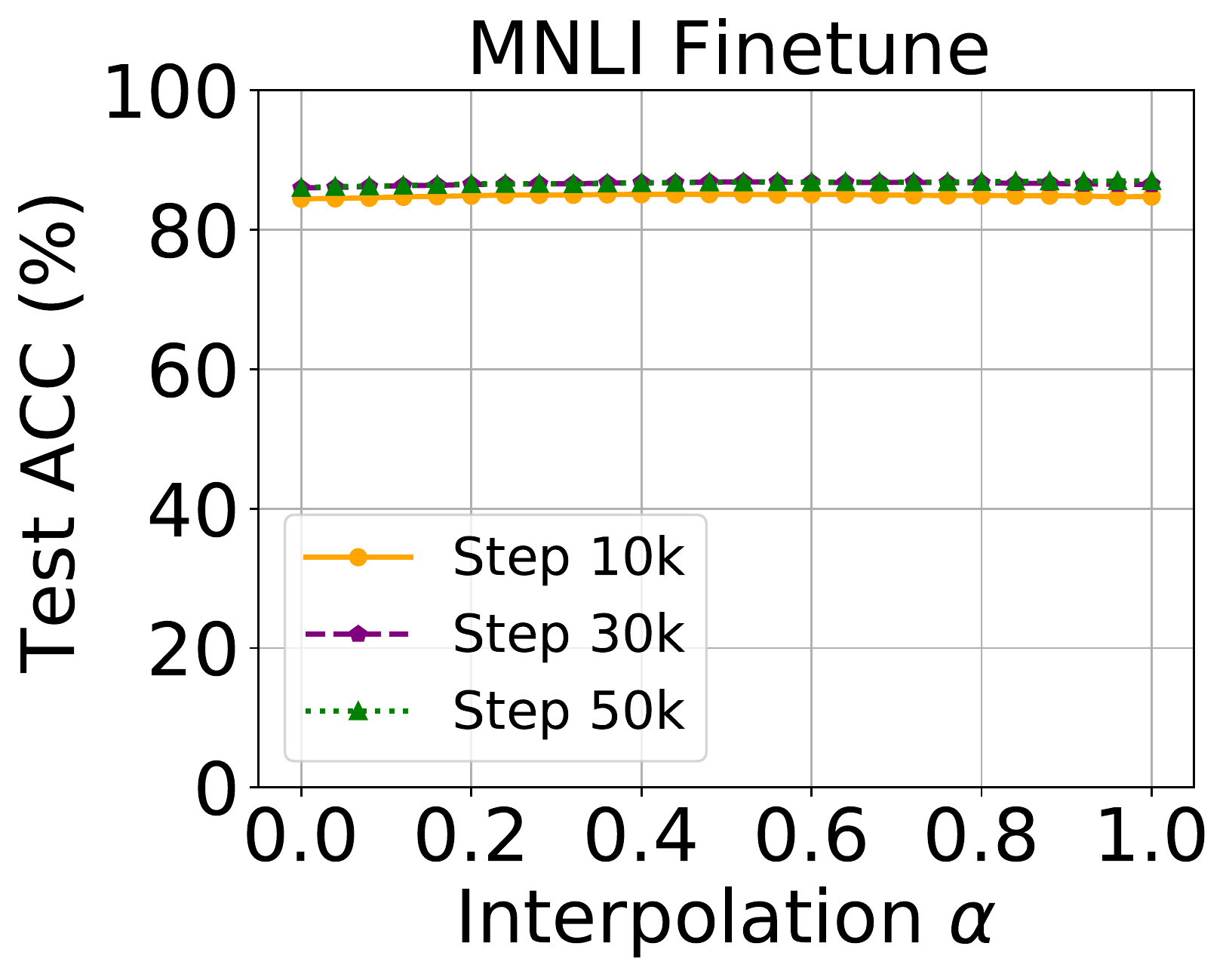}} 
    \subfigure[]{\includegraphics[width=0.24\textwidth]{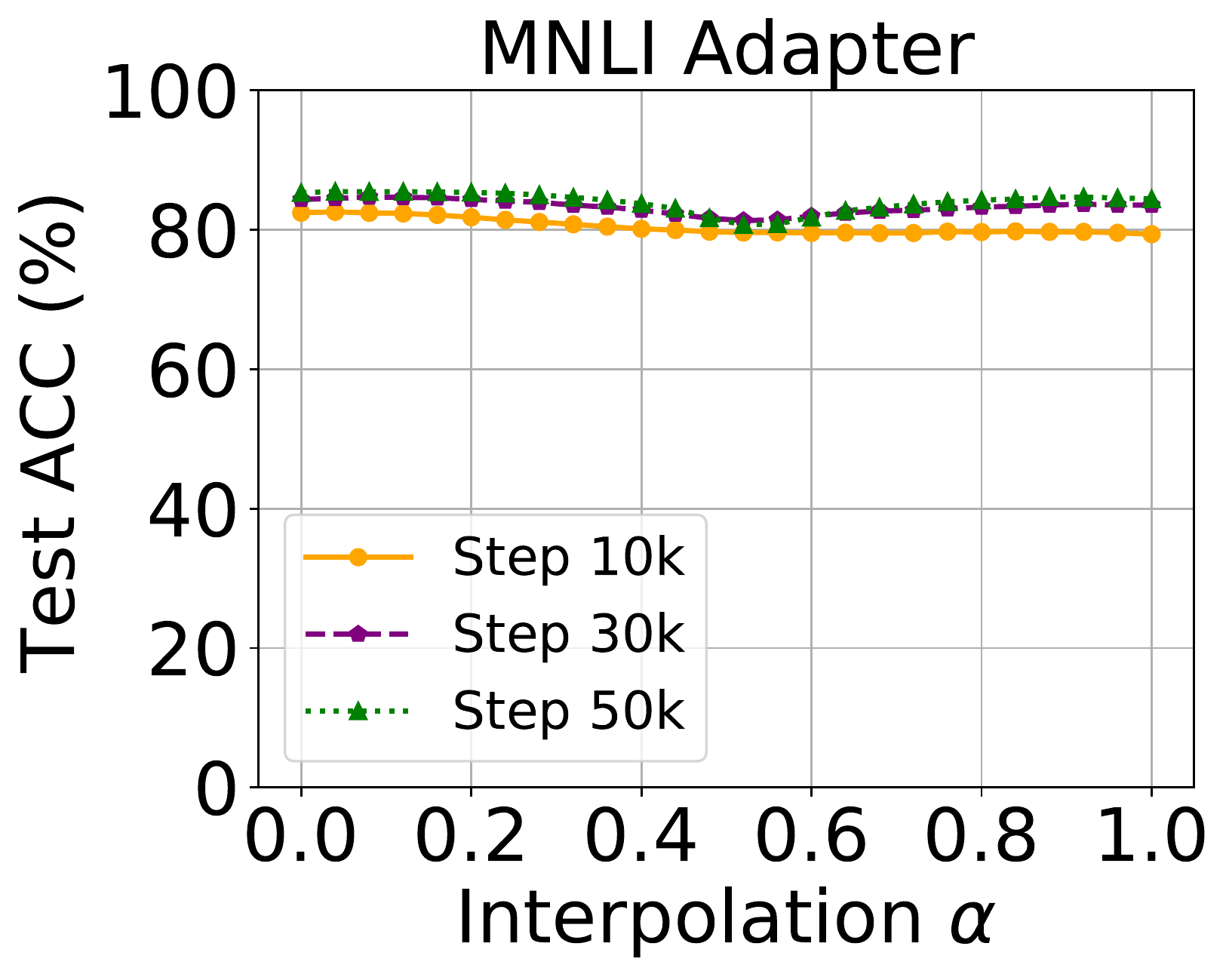}}
    \subfigure[]{\includegraphics[width=0.24\textwidth]{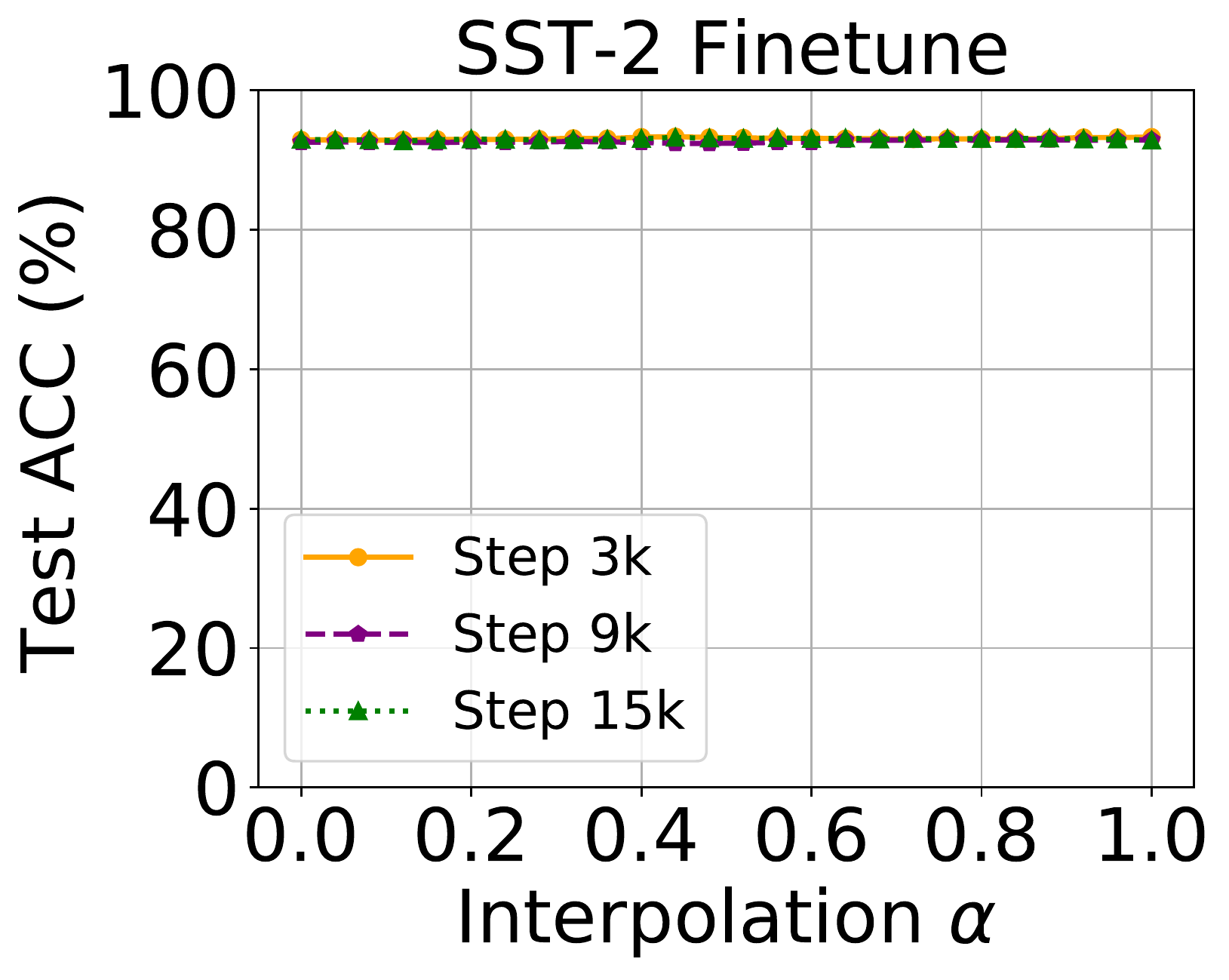}}
    \subfigure[]{\includegraphics[width=0.24\textwidth]{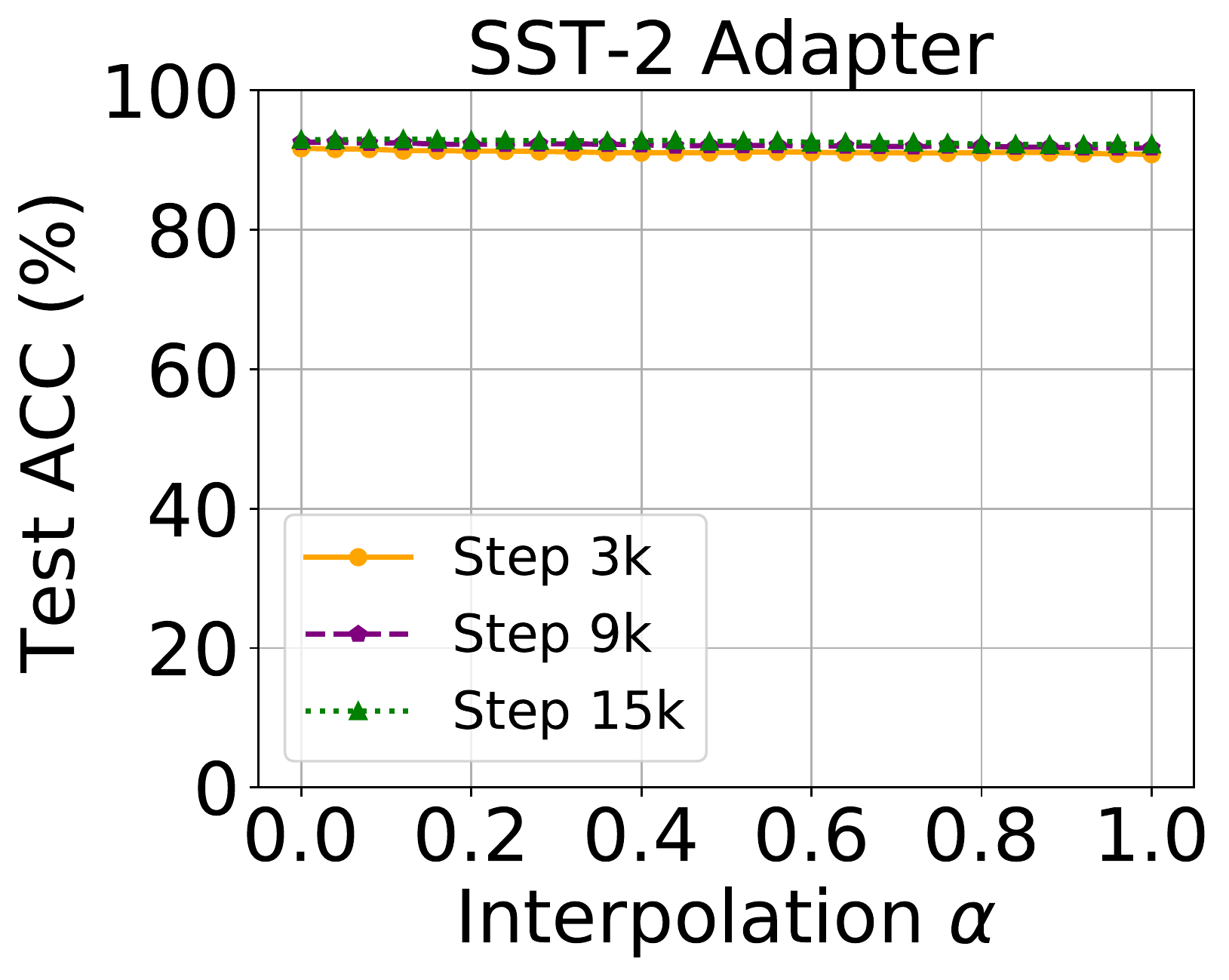}} 
    
    \subfigure[]{\includegraphics[width=0.24\textwidth]{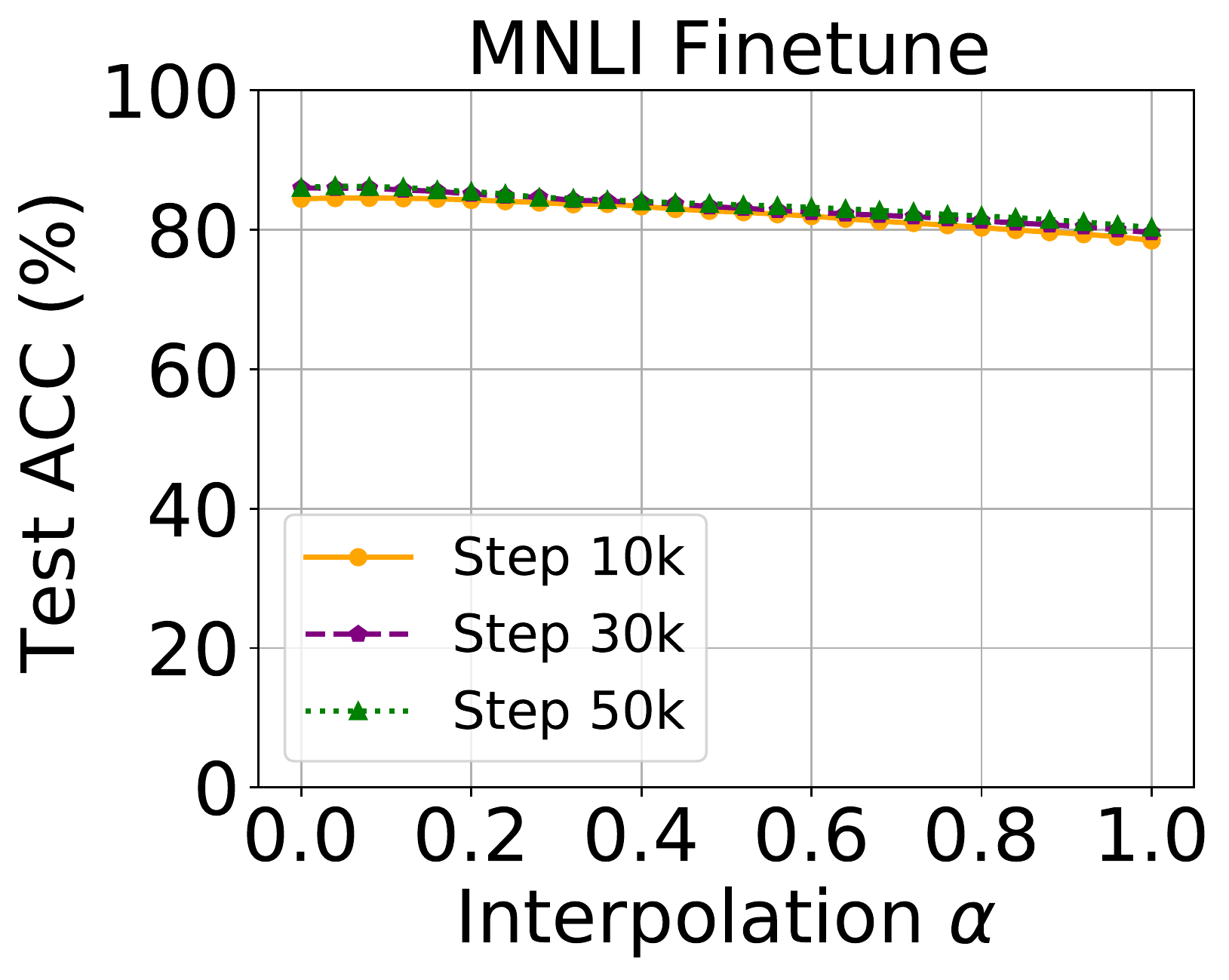}} 
    \subfigure[]{\includegraphics[width=0.24\textwidth]{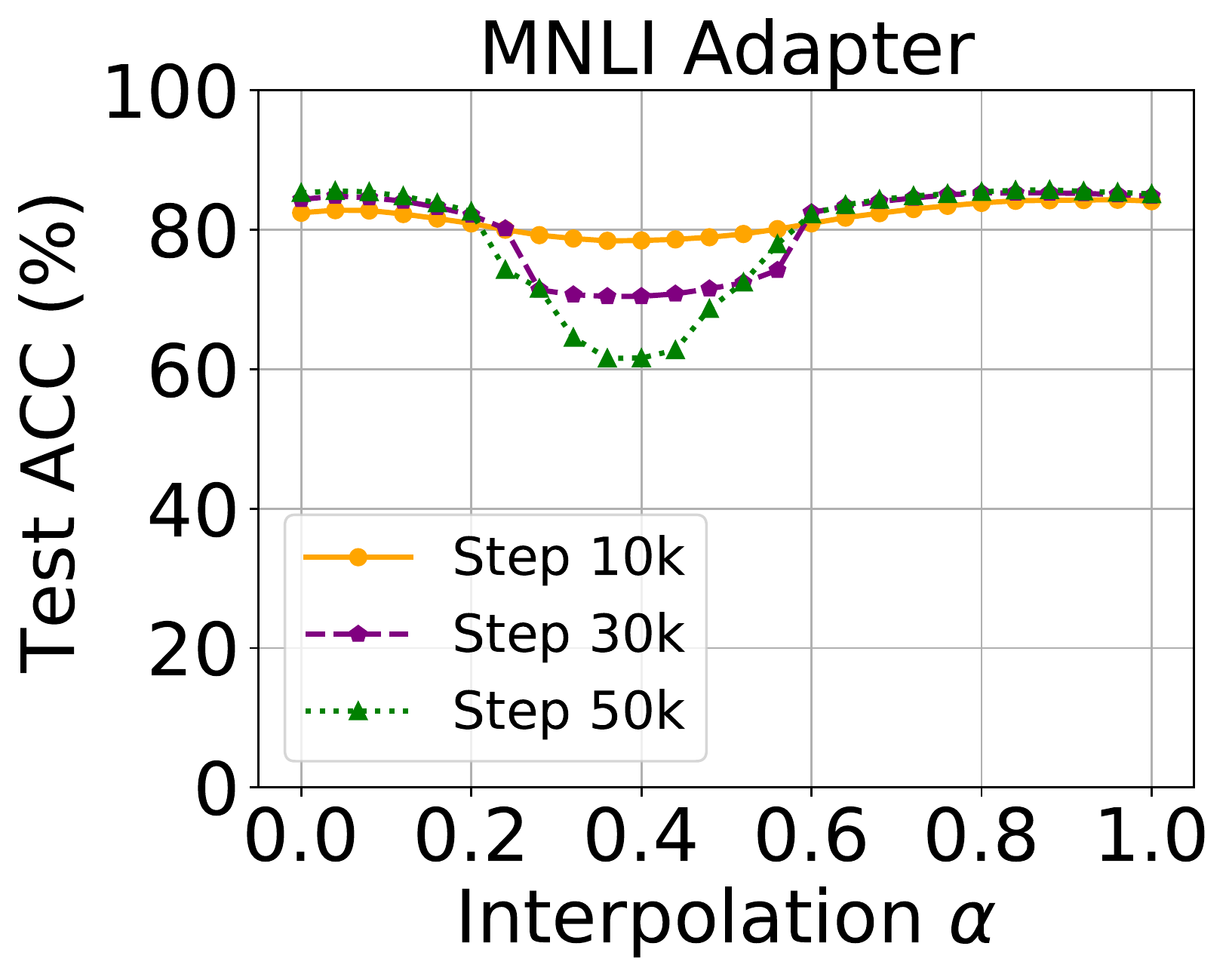}} 
    \subfigure[]{\includegraphics[width=0.24\textwidth]{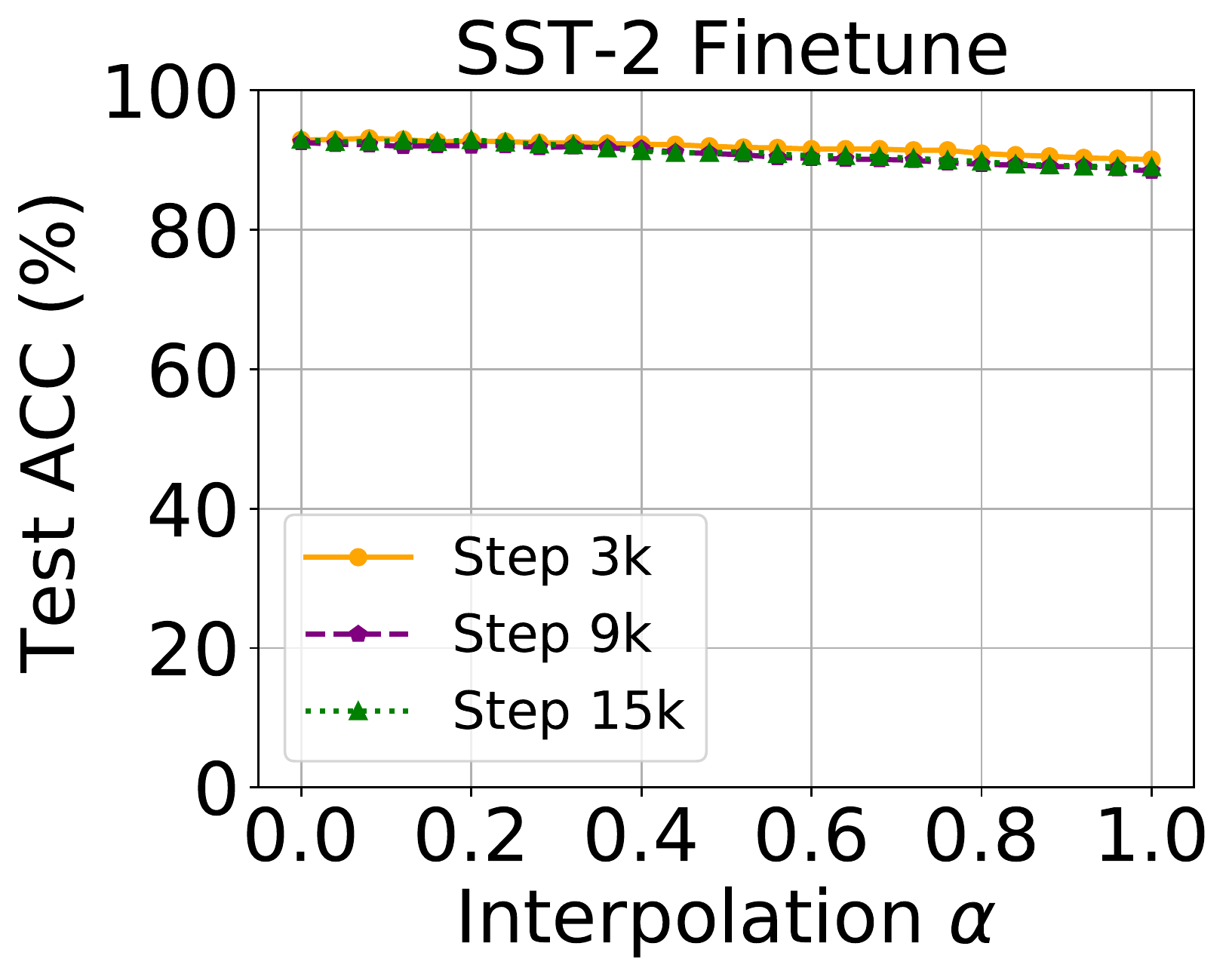}}
    \subfigure[]{\includegraphics[width=0.24\textwidth]{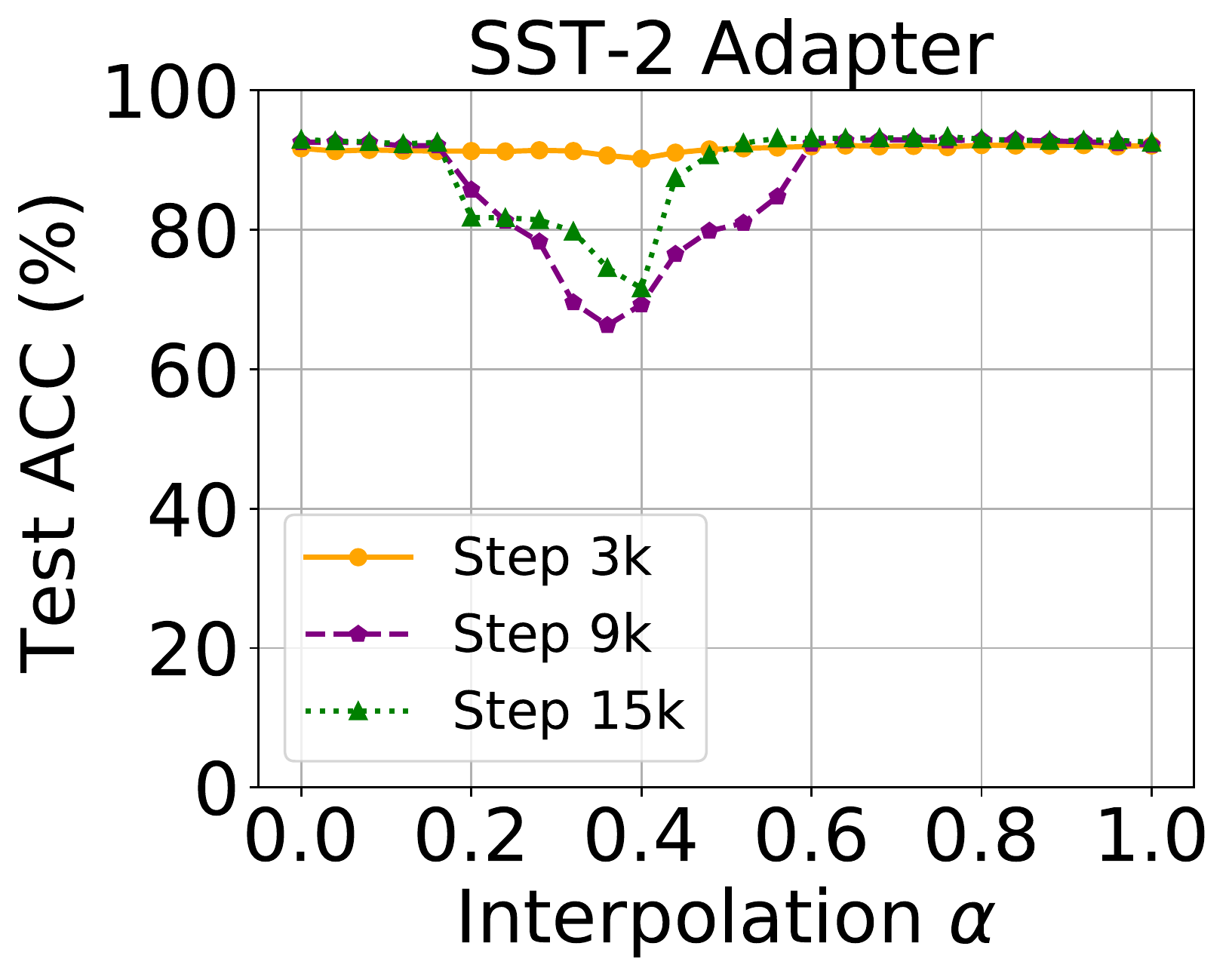}}
    \caption{Experiments of the effects of the learning rate. We conduct linear interpolations for MNLI and SST-2, using both fine-tuning and adapter tuning. For (a-d), both minima are obtained with a learning rate of $1\times10^{-4}$ and $5\times10^{-5}$, respectively; for (e-h), both minima are obtained with a learning rate of $1\times10^{-4}$ and $5\times10^{-4}$, respectively.}
    \label{fig:lr}
\end{figure*}

\begin{figure*}[!t]
    \centering
    \subfigure[]{\includegraphics[width=0.24\textwidth]{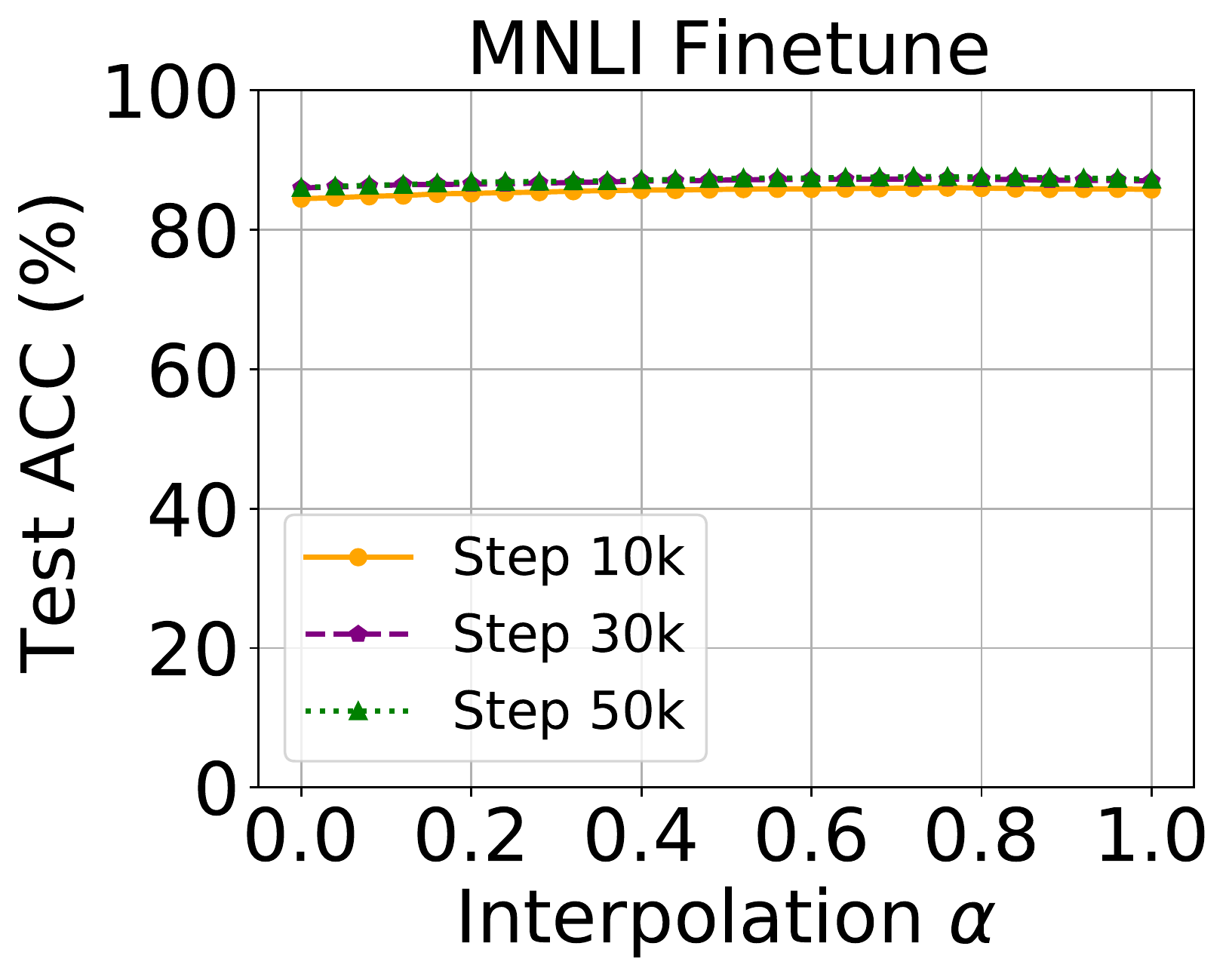}} 
    \subfigure[]{\includegraphics[width=0.24\textwidth]{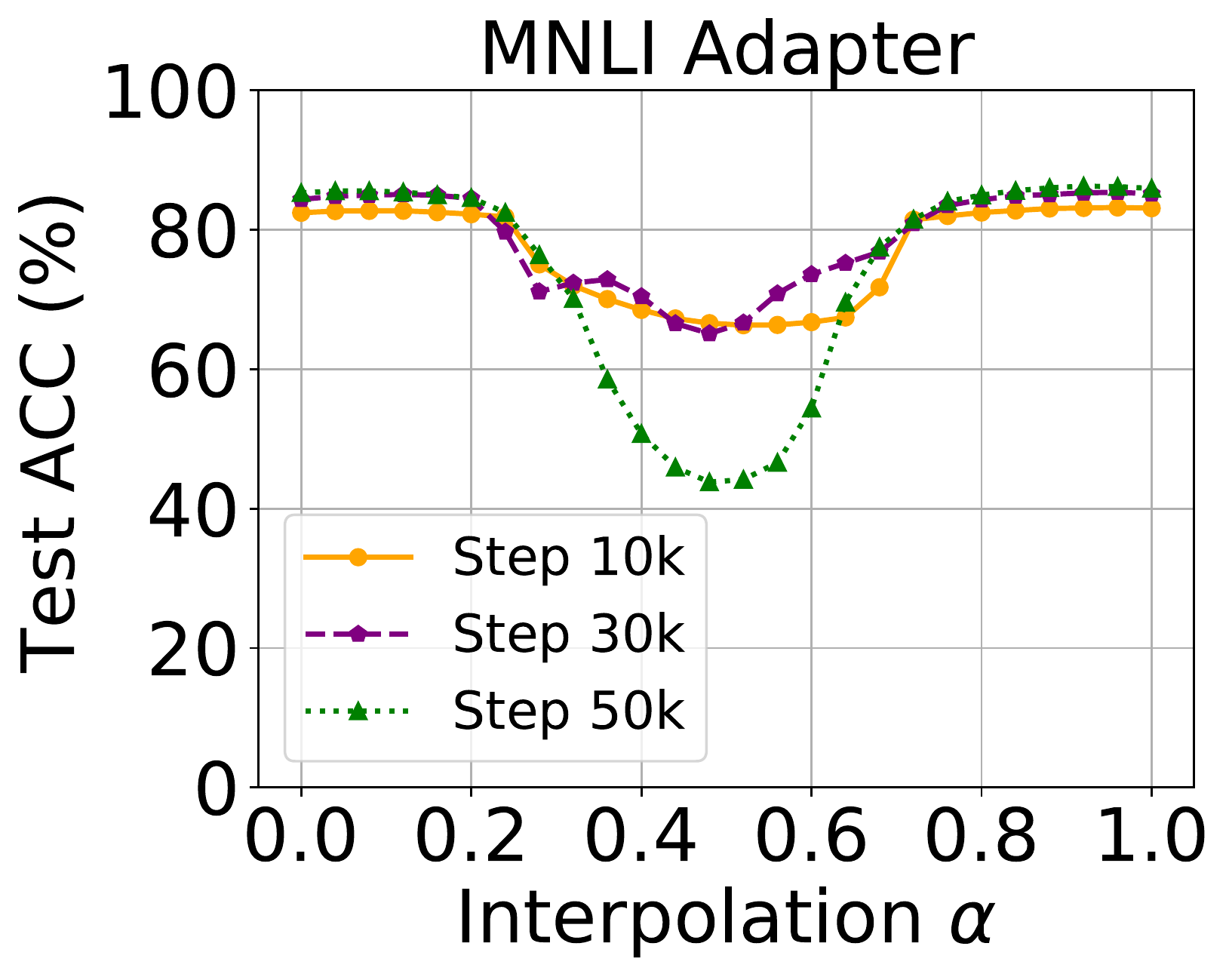}} 
    \subfigure[]{\includegraphics[width=0.24\textwidth]{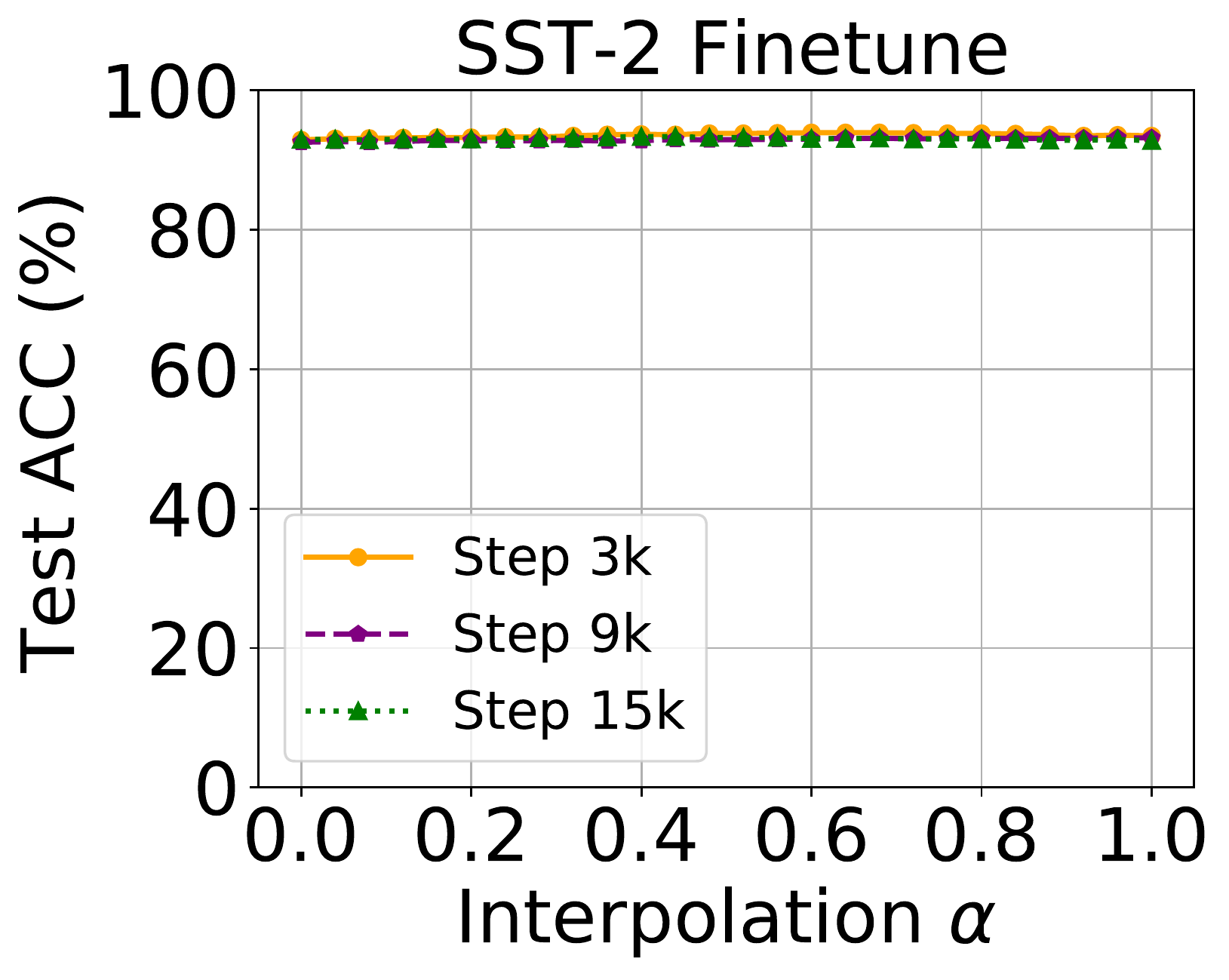}}
    \subfigure[]{\includegraphics[width=0.24\textwidth]{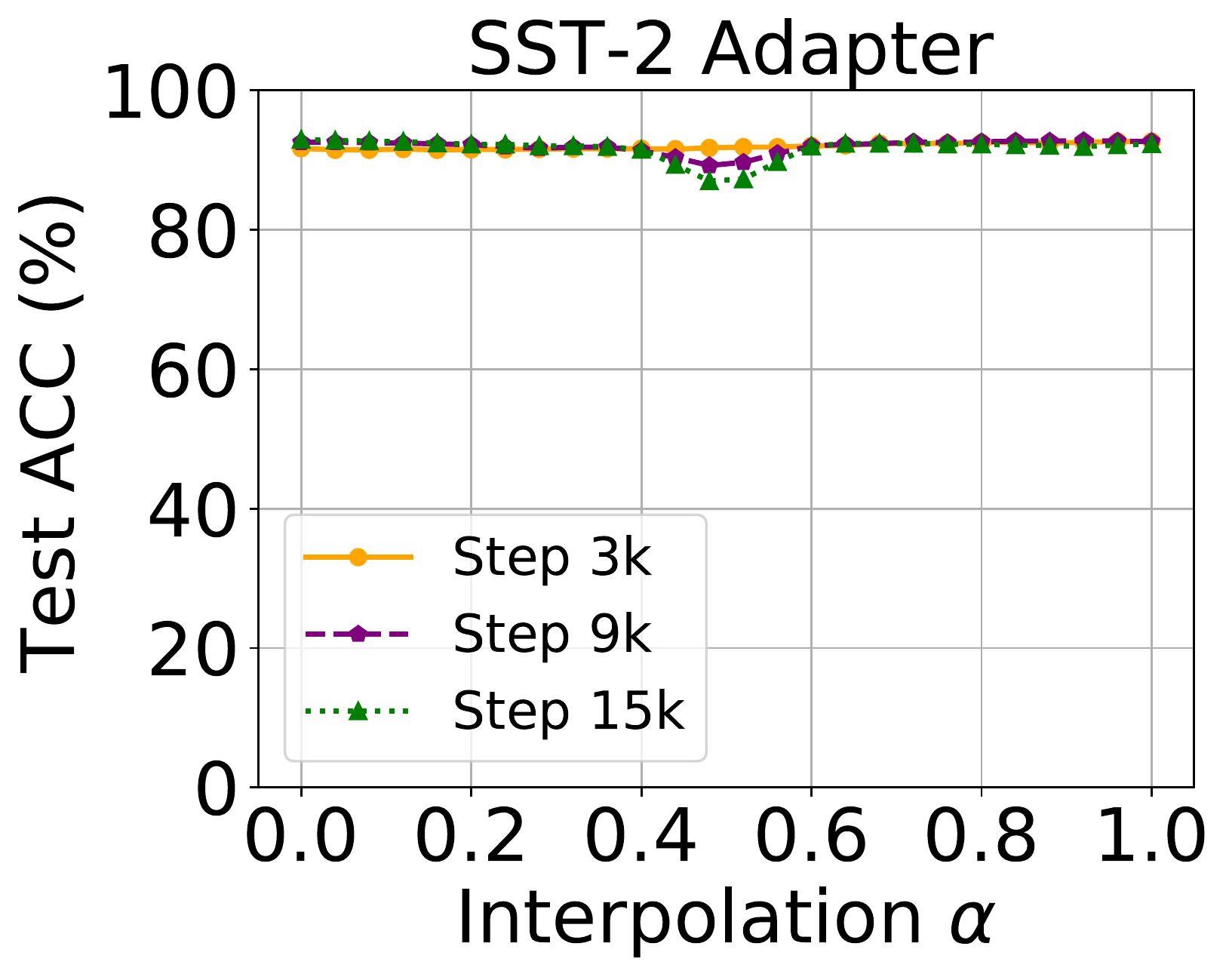}} 
    
    \subfigure[]{\includegraphics[width=0.24\textwidth]{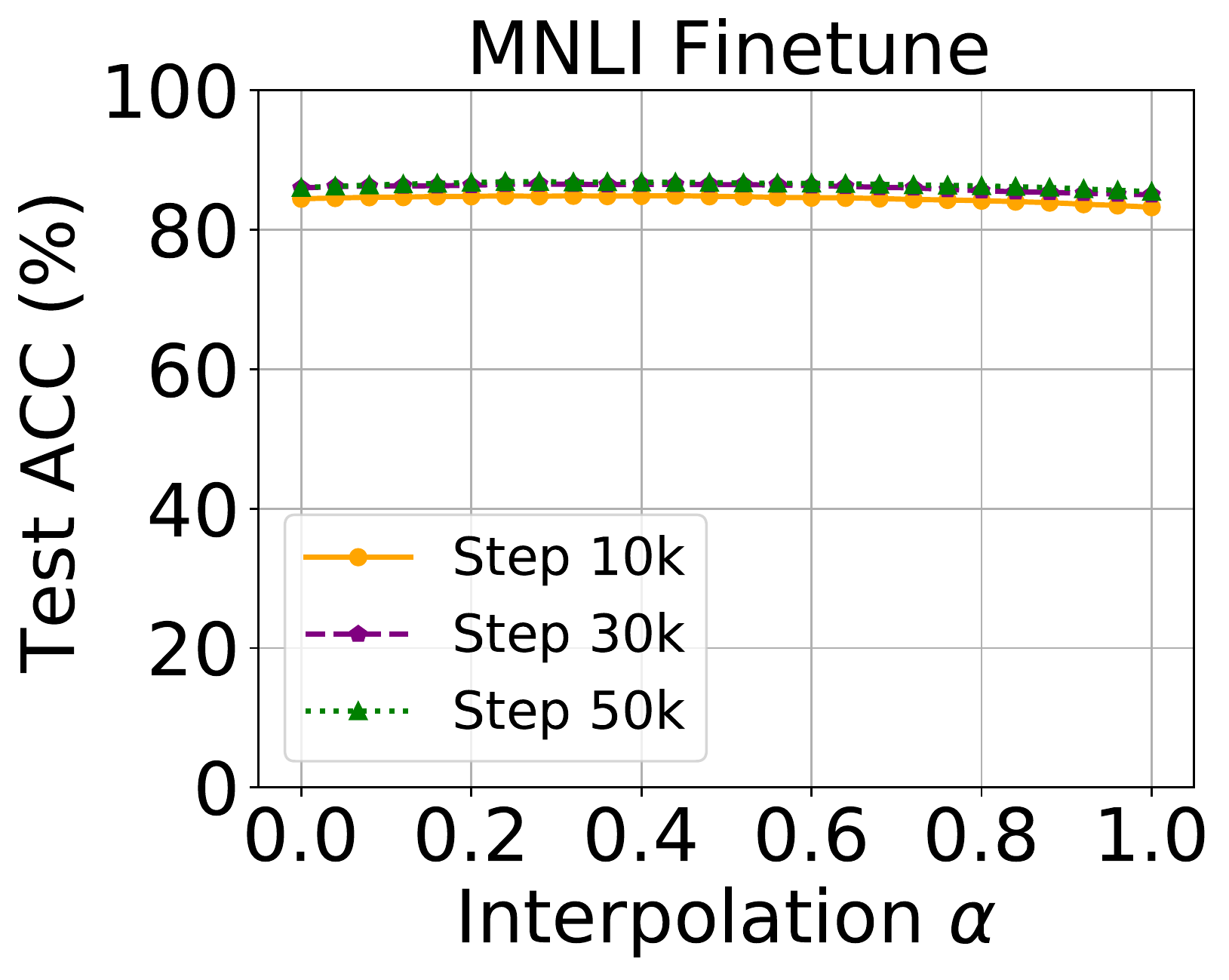}} 
    \subfigure[]{\includegraphics[width=0.24\textwidth]{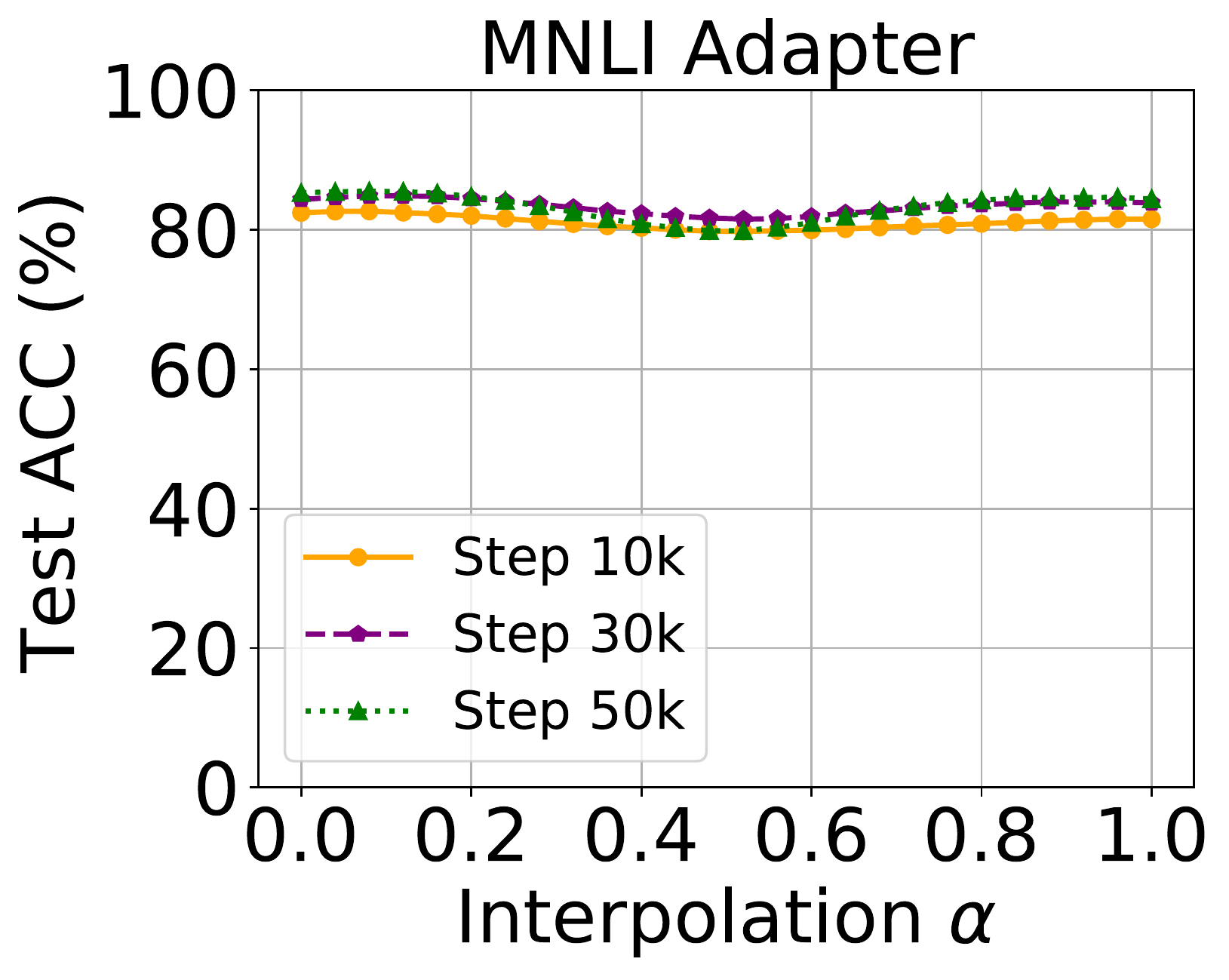}} 
    \subfigure[]{\includegraphics[width=0.24\textwidth]{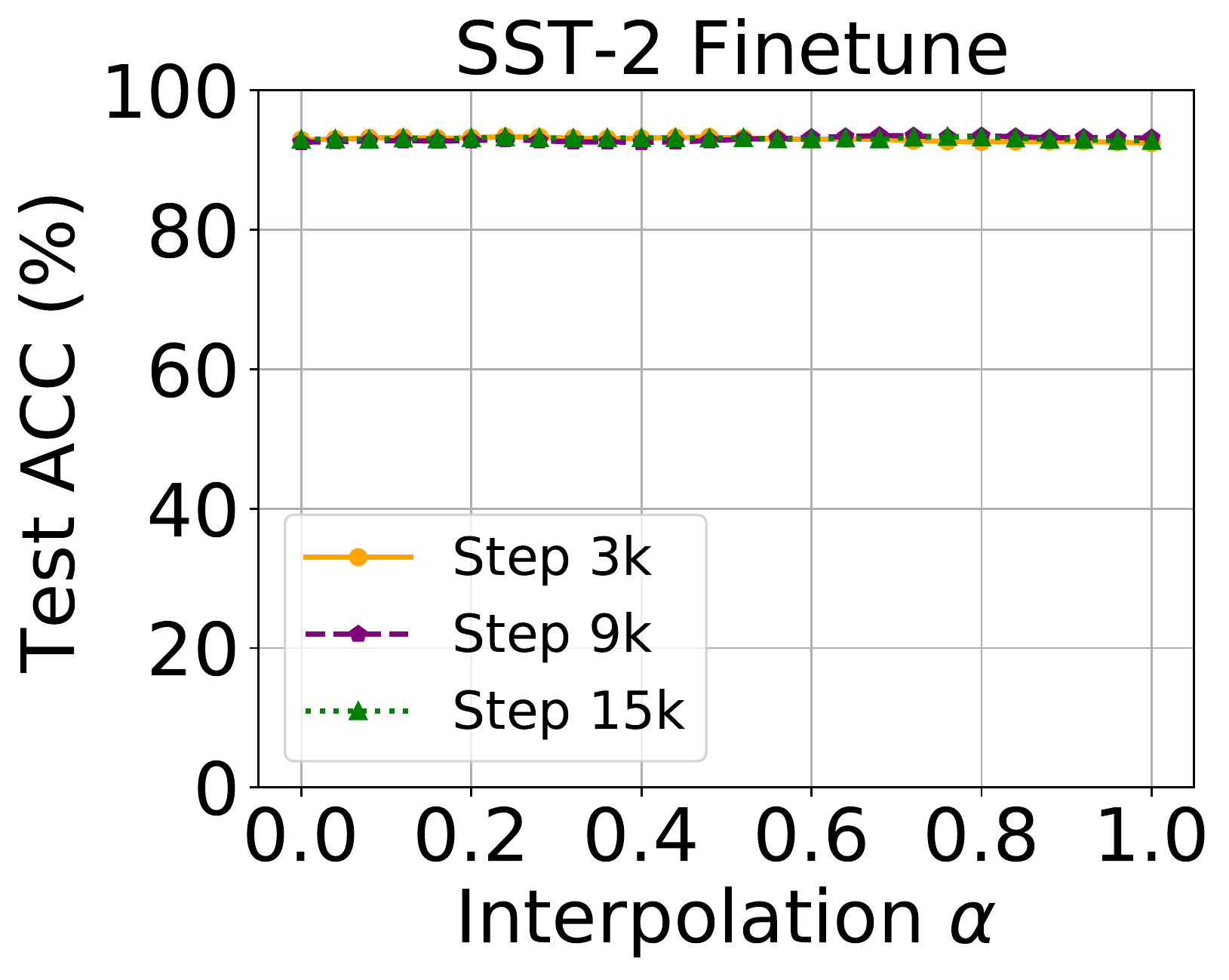}}
    \subfigure[]{\includegraphics[width=0.24\textwidth]{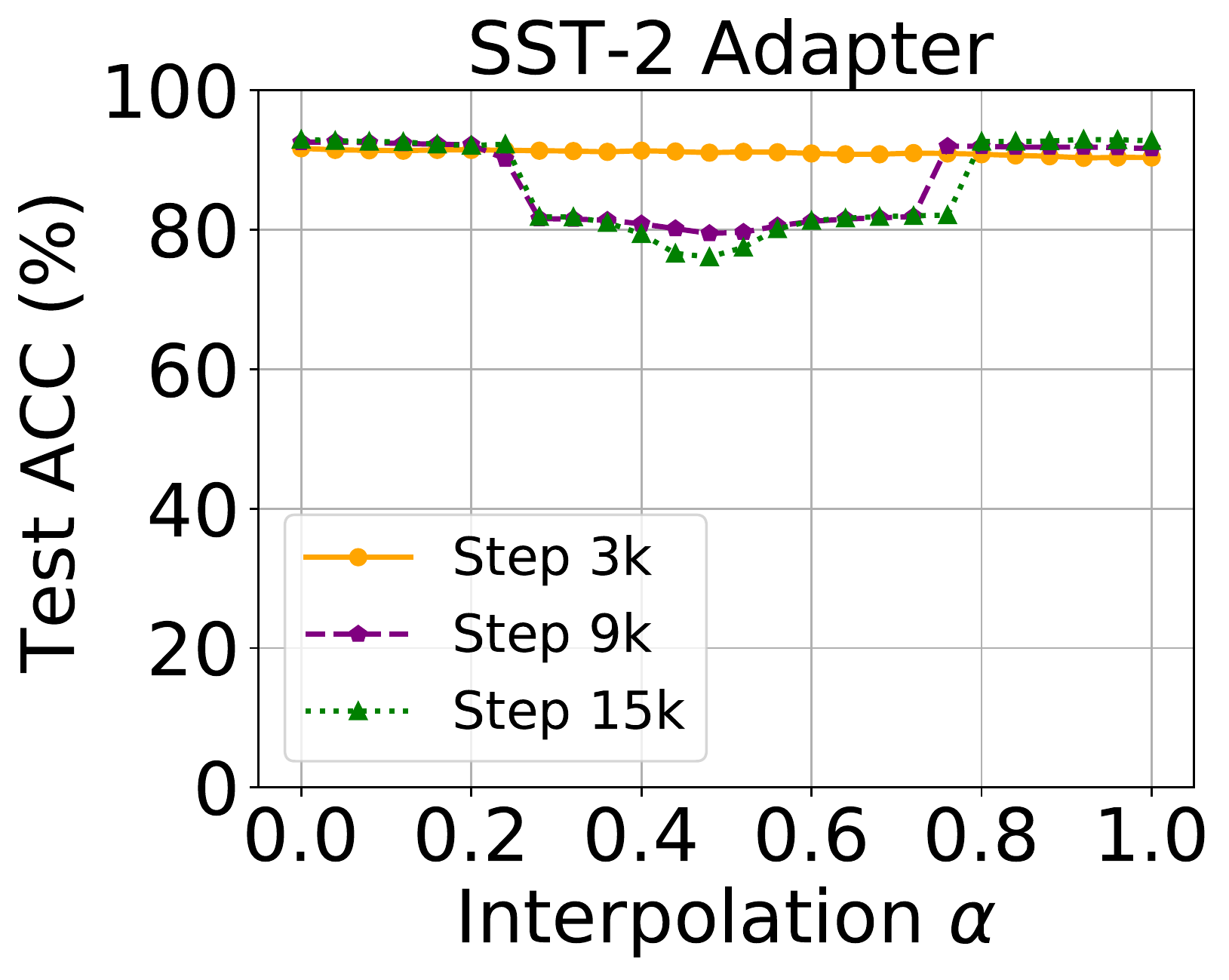}}
    \caption{Experiments of the effects of the batch size. We conduct linear interpolations for MNLI and SST-2, using both fine-tuning and adapter tuning. For (a-d), both minima are obtained with a batch size of $16$ and $32$, respectively; for (e-h), both minima are obtained with a batch size of $16$ and $8$, respectively.}
    \label{fig:bs}
\end{figure*}

We perform experiments to explore the effects of both learning rates and batch sizes. For the former, we evaluate when both endpoints are trained with a learning rate of $\{1\times10^{-4}, 5\times10^{-4}\}$ and $\{1\times10^{-4}, 5\times10^{-5}\}$, and the batch size is set to $16$; for the latter, we chose a batch size of $\{16, 8\}$ and $\{16, 32\}$, and the learning rate is set to $1\times10^{-4}$. The experiments are conducted using both full-parameter fine-tuning and adapter tuning on MNLI and SST-2 with $\text{T5}_\texttt{BASE}$. For MNLI, we experiment when both endpoints are trained for \{10\text{k}, 30\text{k}, 50\text{k}\} steps; for SST-2, both endpoints are trained for \{3\text{k}, 9\text{k}, 15\text{k}\} steps. We illustrate the results of linear interpolation in Figure~\ref{fig:lr} and Figure~\ref{fig:bs}. We could conclude from both figures that, the minima obtained by fine-tuning are always well-connected by the linear path; however, the connectivity of adapter is poor under certain cases. This is aligned with the finding in the main paper that the mode connectivity of fine-tuning is generally better than delta tuning. We also observe that with the training steps becoming larger, the connectivity of adapter tuning sometimes becomes poorer.

\subsection{Additional Experiments for the Effects of Training Steps}
\label{sec:exp_training_step}

\begin{figure*}[!t]
    \centering
    \subfigure{\includegraphics[width=0.24\textwidth]{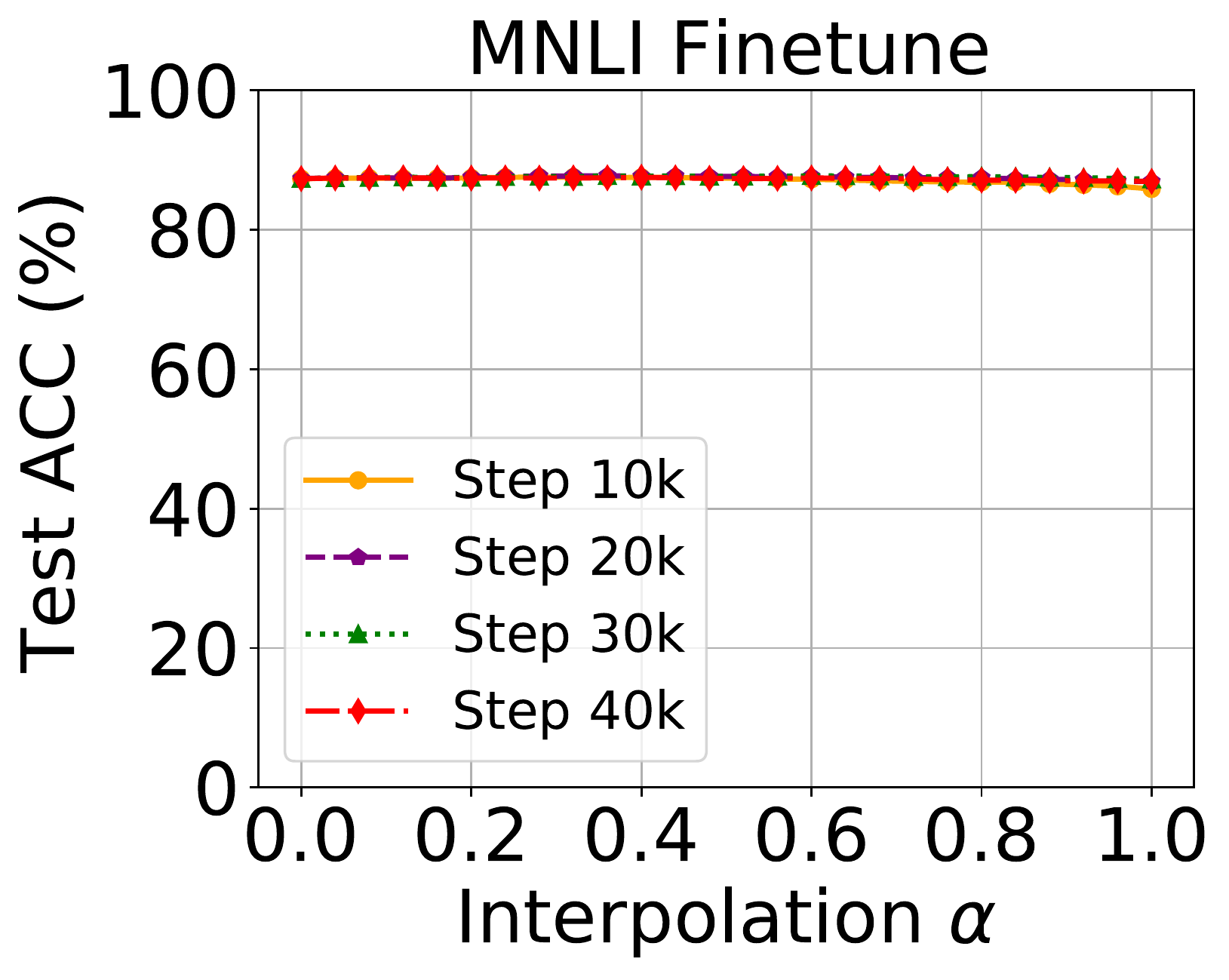}} 
    \subfigure{\includegraphics[width=0.24\textwidth]{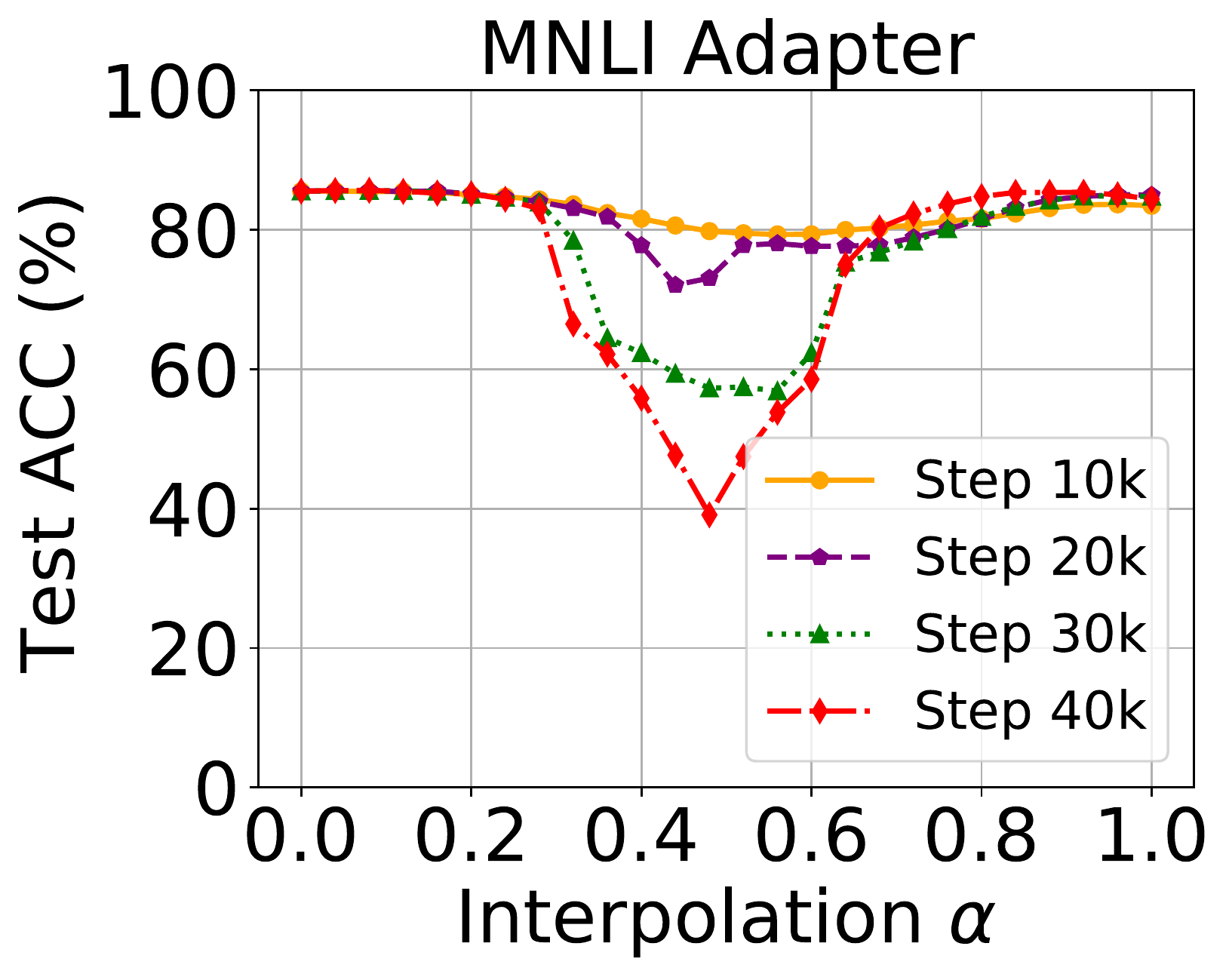}}
    \subfigure{\includegraphics[width=0.24\textwidth]{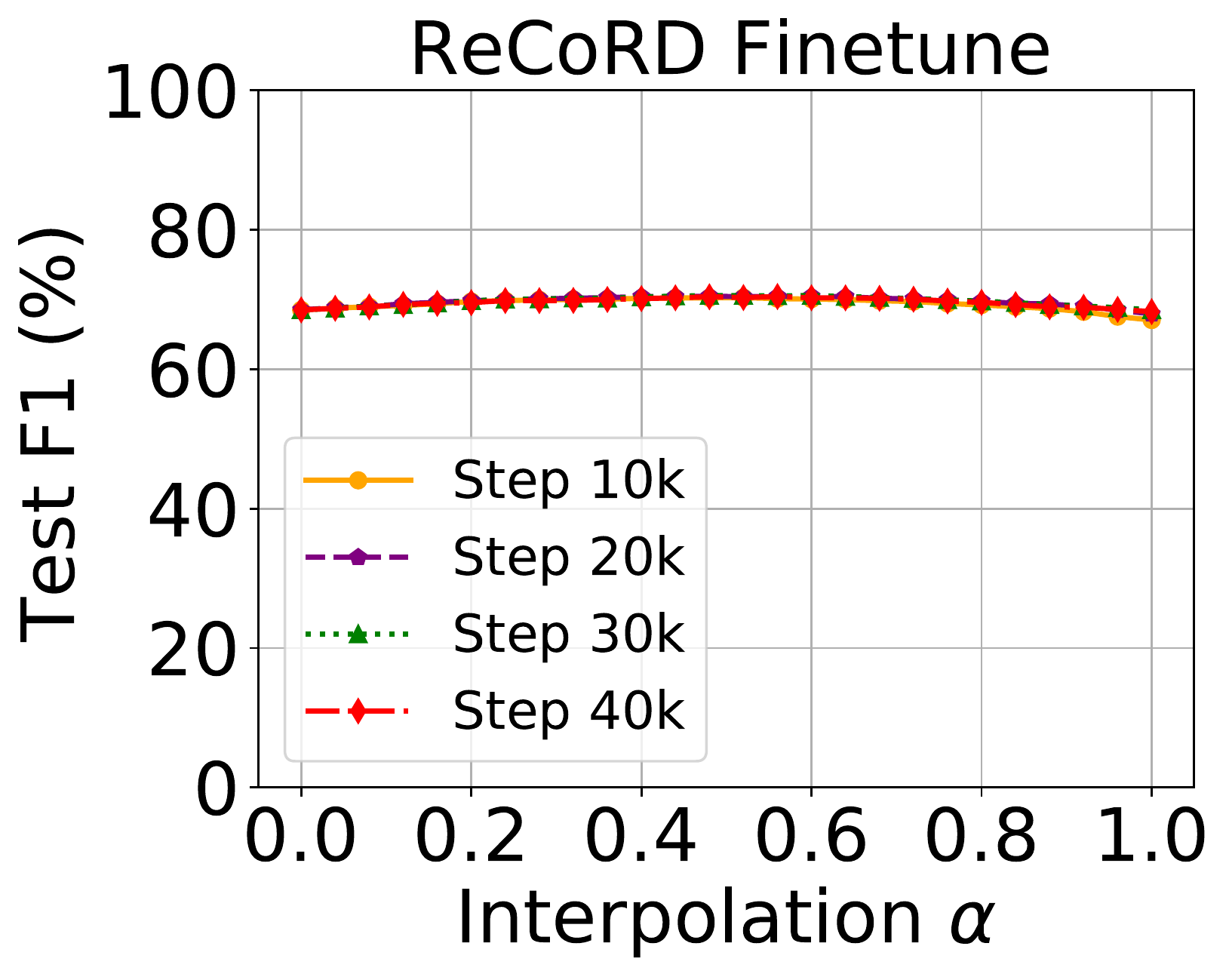}}
    \subfigure{\includegraphics[width=0.24\textwidth]{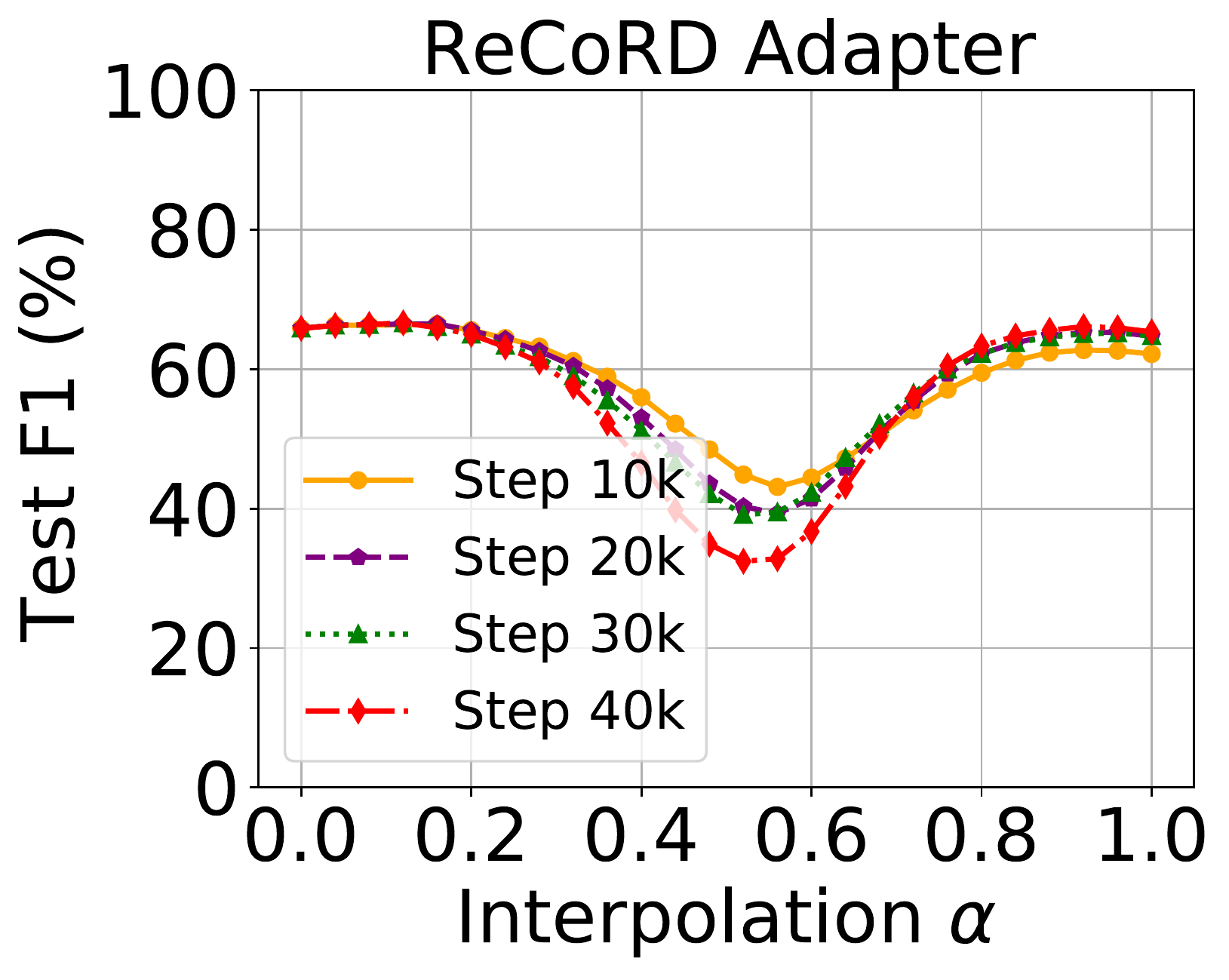}} 
    \caption{Experiments of the effects of the training steps. We conduct linear interpolations for MNLI and ReCoRD, using both fine-tuning and adapter tuning. One endpoint is trained for $50$k steps, while the other endpoint is trained for \{$10$k, $20$k, $30$k, $40$k\} steps, respectively.}
    \label{fig:step}
\end{figure*}

\begin{figure*}[!t]
    \centering
    \subfigure{\includegraphics[width=0.24\textwidth]{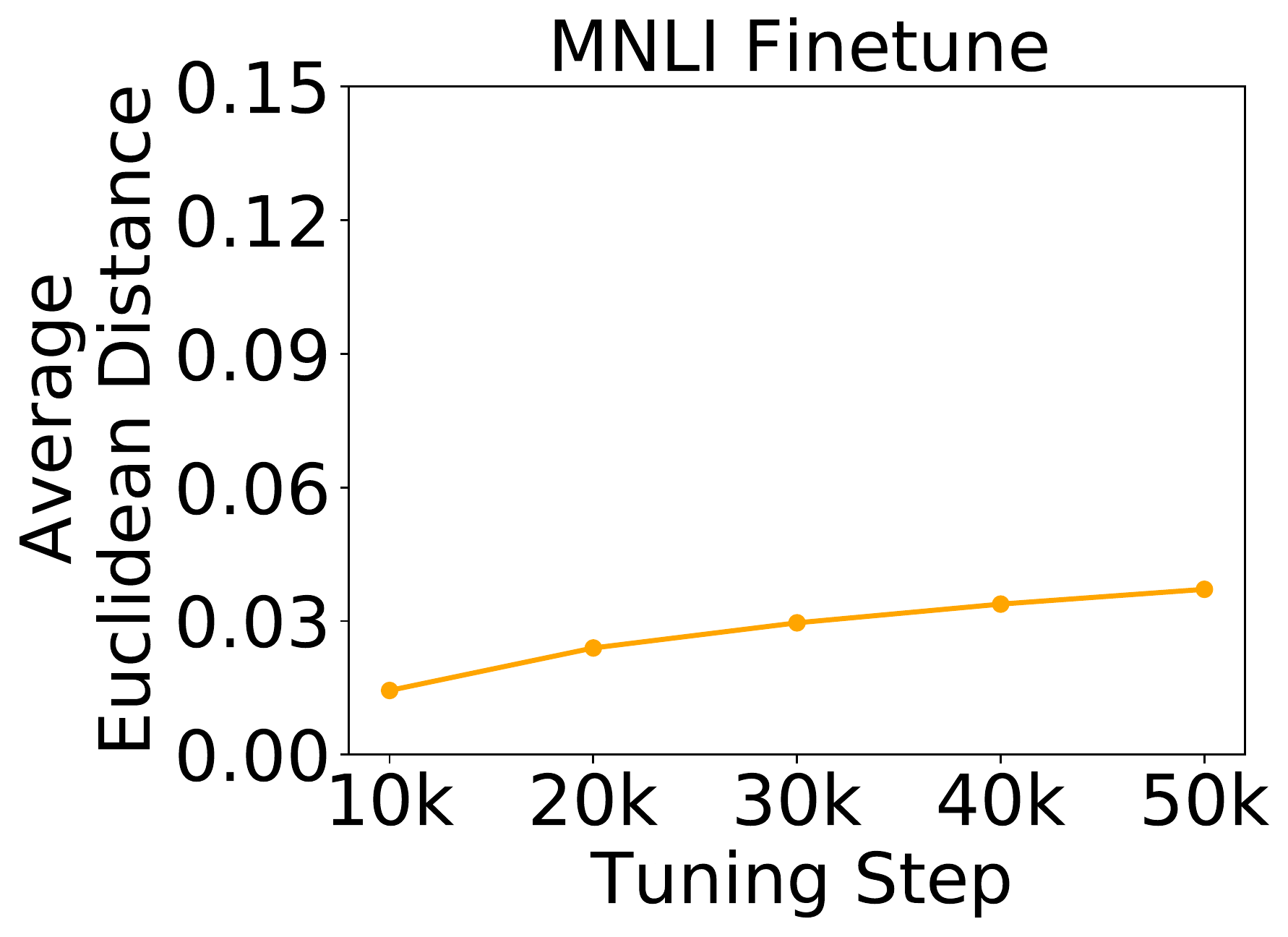}} 
    \subfigure{\includegraphics[width=0.24\textwidth]{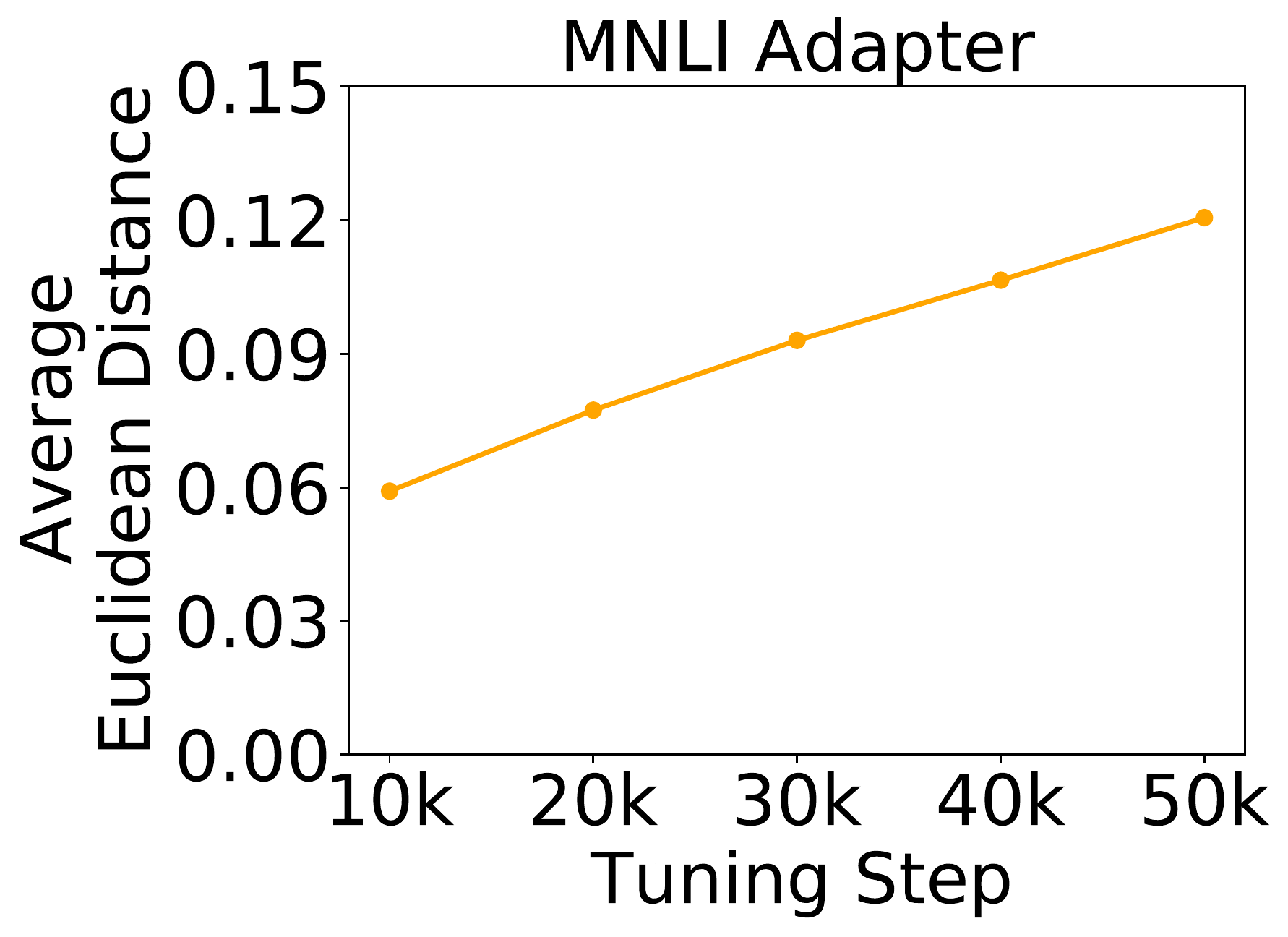}}
    \subfigure{\includegraphics[width=0.24\textwidth]{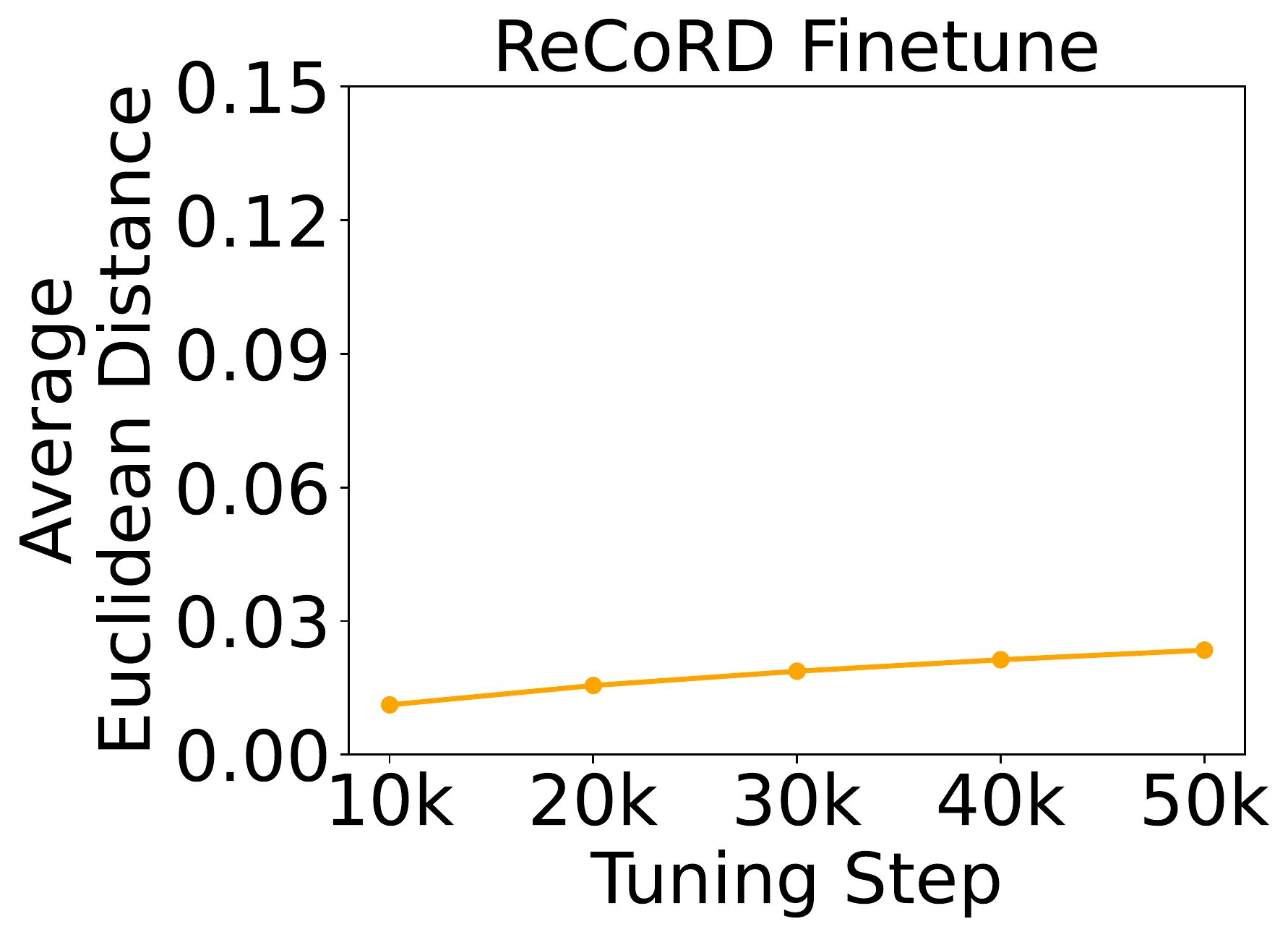}}
    \subfigure{\includegraphics[width=0.24\textwidth]{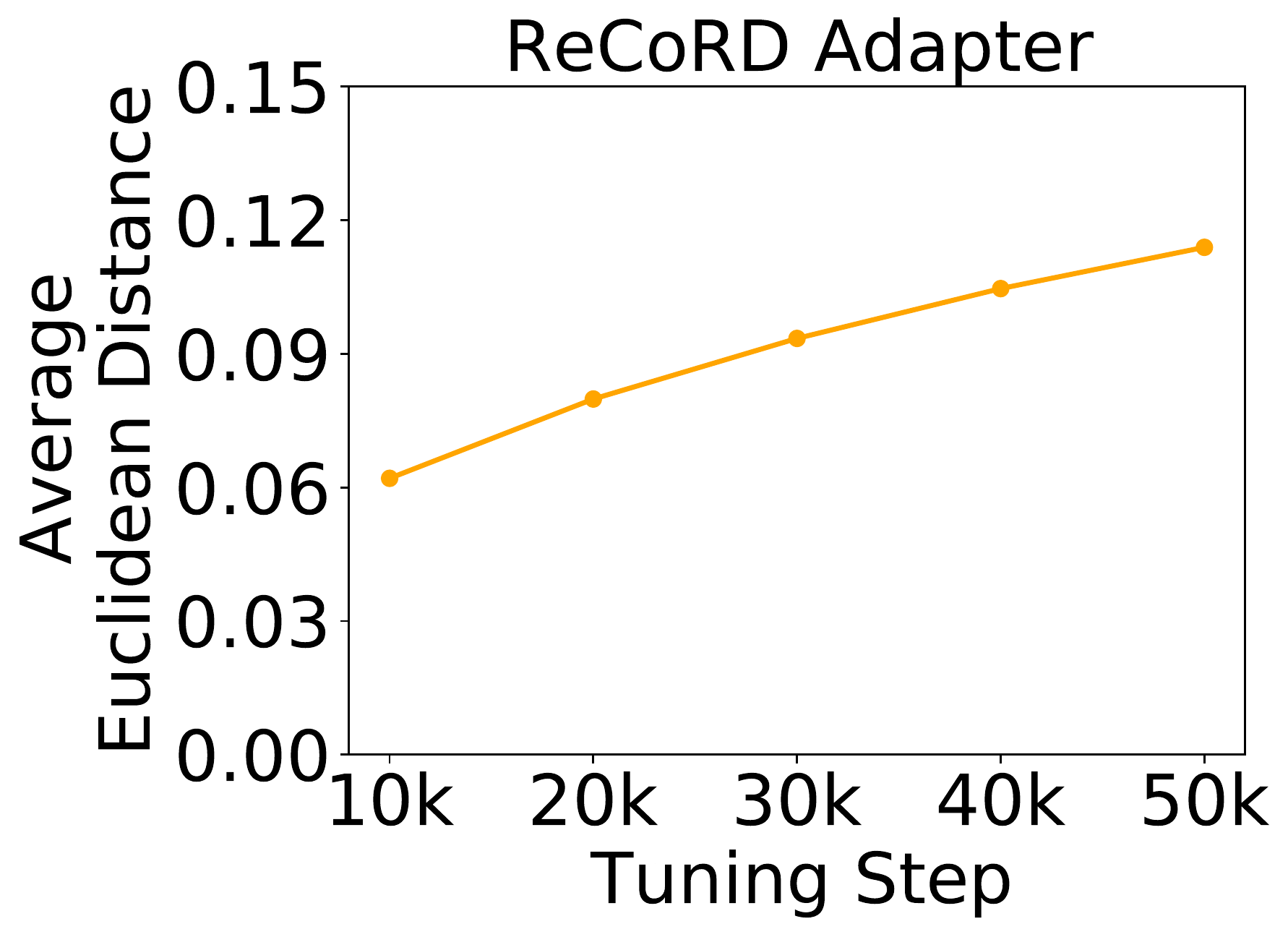}} 
    \caption{Euclidean distance (per neuron) of two minima at different training steps (\{$10$k, $20$k, $30$k, $40$k\}) during downstream adaptation. Two minima are trained from different initialization.}
    \label{fig:l2_distance}
\end{figure*}
In the main paper, when exploring the effects of training steps, we experiment with the setting where both endpoints are trained with the same number of steps. We show that mode connectivity could be poorer when both endpoints are trained for longer steps. To more rigorously investigate the effects of training steps, we experiment when both minima are obtained when using different training steps. Specifically, we adapt $\text{T5}_\texttt{BASE}$ model on MNLI and ReCoRD using both fine-tuning and adapter tuning. We train two endpoints with different initialization, which is implemented by utilizing different random seeds, but keeping the configuration of the initialization (mean and standard deviation of the normal distribution) the same. After that, one endpoint (denoted as $\theta_{C_1}$) is adapted for $50$k steps, while the other endpoint is adapted for \{$10$k, $20$k, $30$k, $40$k\} steps, and denoted as \{$\theta_{C_2}^{10\text{k}}$, $\theta_{C_2}^{20\text{k}}$, $\theta_{C_2}^{30\text{k}}$, $\theta_{C_2}^{40\text{k}}$\}, respectively. Then we evaluate the linear interpolations between \{($\theta_{C_1}$ and $\theta_{C_2}^{10\text{k}}$), ($\theta_{C_1}$ and $\theta_{C_2}^{20\text{k}}$), ($\theta_{C_1}$ and $\theta_{C_2}^{30\text{k}}$), ($\theta_{C_1}$ and $\theta_{C_2}^{40\text{k}}$)\}, respectively. The results are shown in Figure~\ref{fig:step}, from which we observe that, the mode connectivity of fine-tuning is generally good, while two minima of adapter are not well-connected. In addition, with the gap of the training steps between two endpoints becoming larger, the mode connectivity of adapter becomes poorer. These results suggest that the number of training steps can affect PLM's mode connectivity under certain cases.

\paragraph{Euclidean Distance Analysis.}
To better understand the reason why training steps could affect mode connectivity, we record the Euclidean distance variation of two endpoints during downstream adaptation. Both endpoints start from different initialization. We adapt the PLM on MNLI using both fine-tuning and adapter tuning for \{$10$k, $20$k, $30$k, $40$k, $50$k\} steps, and obtain a series of checkpoints: \{($\theta_{C_1}^{10\text{k}}$ and $\theta_{C_2}^{10\text{k}}$), ($\theta_{C_1}^{20\text{k}}$ and $\theta_{C_2}^{20\text{k}}$), ($\theta_{C_1}^{30\text{k}}$ and $\theta_{C_2}^{30\text{k}}$), ($\theta_{C_1}^{40\text{k}}$ and $\theta_{C_2}^{40\text{k}}$), ($\theta_{C_1}^{50\text{k}}$ and $\theta_{C_2}^{50\text{k}}$)\}. Then we calculate the Euclidean distance of two endpoints as: $||\theta_{C_1}^{*} - \theta_{C_2}^{*}||^2$. The change of Euclidean distance is visualized in Figure~\ref{fig:l2_distance}, from which we observe that, with the training steps becoming larger, the distance between two endpoints is also prolonged. This may partially explain the poorer mode connectivity with the increasing of training steps. We have shown in the main paper that PLMs have multiple loss basins connected by a non-linear path, instead of a linear path. Within the same loss basins, most of the solutions have good linear mode connectivity. However, since the loss basin has a boundary, when the distance between two minima becomes large enough, they may finally cross the border of the loss basin. Under this scenario, the linear path connecting both endpoints would incur a high loss.

\subsection{More Experiments for Mode Connectivity along a Non-linear Path}
\label{sec:exp_curved}
In the main paper, to explore the connectivity of two minima along a non-linear path, we experiment on the setting of tuning adapters with different initialization. This setting has been shown to have poor linear mode connectivity but good non-linear mode connectivity. In fact, in our pilot experiments, we find that such a low-loss non-linear curve exists for minima reached under various different settings. In this section, we provide some of the experiments to demonstrate the above finding using $\text{T5}_\texttt{BASE}$.

Specifically, we experiment with three tasks: MNLI, ReCoRD, and SST-2. For adapter tuning, we choose the setting where two minima are trained with (1) different training data order and (2) data from the same distribution but without specific overlap of training instances. For (2), same as before, we randomly partition the original training dataset into two equal splits, and adapt two copies of PLM on each split. For fine-tuning, we choose the setting where two minima are trained with (1) different training steps (the setting is the same as \cref{sec:exp_training_step}), and (2) different initialization.

\begin{figure*}[!t]
    \centering
    \subfigure[]{\includegraphics[width=0.24\textwidth]{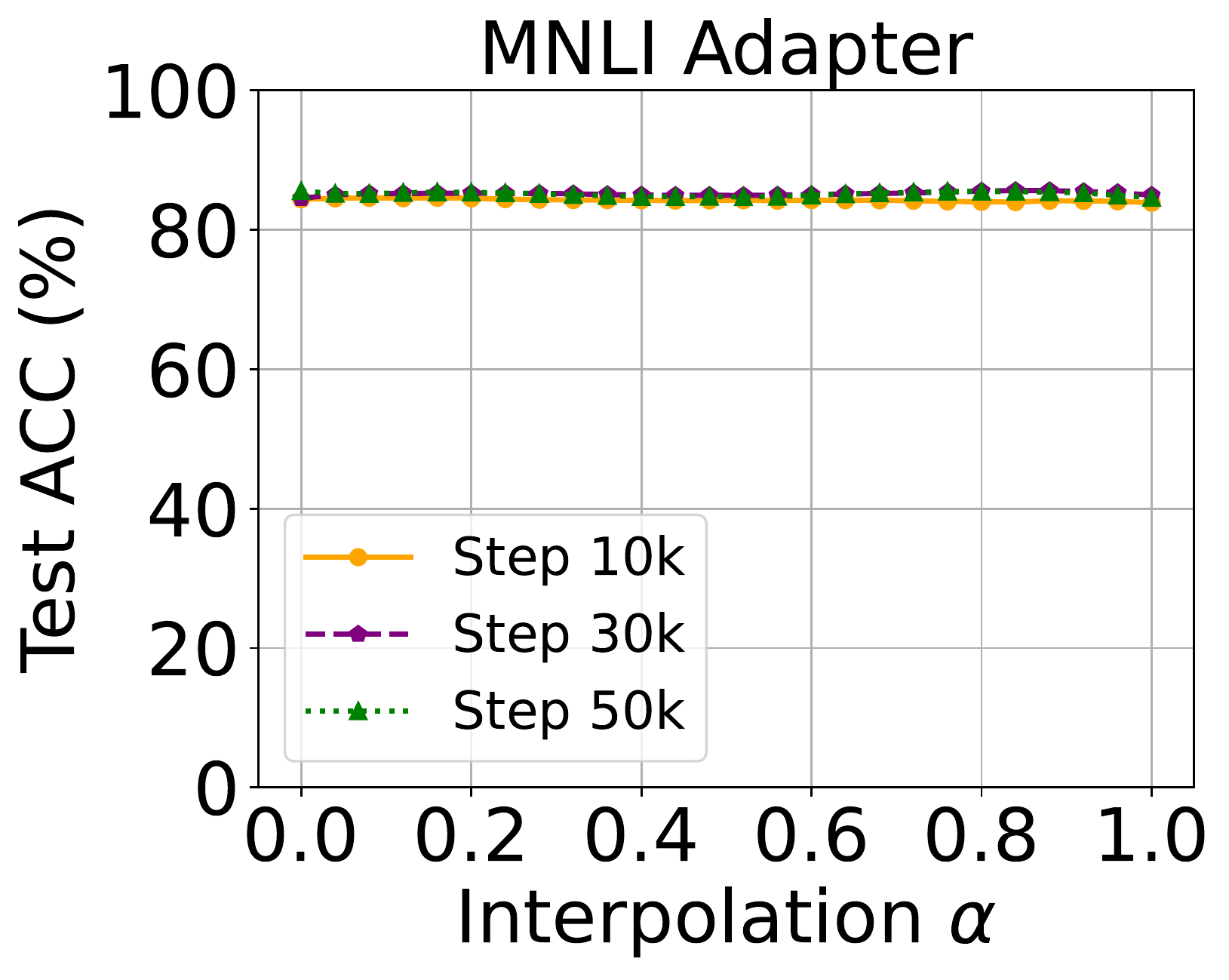}}
    \subfigure[]{\includegraphics[width=0.24\textwidth]{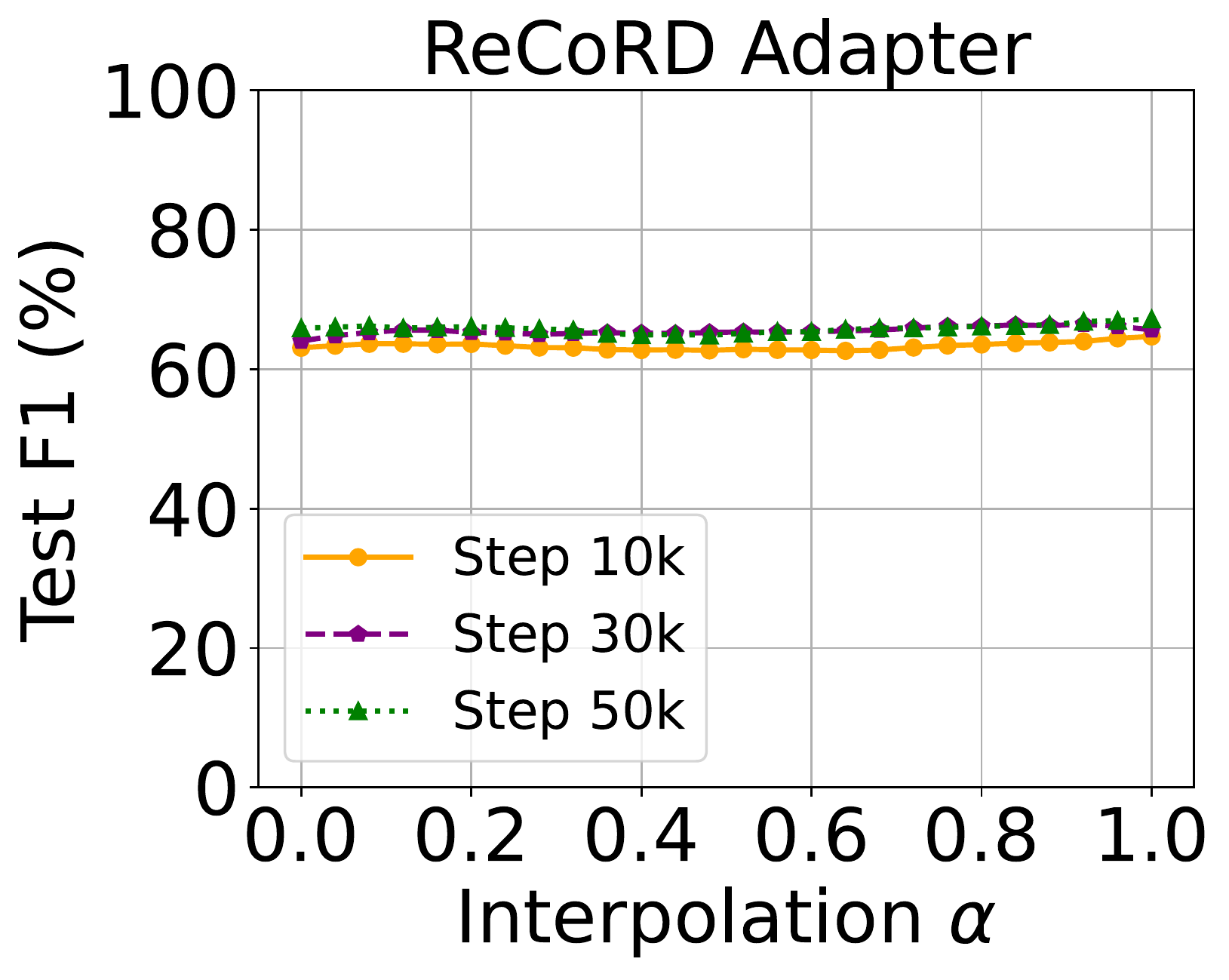}}
    \subfigure[]{\includegraphics[width=0.24\textwidth]{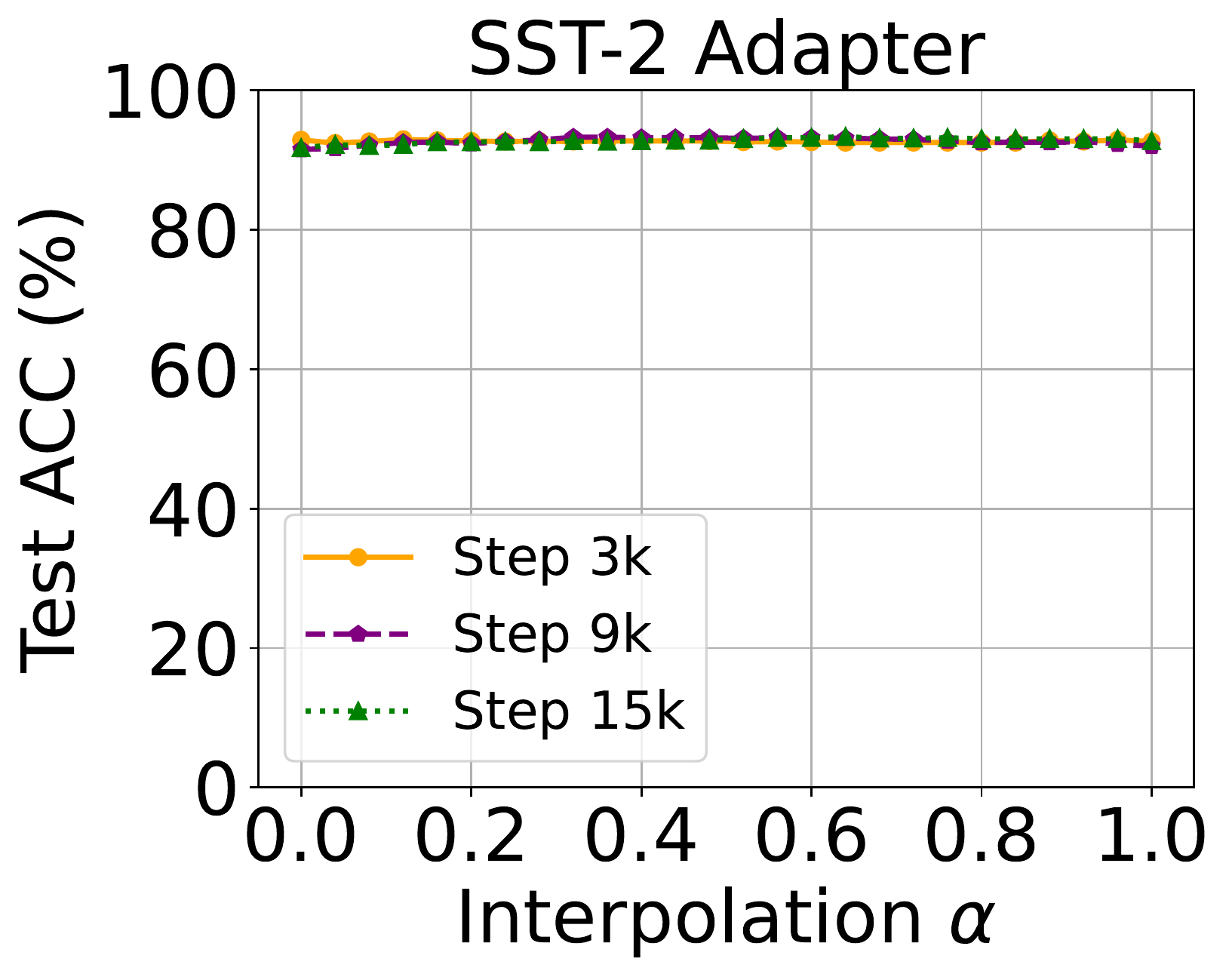}}
    
    \subfigure[]{\includegraphics[width=0.24\textwidth]{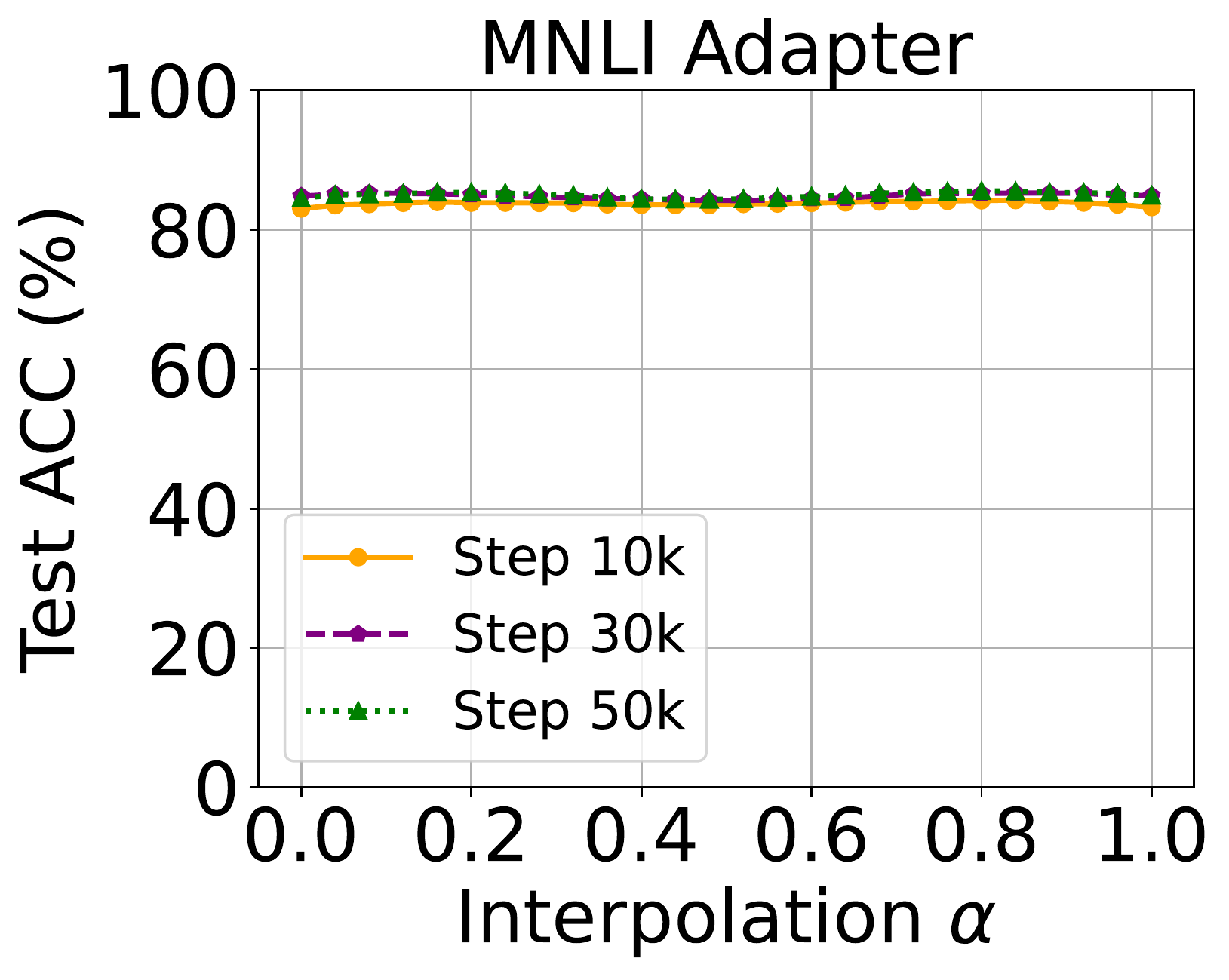}}
    \subfigure[]{\includegraphics[width=0.24\textwidth]{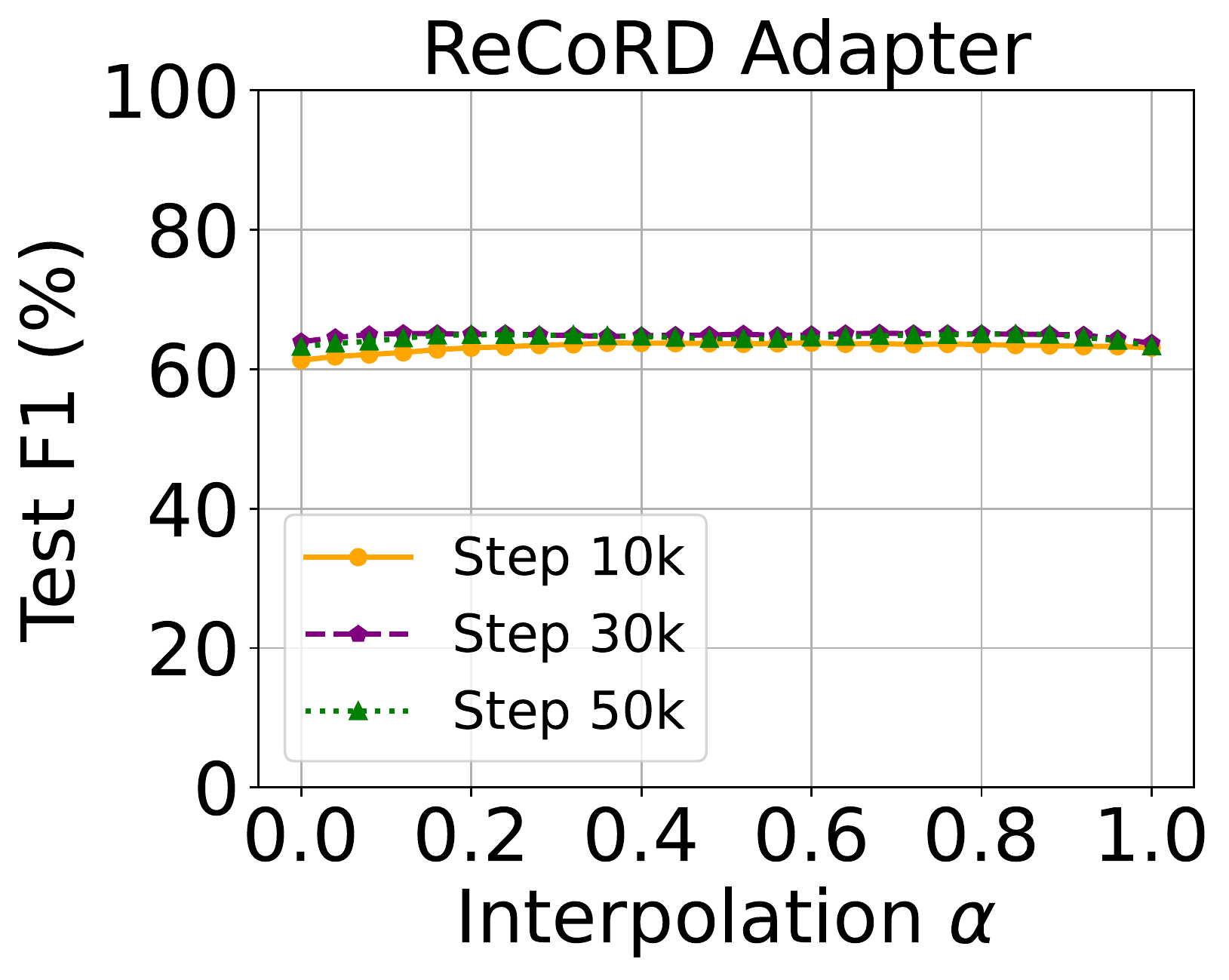}}
    \subfigure[]{\includegraphics[width=0.24\textwidth]{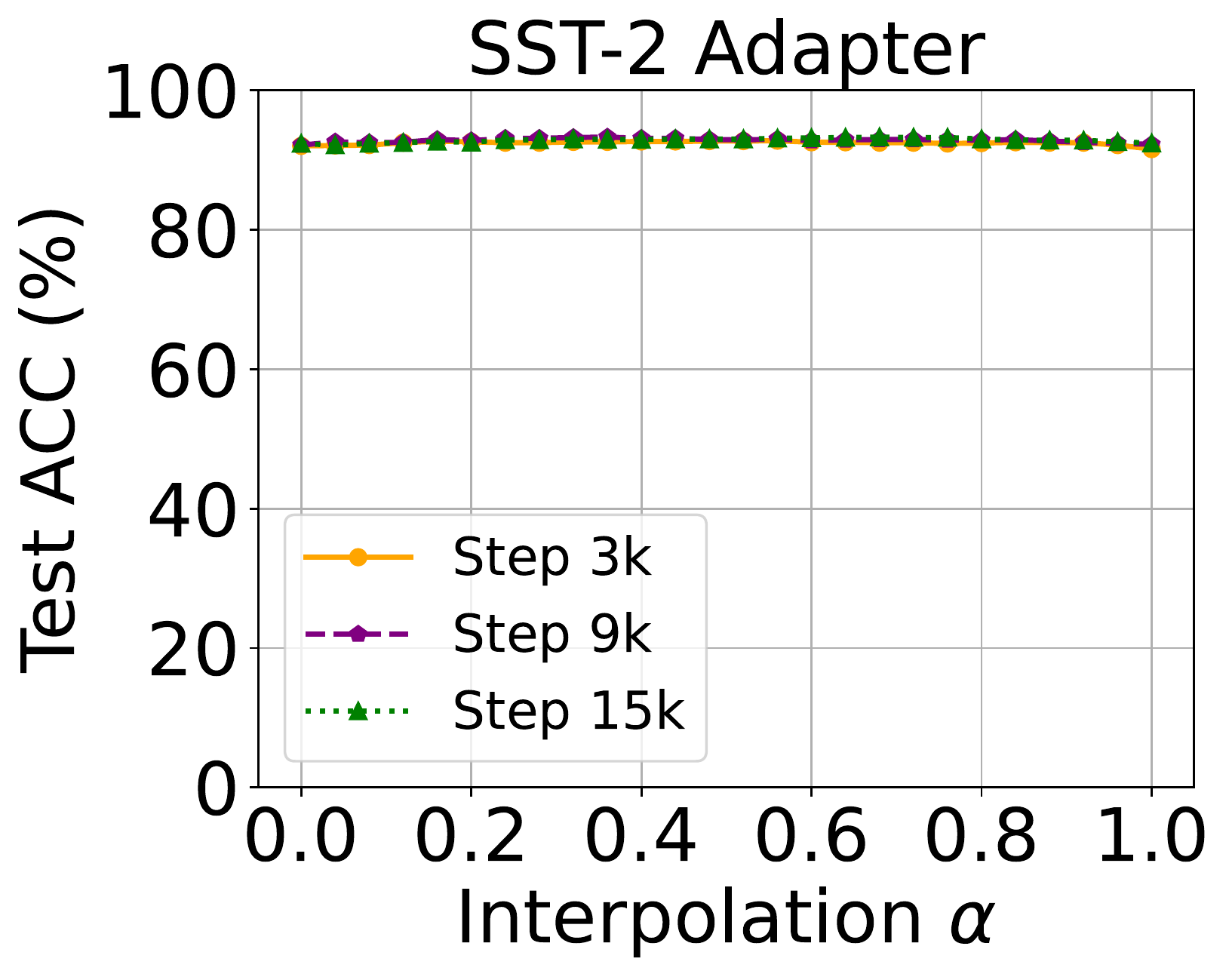}}
    \caption{The performance of interpolations along a non-linear path connecting two minima trained with adapter tuning. For (a-c), two minima are trained with different training data order. For (d-f), two minima are trained with in-distribution data of the same task.}
    \label{fig:curved_more_1}
\end{figure*}

\begin{figure*}[!t]
    \centering
    \subfigure[]{\includegraphics[width=0.24\textwidth]{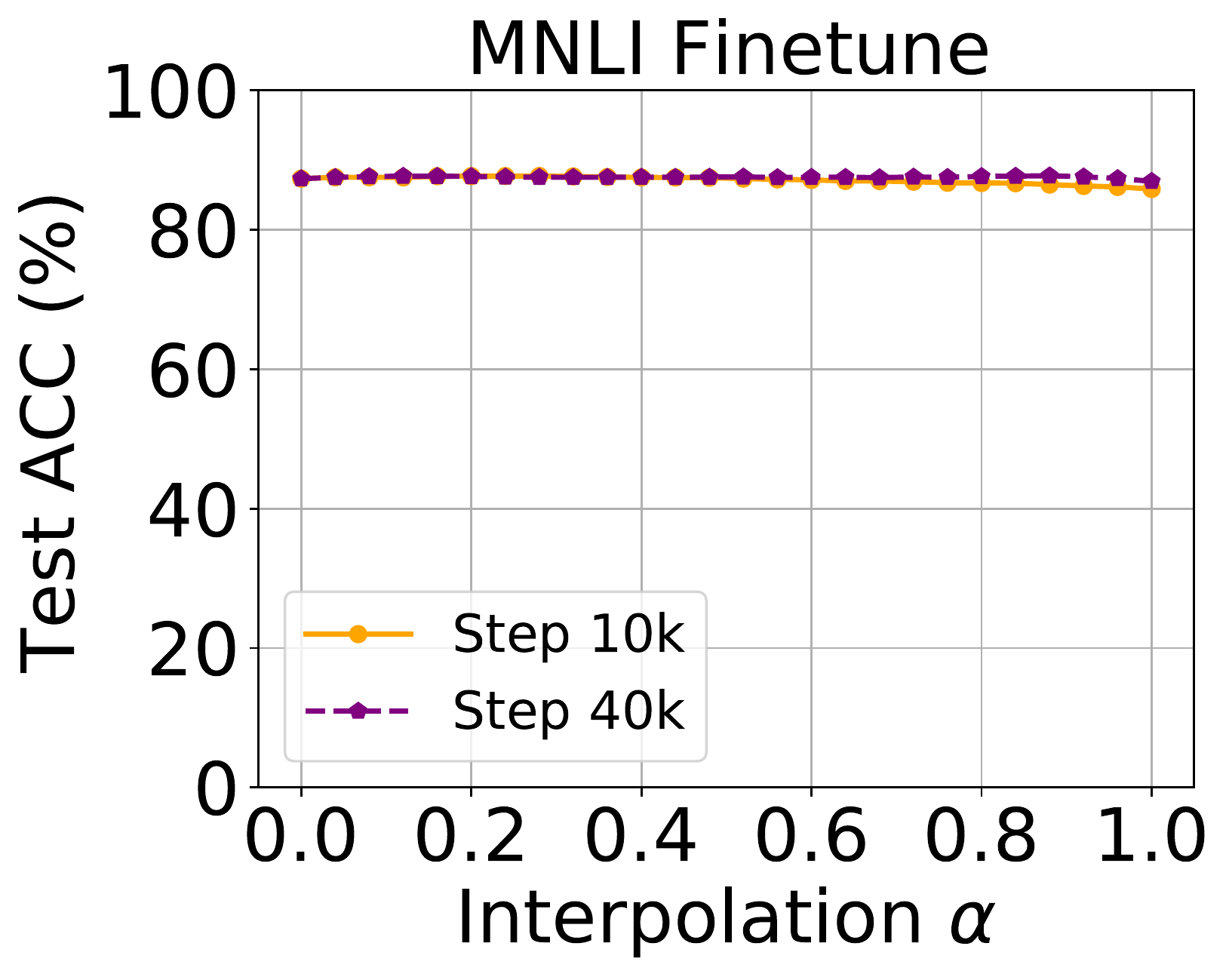}}
    \subfigure[]{\includegraphics[width=0.24\textwidth]{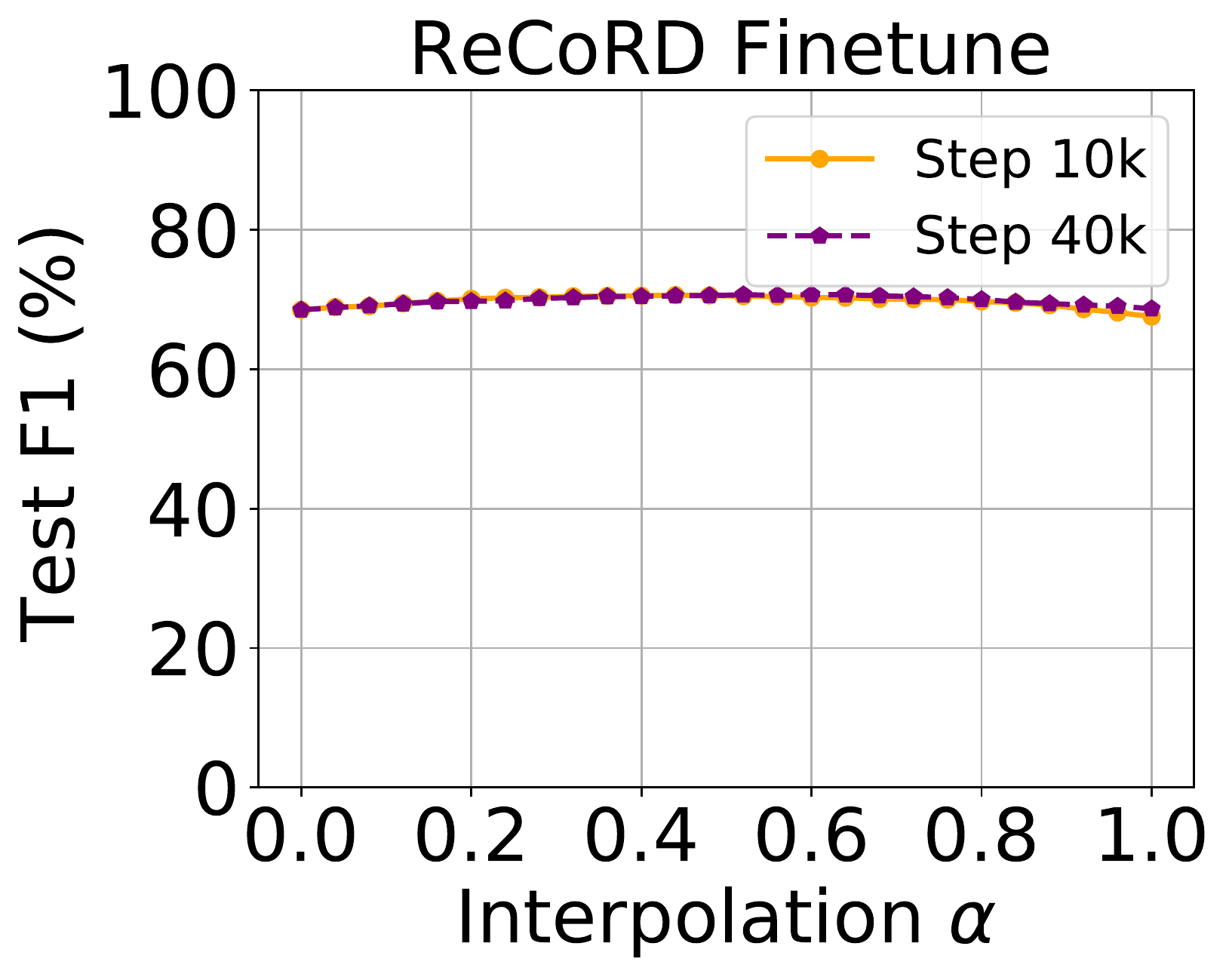}}
    \subfigure[]{\includegraphics[width=0.24\textwidth]{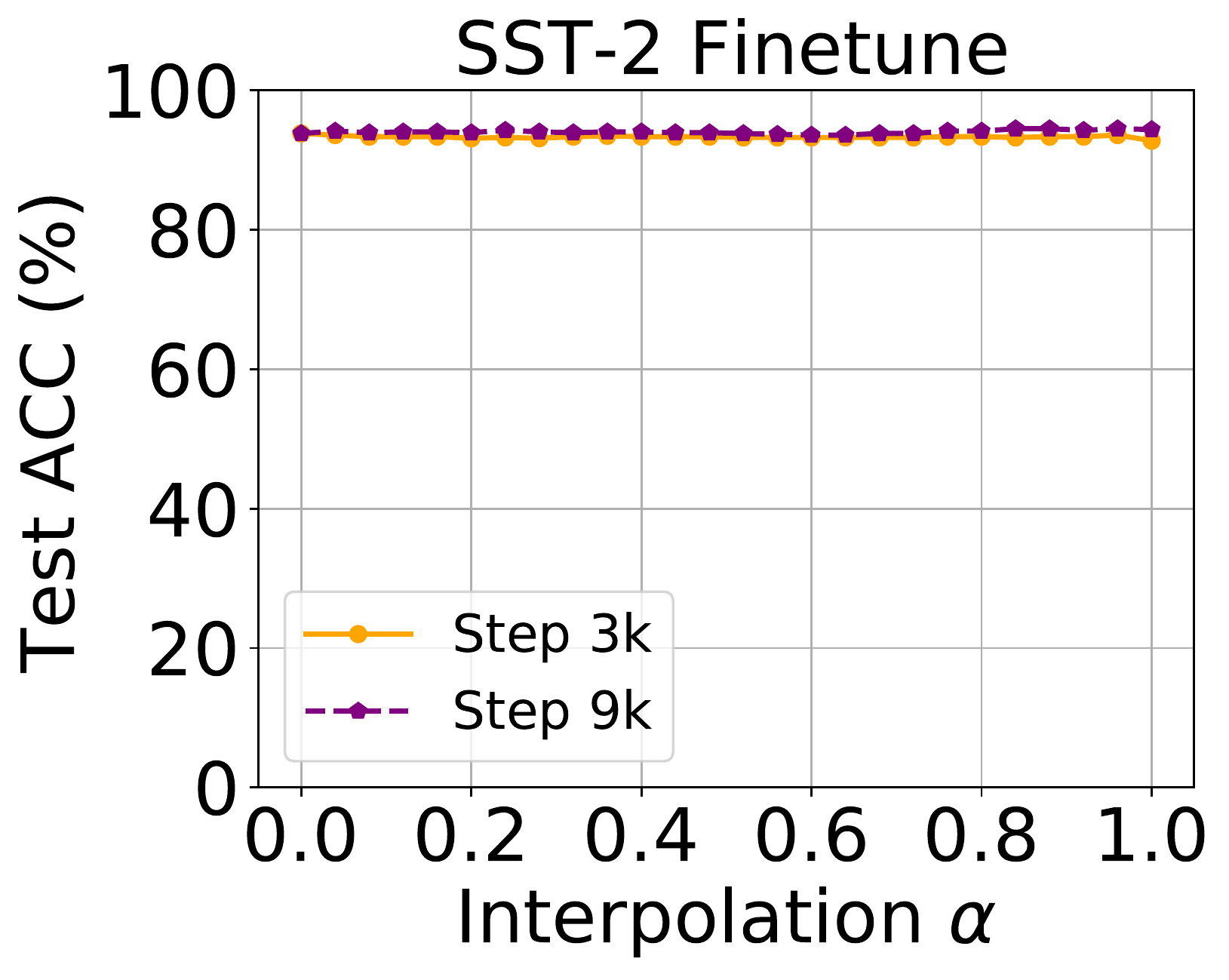}}
    
    \subfigure[]{\includegraphics[width=0.24\textwidth]{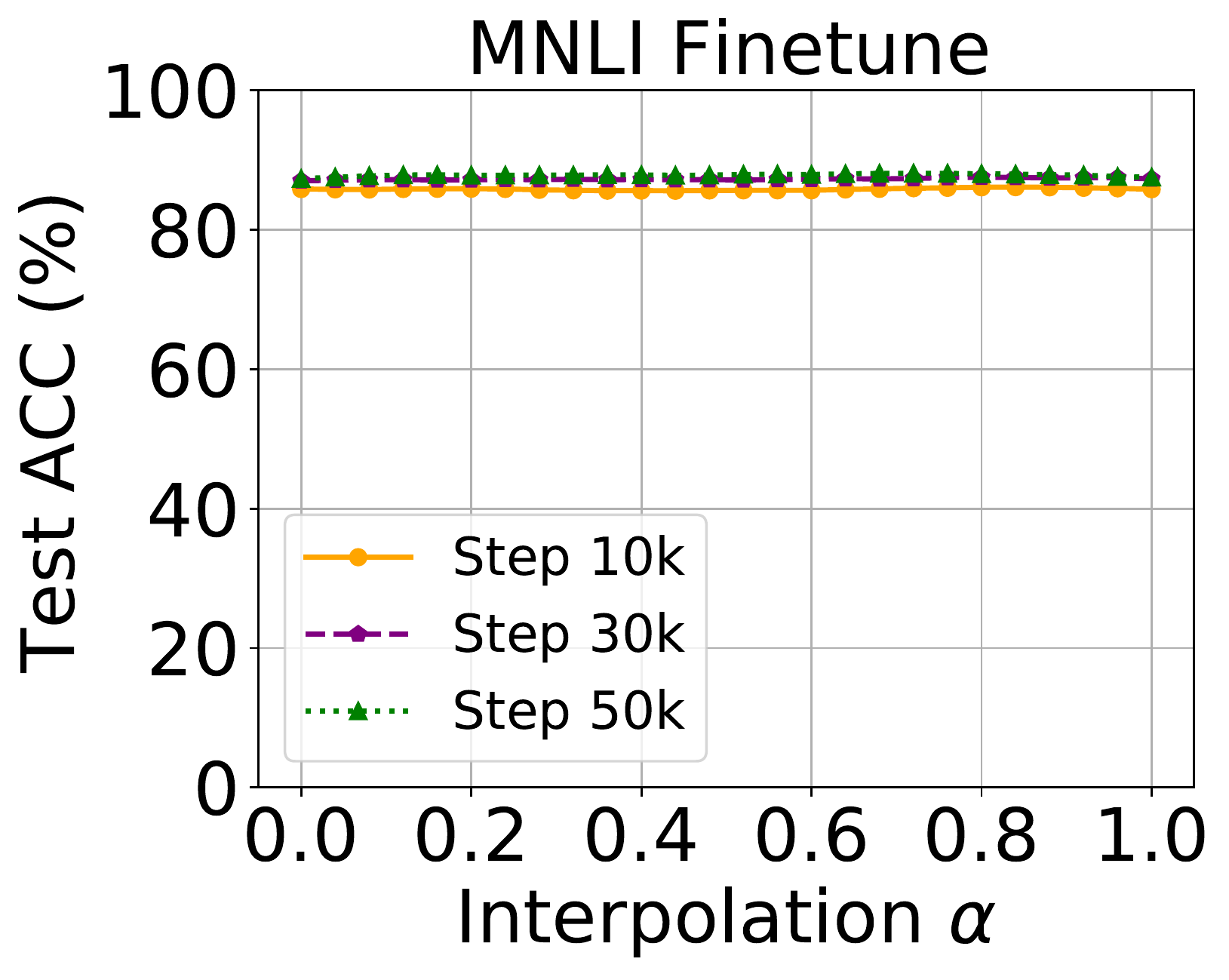}}
    \subfigure[]{\includegraphics[width=0.24\textwidth]{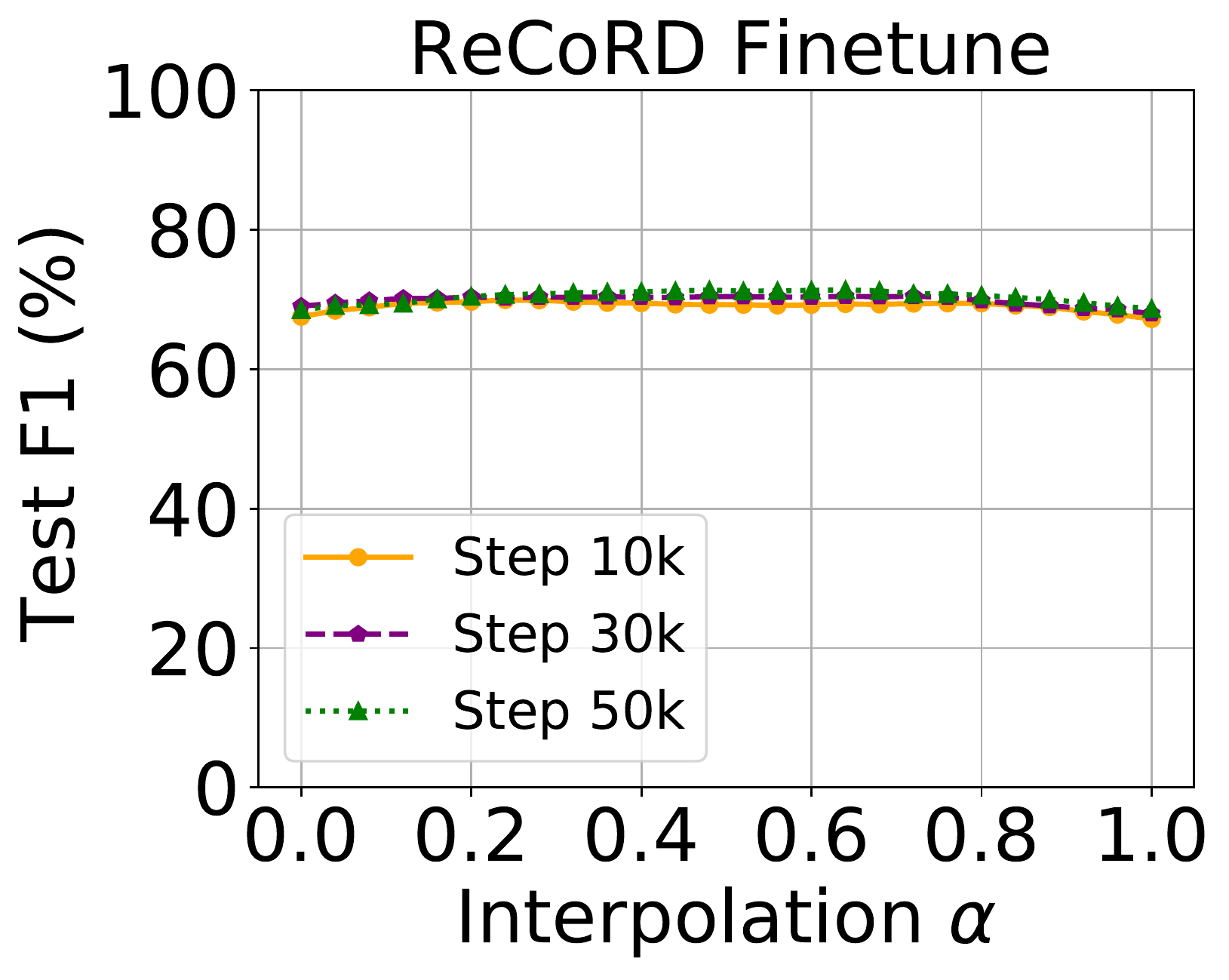}}
    \subfigure[]{\includegraphics[width=0.24\textwidth]{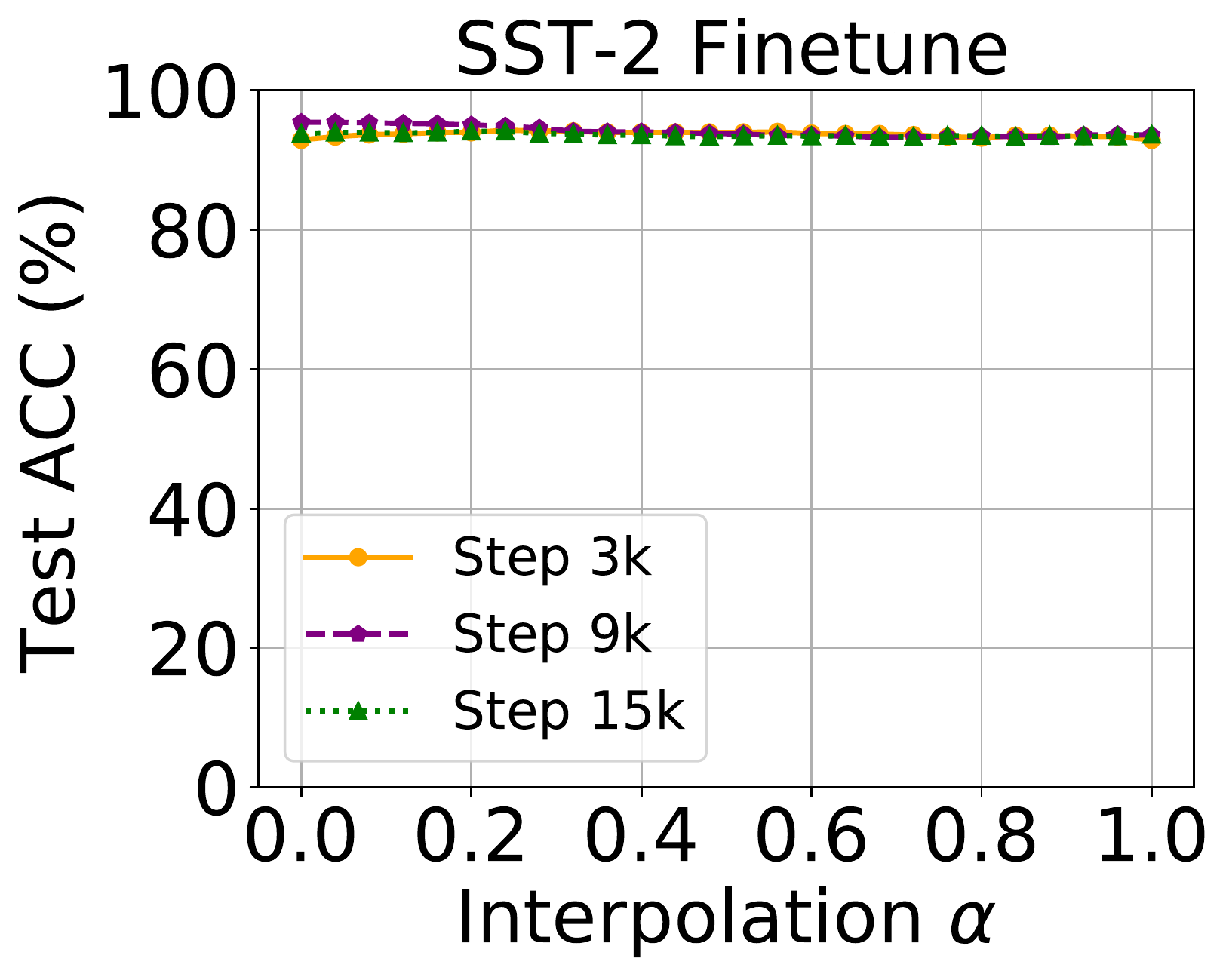}}
    \caption{The performance of interpolations along a non-linear path connecting two minima trained with fine-tuning. For (a-c), two minima are trained with different training steps. For (d-f), two minima are trained with different initialization.}
    \label{fig:curved_more_2}
\end{figure*}

The performance of the interpolations are visualized in Figure~\ref{fig:curved_more_1} for adapter tuning and Figure~\ref{fig:curved_more_2} for fine-tuning. We observe that under all the settings, we do not observe a significant performance drop along the non-linear curve, showing that the connectivity is good. The above results demonstrate that PLMs may have multiple loss basins which are connected via a low-loss non-linear path. In this paper, following \citet{neyshabur2020being}, we spend most of the efforts on convex hull and linear interpolation to avoid possibly trivial connectivity results.

\subsection{Additional Experiments for the Effects of Data Overlap}
\label{sec:additional_overlap}
\begin{figure*}[!t]
    \centering
    \subfigure[]{\includegraphics[width=0.24\textwidth]{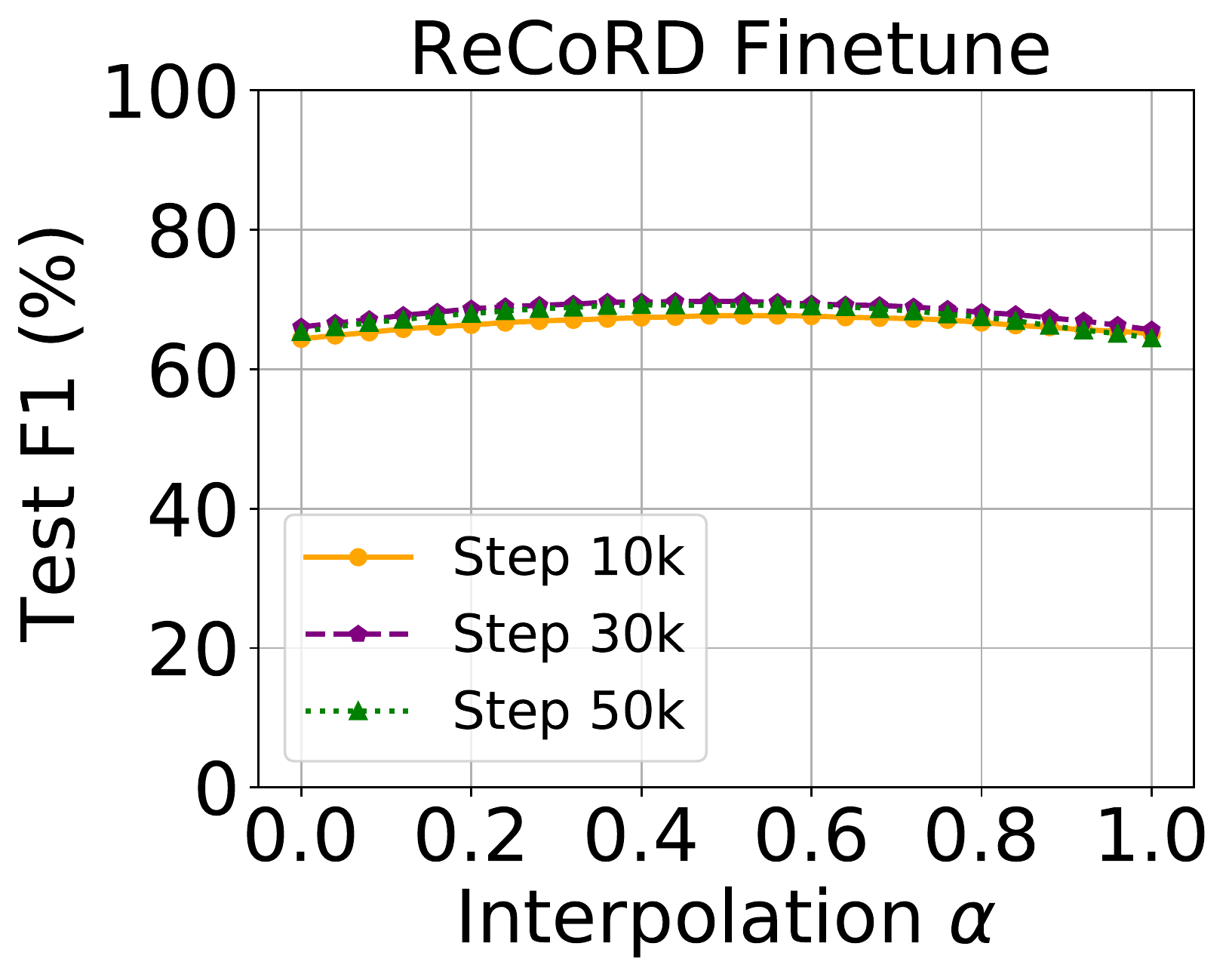}}
    \subfigure[]{\includegraphics[width=0.24\textwidth]{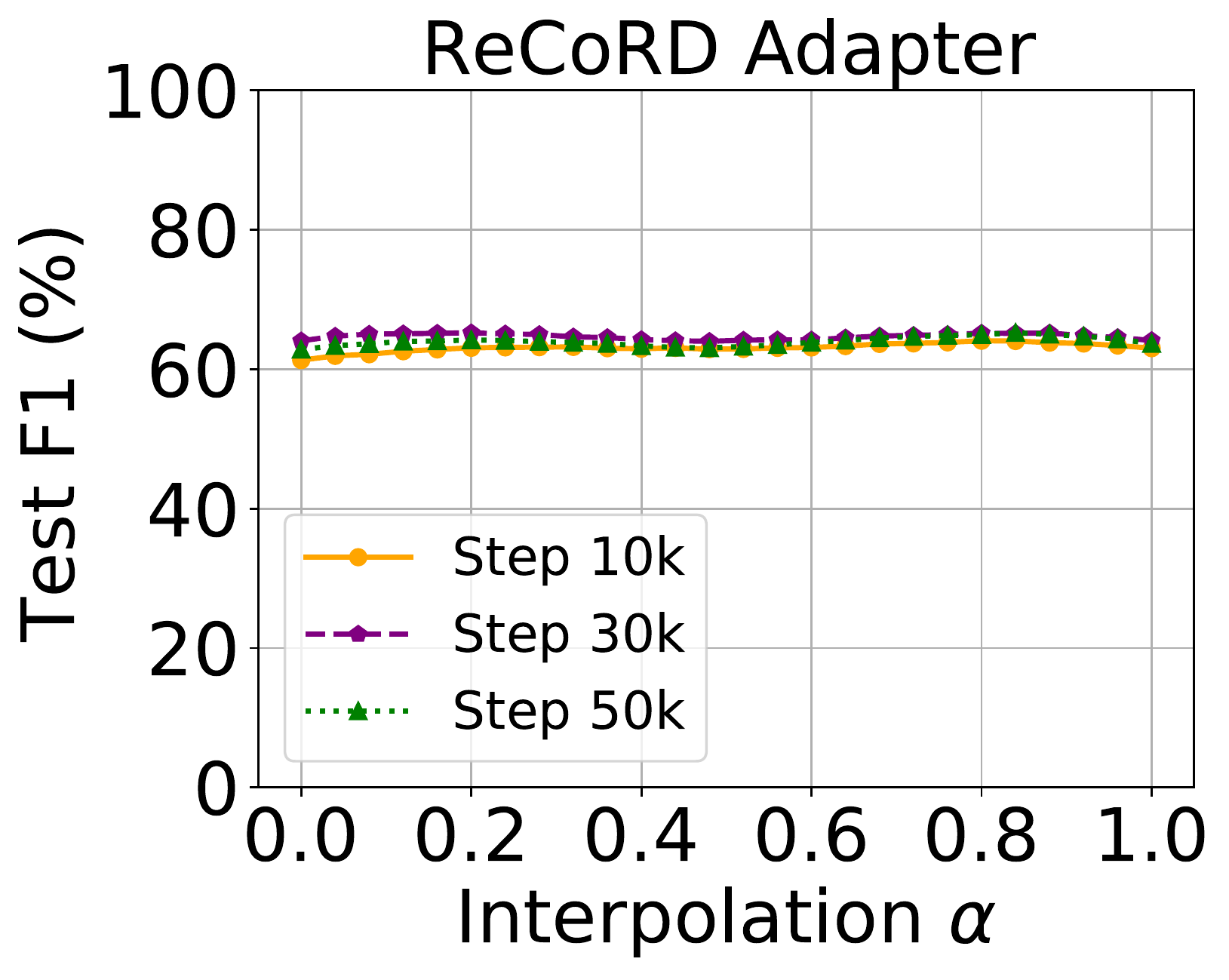}}
    \subfigure[]{\includegraphics[width=0.24\textwidth]{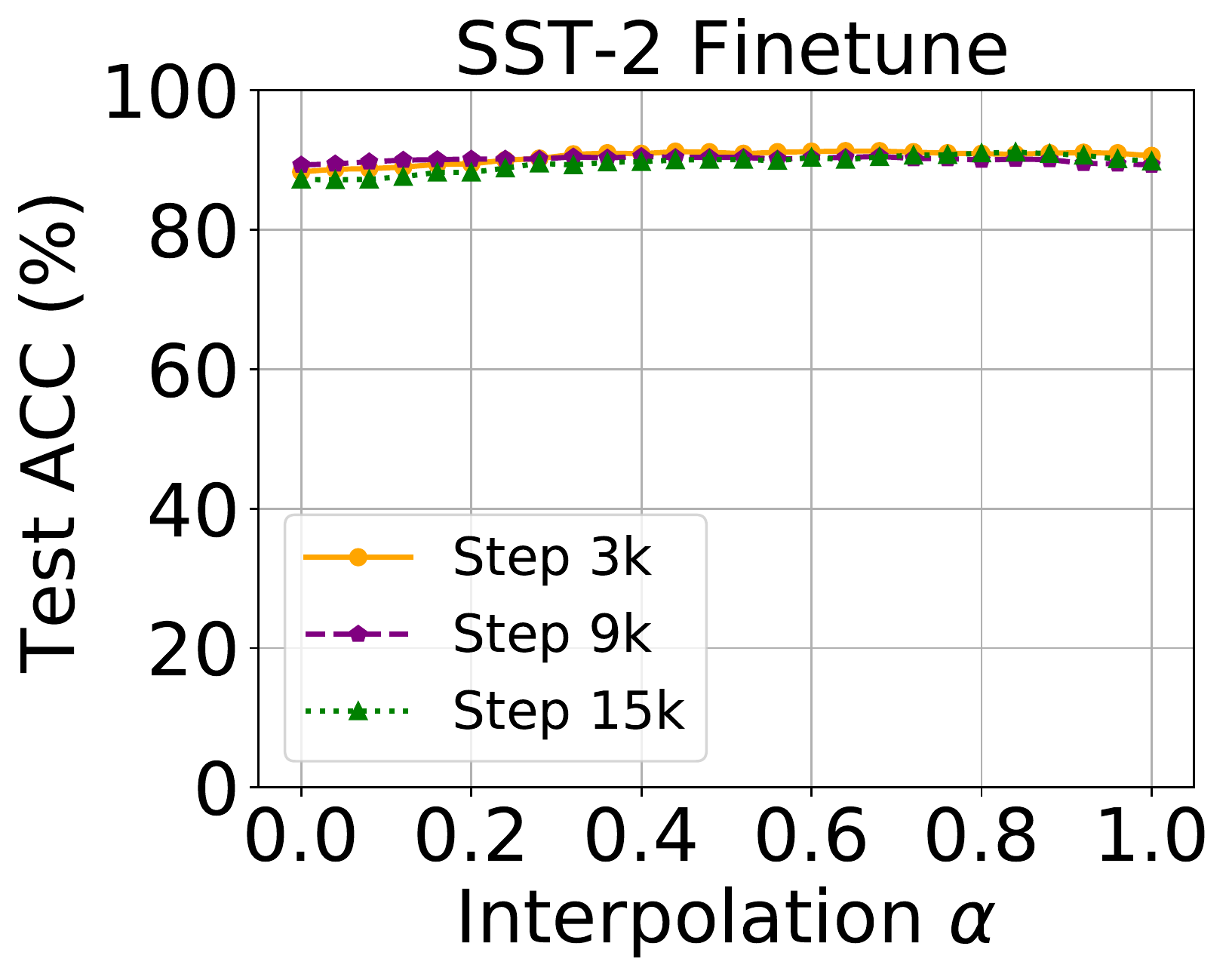}}
    \subfigure[]{\includegraphics[width=0.24\textwidth]{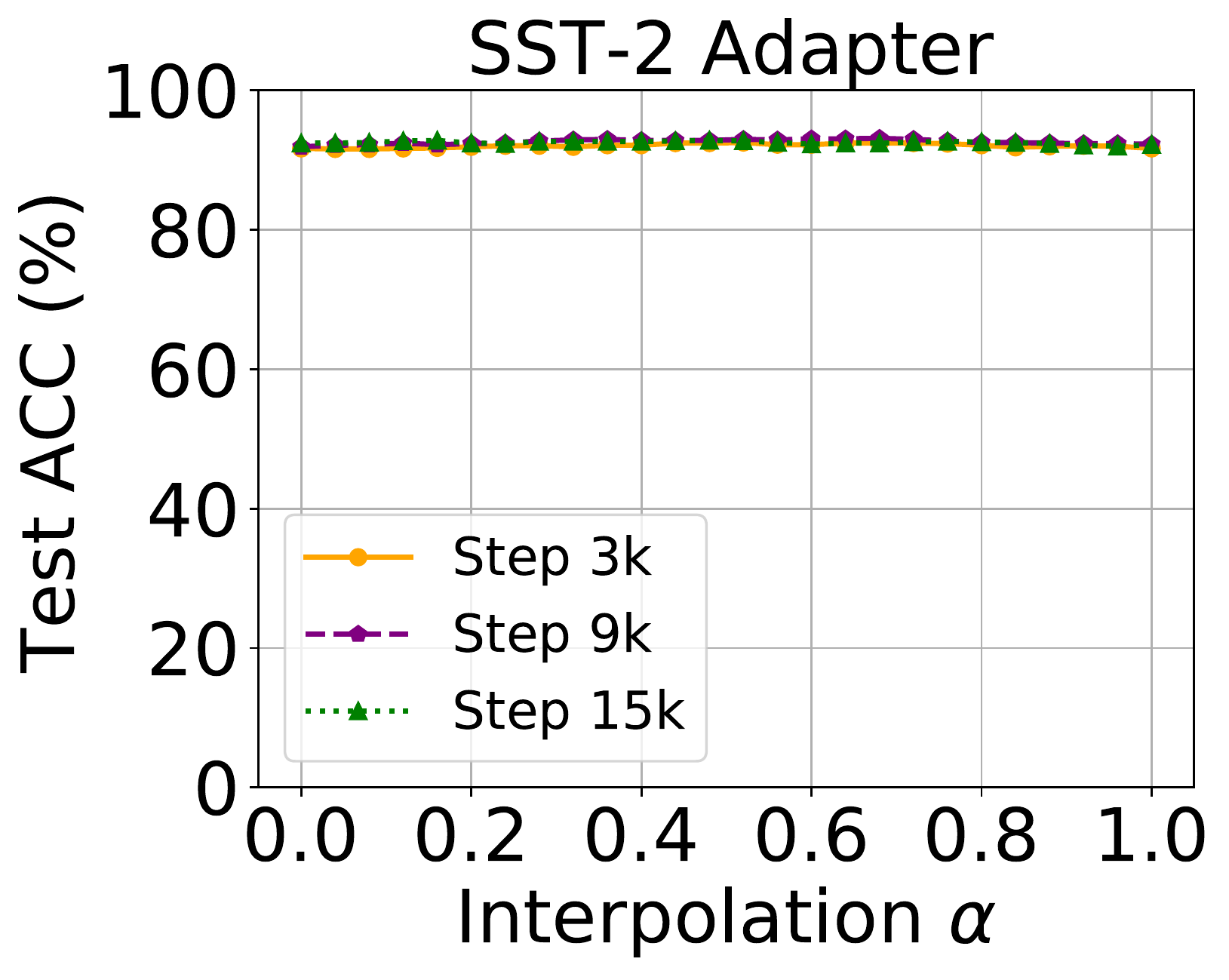}}
    \caption{Linear mode connectivity analysis for two minima trained with in-distribution data. In (a-b), we experiment with ReCoRD. In (c-d), we experiment with SST-2.}
    \label{fig:overlap_2}
\end{figure*}

In the main paper, when evaluating the effects of data overlap, we present the results on MNLI. In this section, we visualize the results when using ReCoRD and SST-2 in Figure~\ref{fig:overlap_2}. Other settings are kept the same as the main paper.

\subsection{Additional Experiments for the Effects of Data Domain}
\label{sec:additional_ood}
\begin{figure}[!t]
    \centering
    \includegraphics[width=0.235\textwidth]{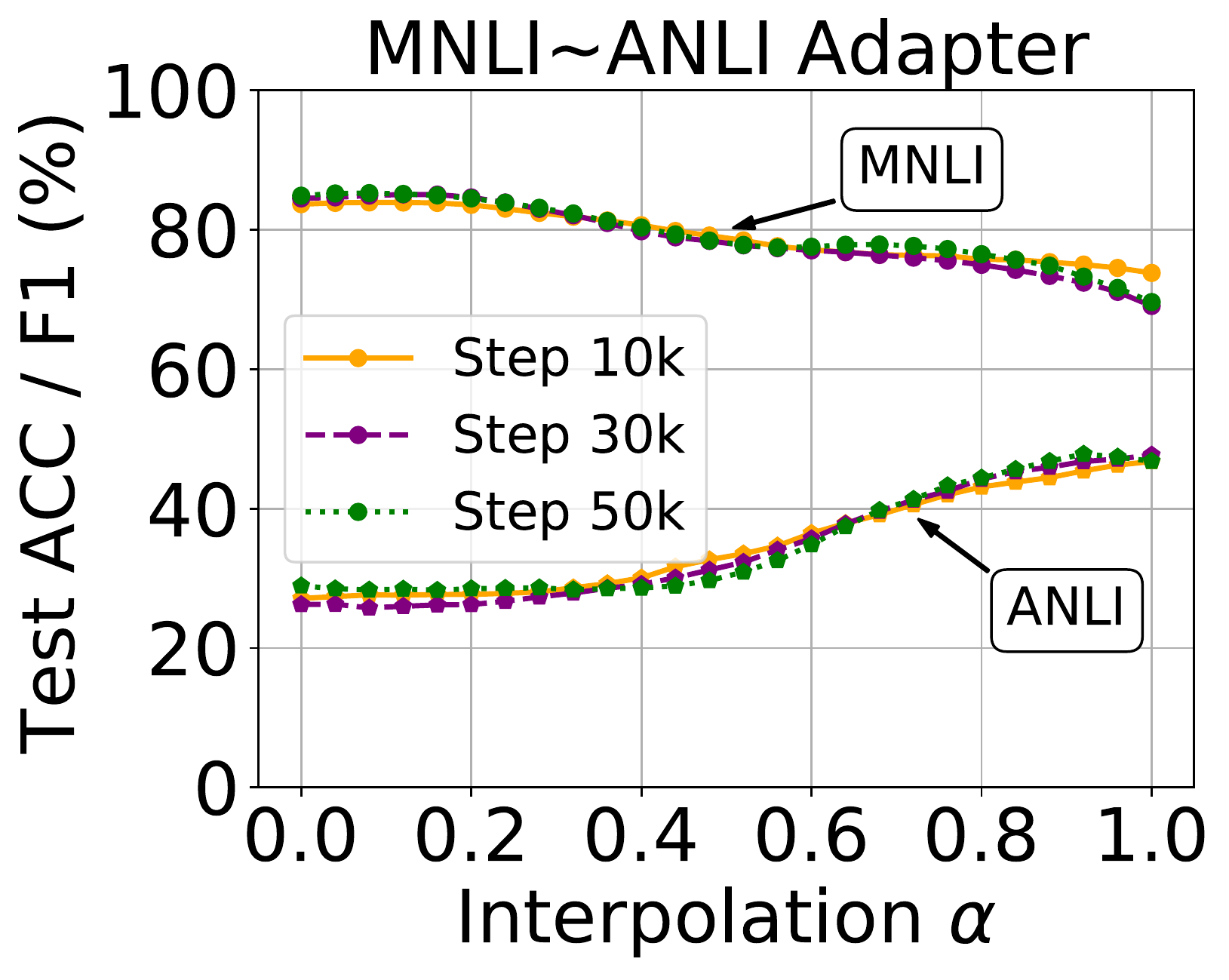} 
    \subfigure{\includegraphics[width=0.235\textwidth]{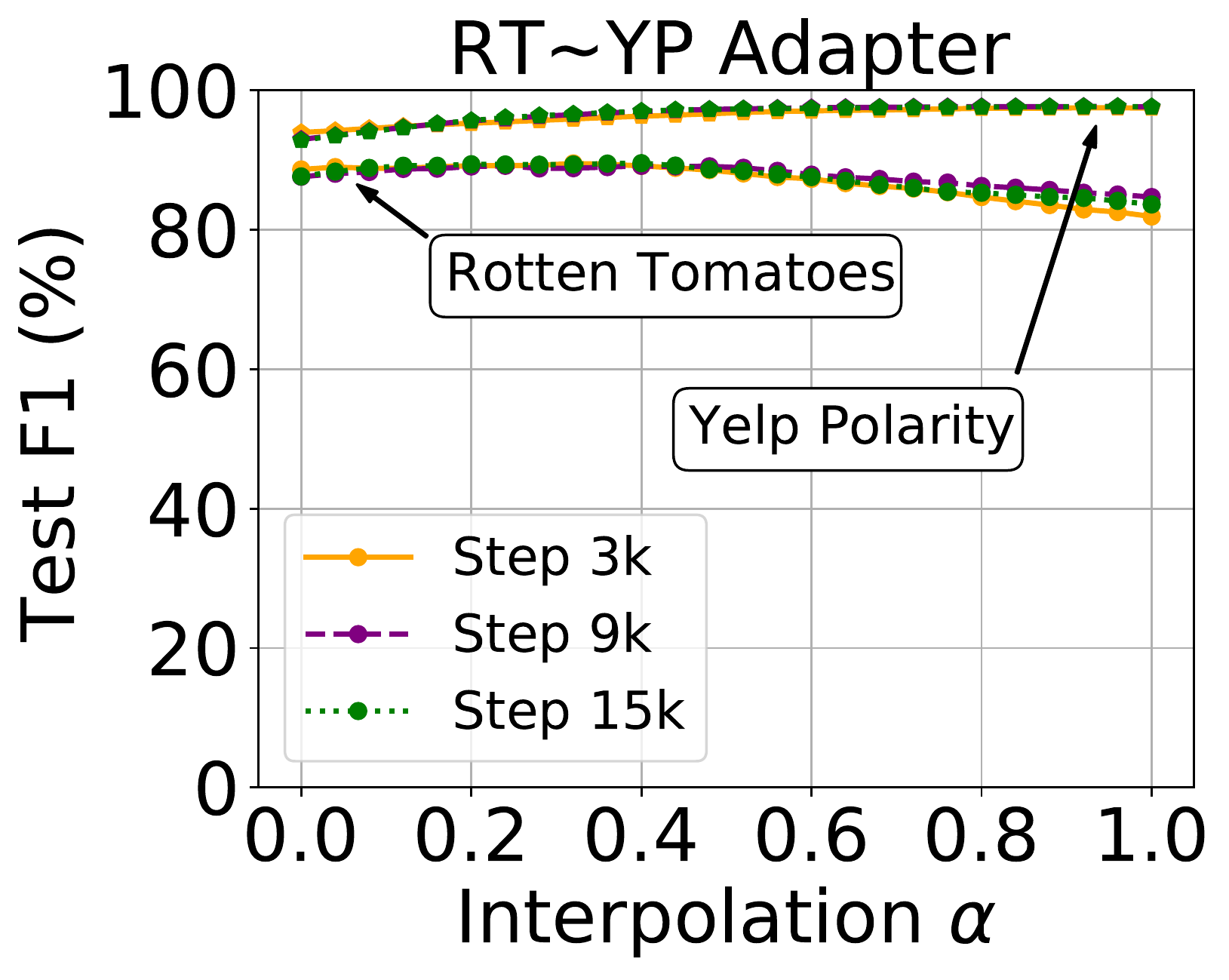}}
    \caption{Linear mode connectivity for two minima trained on different data distributions of the same task using adapter tuning. Left: $\alpha = 0$ / $\alpha = 1$ denotes the minimum of MNLI / ANLI. Right: $\alpha = 0$ / $\alpha = 1$ denotes the minimum of Rotten Tomatoes / Yelp Polarity.}
    \label{fig:domain_2}
\end{figure}

In the main paper, when evaluating the effects of data distributions (data domain), we present the results when using fine-tuning. In this section, we visualize the results when using adapter tuning in Figure~\ref{fig:domain_2}. Other settings are kept the same as the main paper.

\subsection{Additional Experiments for the Change of Mode Connectivity during Pre-training}
\label{sec:additional_diff_task}

\begin{figure}[!t]
    \centering
    \subfigure{\includegraphics[width=0.235\textwidth]{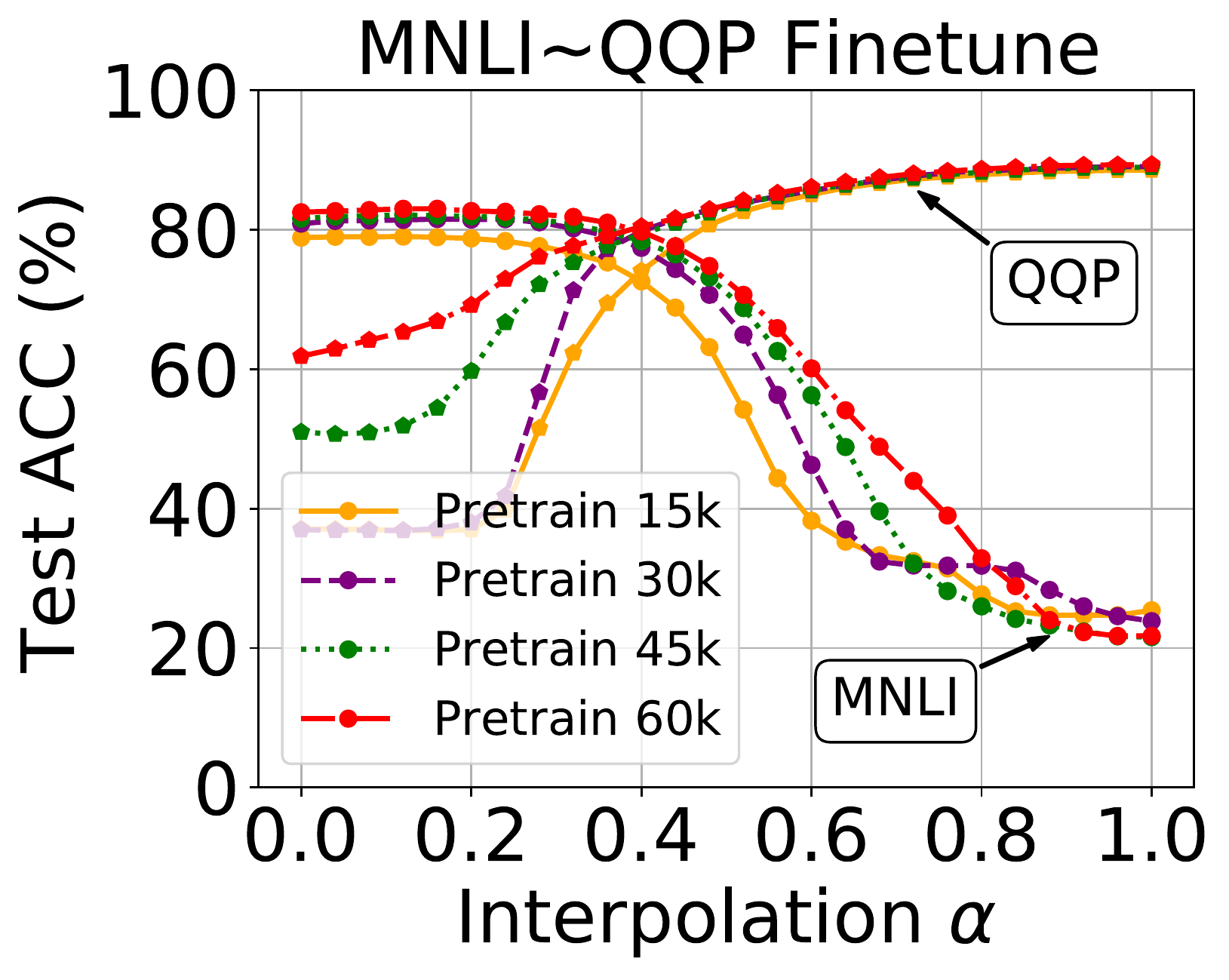}} 
    \subfigure{\includegraphics[width=0.235\textwidth]{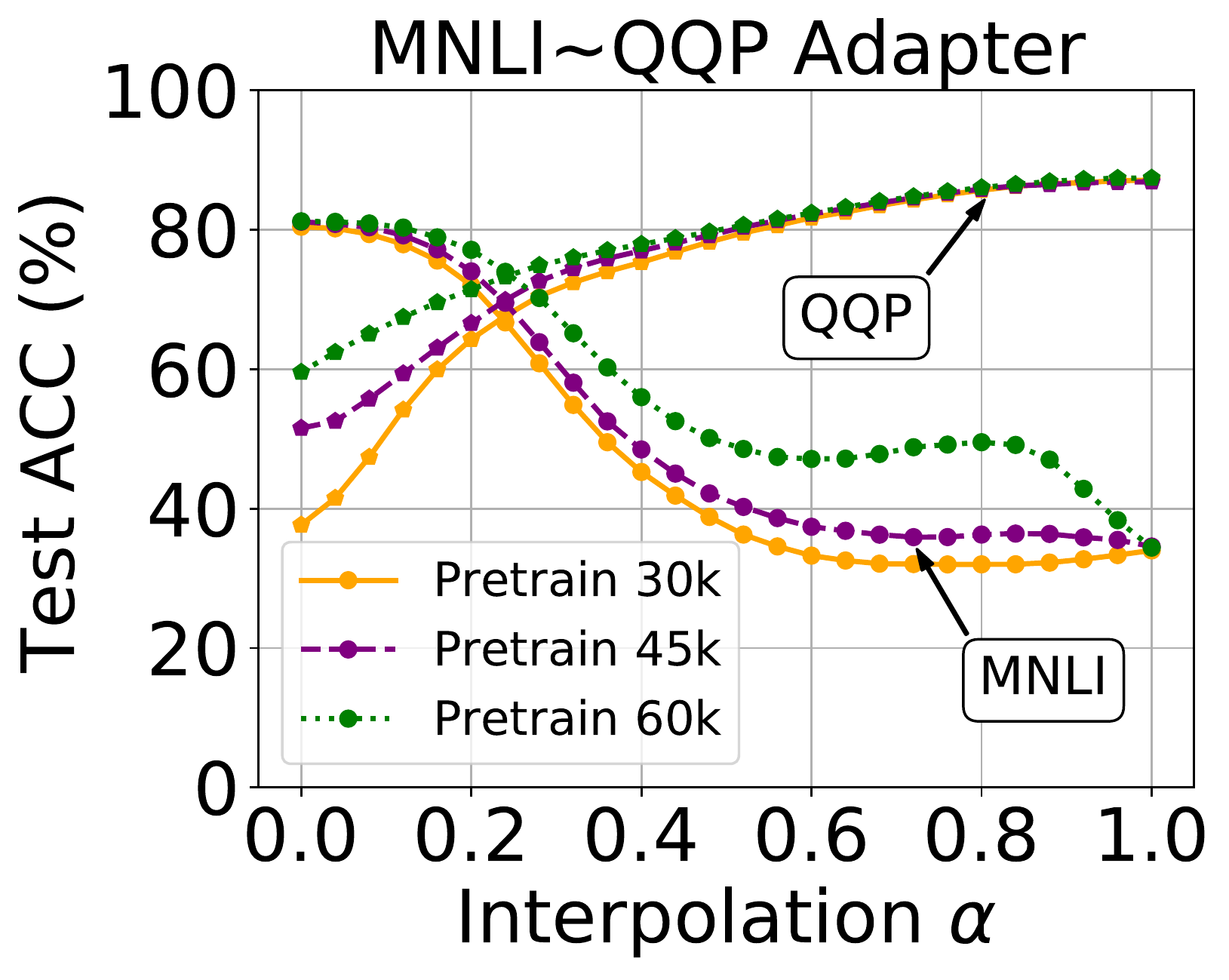}}
    \caption{The change of linear mode connectivity at different pre-training steps. We illustrate the performance of linear interpolations of two minima trained on MNLI and QQP. $\alpha = 0$ / $\alpha = 1$ denotes the minimum of MNLI / QQP.}
    \label{fig:diff_task_2}
\end{figure}

In the main paper, when evaluating the performance variation between two minima trained on two different tasks, we report the results of MNLI and SST-2. In this section, we present the results of MNLI and QQP in Figure~\ref{fig:diff_task_2}. In fact, in our pilot studies, we find that the conclusions in diverse tasks are very consistent. Due to the concern about the energy cost, we only report the performance of two pairs of tasks.

\subsection{Performance along the Connecting Path}

We show that better performance could be achieved by interpolating two independently trained weights in the parameter space. Specifically, we choose the scenario where two copies of PLMs are trained with different training data order. As mentioned in Q1. (a) in the main paper, PLMs have excellent mode connectivity under this setting. We experiment with $\text{T5}_{\texttt{BASE}}$ using fine-tuning and adapter tuning on MNLI, and conduct both linear interpolation and curved interpolation. We evaluate the performance of $24$ evenly distributed points on the curve on a development set, select the best-performing one and evaluate its performance on the test set. We also compare the interpolation with the endpoints (we report the best performance of the two endpoints). All experiments are conducted $3$ times and we report the average test results in Table~\ref{tab:interpolation_results}. We observe that by traversing along the connecting curve between two minima, we could find a solution that performs better than both endpoints. In addition, traversing along a linear path finds an interpolation with higher performance than traversing along a curved path\footnote{Although we have shown that the non-linear mode connectivity is generally good for different minima, it does not mean that the best performance on a non-linear curve is always better than that on a linear curve.}. In general, this finding demonstrates that it is promising to combine the knowledge of multiple models through weight averaging.

\subsection{Other Strategies for Weight Ensemble}
\label{sec:weight_ensemble}
As demonstrated in the main paper, the connectivity on a non-linear path may be better than a linear path under certain cases. This phenomenon demonstrates that despite the simplicity of linear interpolation, there may exist better ways to interpolate two independently trained minima. To demonstrate the existence of a better combination for two minima, we take an initial step and propose to optimize such a combination. Specifically, suppose all the tunable parameters of a PLM can be divided into $M$ components, i.e., $\theta_{C_1} = \{\theta_{C_1}^i\}_{i=1}^M$, $\theta_{C_2} = \{\theta_{C_2}^i\}_{i=1}^M$. We optimize a tunable vector $\bm{\alpha} \in \mathbb{R}^M$ to combine $\theta_{C_1}$ and $\theta_{C_2}$ as follows:
\begin{equation*}
\small
\begin{aligned}
    \theta (\bm{\alpha}) = \{\sigma(\bm{\alpha}_i) \cdot \theta_{C_1}^i + (1 - \sigma(\bm{\alpha}_i)) \cdot \theta_{C_2}^i\}_{i=1}^M,
\end{aligned}
\end{equation*}
where $\sigma$ denotes a \textit{sigmoid} function. During training, both $\theta_{C_1}$ and $\theta_{C_2}$ are kept frozen, and only $\bm{\alpha}$ is tuned. We design three intuitive strategies for parameter division: (1) layer-wise division, where the parameters within the same layer share the same combination ratio; (2) module-wise division, where we discriminate the combination ratio of the feed-forward network module and multi-head attention module in each layer. This means each module in each layer is assigned with an individual combination ratio; (3) matrix-wise division, where each weight matrix in each module is combined individually. Matrix-wise division is the most fine-grained one among the above three strategies. Since we use the $\text{T5}_\texttt{BASE}$ model, which consists of $12$ encoders and $12$ decoders, the number of components $M$ for block-wise, layer-wise, and matrix-wise divisions are $24$, $60$, and $120$ for adapter; and $28$, $64$, and $282$ for fine-tuning. During training, we perform grid search on a series of learning rates $\{0.1, 0.05, 0.01\}$ and set the batch size to $8$, and max training steps to $100$k.

Two endpoints are obtained by fine-tuning $\text{T5}_\texttt{BASE}$ on MNLI using different training data order. The results are shown in Table~\ref{tab:better_interpolation}, from which we find that, among the proposed three combination strategies, matrix-wise division achieves the best performance. The performance is also better than using linear interpolation. This phenomenon demonstrates that there exist better ways for combining two minima's knowledge than linear interpolation. We hope our findings on mode connectivity in this paper could inspire future works to design better weight ensemble methods.

\begin{table}[!t]
  \centering
  \small
    \begin{tabular}{cl|rrr}
    \toprule
    \multicolumn{1}{c}{\textbf{Interpolation}} &
    \multicolumn{1}{c}{\textbf{Step}} &
    \multicolumn{1}{l}{\textbf{Linear}} & \multicolumn{1}{l}{\textbf{Curved}} & \multicolumn{1}{l}{\textbf{Endpoint}} \\
    \midrule
    \multirow{3}[2]{*}{\textbf{Adapter}} & $10$k    & $\textbf{84.58}$ & $84.14$  & $84.37$  \\
          & $30$k    & $\textbf{85.27}$ & $85.14$  & $84.96$  \\
          & $50$k    & $\textbf{85.70}$ & $85.23$  & $85.55$  \\
    \midrule
    \multirow{3}[2]{*}{\textbf{Fine-tuning}} & $10$k    & $\textbf{85.97}$ & $85.77$  & $85.85$  \\
          & $30$k    & $\textbf{87.36}$ & $87.27$  & $87.29$  \\
          & $50$k    & $87.88$  & $\textbf{87.88}$ & $87.52$  \\
    \bottomrule
    \end{tabular}%
  \caption{Test performance for interpolations of two minima trained on MNLI using different training data order. For both linear interpolation and curved interpolation, we choose the best checkpoint based on the development set performance. For two endpoints, we report the endpoint that performs better on the test set.}
  \label{tab:interpolation_results}%
\end{table}%

\begin{table}[!t]
  \centering
  \small
    \begin{tabular}{c@{~~~}|c@{~~~}c@{~~~}c@{~~~}c@{~~~}c}
    \toprule
    \textbf{Step} &
    \textbf{Linear} & \textbf{Layer} & \textbf{Module} & \textbf{Matrix} & \textbf{Endpoint} \\
    \midrule
     $10$k    & $86.0$ & $86.2$ & \textbf{86.3} &  $86.3$  & $85.9$  \\
          $30$k    & $87.4$   & $87.4$  & $87.4$ & \textbf{87.5} & $87.3$  \\
          $50$k    & $87.9$ & $87.7$ & $87.5$ & \textbf{88.0} & $87.5$  \\
    \bottomrule
    \end{tabular}%
  \caption{Test performance for interpolations of two minima fine-tuned on MNLI using different training data order. For linear interpolation (\textbf{Linear}), we choose the best checkpoint based on the development set performance. For two endpoints (\textbf{Endpoint}), we report the endpoint that performs better on the test set. \textbf{Layer}, \textbf{Module}, and \textbf{Matrix} denote layer-wise, module-wise, and matrix-wise divisions we proposed.}
  \label{tab:better_interpolation}%
\end{table}%

\section{Training Details}
\label{sec:training_detail}
For the $\text{T5}_\texttt{BASE}$ model, we use the checkpoint provided by \citet{lester-etal-2021-power}, who conducted additional $100$k steps of language modeling adaption on the official checkpoints released by \citet{2020t5}. Such adaptation is demonstrated to help stabilize downstream adaptation and improve the performance, especially for delta tuning methods~\citep{lester-etal-2021-power}. We use AdamW~\citep{loshchilov2017decoupled} as the optimizer for all the experimented PLMs. All the implementation codes, trained checkpoints and used datasets would be released after publication.

We download all the experimented datasets from \textit{Huggingface Datasets}~\citep{lhoest2021datasets}. Since some datasets do not contain a test set, we first merge all the data points, and then split them into the new training split, development split, and test split with an approximate ratio of $8:1:1$. The above procedure is conducted on all the experimented datasets.

For different tasks fine-tuned on $\text{T5}_\texttt{BASE}$, we first conduct grid search to find an optimal hyperparameter combination. Specifically, the chosen hyperparameter of different tasks for fine-tuning is shown in Table~\ref{tab:hyperparameter_fine-tune}; for adapter tuning, in our prior experiments, we find that a learning rate of $5\times10^{-4}$ and a batch size of $16$ performs good on all tasks, thus we set them as the default configuration. For both tuning methods, we save $5$ checkpoints during training, with different saving intervals for different tasks as shown in Table~\ref{tab:hyperparameter_fine-tune}.

\begin{table}
  \centering
  \small
    \begin{tabular}{c@{~~~~~}c@{~~~~~}c@{~~~~~}c}
    \toprule
    \textbf{Task} & \textbf{LR} & \textbf{BS} & \textbf{SI} \\
    \midrule
    MNLI & $5\times10^{-5}$ & $32$ & $10$k \\
    ReCoRD & $1\times10^{-4}$ & $32$ & $10$k \\
    ANLI & $1\times10^{-4}$ & $32$ & $10$k \\
    SST-2 & $5\times10^{-5}$ & $8$ & $3$k \\
    Rotten Tomatoes & $5\times10^{-5}$ & $32$ & $3$k \\
    Yelp Polarity & $5\times10^{-4}$ & $8$ & $3$k \\
    \bottomrule
    \end{tabular}%
  \caption{Hyperparameters (LR: learning rate, BS: batch size, SI: saving interval) for different tasks during the fine-tuning of $\text{T5}_\texttt{BASE}$ model.}
  \label{tab:hyperparameter_fine-tune}%
\end{table}%

\subsection{Additional Details for the Effects of Initialization}
For adapter tuning, all the modules newly introduced are initialized using a Gaussian distribution. As for fine-tuning, we add Gaussian noise to all the tunable parameters. The mean and standard deviation of the Gaussian distribution are set to $0$ and $0.0002$, respectively. We use different random seeds to generate different initialization.

\subsection{Additional Details for Curve Finding}
\label{sec:curve_find_detail}
When optimizing the Bezier curve, we set the learning rate to $1\times10^{-4}$, batch size to $8$, max training steps to $5$k for fine-tuning; and set the learning rate to $1\times10^{-4}$, batch size to $16$, max training steps to $10$k for adapter tuning. During curve finding, we evaluate the development performance of the current curve for every $100$ steps, using a series interpolations with $\alpha \in \{0.25,0.5,0.75\}$.

\subsection{Additional Details for Calculating Confidence and Variability}
\label{sec:ood_detail}

As mentioned in the main paper, we use the training dynamics to characterize each training sample. For MNLI / ANLI, we adapt the model for $8$ epochs / $20$ epochs to calculate both confidence and variability. We tune more epochs for ANLI because the size of its training dataset is far smaller than that of MNLI.
 
\begin{figure}[!t]
    \centering
    \subfigure{\includegraphics[width=0.235\textwidth]{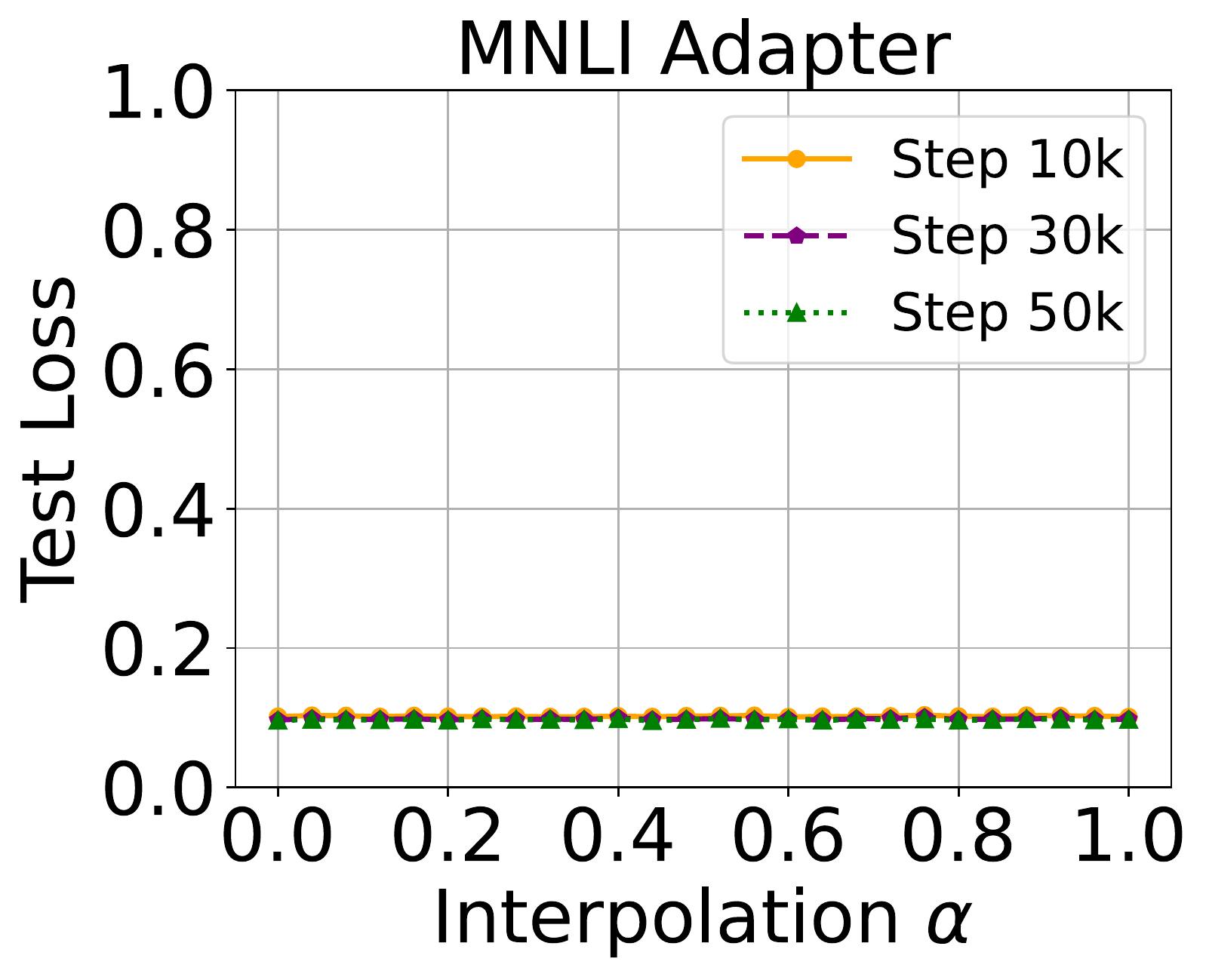}} 
    \subfigure{\includegraphics[width=0.235\textwidth]{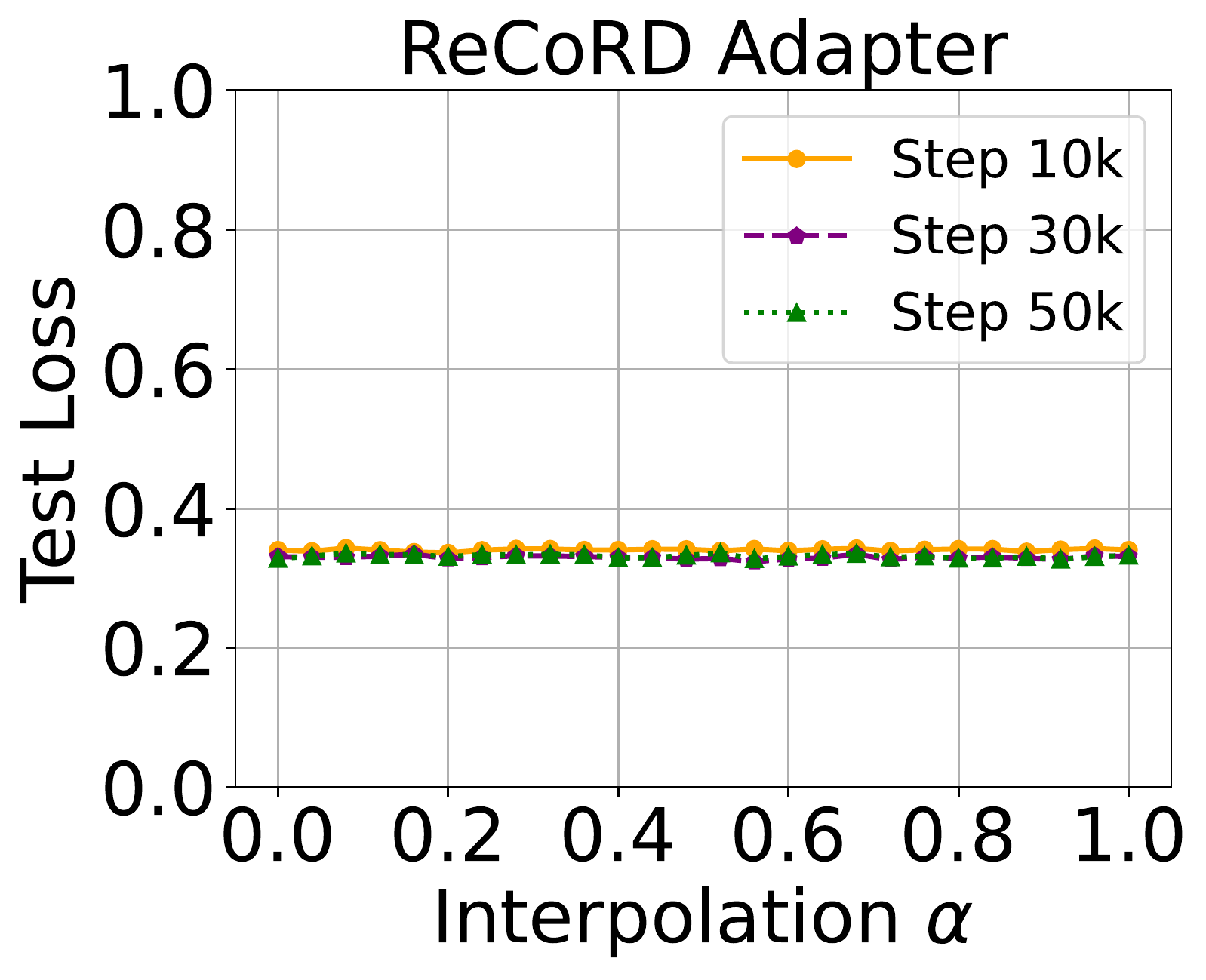}} 
    \caption{The loss of curved interpolations between two minima trained with adapter tuning from different initialization. The corresponding performance visualization is Figure~\ref{fig:curved_main_paper}.}
    \label{fig:curved_main_paper_loss}
\end{figure}
 
\subsection{Additional Details for Pre-training $\text{RoBERTa}_\texttt{BASE}$}
\label{sec:pre-training_roberta}
We closely follow the pre-training setting of \citet{liu2019roberta}, except that for pre-training data, we use the concatenation of Wikipedia and BookCorpus~\cite{zhu2015aligning} same as BERT~\citep{devlin2018bert}, and we pre-train our model with a batch size of $2048$. The pre-training implementations for $\text{RoBERTa}_\texttt{BASE}$ are based on those of \citet{qin-etal-2022-knowledge,qin-etal-2022-elle}. Adam~\citep{loshchilov2017decoupled} is chosen as the optimizer. The hyperparameters for the optimizer is set to $1\times10^{-6}$, $0.9$, $0.98$ for $\epsilon$, $\beta_1$, $\beta_2$, respectively. We set the dropout rate to $0.1$, weight decay to $0.01$ and use linear learning rate decay. The model architecture is the same as the official $\text{RoBERTa}_\texttt{BASE}$ model~\citep{liu2019roberta}. The pre-training is conducted using $8$ NVIDIA V100 GPUs.

\begin{figure}[!t]
    \centering
    \subfigure{\includegraphics[width=0.235\textwidth]{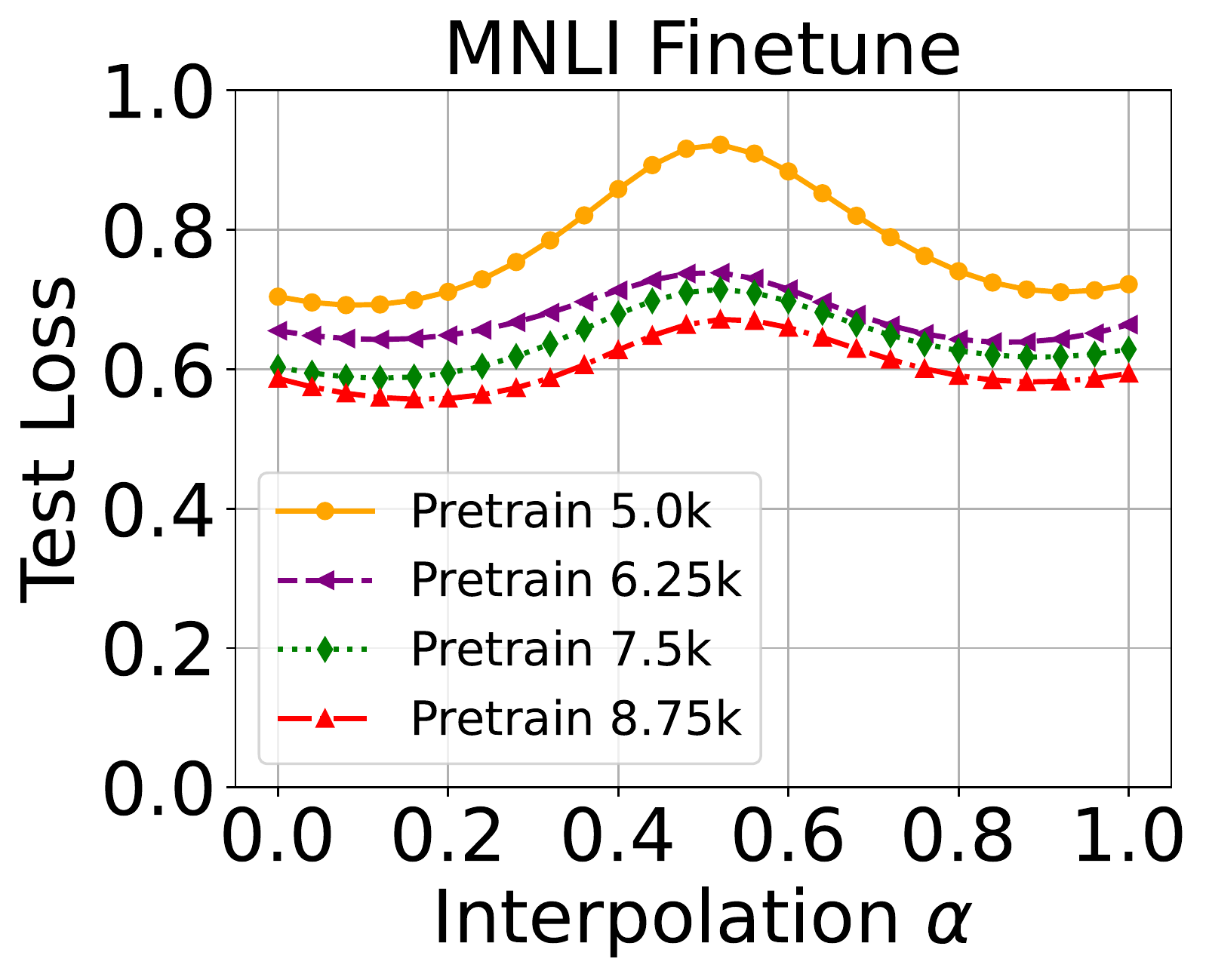}}
    \subfigure{\includegraphics[width=0.235\textwidth]{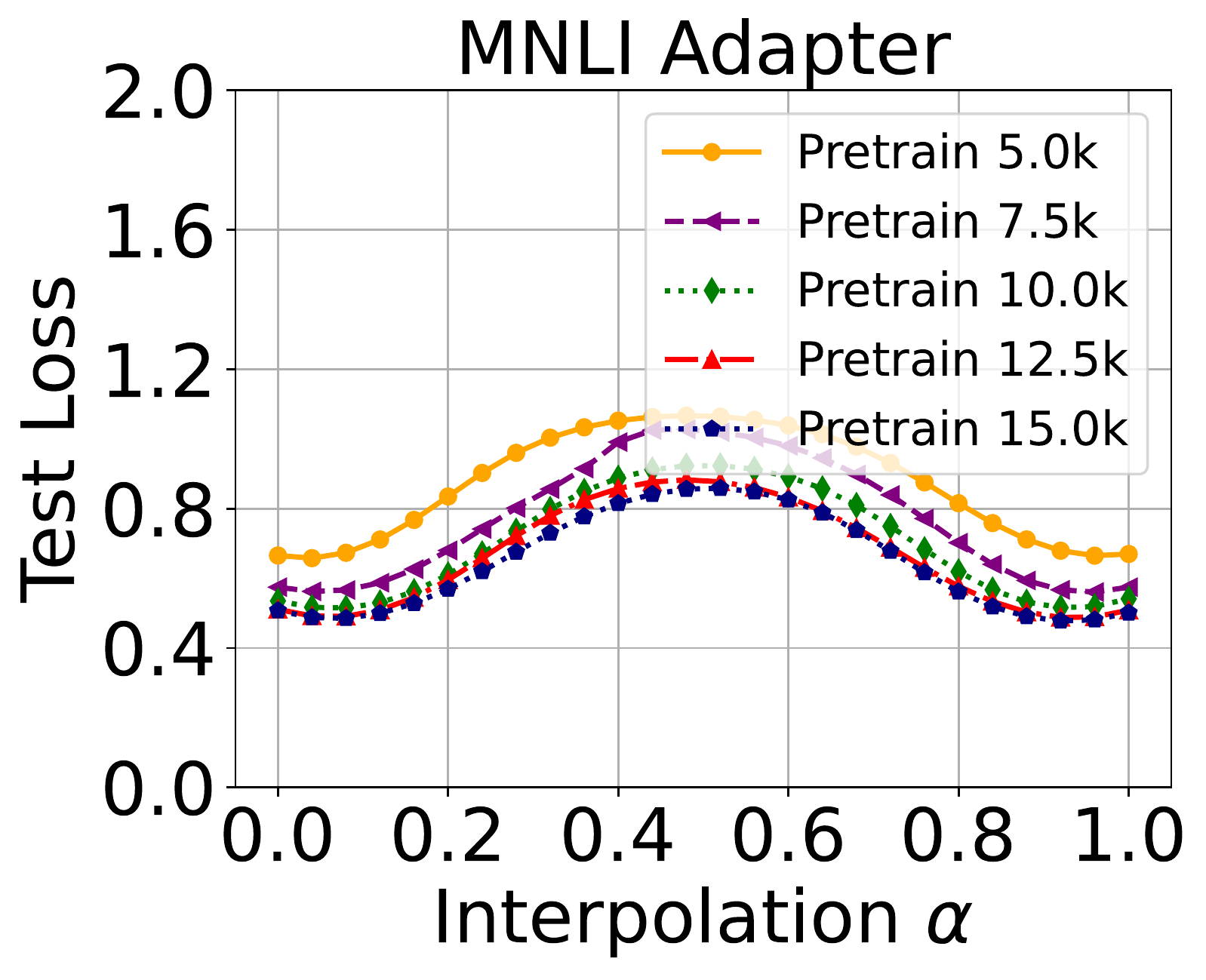}}
    \caption{The loss of the change of mode-connectivity at different pre-training steps. We report the results of linear interpolations of two minima trained on MNLI using different initialization. The corresponding performance visualization is Figure~\ref{fig:pretrain_same_task}.}
    \label{fig:pretrain_same_task_loss}
\end{figure}

\section{The Visualization of Loss for Interpolations}
\label{sec:exp_loss}
\begin{figure*}[!t]
    \centering
    \subfigure{\includegraphics[width=0.24\textwidth]{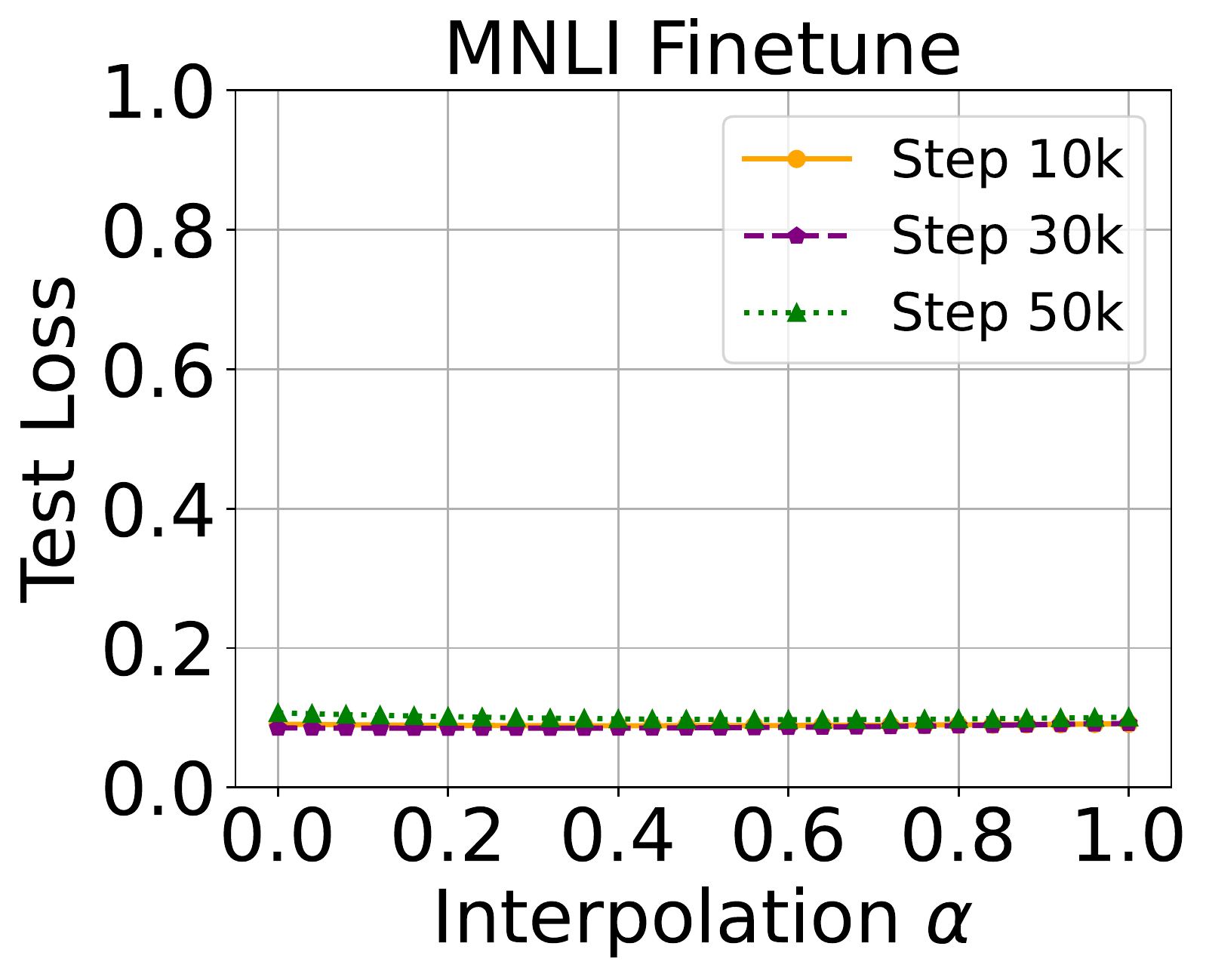}} 
    \subfigure{\includegraphics[width=0.24\textwidth]{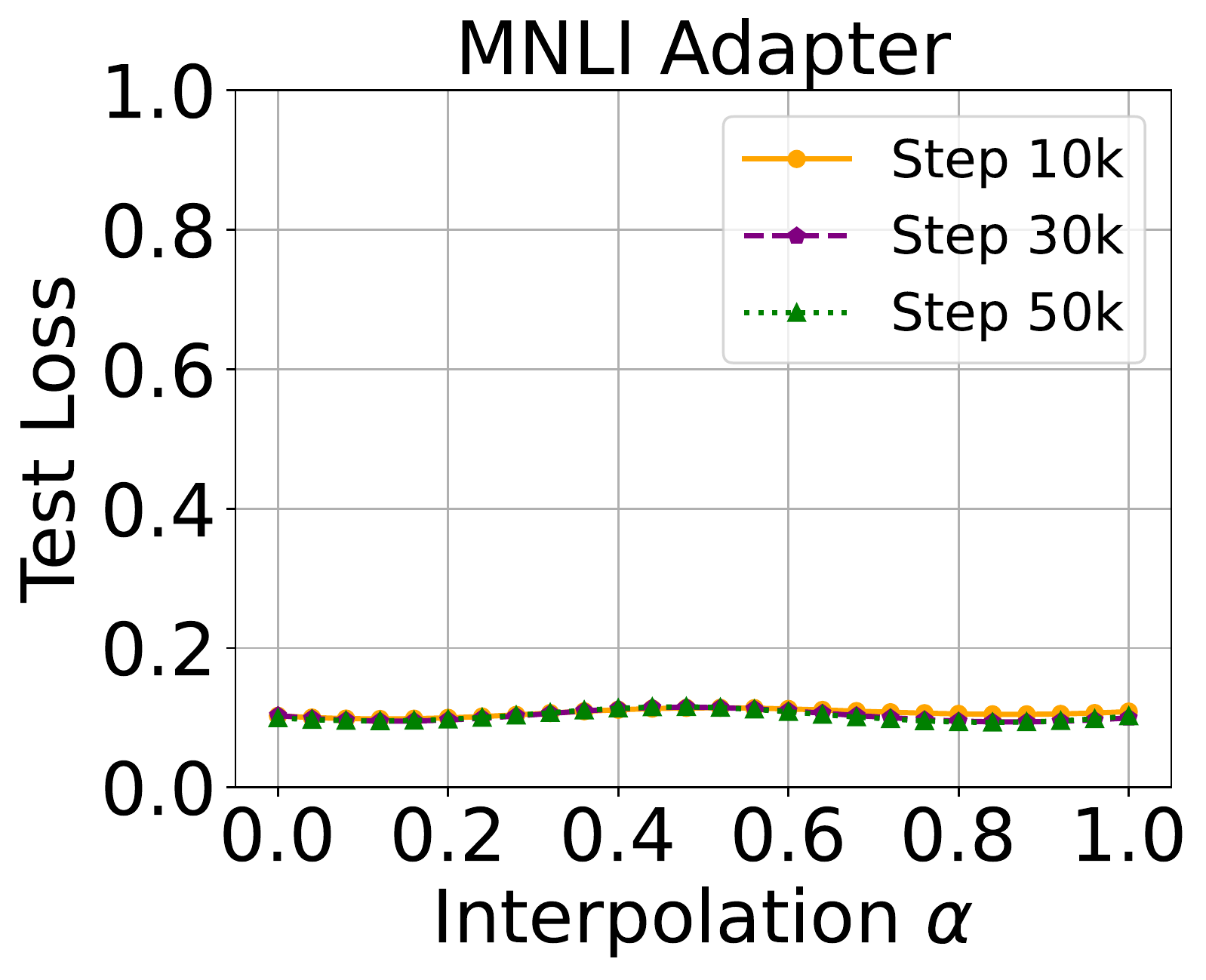}} 
    \subfigure{\includegraphics[width=0.24\textwidth]{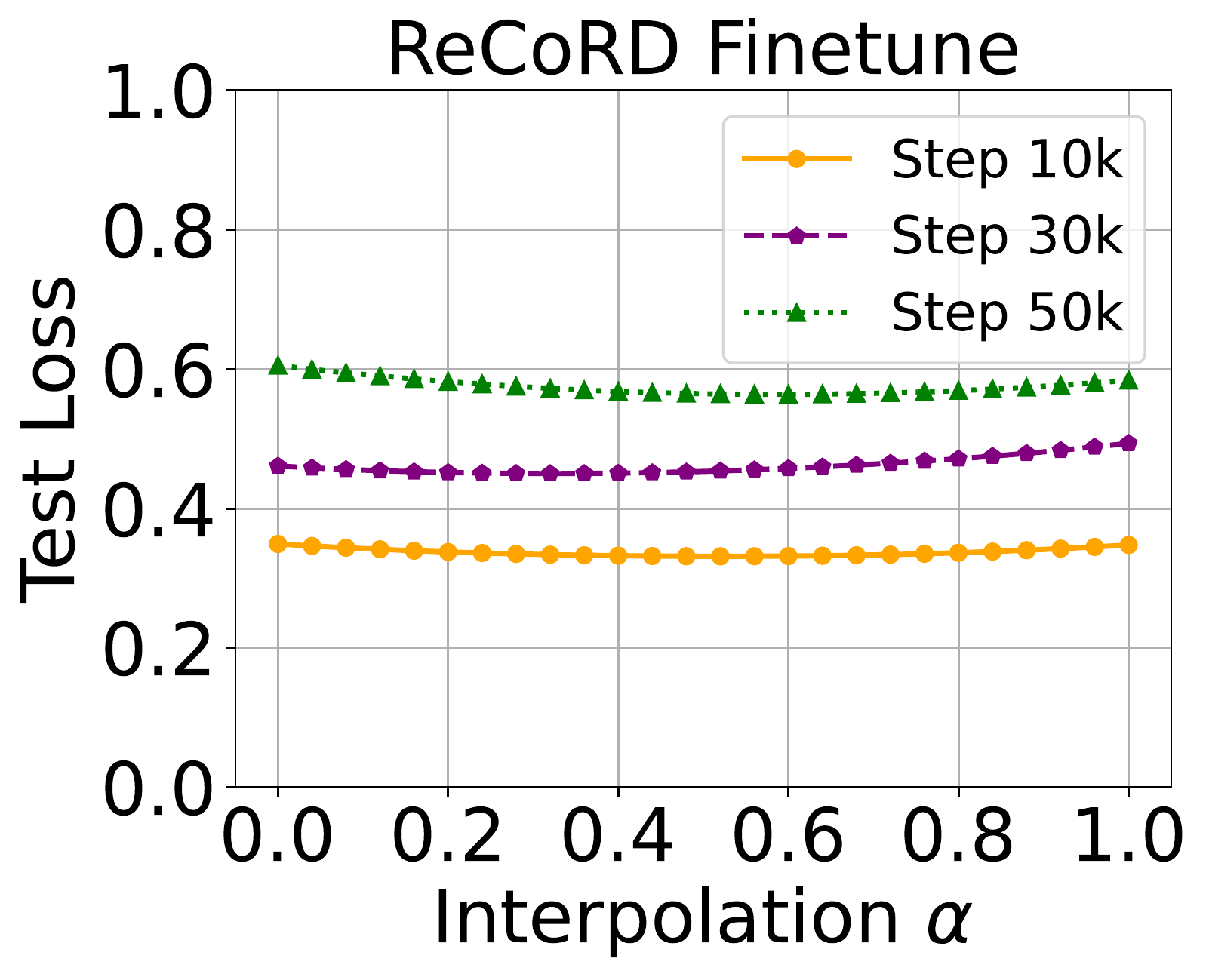}} 
    \subfigure{\includegraphics[width=0.24\textwidth]{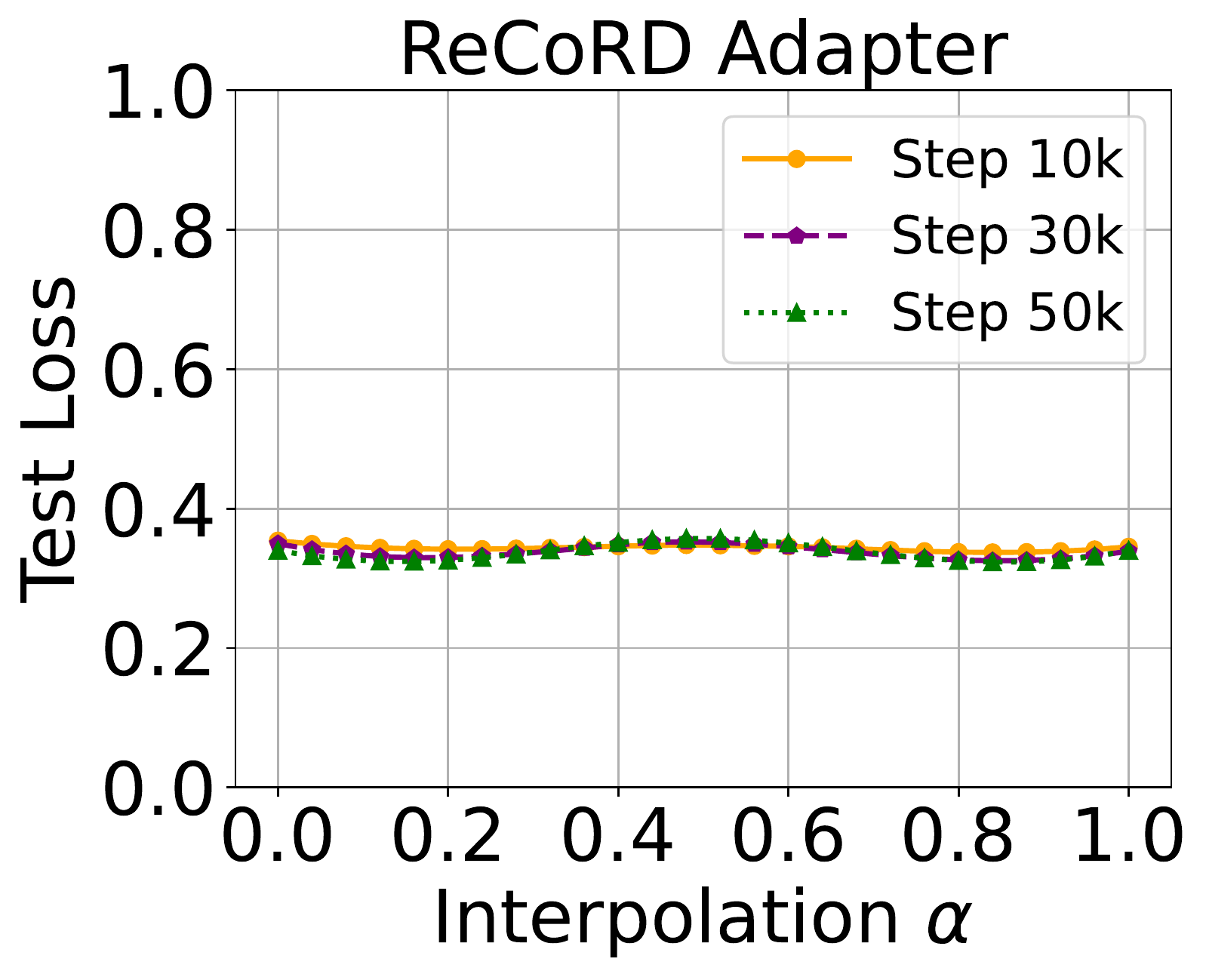}}
    \caption{The loss of linear interpolations between two minima trained with different training data order. The corresponding performance visualization is Figure~\ref{fig:order}.}
    \label{fig:order_loss}
\end{figure*}

\begin{figure*}[!t]
    \centering
    \subfigure{\includegraphics[width=0.24\textwidth]{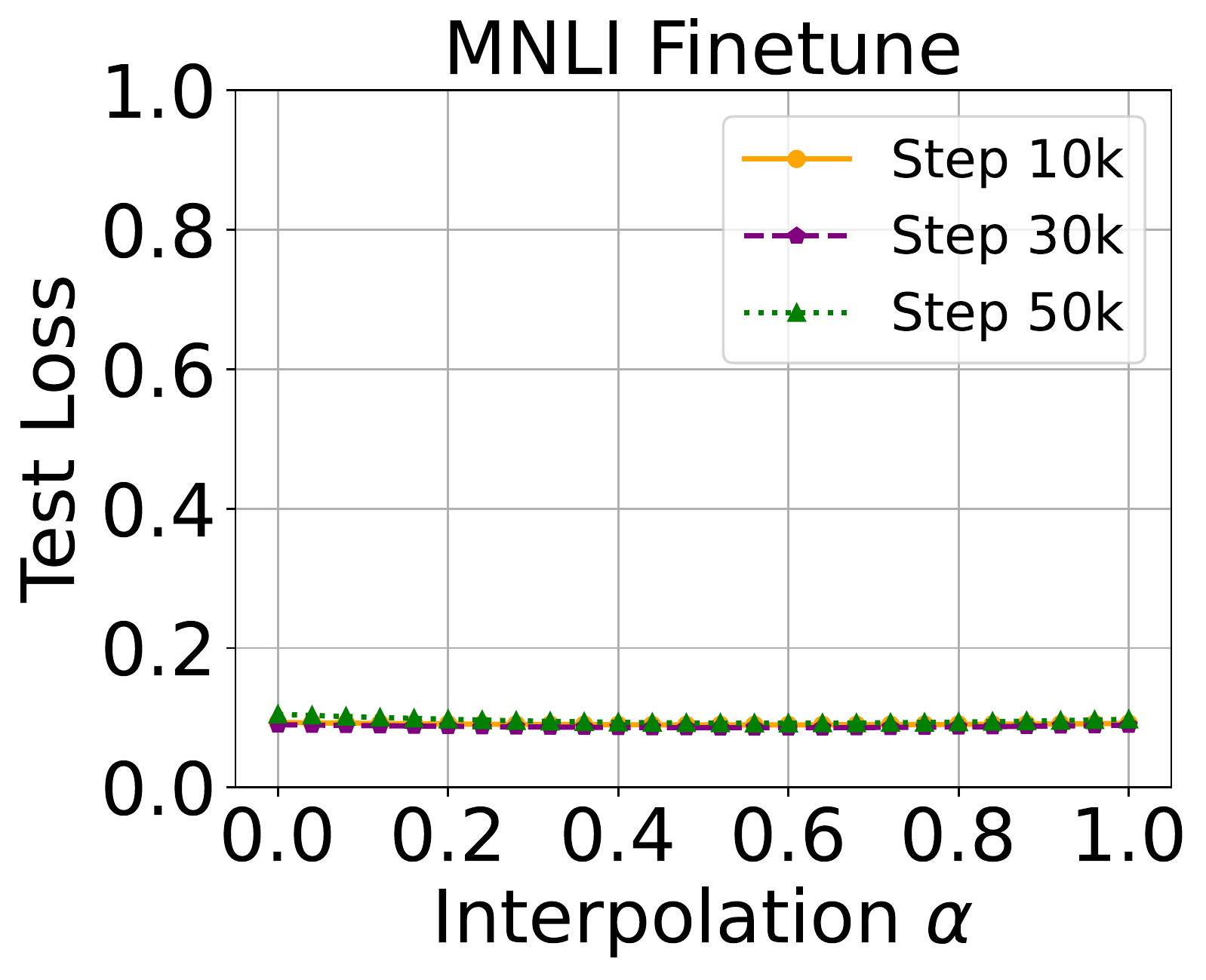}}
    \subfigure{\includegraphics[width=0.24\textwidth]{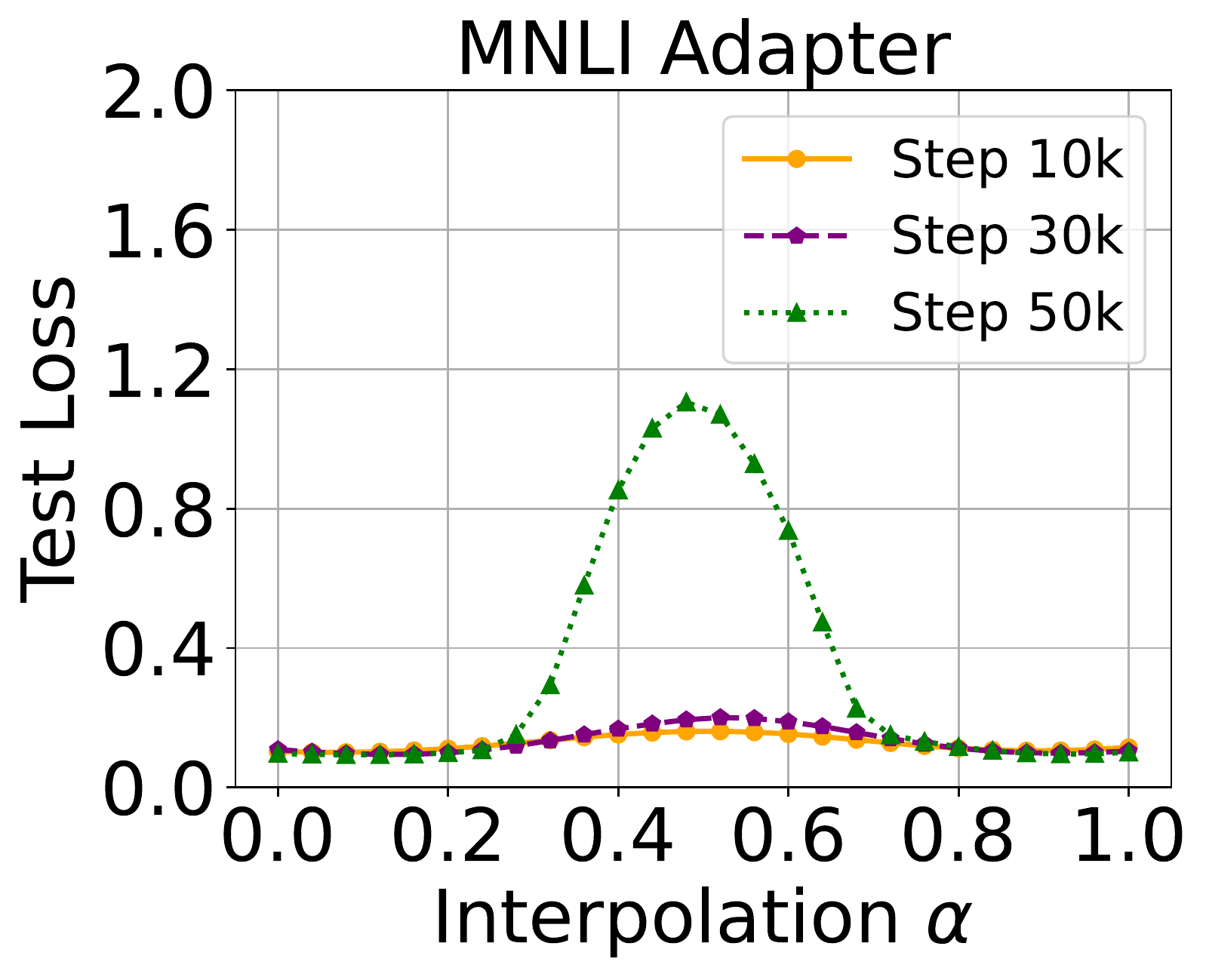}} 
    \subfigure{\includegraphics[width=0.24\textwidth]{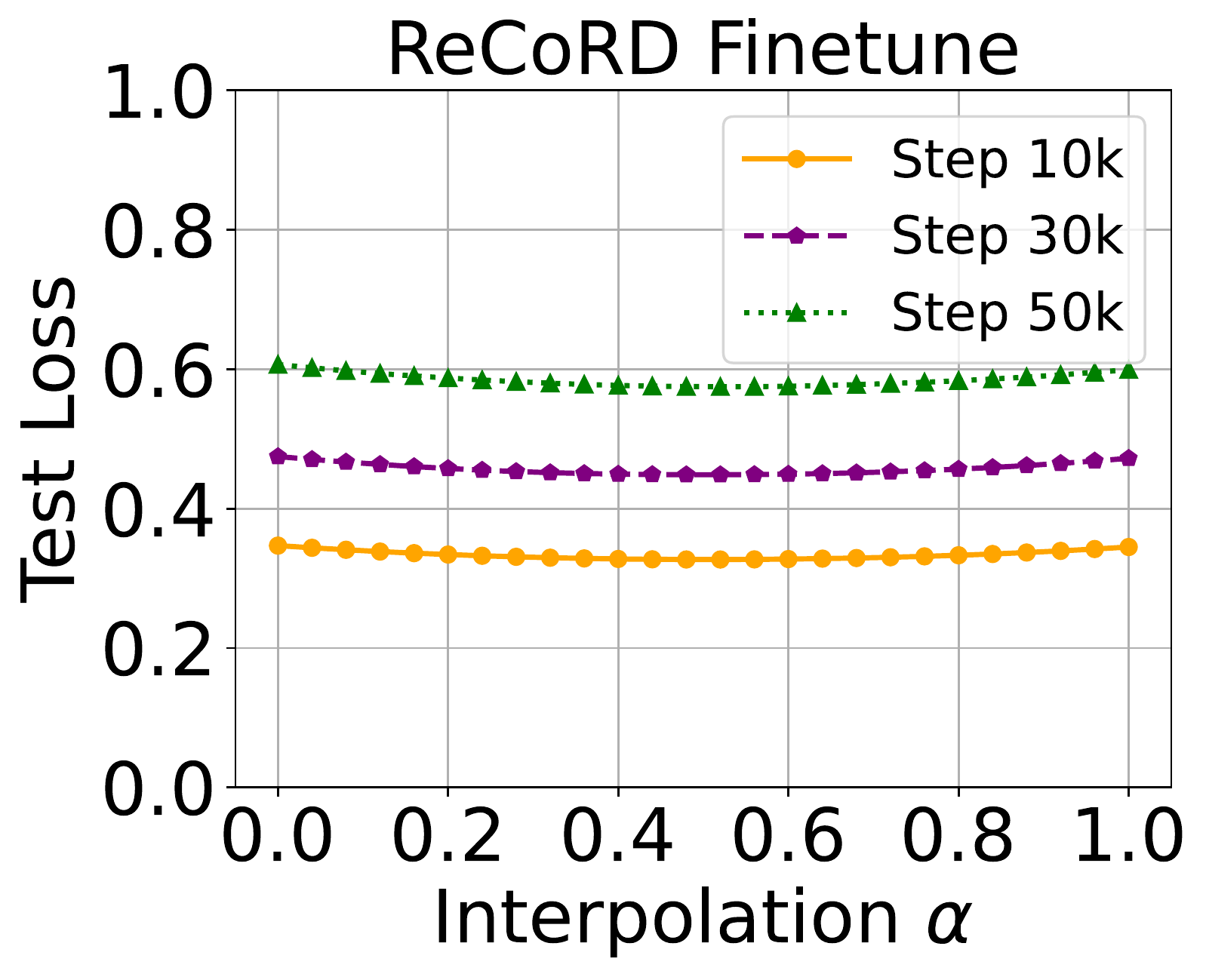}} 
    \subfigure{\includegraphics[width=0.24\textwidth]{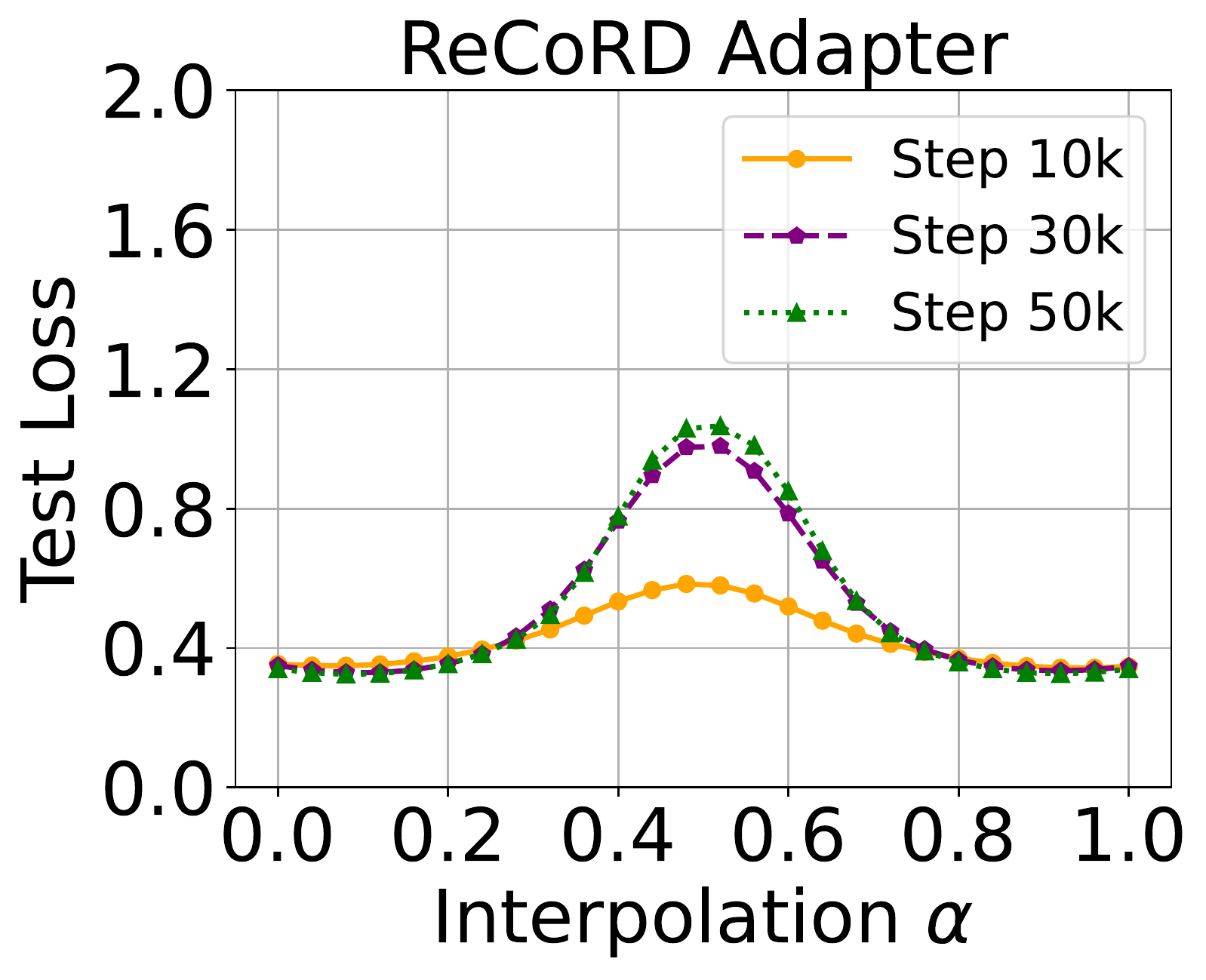}} 
    \caption{The loss of linear interpolations between two minima trained with different initialization. The corresponding performance visualization is Figure~\ref{fig:initialization}.}
    \label{fig:initialization_loss}
\end{figure*}

\begin{figure*}[!t]
    \centering
    \subfigure{\includegraphics[width=0.24\textwidth]{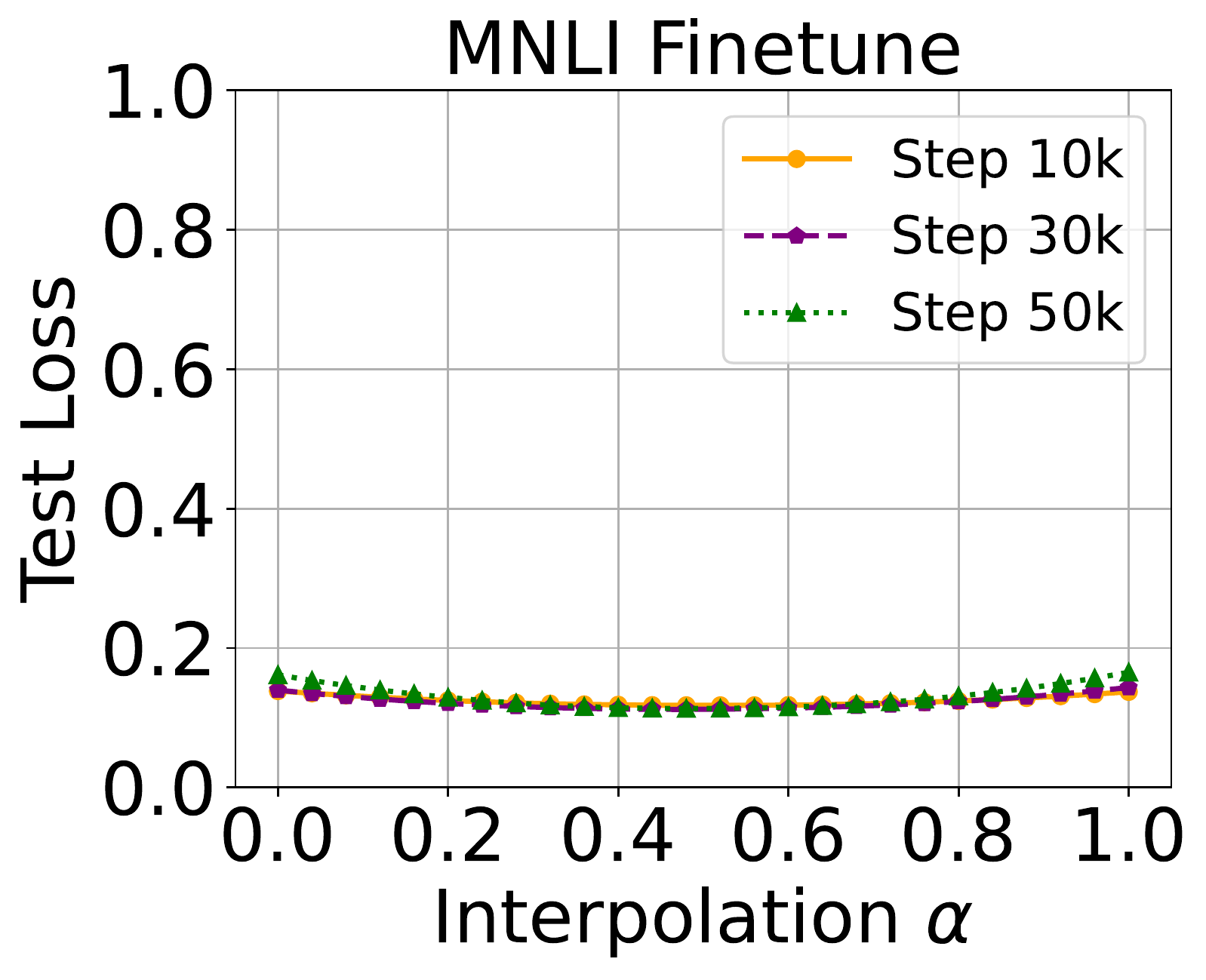}} 
    \subfigure{\includegraphics[width=0.24\textwidth]{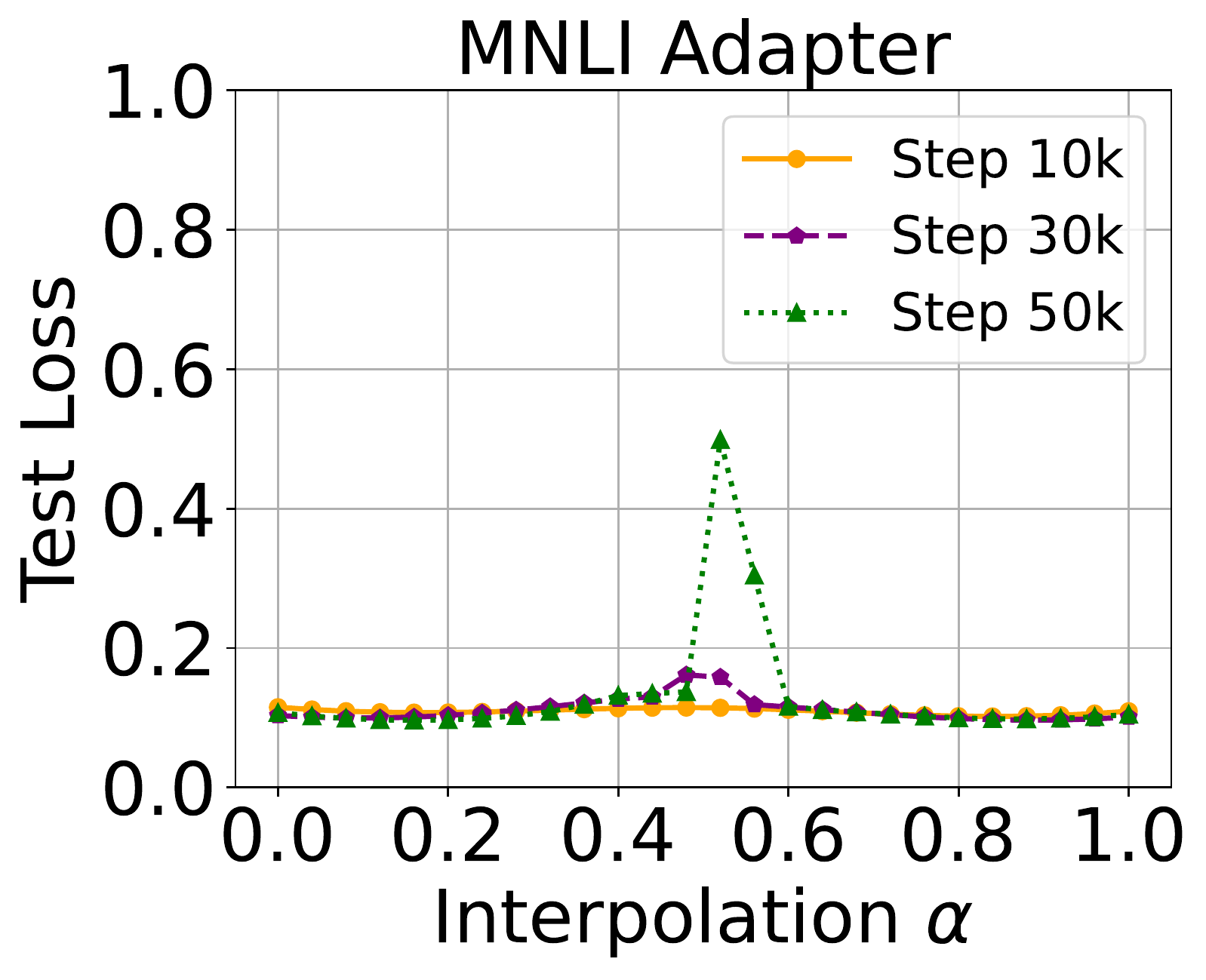}} 
    \subfigure{\includegraphics[width=0.24\textwidth]{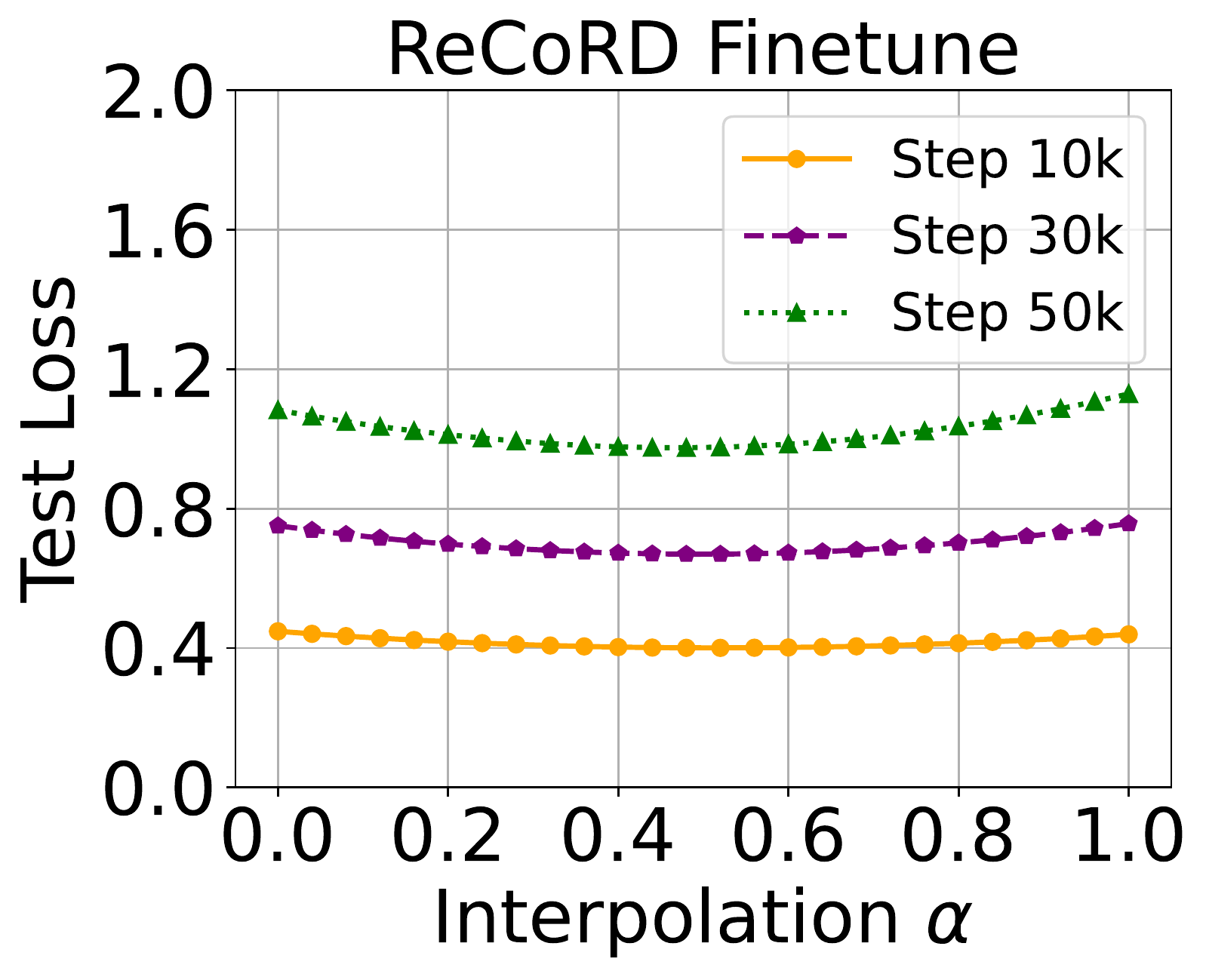}}
    \subfigure{\includegraphics[width=0.24\textwidth]{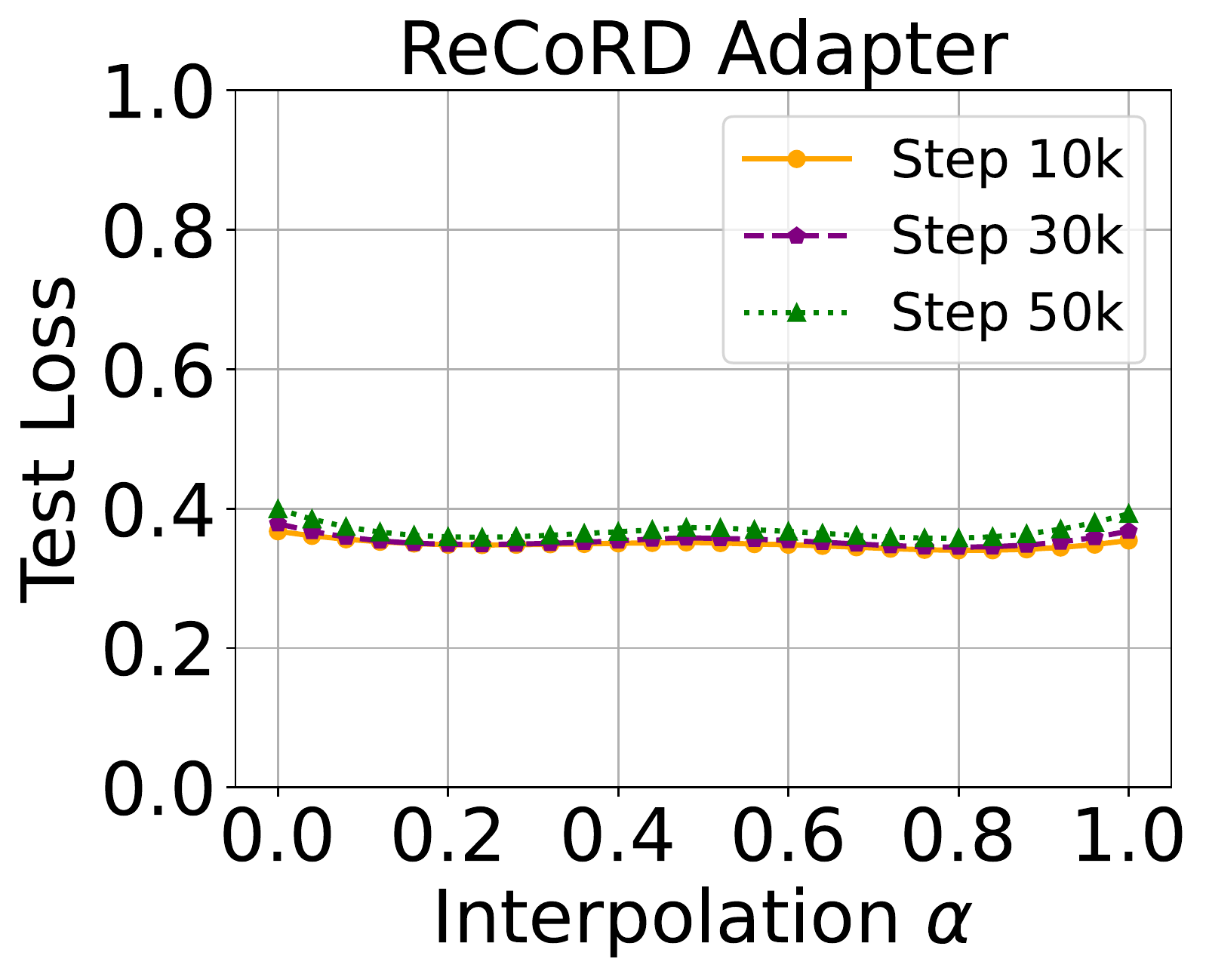}}
    \caption{Linear mode connectivity analysis (loss) for two minima trained with in-distribution data. The corresponding performance visualization is Figure~\ref{fig:overlap} and Figure~\ref{fig:overlap_2}.}
    \label{fig:overlap_loss}
\end{figure*}

\begin{figure*}[!t]
    \centering
    \subfigure[]{\includegraphics[width=0.235\textwidth]{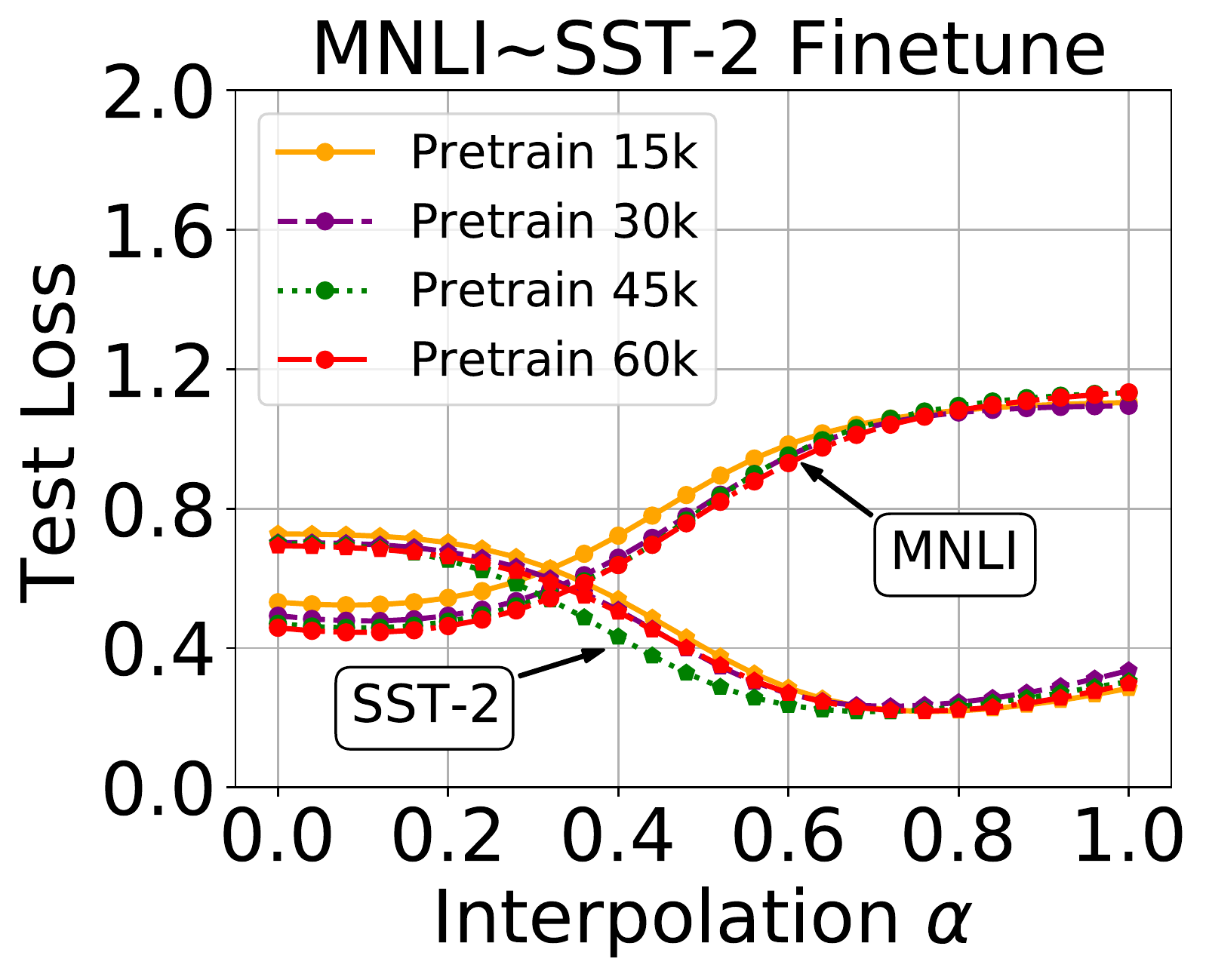}}
    \subfigure[]{\includegraphics[width=0.235\textwidth]{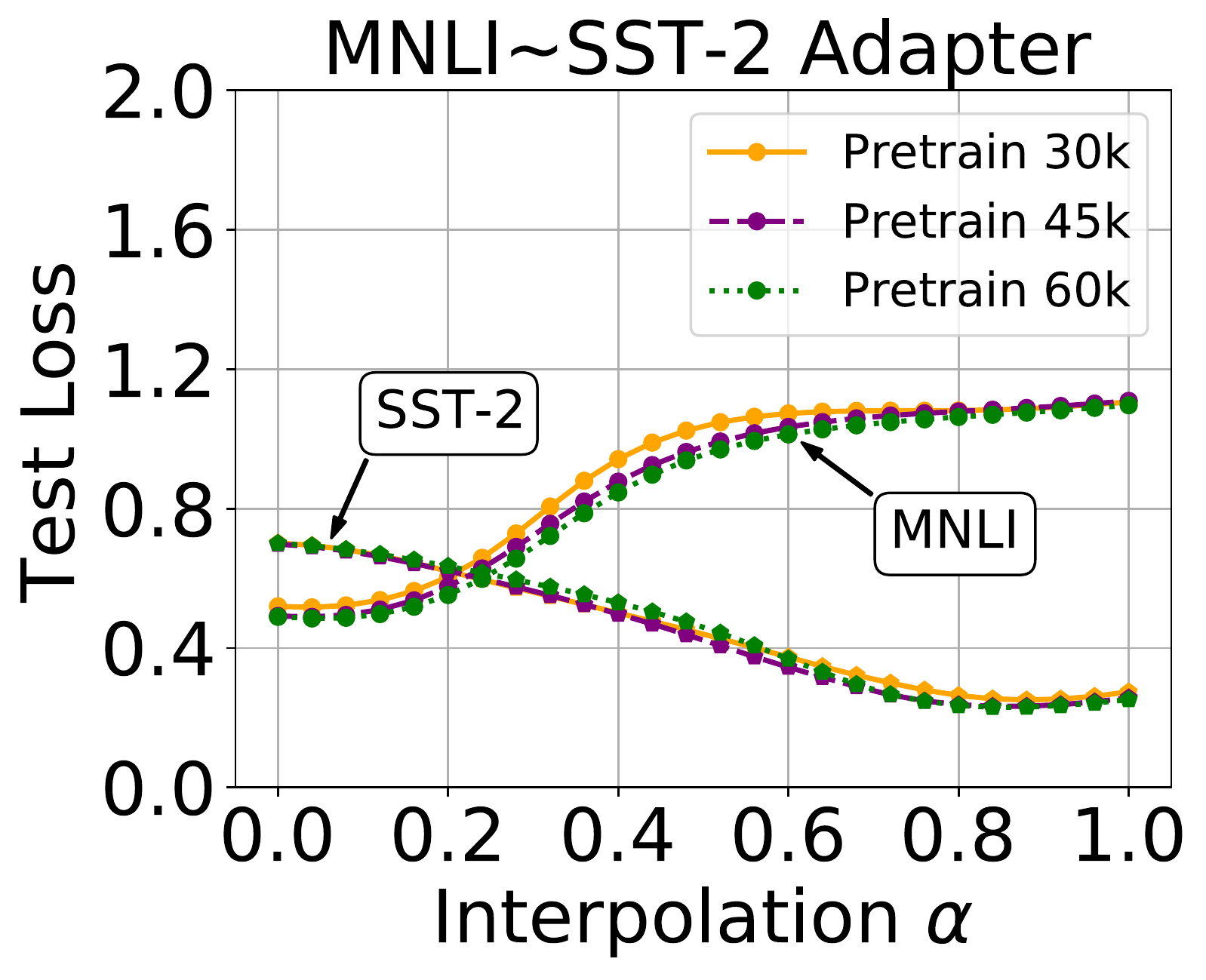}}
    \subfigure[]{\includegraphics[width=0.235\textwidth]{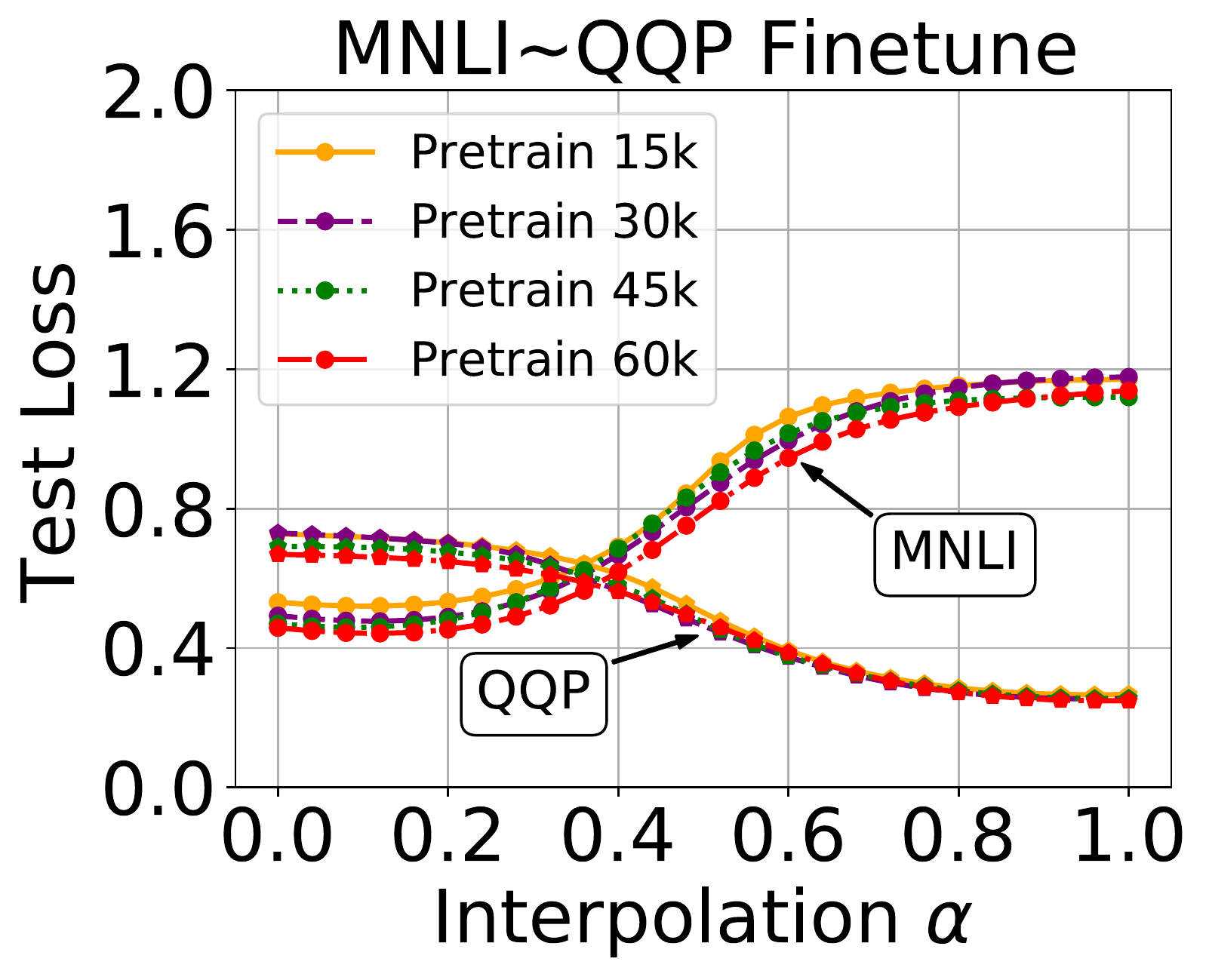}}
    \subfigure[]{\includegraphics[width=0.235\textwidth]{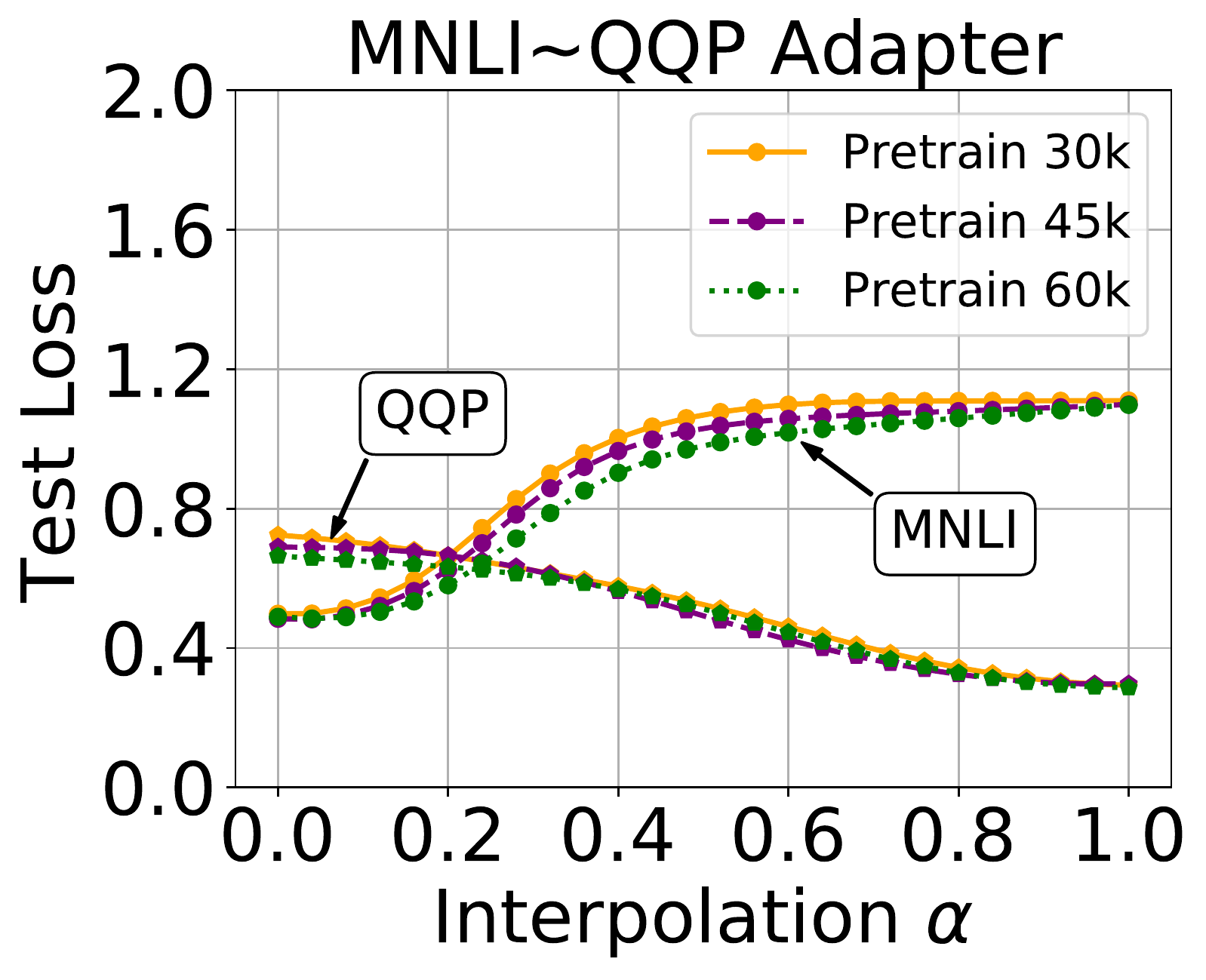}}
    \caption{The results (loss) of the change of mode-connectivity at different pre-training steps. (a-b) record the results of linear interpolations of two minima trained on MNLI and SST-2, (c-d) record the results of linear interpolations of two minima trained on MNLI and QQP. The corresponding performance visualization is Figure~\ref{fig:pretrain_diff_task} and Figure~\ref{fig:diff_task_2}.}
    \label{fig:pretrain_diff_task_loss}
\end{figure*}

\begin{figure*}[!t]
    \centering
    \subfigure[]{\includegraphics[width=0.24\textwidth]{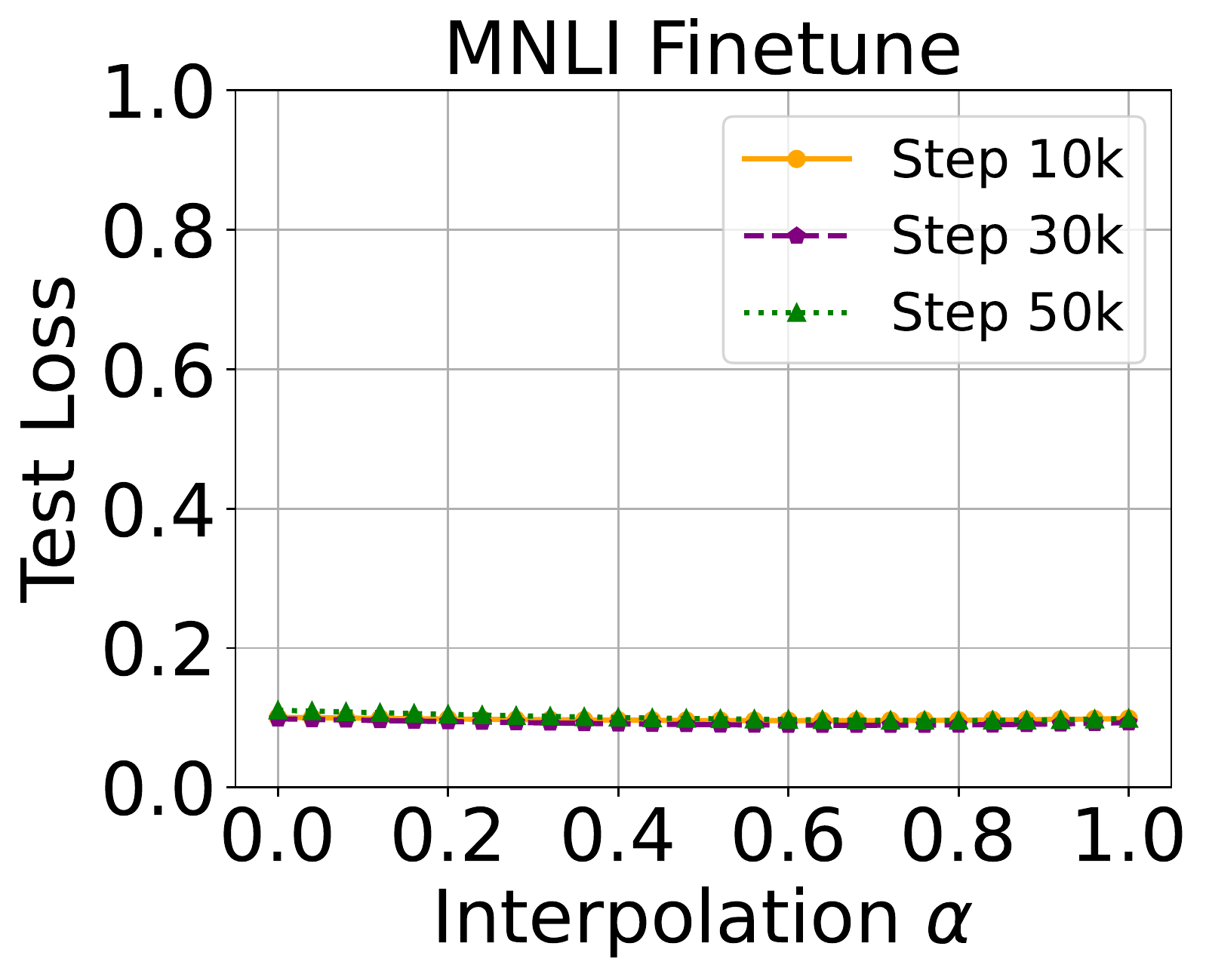}} 
    \subfigure[]{\includegraphics[width=0.24\textwidth]{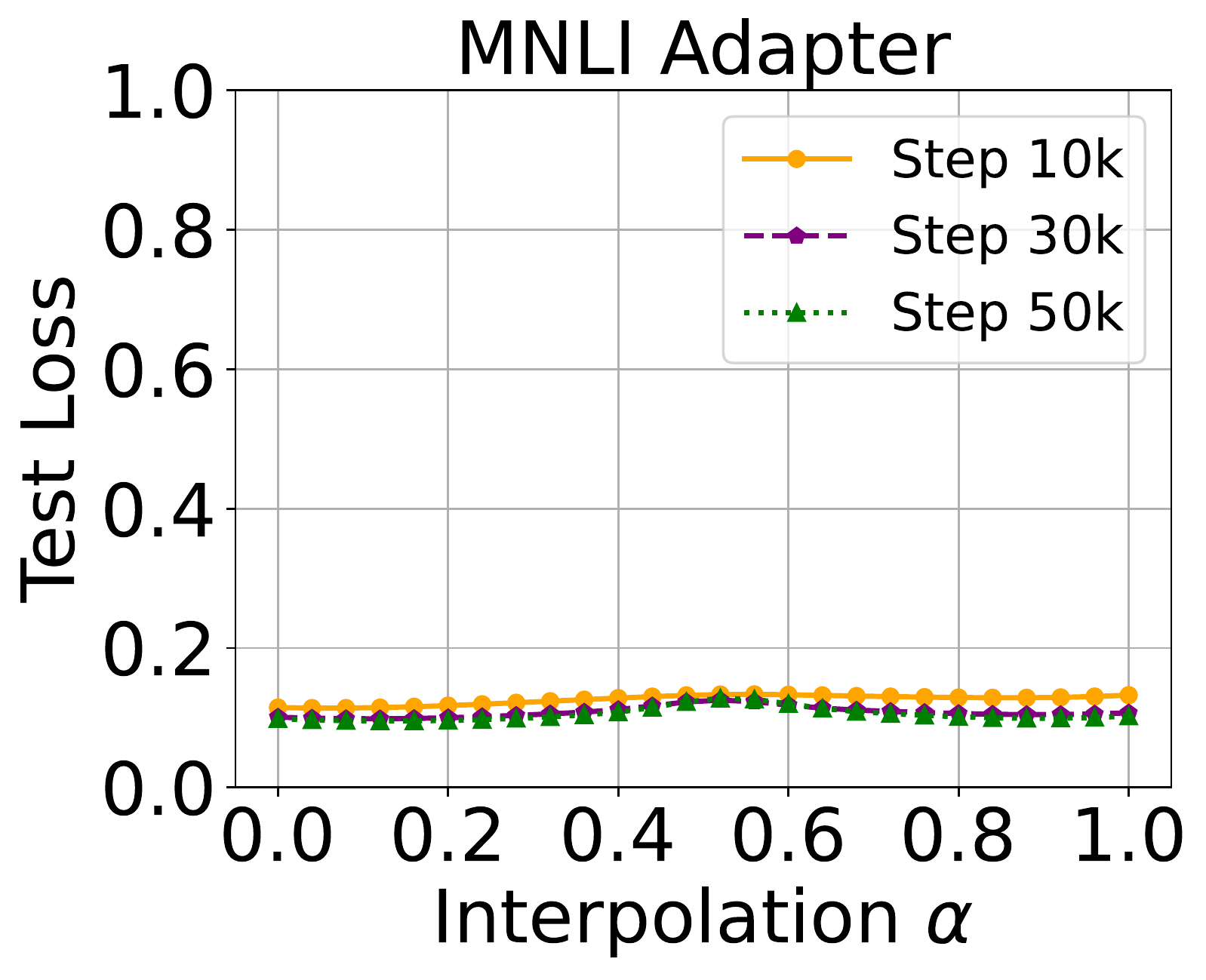}}
    \subfigure[]{\includegraphics[width=0.24\textwidth]{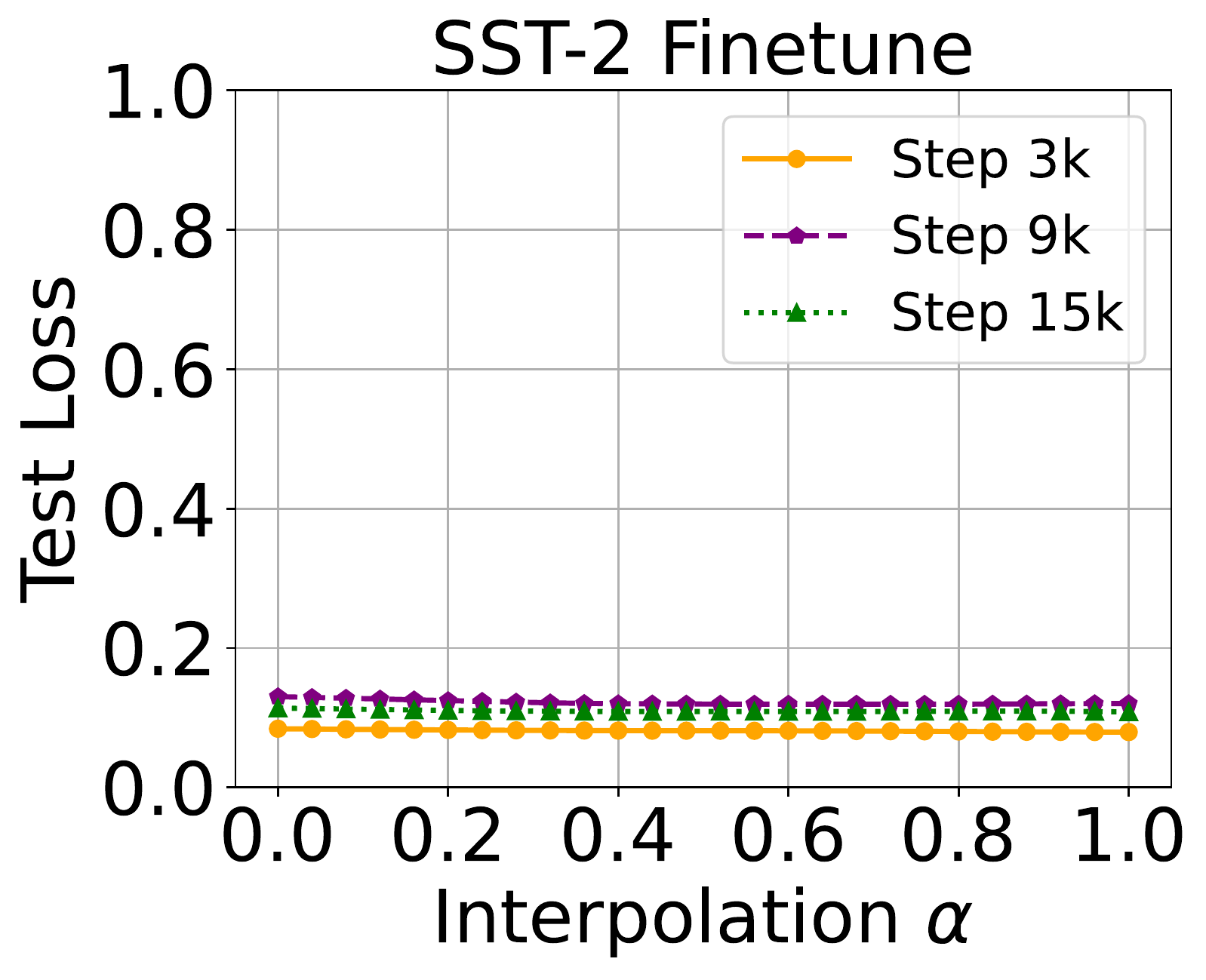}}
    \subfigure[]{\includegraphics[width=0.24\textwidth]{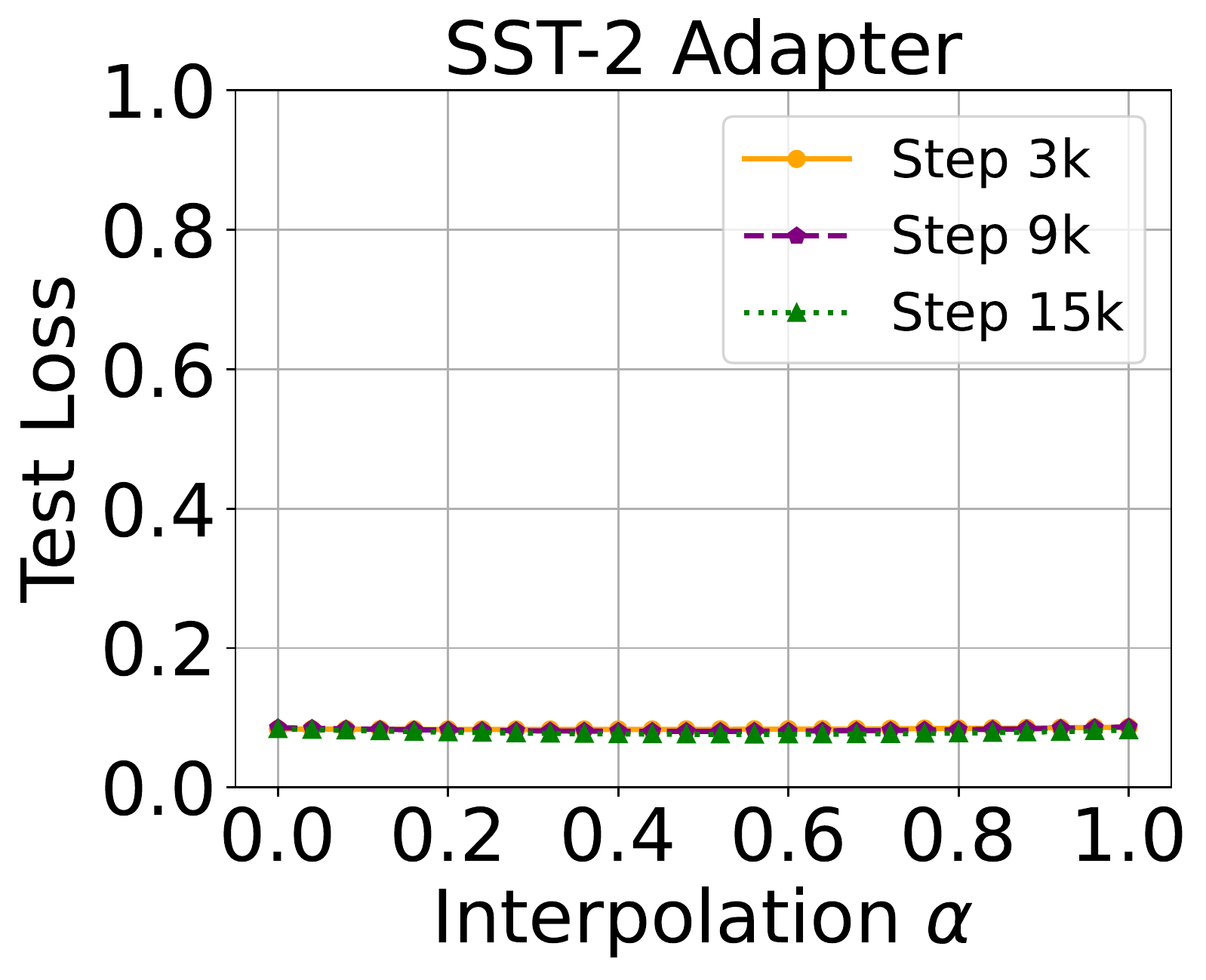}} 
    
    \subfigure[]{\includegraphics[width=0.24\textwidth]{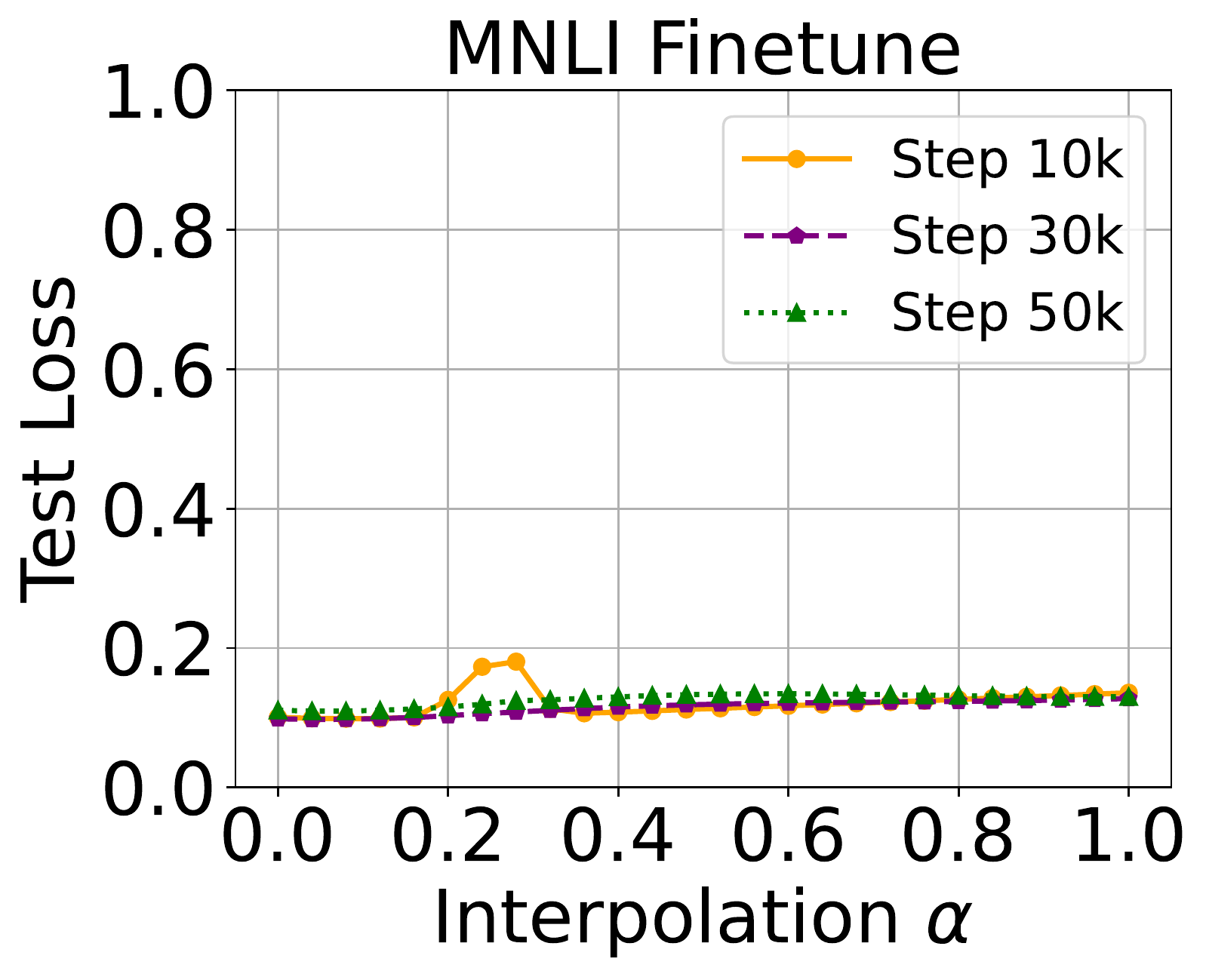}} 
    \subfigure[]{\includegraphics[width=0.24\textwidth]{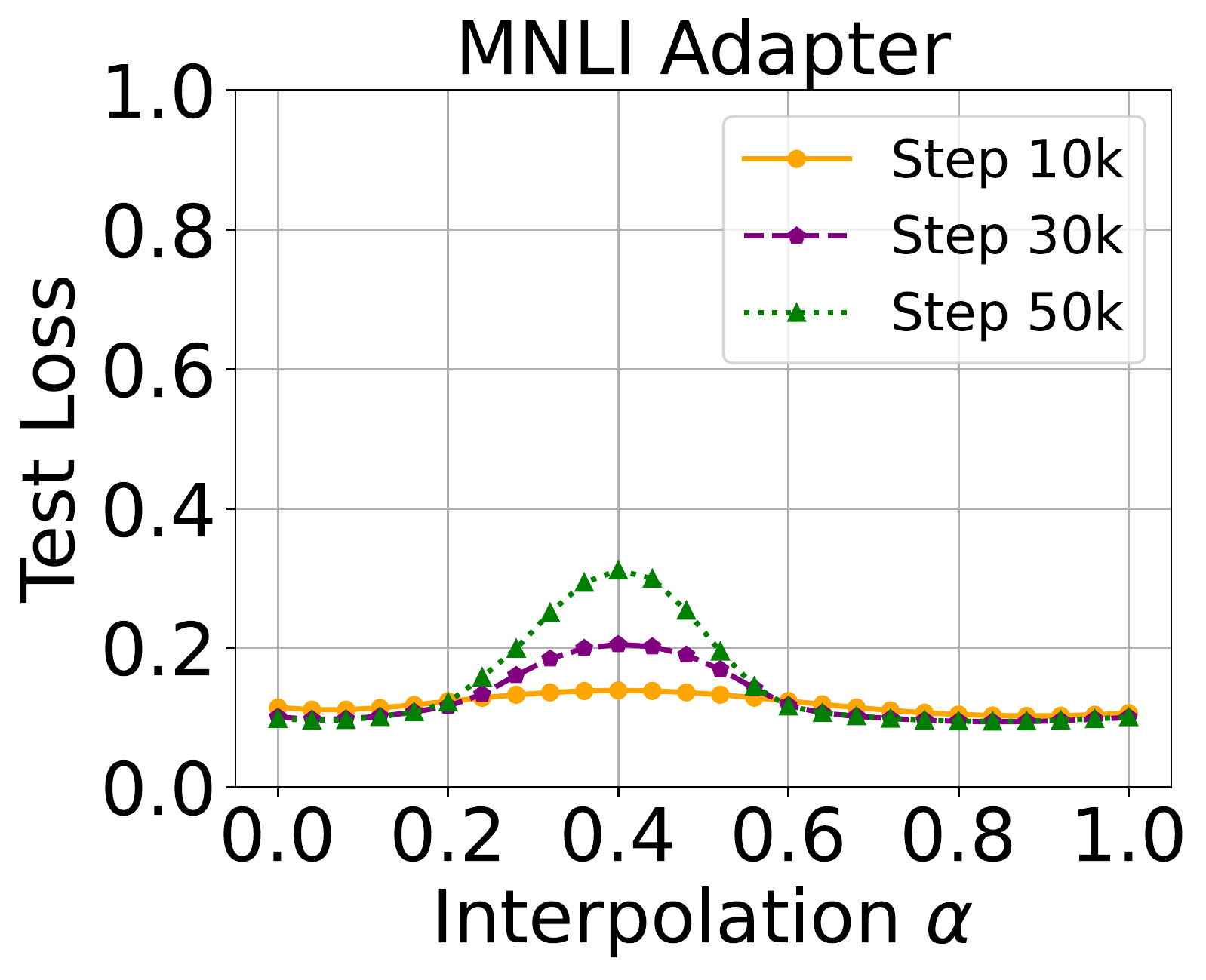}} 
    \subfigure[]{\includegraphics[width=0.24\textwidth]{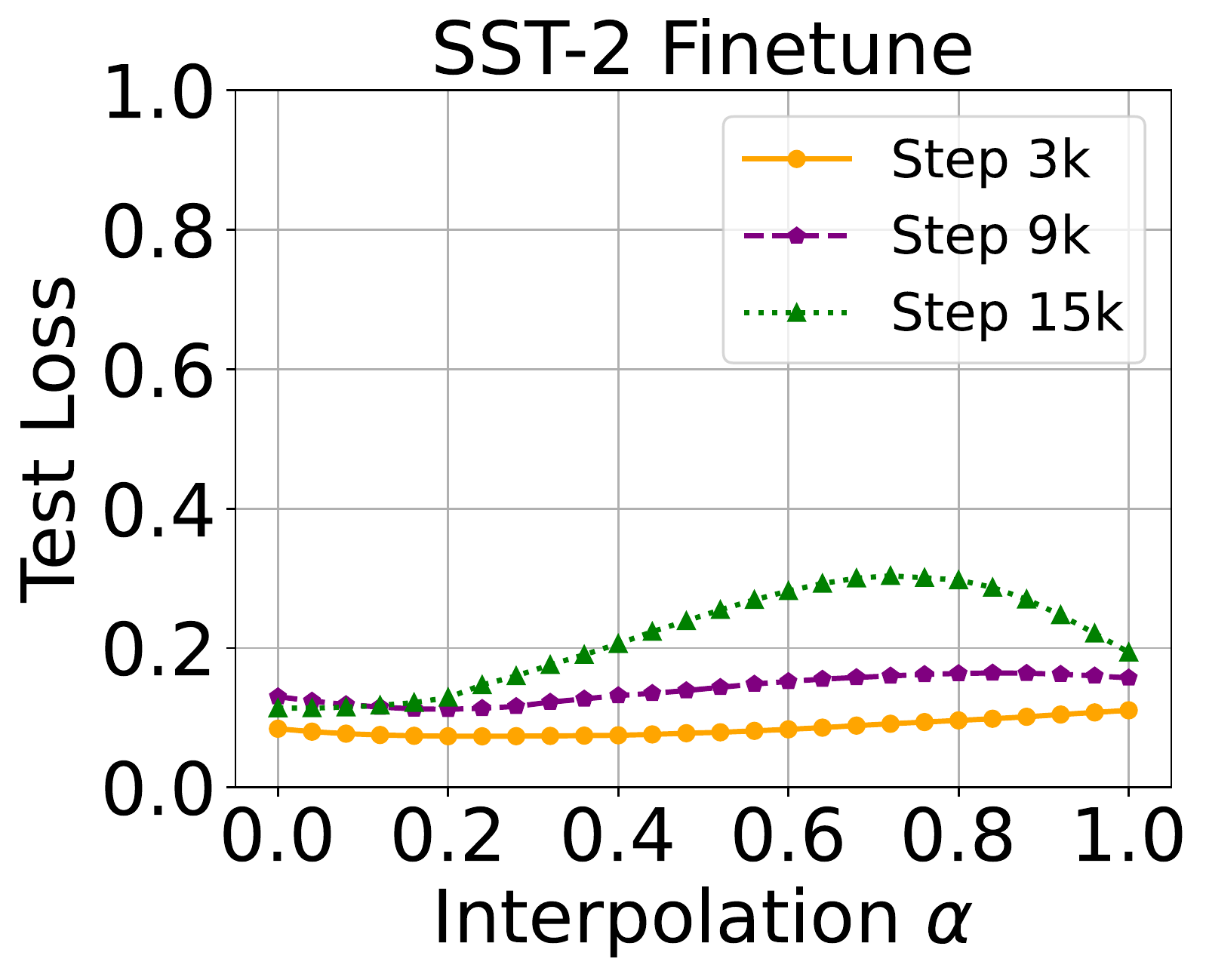}}
    \subfigure[]{\includegraphics[width=0.24\textwidth]{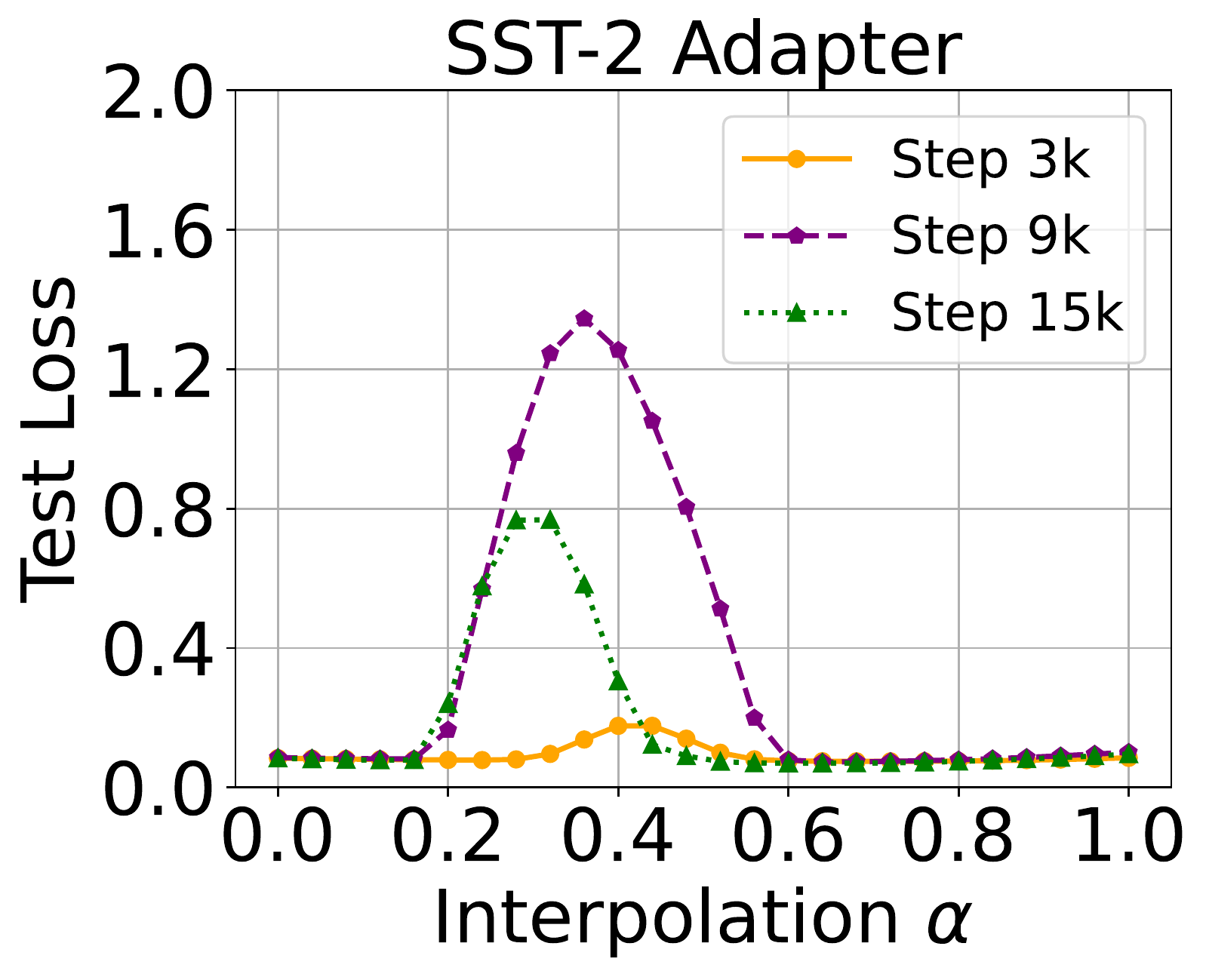}}
    \caption{Experiments of the effects of the learning rate. We conduct linear interpolations for MNLI and SST-2, using both fine-tuning and adapter tuning, and visualize their loss. For (a-d), both minima are obtained with a learning rate of $1\times10^{-4}$ and $5\times10^{-5}$, respectively; for (e-h), both minima are obtained with a learning rate of $1\times10^{-4}$ and $5\times10^{-4}$, respectively. The corresponding performance visualization is Figure~\ref{fig:lr}.}
    \label{fig:lr_loss}
\end{figure*}

\begin{figure*}[!t]
    \centering
    \subfigure[]{\includegraphics[width=0.24\textwidth]{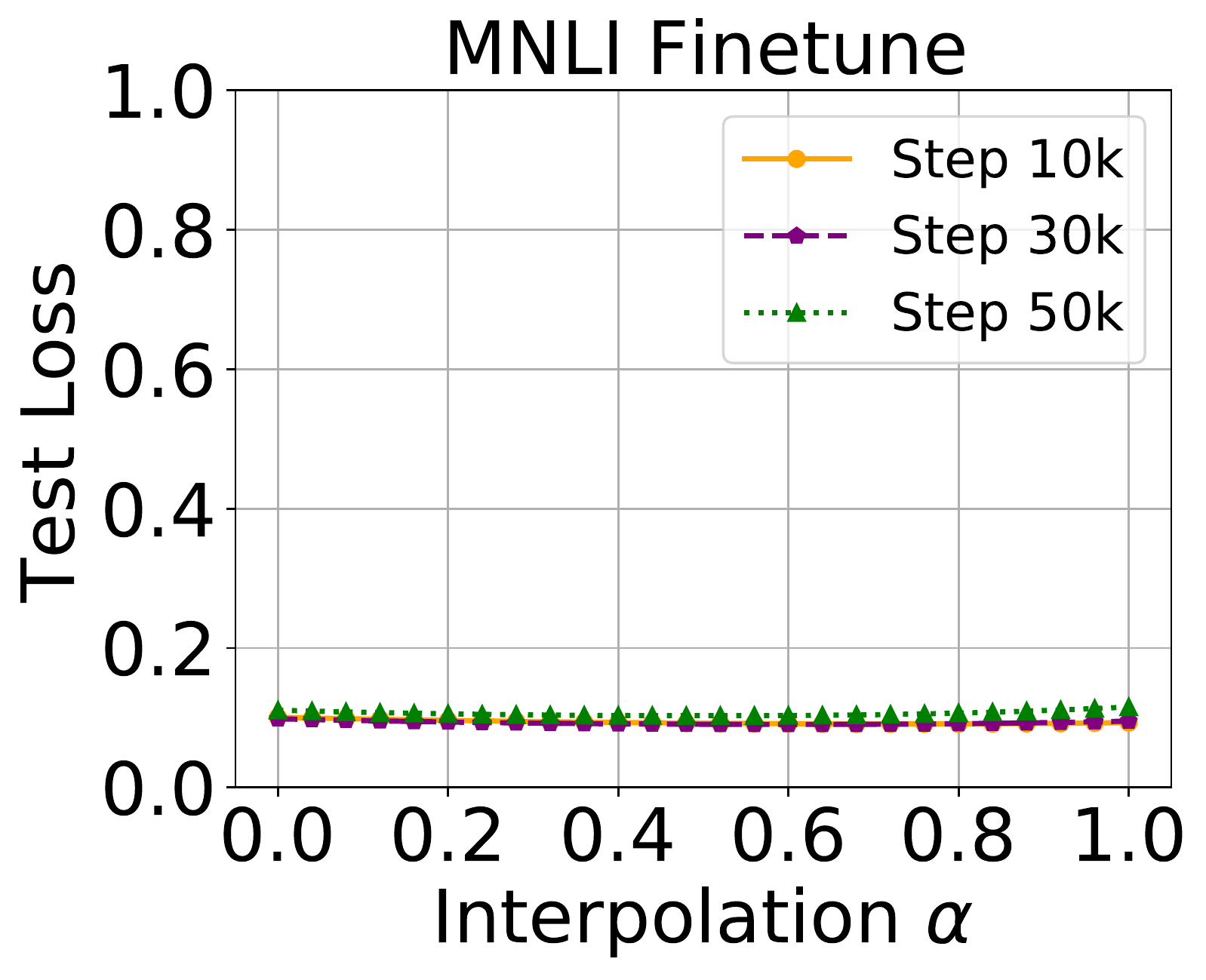}} 
    \subfigure[]{\includegraphics[width=0.24\textwidth]{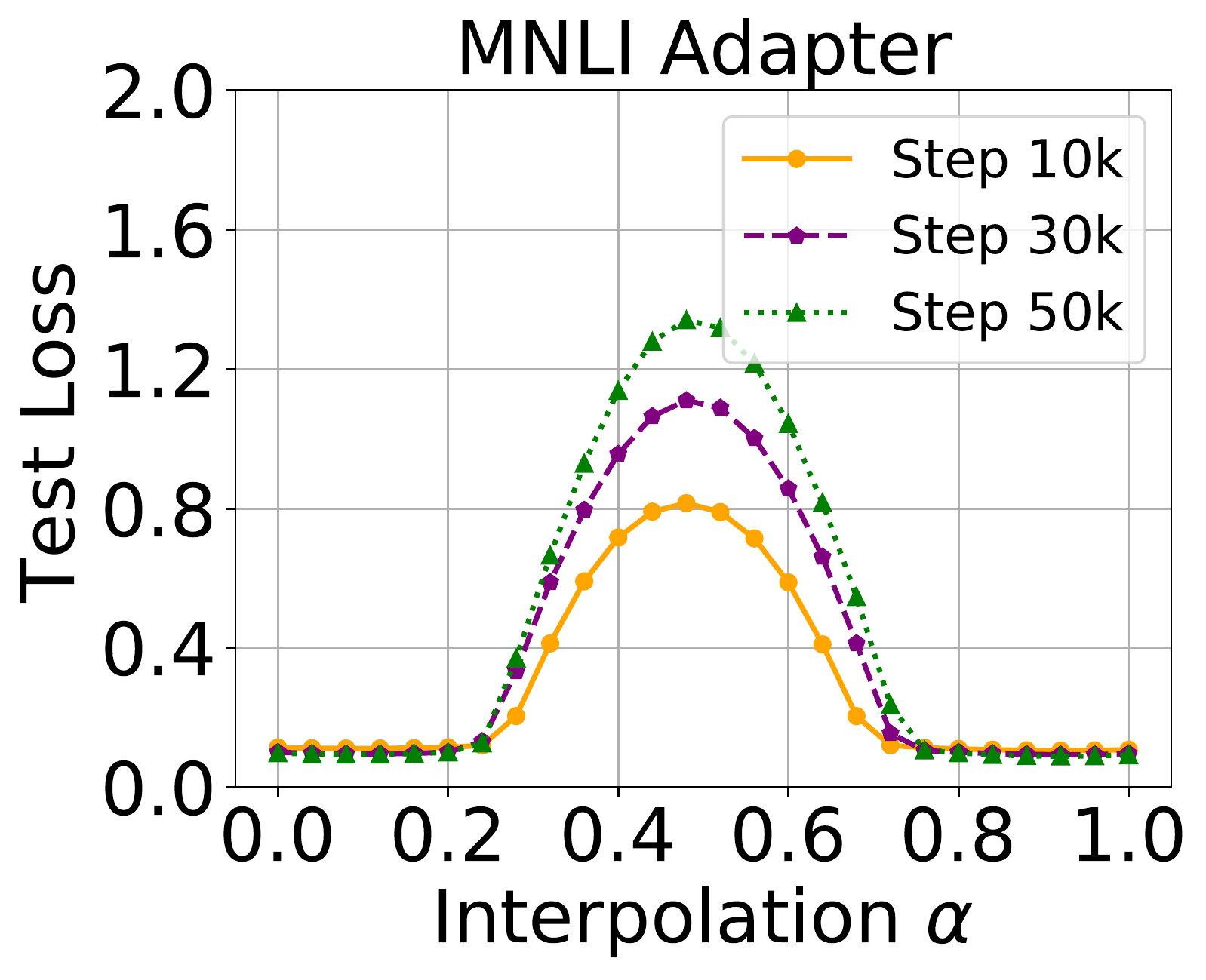}} 
    \subfigure[]{\includegraphics[width=0.24\textwidth]{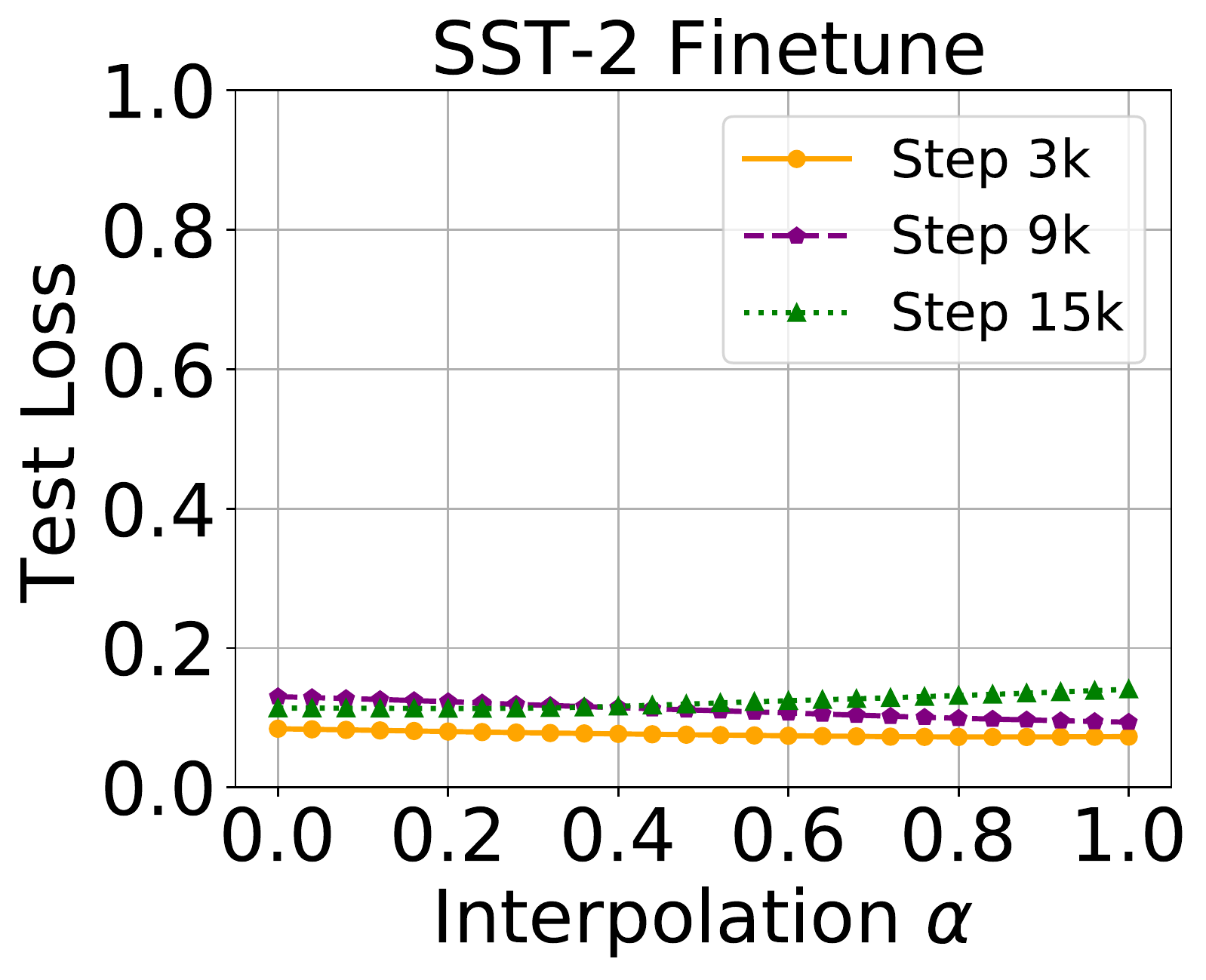}}
    \subfigure[]{\includegraphics[width=0.24\textwidth]{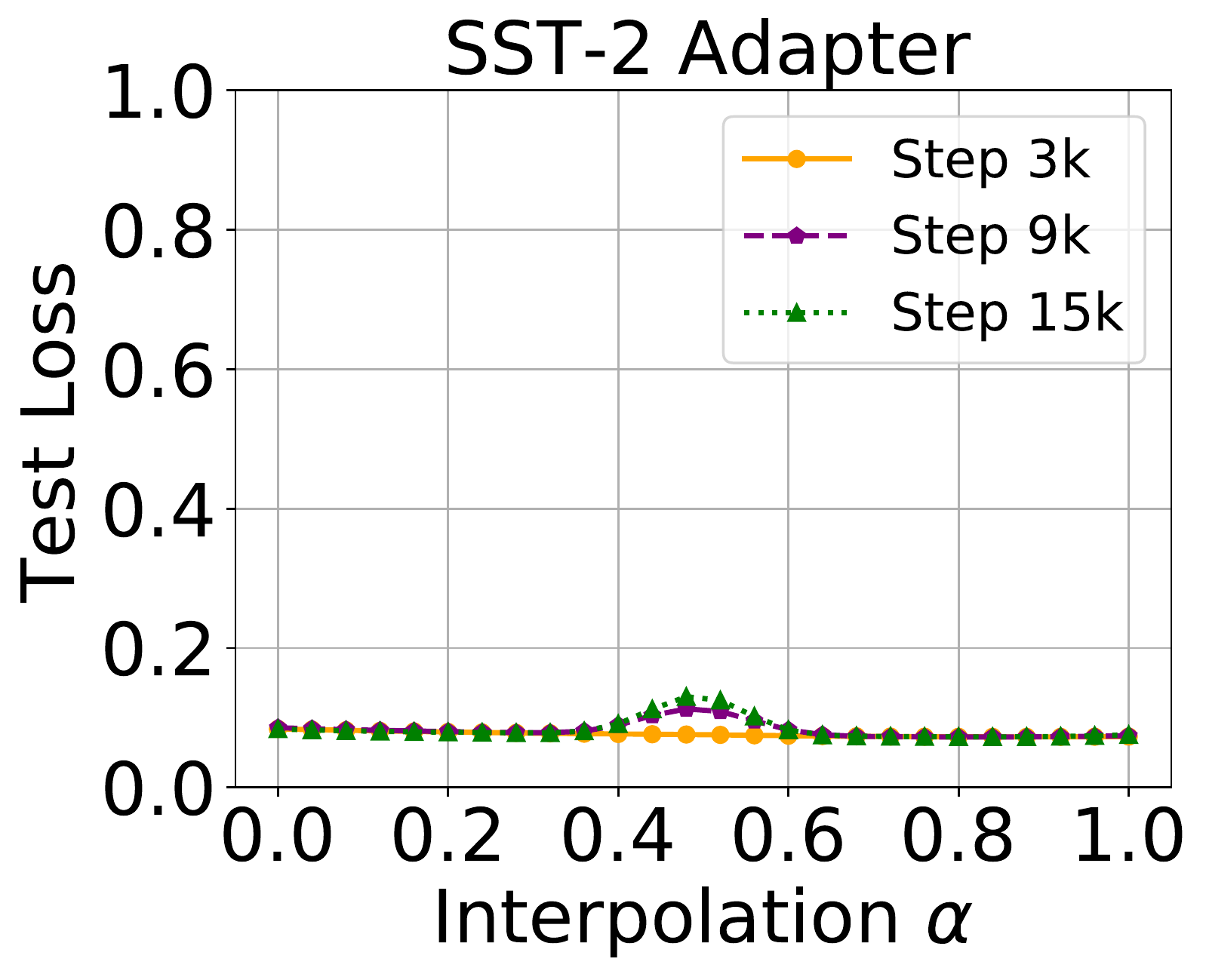}} 
    
    \subfigure[]{\includegraphics[width=0.24\textwidth]{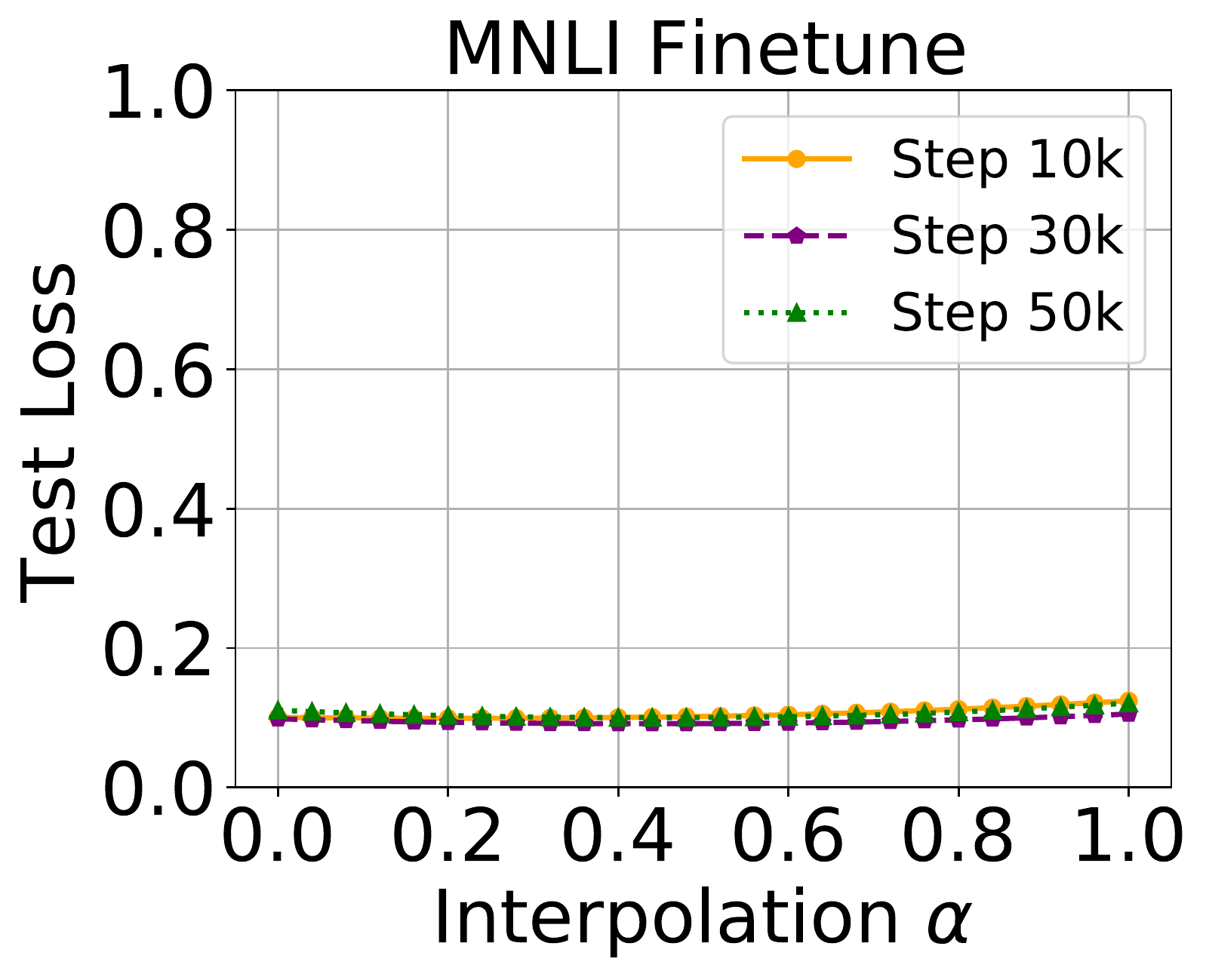}} 
    \subfigure[]{\includegraphics[width=0.24\textwidth]{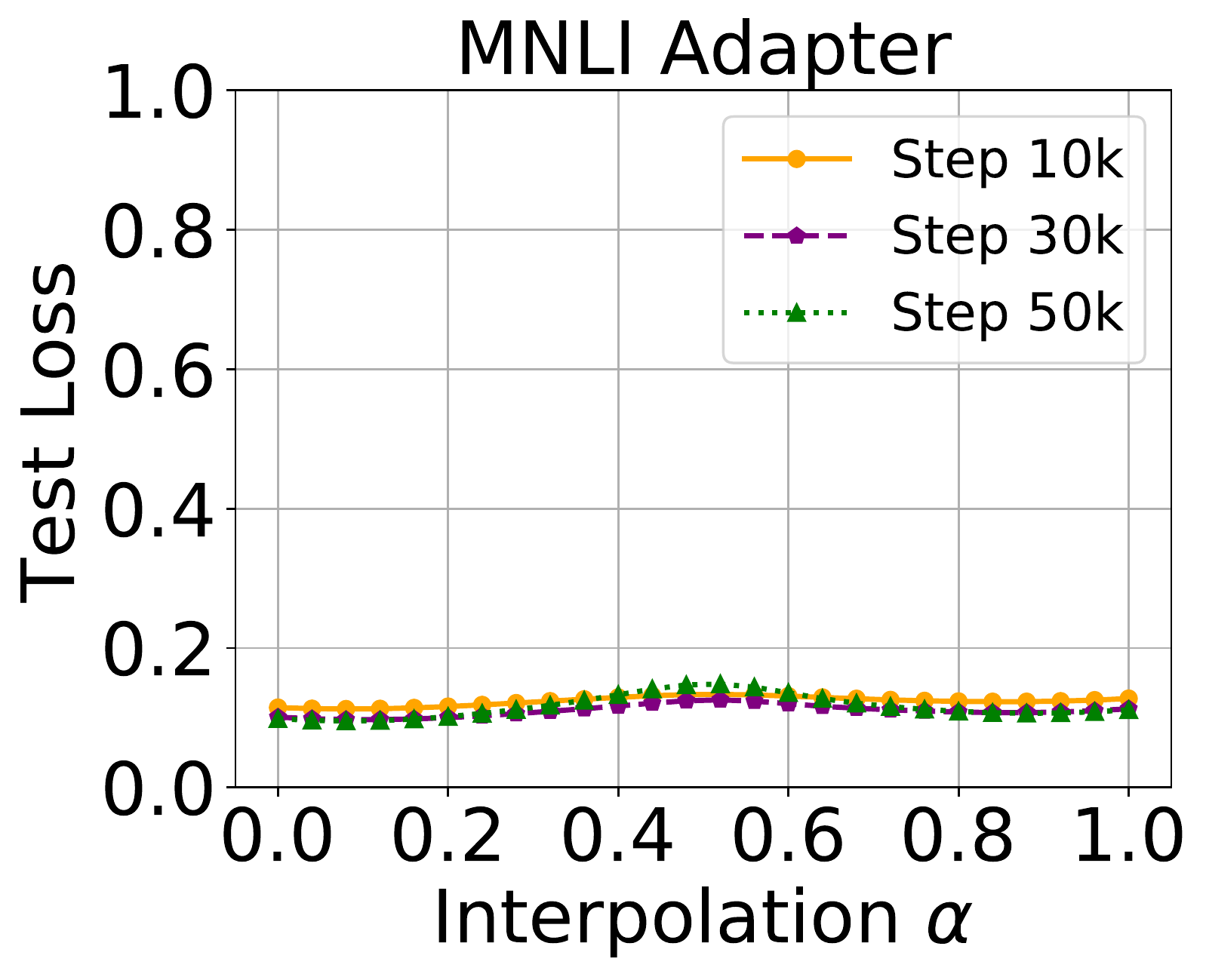}} 
    \subfigure[]{\includegraphics[width=0.24\textwidth]{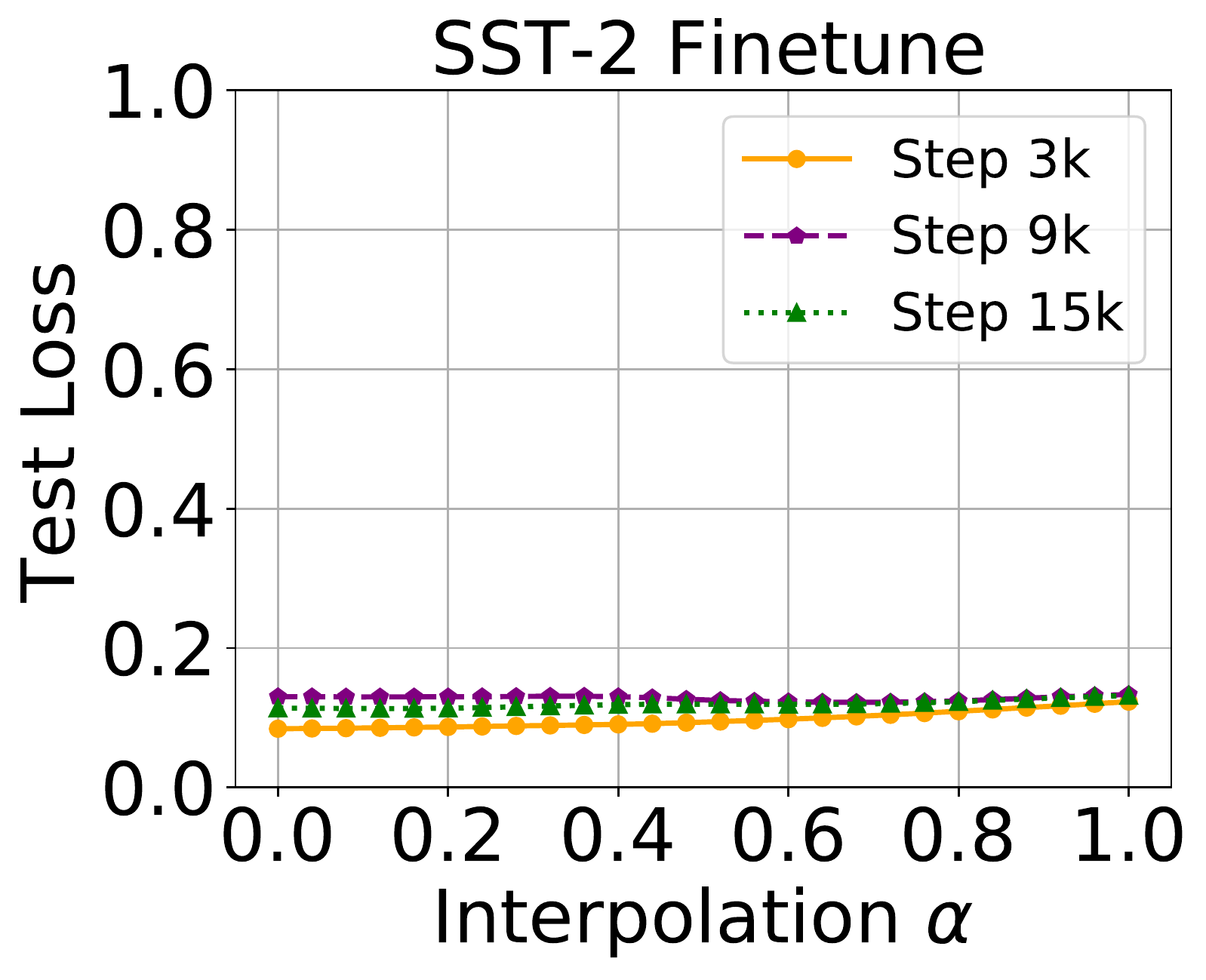}}
    \subfigure[]{\includegraphics[width=0.24\textwidth]{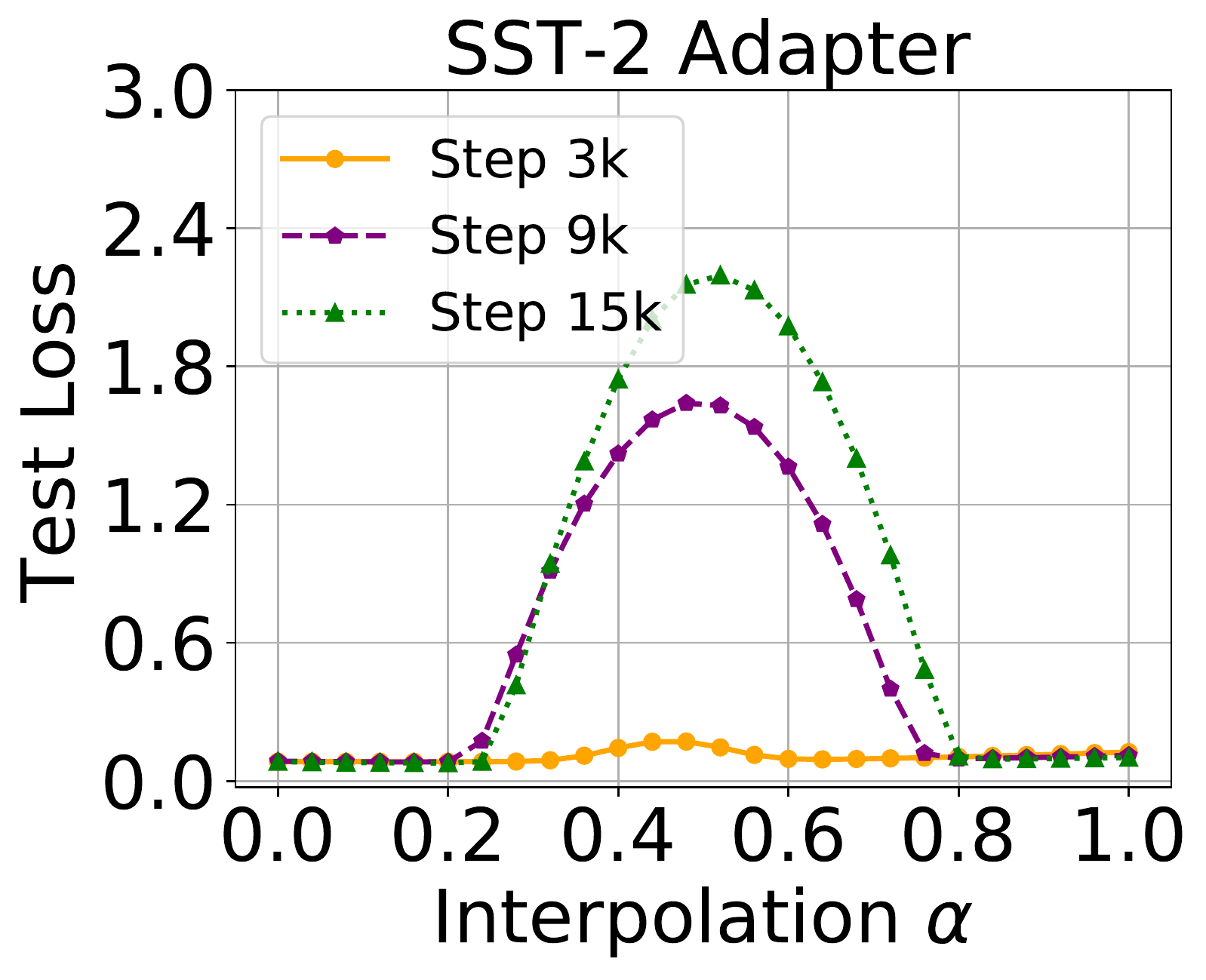}}
    \caption{Experiments of the effects of the batch size. We conduct linear interpolations for MNLI and SST-2, using both fine-tuning and adapter tuning, and visualize their loss. For (a-d), both minima are obtained with a batch size of $16$ and $32$, respectively; for (e-h), both minima are obtained with a batch size of $16$ and $8$, respectively. The corresponding performance visualization is Figure~\ref{fig:bs}.}
    \label{fig:bs_loss}
\end{figure*}

As mentioned before, we record both loss and performance for each interpolation. Since we find that the trends of loss and performance are generally highly correlated, due to the length limit, we only report the performance in the main paper. In this section, we visualize the loss for most of the experiments conducted in this paper, see Figure~\ref{fig:curved_main_paper_loss}, Figure~\ref{fig:pretrain_same_task_loss}, Figure~\ref{fig:order_loss}, Figure~\ref{fig:initialization_loss}, Figure~\ref{fig:overlap_loss}, Figure~\ref{fig:pretrain_diff_task_loss}, Figure~\ref{fig:lr_loss}, and Figure~\ref{fig:bs_loss}.

